\documentclass[twoside, openright,letterpaper,11pt]{report}
\input{Preamble/preamble}

\begin{document}


\thispagestyle{empty}

\begin{titlepage}
\centering
\setstretch{1.0}
\null
\vspace*{0.2in}
\huge
Predictive Whole-Body Control of Humanoid Robot Locomotion\\
\normalfont\large
\Large
\vspace{2em}
Stefano Dafarra
\\
\vspace{2em}
\begin{figure}[ht!]
    \centering
    \subfloat{\includegraphics[width=0.3\textwidth]{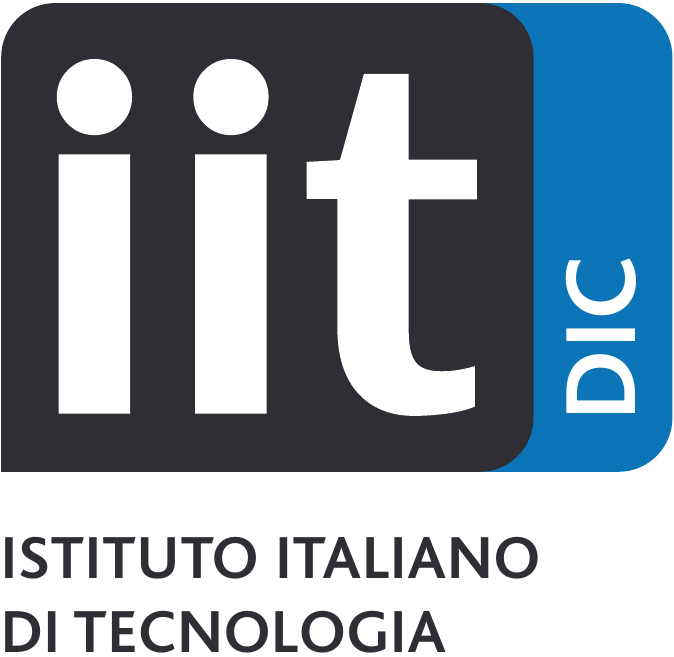}}
    \hspace{0.1\textwidth}
    \subfloat{\includegraphics[width=0.3\textwidth]{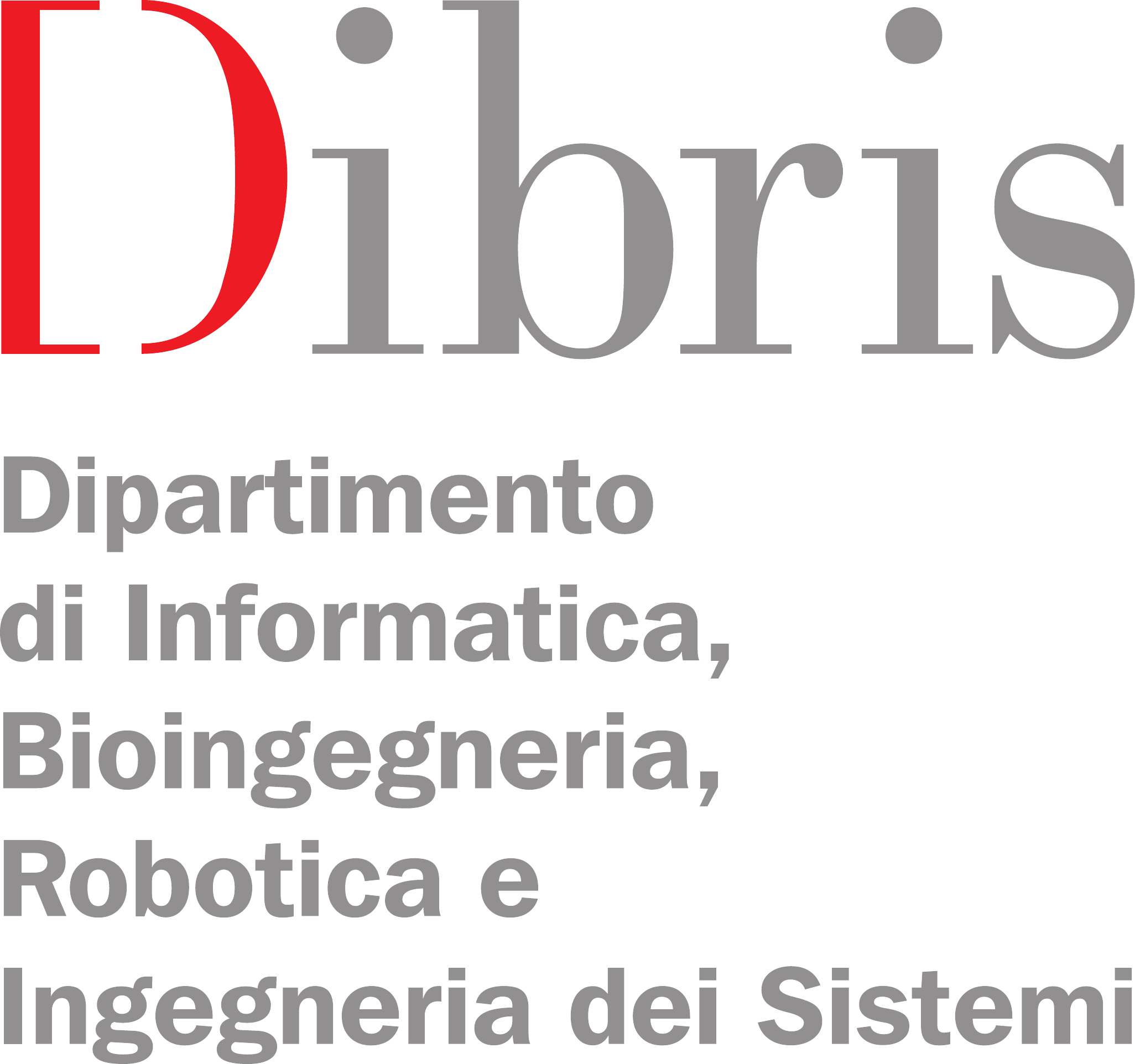}}
\end{figure}
\vspace{2em}
Supervisors: Daniele Pucci, Giorgio Metta\\
\vspace{1em}
\large
Jury Members and Reviewers$^*$\\
\begin{center}
	\begin{tabular}{ l r}
		Rachid Alami & Senior Scientist at LAAS-CNRS, Toulouse \\
		Antonio Franchi & Associate Professor at University of Twente \\
		Robert Griffin$^*$ & Research Scientist at IHMC, Pensacola \\
		Ludovic Righetti & Associate Professor at New York University \\
		Olivier Stasse$^*$ & Senior Researcher at LAAS-CNRS, Toulouse 	
	\end{tabular}
\end{center}
\vspace{1em}
\normalsize
Fondazione Istituto Italiano di Tecnologia\\
Dynamic Interaction Control Lab\\
\vspace{0.5em}
Dipartimento di Informatica, Bioingegneria, Robotica e Ingegneria dei Sistemi, Università di Genova
\\
\vspace{0.5em}
Genova, Italy\\
2020
\\
\par
\end{titlepage}
\clearpage\null\pagenumbering{gobble}\newpage
\begin{abstract}
Humanoid robots are machines built with an anthropomorphic shape. Despite decades of research into the subject, it is still challenging to tackle the robot locomotion problem from an algorithmic point of view. For example, these machines cannot achieve a constant forward body movement without exploiting contacts with the environment. The reactive forces resulting from the contacts are subject to strong limitations, complicating the design of control laws. As a consequence, the generation of humanoid motions requires to exploit fully the mathematical model of the robot in contact with the environment or to resort to approximations of it.

This thesis investigates predictive and optimal control techniques for tackling humanoid robot motion tasks. They generate control input values from the system model and objectives, often transposed as cost function to minimize. 
In particular, this thesis tackles several aspects of the humanoid robot locomotion problem in a crescendo of complexity. First, we consider the single step push recovery problem. Namely, we aim at maintaining the upright posture with a single step after a strong external disturbance. Second, we generate and stabilize walking motions. In addition, we adopt predictive techniques to perform more dynamic motions, like large step-ups.

The above-mentioned applications make use of different simplifications or assumptions to facilitate the tractability of the corresponding motion tasks. Moreover, they consider first the foot placements and only afterward how to maintain balance. We attempt to remove all these simplifications. We model the robot in contact with the environment explicitly, comparing different methods. In addition, we are able to obtain whole-body walking trajectories automatically by only specifying the desired motion velocity and a moving reference on the ground. We exploit the contacts with the walking surface to achieve these objectives while maintaining the robot balanced.

Experiments are performed on real and simulated humanoid robots, like the Atlas and the iCub humanoid robots.

\end{abstract}

\clearpage\null\pagenumbering{gobble}\newpage

\setcounter{page}{1}
\renewcommand{\thepage}{\roman{page}}%

\chapter*{Acknowledgements}
While I am writing these acknowledgments, Italy and several other countries in the world are in lock-down since a couple of weeks due to a pandemic disease. Yet, this difficult period made me realize once again the beauty of research, and the importance that the robotic field ought to have in the future if a similar situation happens again. Looking three years back, I would have never imagined to defend my thesis through a camera, but, at the same time, this Ph.D. taught me the hard way that things rarely go as planned.  

The past three years have been intense and incredibly formative. For this reason I have to deeply thank all the Dynamic Interaction Control lab members, pasts and presents, and the iCub Facility as a whole. I had the invaluable possibility of simply standing up to find interesting sources of discussion, answers to any of my questions or suggestions. All these have been fundamental in my path. It is no surprise that the direction I usually took after standing up was the one towards the Dani's desk. He always had this amazing ability of getting me out of the mud when I got stuck, while being supportive. It is no surprise either that we argued a lot, but this has been crucial to develop a critical thinking.  

In my opinion, one of the best parts of doing a Ph.D. is the possibility of meeting people from all around the world. I had awesome experiences of which I am extremely jealous. Many of these are from my secondment at IHMC, in Pensacola, and I have to thank Dr. Jerry Pratt for the hospitality. I met such great people there and I have great memories. That period was once again the proof that things don't go as you expect, and I am glad they did not. To all those people I met in the Robotics Lab and in Pensacola, I just want to say: ``Thank y'all, Yes''.

These three years have been incredibly challenging from the personal point of view. Being constantly far from home, and having to deal with all the stress of a being Ph.D. student, requires to have great support. I need to thank my love Erica for this. We grew up together and we can easily understand each other with a single look, despite being far away for the most part of the week. She has been incredibly supporting in any of the choices I wanted to take, despite the cost, and what I have is also because of her. Again, I have been pretty lucky, cause I can always count on a great family. My mother Anna and my father Claudio have been a continuous source of inspiration. I also need to thank my two brothers Davide and Matteo. I owe them a lot. Finally, I need to thank all the friends I have. They can always make me smile.

To conclude, I would like to express my gratitude to the reviewers for their valuable feedbacks on the development of this manuscript.

\begin{flushright}
	Stefano.
\end{flushright}


\tableofcontents
\cleardoublepage



\newpage
\setcounter{page}{1}
\renewcommand{\thepage}{\arabic{page}}%

\phantomsection
\addcontentsline{toc}{chapter}{Prologue}
\chapter*{Prologue}
\begin{quotation}
	{\footnotesize
		\noindent\emph{As the births of living creatures, at first, are ill-shapen: so are all Innovations, which are the births of time.}
		\begin{flushright}
			Francis Bacon
		\end{flushright}
	}
\end{quotation}
\vspace{0.5cm}

The term \emph{robot} has been coined by Czech writer Karel \v{C}apek to describe creatures with a human-like appearance, but used for tedious work. Nowadays, humanoid robotics focuses on the study and development of machines where the resemblance to humans extends to the ability of sensing, reasoning and acting. The research efforts aim not only at substituting tedious tasks, but rather to have machines able to replace humans in a multitude of scenarios, including disaster situations. In 2015, the DARPA Robotics Challenge (DRC) took place in Pomona, California, to test 23 humanoid robots in a scenario representative of a nuclear disaster. The tasks to be performed involved, for example, driving a utility vehicle, walking through a door or manipulate a tool to cut a hole in a wall. All the robots were completely untethered, while teams were not allowed to see their robot directly. Many of the biped robots incurred in a fall during the trials \citep{guizzo2015hard}, emphasizing the unripeness of this technology.

More than 40 years after the first moving humanoid robot, walking still proves to be a challenging task for humanoid robots.
Recent developments of optimal control techniques have largely increased their applicability to robotic applications such as humanoid locomotion. 
Optimal control provides a high level of abstraction where control actions are automatically obtained starting from the model of the system under control and a mathematical description of the task to be achieved. In addition, the availability of a prediction window allows performing anticipatory actions which may not be available to classic feedback control laws. This proves to be useful when a humanoid robot has to perform a step, for example. This motion requires it to instantiate a \emph{stable} contact with the environment while absorbing eventual external disturbances. Hence, the mechanical structure behaves differently before and after the impact, requiring the control law to be robust to this critical transition. 

In other cases, the external disturbance is the result of an interaction with a human. Indeed, humanoid robots are meant to share their environment with humans, rather than being constricted behind safety fences. The physical human-robot interaction poses new challenges with social and ethical implications. When performing dynamic tasks, the robot needs to absorb impacts through intrinsic compliance while maintaining balance. As a consequence, it is necessary to consider the full dynamic model of the robot. Nevertheless, when planning motions, it may be necessary to adopt assumptions and simplifications to ease the tractability of the problem at the cost of losing descriptiveness of the model under consideration. 

In this thesis, we apply optimal control techniques to humanoid robot locomotion with a focus on compliance. We adopt a spectrum of models ranging from \emph{simplified} to \emph{complete}, presenting a series of applications to real robotic platforms. In particular, we explore different choices of model depending on the specific context. We finally study the implications of using a kinematically and dynamically detailed model. This research work has been carried out during my Ph.D within the Dynamic Interaction Control laboratory of the Italian Institute of Technology in Genova, Italy. My Ph.D. secondment has been carried out at the Robotics Lab within the Institute of Human and Machine Cognition located in Pensacola, Florida.

This thesis is divided into three parts.

\begin{itemize}
\item Part \ref{part:background} provides the reader with a small background about the concepts exploited in the thesis.
\begin{list}{$\circ$}{}  
	\item Chapter \ref{chap:intro} introduces the thesis with few historical curiosities. It also presents the robots adopted as implementation platforms for the concepts presented in the following parts.
	\item Chapter \ref{chap:oc} introduces a narrow set of optimal control techniques and their terminology. They are presented assuming a generic system under control.
	\item Chapter \ref{chap:robot_model} describes the modeling tools in case the system under control is a humanoid robot.
	\item Chapter \ref{chap:soa_context}  presents a small state of the art overview, defining the thesis context.
\end{list}

\item In Part \ref{part:applications} we exploit the prediction capabilities of optimal control techniques to perform dynamic motions, especially those involving activation and deactivation of contacts.
\begin{list}{$\circ$}{}  
	\item Chapter \ref{chap:steprecovery} presents a technique which leverages the prediction capabilities of the controller to maintain balance by stepping after a large push. It uses an approximated model to generate a set of desired contact forces. Results are shown in simulation.
	\item Chapter \ref{chap:iros_walking} exploits prediction to maintain the tracking of a reference trajectories while taking into account several constraints arising during each walking phase. This controller is part of a layered architecture which allows the iCub robot to walk both in position and torque mode. 
	\item Chapter \ref{chap:stepups} uses trajectory optimization to perform large step-ups. We show that these techniques can generate dynamic motions obtaining a consistent maximum joint torque reduction. Experiments are performed on the Atlas humanoid robot.
\end{list}

\item In Part \ref{part:dynamic_planner} we frame the locomotion problem into the optimal control framework, showing that it is possible to obtain walking motions automatically. Indeed, locomotion is not a trivial task. Simply moving forward may be considered a straightforward objective. The design of a controller which follows directly a forward velocity reference may guarantee a high reward in the short term. The robot may simply lean forward, achieving a perfect tracking of the velocity reference. Clearly, this can be incompatible with the constraints imposed by contacts. Hence, it may be necessary to sacrifice some of the short term reward to achieve a better objective in the long term. For this reason, we believe that in case of highly constrained systems with non-trivial objectives, such as the locomotion of humanoids, prediction is fundamental. 

At the same time, the definition of references plays an important role. Instead of specifying the full path the robot is supposed to follow, our goal is to define only a minimum set of references, thus limiting the effort from the operator. 
\begin{list}{$\circ$}{}  
	\item Chapter \ref{chap:modeling_dp} is mainly devoted to the definition of contacts, which is critical for locomotion planning. It presents the model adopted in the non-linear optimal control problem.
	\item Chapter \ref{chap:tasks} shows how we shape the cost function to obtain walking motions with a minimum set of references.
	\item Chapter \ref{chap:experiments} validates the walking planner and presents results on the real robot.
\end{list}
\end{itemize}

\section*{Summary of publications}
\nobibliography*

\noindent The content of Chapter \ref{chap:steprecovery} appears in:
\fcite{dafarra2017receding}
\vspace{5mm}

\noindent The content of Chapter \ref{chap:iros_walking} appears in:
\fcite{dafarra2018control}
\vspace{5mm}

\noindent The content of Chapter \ref{chap:stepups} will eventually appear in:
\fcite{dafarra2020stepup}
\vspace{5mm}

\noindent The content of Part \ref{part:dynamic_planner} partially appears in:
\fcite{dafarra2020dynamic}
\vspace{5mm}
\vspace{3mm}
All the videos associated with the papers and to the results included in this thesis are in the following playlist: \url{https://www.youtube.com/playlist?list=PLBOchT-u69w9hJ6BmqPf06r0zWmungOrc}.

\vspace{3mm}
\noindent
List of other contributions:

\begin{leftbar}
	\begin{quote}%
		\bibentry{shafiee2019online} \vspace{5mm}\newline 
		\bibentry{romualdi2018benchmarking} \vspace{5mm}\newline 
		\bibentry{romualdibenchmarking} \vspace{5mm}\newline
	    \bibentry{romano2017codyco} \vspace{5mm}\newline 
	    \bibentry{nava2017modeling} \vspace{5mm}\newline 
		\bibentry{elobaid2019telexistence}
	\end{quote}
\end{leftbar}
\epigraphhead[500]{\begin{quotation}
		{\footnotesize
			\noindent\emph{All models are wrong; some models are useful.}
			\begin{flushright}
				George E. P. Box 
			\end{flushright}
		}
	\end{quotation}}
\part{Background and Fundamentals}\label{part:background}
\chapter{Introduction} \label{chap:intro}
In the June of 1696, Johann Bernoulli, professor of mathematics at the University of Groningen, posed the following challenge in the issue of the \emph{Acta Eruditorum} journal \citep{sussmann2002brachistochrone}: 
\begin{quote}
\textit{If in a vertical plane two points A and B are given, then it is required to specify the orbit AMB of the movable point M, along which it, starting from A, and under the influence of its own weight, arrives at B in the shortest possible time. [...]}
\end{quote}

\begin{figure}[tpb]
	\centering
	\includegraphics[width=.9\columnwidth]{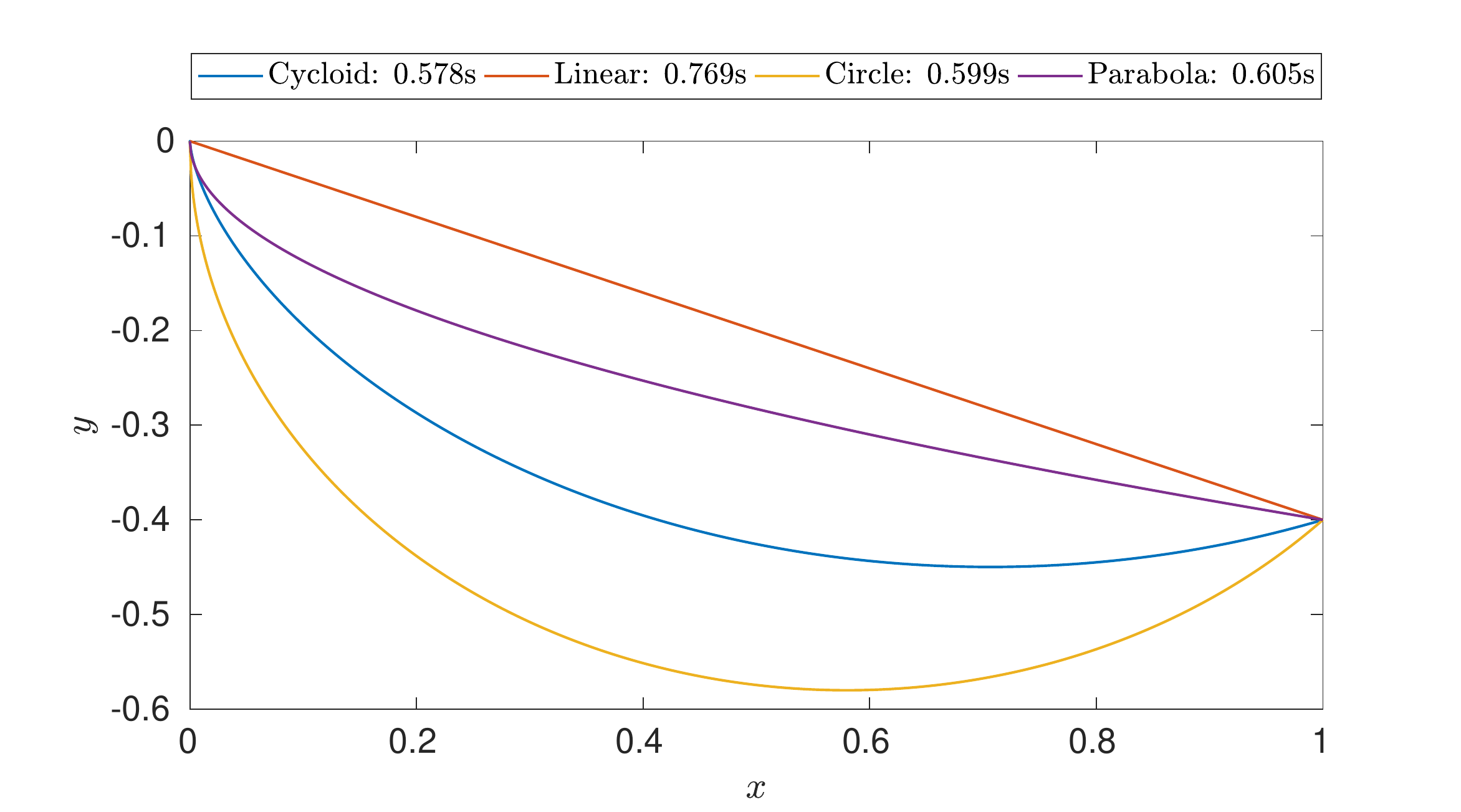}
	\caption{Different trajectories of a point with unit mass moving from (0,0) to (1.0, -0.4) according to the brachystochrone problem. The duration of each trajectory is shown in the legend. The ``Circle'' and ``Parabola'' are computed such that the tangent in the first point is vertical, thus having the highest initial acceleration.}	
	\label{fig:brachystochrone}
\end{figure}

\begin{figure}[tpb]
	\centering
	\includegraphics[width=.9\columnwidth]{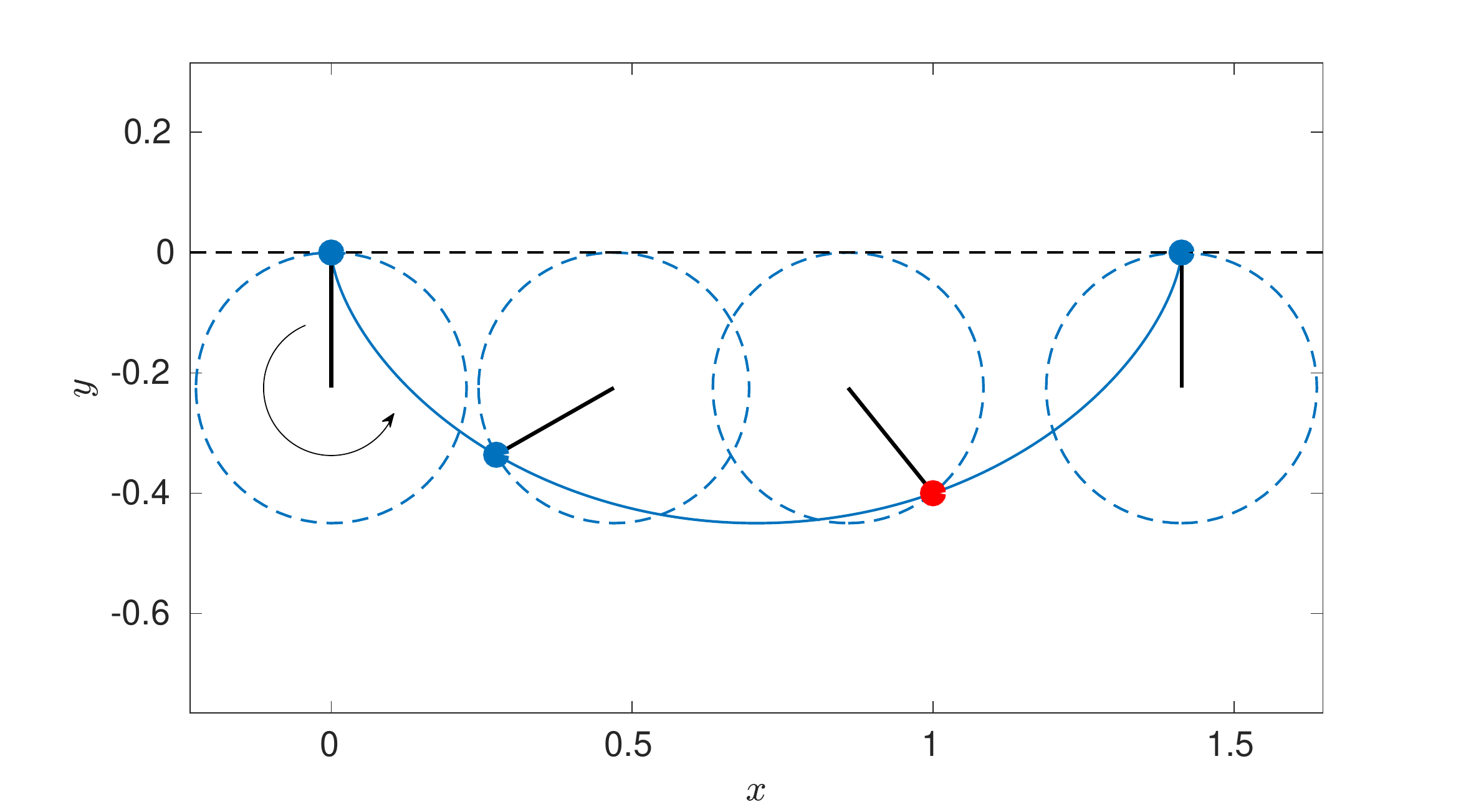}
	\caption{Graphical representation of the circle generating the cycloid of Fig. \ref{fig:brachystochrone}. It rolls counterclockwise along the black dashed line performing a full turn. The blue point, rigidly attached to the circle outer edge, draws the cycloid. The red point is the end point of the trajectories in Fig. \ref{fig:brachystochrone}.}	
	\label{fig:cycloid}
\end{figure}

This is known as the \emph{Brachystochrone problem}, from the Greek ``shortest time''. Johann himself, Leibniz, Johann's elder brother Jakob, Tschirnhaus, l'Hopital and Newton (even if in anonymous form) provided a solution to this problem, indicating the cycloid to be the trajectory with minimum time. We show an example of this curve in Fig. \ref{fig:brachystochrone}. In particular, Johann Bernoulli found this solution using the Fermat's principle of least time \citep{erlichson1999johann}. The cycloid is the curve drawn by a point fixed to the outer edge of a circle ``rolling'' without slipping on a horizontal line, as depicted in Fig. \ref{fig:cycloid}.

Even if the solution found by Bernoulli exploited a method well known in optics, the formulation of the challenge corresponds to an \emph{optimal control problem}. Given the system dynamics, a ball moving under the effect of its own weight, the goal is to transfer its state between two points minimizing a metric, the total duration. Hence, 1696 can be considered as the year of birth of optimal control \citep{sussmann1997300}. In addition, the \emph{Brachystochrone problem} is inherently related to \emph{locomotion} (from the Latin \emph{moving from a place}) since it involves the motion of a body.

Leonhard Euler, who became a student of Johann Bernoulli at the age of 13, published in 1744 a book entitled \emph{The Method of Finding Plane Curves that Show Some Property of Maximum and Minimum}, providing the foundations of what is known as Euler's equation. Eleven years later, a 19-year-old Joseph-Louis Lagrange wrote Euler a brief letter describing a revolutionary idea  which refined Euler's method. Lagrange derived a necessary condition for optimality, known today as the \emph{Euler-Lagrange} equation.

\begin{figure}[tpb]
	\centering
	\subfloat[] {\includegraphics[width=.45\textwidth]{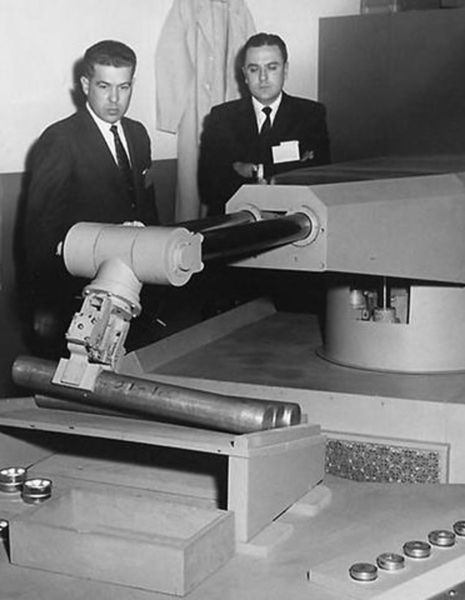}\label{fig:unimate}}
	\hspace{0.05\textwidth}	
	\subfloat[] {\includegraphics[width=.45\textwidth]{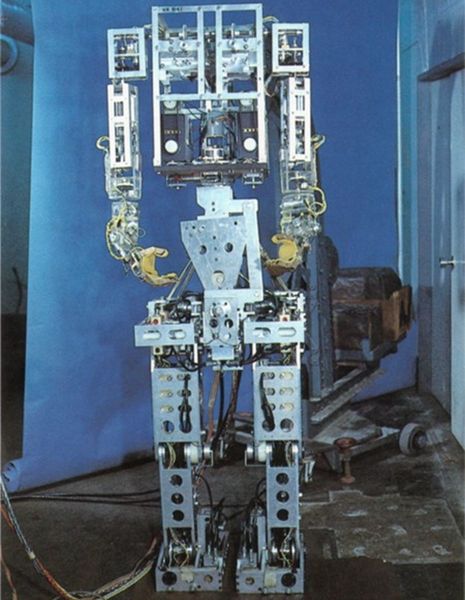}\label{fig:wabot}}
	\caption{(a) A picture of the Unimate first prototype (picture downloaded from \url{www.robotics.org/joseph-engelberger/about.cfm}). (b) The WABOT-1 humanoid robot (picture from \url{www.humanoid.waseda.ac.jp/booklet/kato_2.html}). }
\end{figure}

The \emph{Euler-Lagrange} equation is ubiquitous in robotics too. It allows deriving the equation of motions governing a multi-body system. Hence, the work by Lagrange defines a particular common ground between optimal control and robotics, exploited in the second part of the 20th century. In fact, the first industrial robot, the \emph{Unimate $\#$001} shown in Fig. \ref{fig:unimate}, was developed in 1959, while the first machine capable of human-like locomotion appeared in 1972. Its name is WABOT-1 (Fig. \ref{fig:wabot}) and it is the result of the WABOT project \citep{kato1974information} from the Waseda University started in 1967. It weighs approximately 130\textrm{kg} and it moves thanks to 11 hydraulically powered joints. Walking is achieved by splitting the motion in different phases and by storing the corresponding joint references in the ``mini-computer'' equipped on the robot. An analog circuit tries to match the desired joint value with the one measured by a potentiometer.

Nowadays, humanoid locomotion is still a challenging problem. One of the difficulties is the lack of mathematical tools able to formally describe ``balanced'' and ``falling'' states of a mechanical structure with limbs. Indeed, some of the most successful conditions are drawn under strong assumptions and simplifications \citep{Pratt2006, vukobratovic2004zero}. In addition, walking implies exploiting contacts with the environments. The mathematical description of a mechanical system with impacts is an active research field \citep{olejnik2017modeling, azhmyakov2019relaxation,  rijnen2019sensitivity}. 

In this context, the abstraction and prediction capabilities of optimal control techniques could facilitate the development of algorithms able to maintain the balance of complicated machines like humanoids, while achieving locomotion tasks. The next sections will present the two research platforms exploited in this thesis: the iCub and the Atlas humanoid robots.
\section{The iCub humanoid robot} \label{sec:icub}
\begin{figure*}[tpb]
	\centering
	\subfloat[] {\includegraphics[width=.45\textwidth]{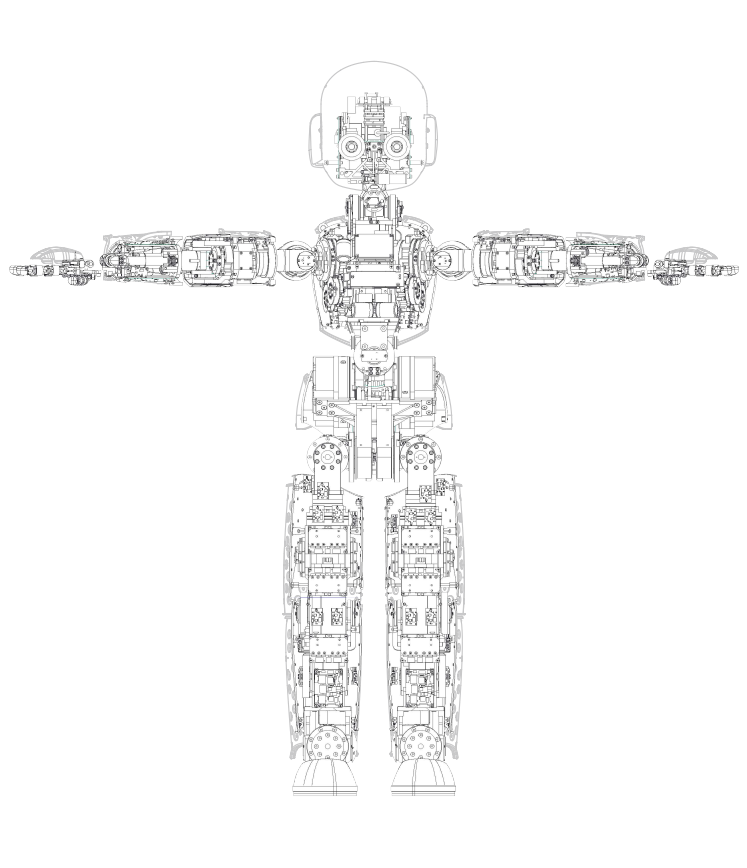}\label{fig:icubWiring}}
	\hspace{0.05\textwidth}	
	\subfloat[] {\includegraphics[width=.45\textwidth]{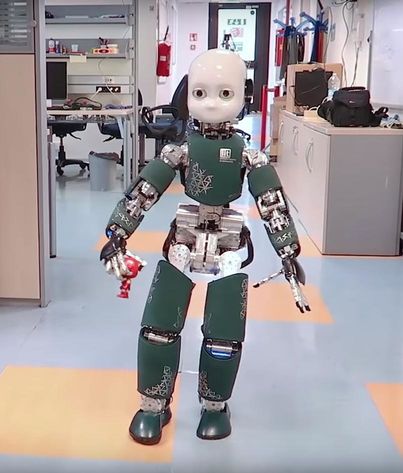}\label{fig:icubWalking}}
	\caption{(a) A wire-frame view of the iCub robot with some of the electronics in transparency. (b) A picture of the robot fully untethered while walking during a tele-operation experiment.}
	\label{fig:icub}
\end{figure*}

The iCub humanoid robot is a state-of-the-art open-source robotic platform  developed at the Italian Institute of Technology as part of the European project RobotCub \citep{metta2005robotcub, Nataleeaaq1026}. More than 40 of these robots have been built in different versions and distributed in several laboratories worldwide. 

\subsection{Hardware}

The iCub humanoid robot, depicted in Fig. \ref{fig:icub}, is 104 cm tall, weighing 33 kg. It possesses in total 53 degrees of freedom including those in the hands and in the eyes. For locomotion purposes, only 23 joints are employed and they are distributed in the following way:
\begin{itemize}
	\item ~6 joints in each leg,
	\item ~3 joints in the torso,
	\item ~4 joints in the arm, 3 of which in the shoulder and one in the elbow.
\end{itemize}

The torso and shoulder joints are mechanically coupled and driven by tendon mechanisms. All 23 joints are powered by brushless electric motors equipped with Harmonic Drive transmissions with a reduction ratio of 1/100.

The robot is powered either by an external supplier or by a custom made battery. In the first case, the operating continuous voltage is 40V with about 3A current, depending also on the task executed by the robot. The battery, instead, provides 36VDC and it has a capacity of 9.3Ah, which allows about 45 minutes of continuous usage.

\begin{figure}[tpb]
\centering
\subfloat[] {\includegraphics[width=.40\textwidth]{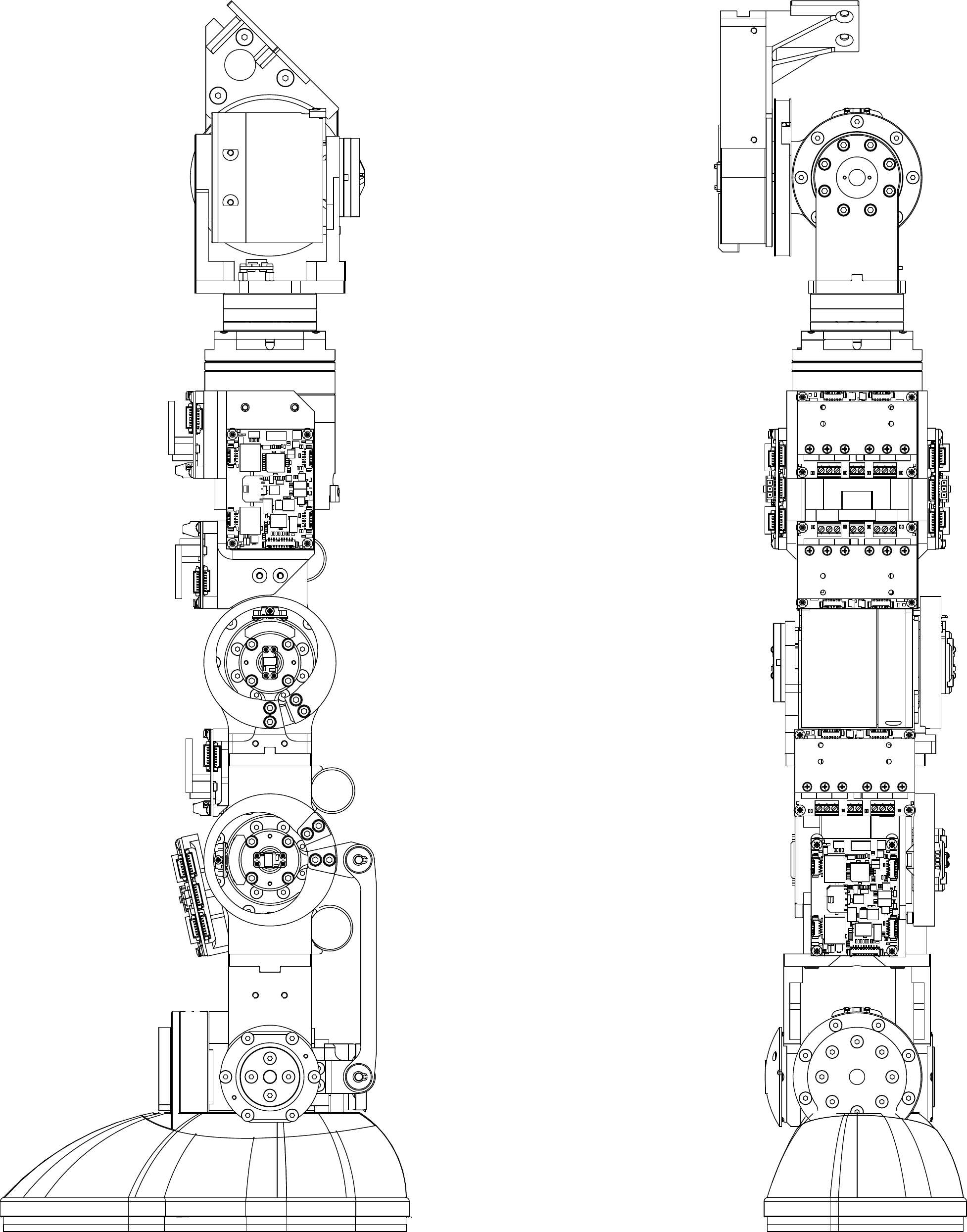}\label{fig:icubleg}}
\hspace{0.1\textwidth}	
\subfloat[] {\includegraphics[width=.32\textwidth]{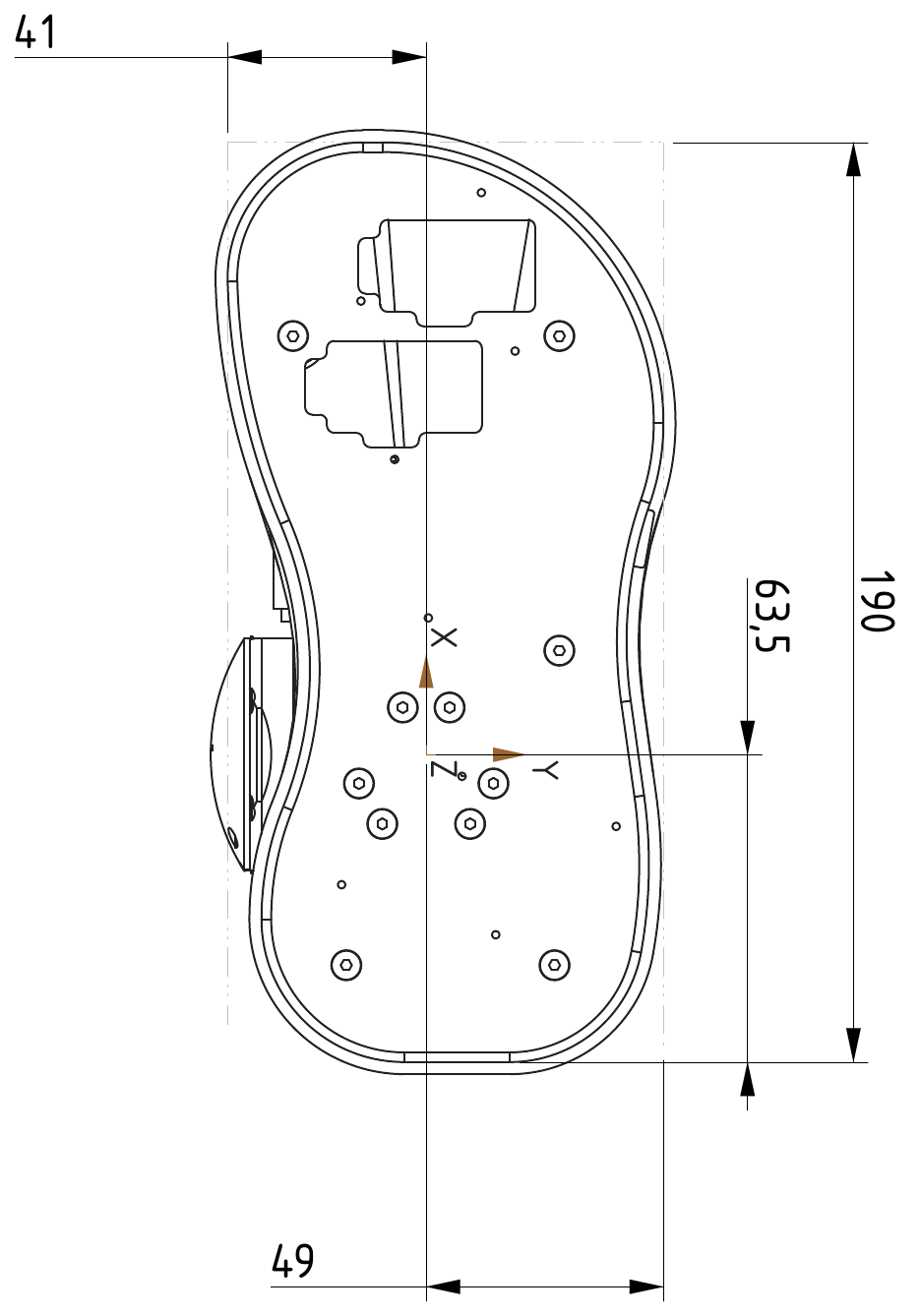}\label{fig:icubfoot}}
\caption{(a) Lateral and front view of the iCub left leg mechanics. (b) Bottom view of the iCub left foot with its dimensions in millimeters. The right foot is specular about the longitudinal axis.}
\end{figure}

A particular feature of iCub is the vast array of sensors available, that includes force/torque sensors, accelerometers, gyroscopes, a distributed tactile skin, two VGA cameras and microphones. More in detail, iCub possesses 6 six-axes force/torque (F/T) sensors \citep{Fumagalli2012}. In particular, two of them are mounted at the shoulder, the other four respectively at the hips and ankles. iCub also possesses distributed tactile sensors as an artificial skin \citep{Cannata2008,Maiolino2013}, which provides informations about both the location and the intensity of the contact forces. Notably, the skin is distributed on the torso, arms, the palm and fingertips of hands and legs.
Because iCub is not endowed with joint torque sensors, the F/T sensors and the skin are used to provide an estimation of both the internal torques and external forces \citep{Fumagalli2012}.
Two inertial sensors are mounted on the robot: one on the head an the other on the waist. They provide acceleration and angular velocity information.

All the motors and sensors are controlled through more than 30 electronic boards spread on the body, as sketched in Fig. \ref{fig:icubWiring}. 
They are connected trough an Ethernet network (CAN in previous versions or in some particular boards) in daisy chain. Fig. \ref{fig:icubleg} presents the mechanical structure of the left leg and some of the motor control boards. An on-board computer, located in the robot head, provides a bridge between the internal network and external computers. It is equipped with a 4$^{th}$ generation Intel\textsuperscript{\textregistered} Core i7@1.7GHz and 8GB of RAM running Ubuntu. The connection to the robot can be established through an Ethernet cable or wirelessly thanks to a standard 5GHz Wi-Fi network.
Lastly, Fig. \ref{fig:icubfoot} shows the robot particular foot shape and its dimensions.

\subsection{Software infrastructure}
\begin{figure}[tpb]
	\includegraphics[width=.9\textwidth]{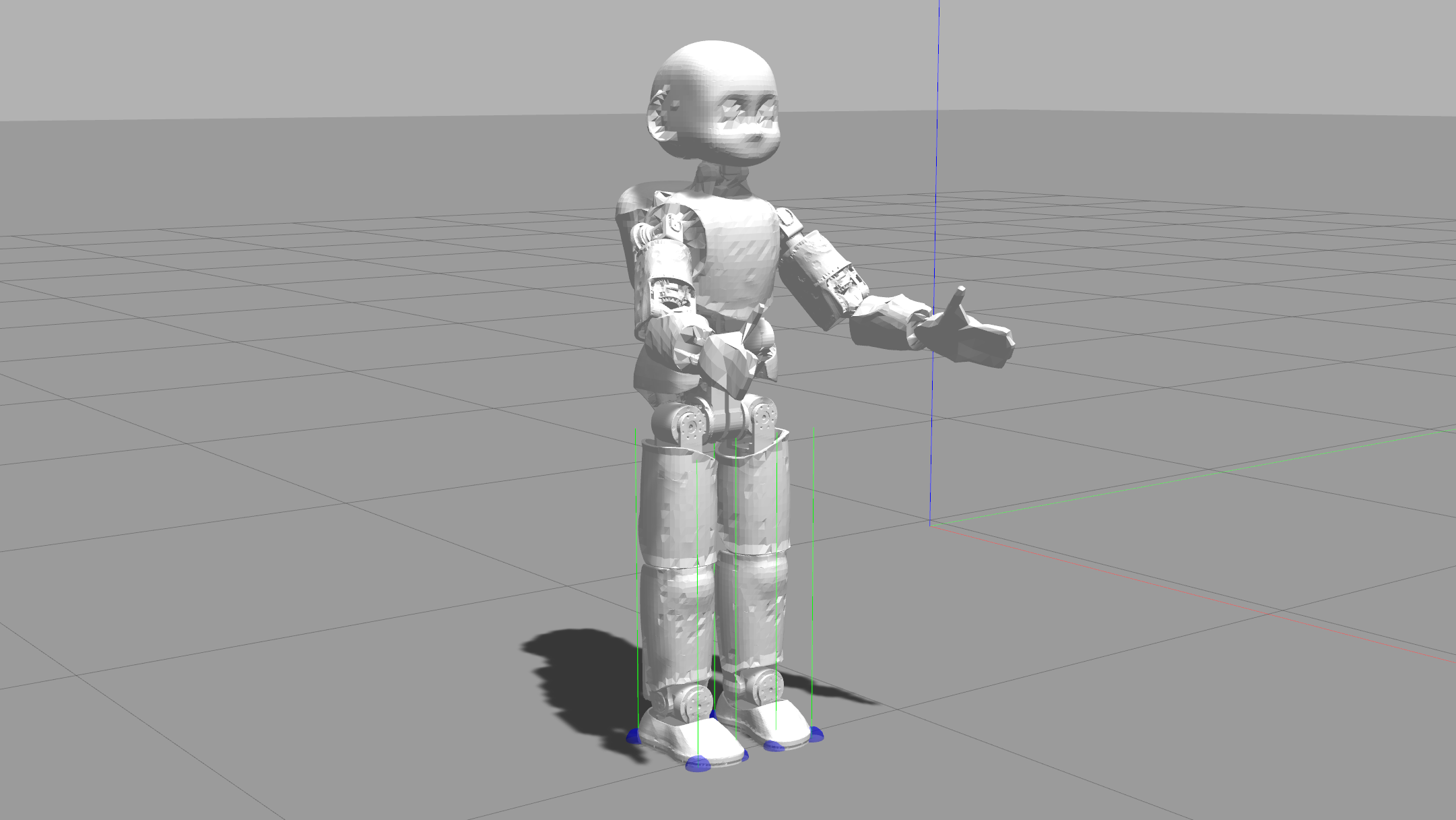}
	\caption{The iCub humanoid robot in the Gazebo environment.}
	\label{fig:icubgazebo}
\end{figure}
To control the robot, an infrastructure of computers is necessary. To this purpose, the software Yet Another Robot Platform (YARP) \citep{metta2006yarp} is employed. It is a software middleware whose main purpose is to allow seamless communication between ``applications'', which can reside on different computers. Furthermore, it provides interfaces to interact with physical devices independently of the actual implementation, thus facilitating code reuse and modularity. The acquisitions from sensors and motor controllers are provided through YARP interfaces.

Apart of the YARP middleware, an additional software layer is available, called iDynTree \citep{Frontiers2015}. Its objective is to simplify the writing of whole-body controllers. It wraps dynamic computations such as the estimation of the mass matrix or inverse dynamics. The interface has been coded in C++. Considerably, these interfaces can also be accessed in  MATLAB\textsuperscript{\tiny\textregistered} functions, or Simulink\textsuperscript{\tiny\textregistered} projects through the use of WB-Toolbox library\citep{RomanoWBI17Journal}. Indeed, the ease of developing a control system with Simulink, together with the possibility of exploiting simple debugging tools and existing toolboxes, makes possible the rapid prototyping of controllers.

Controller prototyping also exploits the Gazebo simulation environment \citep{Koenig04}. It is an open-source simulator for multi-body systems able to detect and handle contacts through a ``compliant'' model. It is highly flexible. A simulated version of the robot, shown in Fig. \ref{fig:icubgazebo}, can be controlled through the corresponding Yarp plugins \citep{YarpGazebo2014} allowing to apply the very same controller on both the simulator and the real robot, thus making the actual implementation transparent.

\section{The IHMC Atlas humanoid robot}
\begin{figure*}[tpb]
	\includegraphics[width=0.9\textwidth]{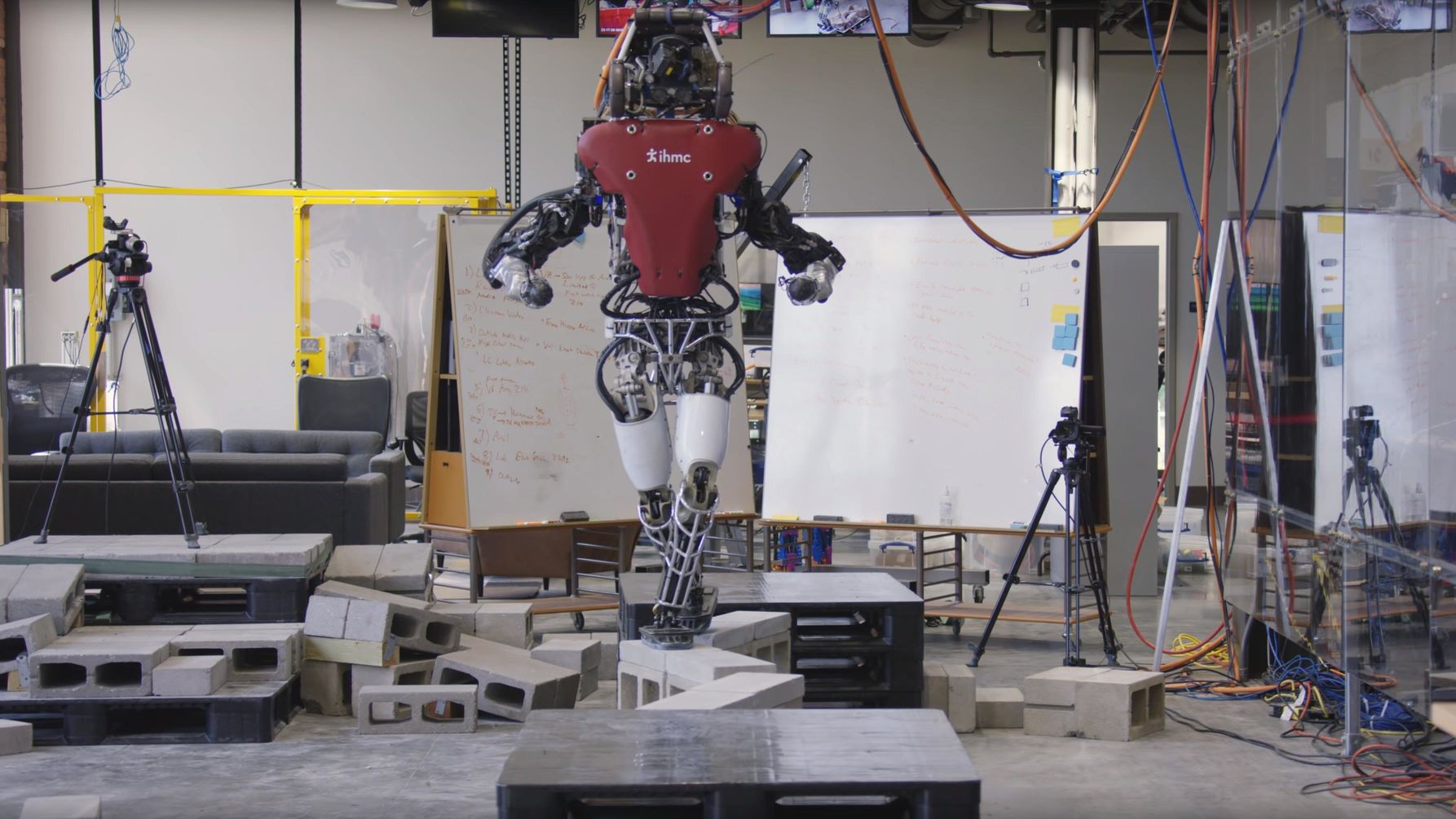}

	\caption{The IHMC Atlas humanoid robot performing a complex walking task on top of a narrow cinder block pile.}
	\label{fig:atlasWalking}
\end{figure*}
The author had the honor and the pleasure of testing some of the algorithms presented in this thesis on the Atlas humanoid robot granted to the Institute of Human and Machine Cognition (IHMC). The IHMC team, located in Pensacola, Florida, won the first place at the DARPA Virtual Robotics Challenge \citep{koolen2013summary} and the second place at the DARPA Robotics Challenge Trials and Finals \citep{johnson2015team, johnson2017team}.

The robot has been designed and built by Boston Dynamics\textsuperscript{\tiny\textregistered}. Nevertheless, the experimental results obtained with this platform have been possible thanks also to the software infrastructure developed by the Robotics Lab at IHMC. Hence, in the following we refer to this robot as IHMC Atlas.

\subsection{Hardware}

\begin{figure*}[tpb]
	\centering
	\subfloat[] {\includegraphics[width=.45\textwidth]{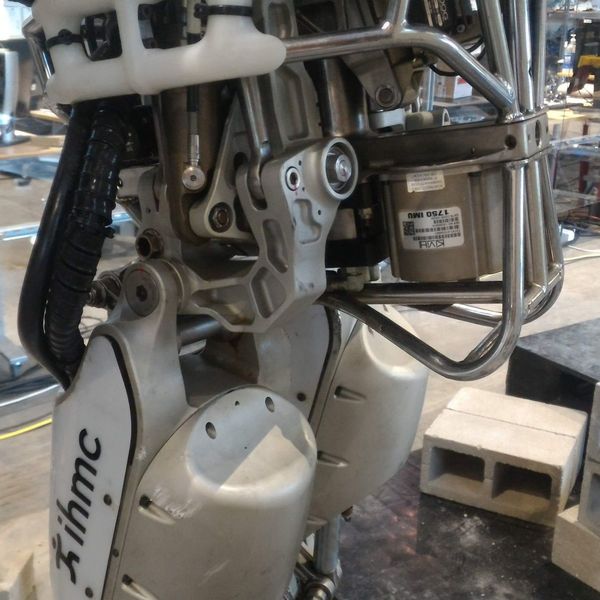}\label{fig:atlasHip}}
	\hspace{0.05\textwidth}	
	\subfloat[] {\includegraphics[width=.45\textwidth]{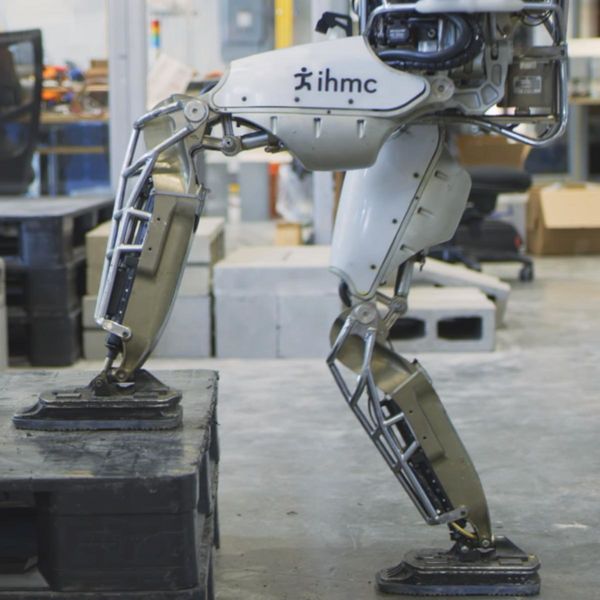}\label{fig:atlasLegs}}
	\caption{A close-up view of the Atlas legs, also showing the KVH\textsuperscript{\tiny\textregistered} IMU located on the pelvis.}
	\label{fig:atlasLeg}
\end{figure*}

The Atlas humanoid robot, showed in Fig. \ref{fig:atlasWalking}, is a hydraulically powered humanoid robot designed by Boston Dynamics\textsuperscript{\tiny\textregistered}. It exploits an electric pump and an accumulator located slightly above the backpack, to maintain a hydraulic fluid at a desired pressure of 2300 PSI. It is possible to control the motion of hydraulic linear actuators located at each joint by controlling the opening of hydraulic valves. The forearms are an exception since they are controlled by electric motors. They are visible in Fig. \ref{fig:atlasWalking} thanks to their gray color, different from the rest of the arms. 
The robot has 30 degrees of freedom with six in each leg, seven in the arms, one in the neck, and an additional three in the pelvis. Fig. \ref{fig:atlasLeg} focuses on the leg mechanics, with a close-up view of the hip roll and pitch joints and on the full leg before performing a large step-up. The force applied by each hydraulic actuator is measured through pressure sensors in the valve chambers. This measurement and a joint acceleration feed-forward term are used to implement joint torque control \citep{koolen2016design}. Indeed, hydraulic actuation is affected by high friction due to the viscous fluid flowing in the valves chambers and to the seals. This feed-forward term and a mechanism to compensate for joint backlash \citep{koolen2016design} allow the implementation of force control at the robot feet, which is fundamental for walking.
Leg joint positions are measured through the elongation of the hydraulic actuators.

Atlas weighs 150$\mathrm{kg}$ and stands 1.88m tall. It requires 480V three-phase voltage provided offboard by an extensible tether. During the DARPA Robotics Challenge, it worked with an on-board battery pack that permitted untethered operation for over an hour.
A Carnegie Robotics MultiSense-SL head provides two forward-facing cameras and an axial rotating
Hokuyo LIDAR. Atlas also has two wide-angle cameras intended to compensate for the robot not being able to yaw its head. An interchangeable end-effector mount terminates each arm, allowing third party hand installations. The robot possess a KVH\textsuperscript{\tiny\textregistered} IMU with fiber optic gyros located on the pelvis, and load cells at the feet to form a 3-axes force-torque sensor.

The sensor and valve electronics are connected via a CAN network. Each appendage has a separate CAN link connected to a main hub placed behind the front cover. The firmware governing these peripheral nodes is proprietary to Boston Dynamics\textsuperscript{\tiny\textregistered}. A cluster of four embedded computers running Ubuntu is placed in the backpack and can be accessed through an Ethernet switch to communicate with the robot.
\subsection{Software infrastructure}
\begin{figure}[tpb]
	\includegraphics[width=.9\textwidth]{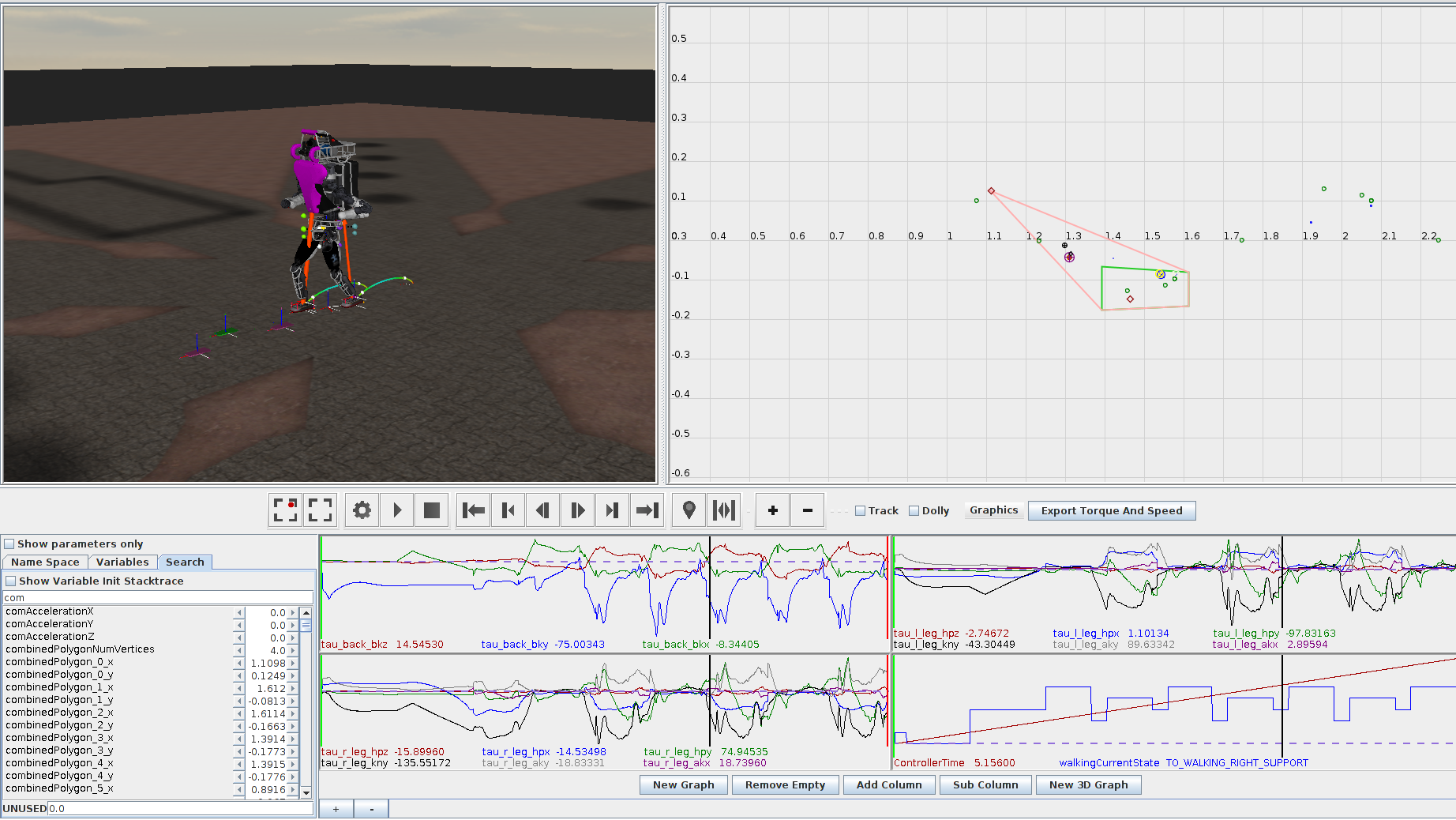}
	\caption{The IHMC Atlas humanoid robot walking in the Simulation Construction Set environment. It presents many graphical elements indicating relevant quantities like desired foot poses, center of mass state, controller phase, joint torques, and others.}
	\label{fig:scs}
\end{figure}
The software infrastructure strongly relies on Java \citep{smith2014real}. The estimation thread and the main balancing controller run on the on-board computer at respectively $1\mathrm{kHz}$ and $333\mathrm{Hz}$ \citep{koolen2016design}. More information on the controller is presented in Sec. \ref{sec_ls:qp_controller}. The communication with these modules is performed by means of a custom Java implementation of ROS2 messages \footnote{See \url{https://github.com/ihmcrobotics/ihmc-java-ros2-communication}.}.

The code-base contains over 3000 unit test \citep{johnson2017team} which could be related to a single class or to a high-level behavior, like walking or even opening doors. Indeed, many of these tests require simulating the robot interacting with the environment. This is done through the \emph{Simulation Construction Set} (SCS) simulator\footnote{The source code is available at the following link: \url{https://ihmcrobotics.github.io/simulation-construction-set/docs/scshome.html}.}. An example of the SCS graphical user interface is shown in Fig. \ref{fig:scs}. In addition to the visualization of the robot in the top left box, it also allows on-line plotting of any of the relevant quantities used inside the code. This simulator has been fundamental in the development of other robots by the IHMC team \citep{pratt2009yobotics, cotton2012fastrunner}. SCS integrates the rigid body dynamics equations \citep{featherstone2014rigid} while using a compliant model for handling the contacts. Simulations are usually deterministic and hence highly reproducible. Nevertheless, it is possible to introduce artificial noise and disturbances in the measurements to mimic a real scenario.

The SCS infrastructure can also be used to replay data logged from the robot during experiments. The values over time of more than 14000 variables are stored after each run of the controller, together with up to four time-synced video streams coming from external cameras. This tool strongly simplifies the debugging process involved when performing experiments on the real robot.

\section{Notation}\label{sec:notation}
Throughout the thesis we will use the following notation.
\begin{itemize}
	\item Vector and matrices are expressed with a bold symbol.
	\item The $i_{th}$ component of a vector $\bm{x}$ is denoted as $x_i$. 
	\item The transpose operator is denoted by $(\cdot)^{\top}$.
	\item The superscript $(\cdot)^*$ indicates desired values.
	\item Given a function of time $f(t)$ the dot notation denotes the time derivative, i.e.
		$\dot{f} := \frac{\dif f}{\dif t}$. Higher order derivatives are denoted with a corresponding amount of dots.
	\item Given a function ${f}$, we define with $\nabla_{\bm{x}}f(\bm{y})$ the partial derivative of $f$ with respect to $\bm{x}$, evaluated in $\bm{y}$.
	\item $\mathcal{I}$ is a fixed inertial frame with respect to (w.r.t.) 
	which the robot's absolute pose is measured. Its $z$ axis is supposed to point against gravity, while the $x$ direction defines the forward direction.
	\item $\mathds{1}_n \in \mathbb{R}^{n \times n}$ denotes the identity matrix of dimension $n$.
	\item $\bm{0}_{n \times n} \in \mathbb{R}^{n \times n}$ denotes a zero matrix while $\bm{0}_n = \bm{0}_{n \times 1}$ is a zero column vector of size $n$.
	\item $\bm{e}_i$ is the canonical base in $\mathbb{R}^n$, i.e. $\bm{e}_i = [0, 0, \dots, 1, 0, \dots, 0]^\top \in \mathbb{R}^n$, where the only unitary element is in position $i$. Throughout the thesis, $n$ will be either 2 or 3 depending on the context. 
	\item The operator $\wedge$ defines the skew-symmetric operation associated with the cross product in $\mathbb{R}^3$. Its inverse is the operator vee $\vee$.
	\item The weighted L2-norm of a vector $\bm{v} \in \mathbb{R}^n$ is denoted by $\|\bm{v}\|_{\bm{W}}$, where $\bm{W} \in \mathbb{R}^{n\times n}$ is a weight matrix.
	\item $^{A}\bm{R}_{B} \in SO(3)$ and $^{A}\bm{H}_{B} \in SE(3)$ denote the rotation and transformation matrices which transform a vector expressed in the $B$ frame, $^B \bm{x}$, into a vector expressed in the $A$ frame, $^A \bm{x}$.
	\item ${}^D\bm{V}_{A,D} \in \mathbb{R}^6$ is the relative velocity between frame $A$ and $D$,  whose coordinates are expressed in frame $D$.
	\item $\bm{x}_\text{CoM} \in  \mathbb{R}^3$ is the position of the center of mass with respect to $\mathcal{I}$.
	\item $\bm{R}_2(\theta) \in SO(2)$ is the rotation matrix of an angle $\theta \in \mathbb{R}$; $\bm{S}_2=R_2(\pi/2)$ is the unitary skew-symmetric matrix.
	\item $\mathbf{n}(\cdot), \mathbb{R}^3 \rightarrow \mathbb{R}^3$ is a function returning the direction normal to the walking plane given the argument $x$ and $y$ coordinates.
	\item $\mathbf{t}(\cdot), \mathbb{R}^3 \rightarrow \mathbb{R}^{3\times2}$ is a function returning two perpendicular directions normal to $\mathbf{n}(\cdot)$. The composition of $\mathbf{t}(\cdot)$ and $\mathbf{n}(\cdot)$, $\left[\mathbf{t}(\cdot)~\mathbf{n}(\cdot)\right]$, defines the rotation matrix ${}^\mathcal{I}\bm{R}_{plane}$.
	\item $h(\bm{p}), \mathbb{R}^3 \rightarrow \mathbb{R}$ defines the distance between $\bm{p}$ and the walking surface. 
	\item $diag(\cdot), \mathbb{R}^n \rightarrow \mathbb{R}^{n \times n}$ is a function casting the argument into the corresponding diagonal function.
	
\end{itemize}

\chapter{Optimal Control and Non-Linear Optimization Basics}\label{chap:oc}
In this chapter, we present the basics and terminology of optimal control and non-linear optimization. In particular, Sec. \ref{sec:oc_basics} defines the elements composing an optimal control problem, while Sections \ref{sec:indirect} and \ref{sec:direct} introduce the methods used to solve it. Sec. \ref{sec:receding_horizon} presents the \emph{receding horizon principle}, which is the base of the main predictive algorithm used in this thesis. Finally, Sec. \ref{sec:nlp} provides a background on non-linear optimization, used as a tool for the solution of optimal control problems.
\section{Optimal control basics}\label{sec:oc_basics}
Optimal control allows achieving a specified objective while minimizing a metric indicating the performances of the control actions. Compared to traditional techniques, there is a paradigm shift: the designer has not to define a control law, but rather to describe the system under control, and the objective to be achieved, in mathematical terms. The tuning process moves from the definition of controller gains to the weights describing the relative importance of each objective. The resulting control law is obtained by applying methods presented in the following. 

Consider a generic dynamical system
\begin{equation}\label{eq:ode}
	\dot{\bm{x}} = \bm{f}\left(\bm{x}(t), \bm{u}(t), t\right),
\end{equation}
where its time evolution is a function $\bm{f}: \mathbb{R}^{n} \times \mathbb{R}^m \times \mathbb{R}  \rightarrow \mathbb{R}^{n}$ depending on the state $\bm{x} \in \mathbb{R}^{n}$, the control variables $\bm{u} \in \mathbb{R}^{m}$ and time. Eq. \eqref{eq:ode} is an \emph{Ordinary Differential Equation} (ODE). The problem of determining the evolution of the system, given an initial value $\bm{x}(t_0) = \bm{x}_0$, is called \emph{Initial Value Problem} (IVP). On the contrary, if the terminal condition is specified, i.e. $\bm{x}(T) = \bm{x}_T$, we have a \emph{Boundary Value Problem} (BVP).

Often, a simple ODE may not be sufficient to model complex dynamical systems. Indeed, it may be necessary to introduce algebraic constraints, like
\begin{equation}\label{eq:constraint}
	{}_g\mathbf{l} \leq \bm{g}\left(\bm{x}(t), \bm{u}(t), t\right) \leq {}_g\mathbf{u}.
\end{equation}
$\bm{g}: \mathbb{R}^{n} \times \mathbb{R}^m \times \mathbb{R}  \rightarrow \mathbb{R}^{n_g}$ is a generic constraint function of dimension $n_g$, bounded by ${}_g\mathbf{l} \in \mathbb{R}^{n_g}$ and ${}_g\mathbf{u} \in \mathbb{R}^{n_g}$. When ${}_g\mathbf{l}$ and ${}_g\mathbf{u}$ coincide, Eq. \eqref{eq:constraint} is an equality constraint. Then, together with Eq. \eqref{eq:ode}, they define a \emph{Differential Algebraic Equation} (DAE). On the contrary, when the bounds do not coincide, the set of equations \eqref{eq:ode}$-$\eqref{eq:constraint} is defined as \emph{inequality constrained DAE}. The bounds on the control inputs, which are usually present in physical systems, represent a common example of constraint.

The DAE consisting of Eq.s \eqref{eq:ode} and \eqref{eq:constraint} represents a first ingredient of an optimal control problem, defining a mathematical description of the system under control. In addition, the initial condition $\bm{x_0}$ is also specified (e.g. feedback from the system). The second ingredient is the metric measuring how well the control law is performing according to the designer's objective, i.e. the \emph{cost function}. Consider applying the optimal controller in a fixed time window, i.e. $t \in \left[t_0, \, T\right]$. Then, the cost function is the following:
\begin{equation}\label{eq:cost}
	\mathcal{J}\left(\bm{x}_0, \bm{u}(\cdot), t\right) = \text{m}\left(\bm{x}(T)\right) + \int_{t_0}^T \ell \left(\bm{x}(\tau), \bm{u}(\tau)\right) \mathrm{d}\tau.
\end{equation}
The function $\text{m} : \mathbb{R}^{n} \rightarrow \mathbb{R}$ is called \emph{Mayer} term and weights the terminal state, while the function $\ell : \mathbb{R}^{n} \times \mathbb{R}^m \times \mathbb{R} \rightarrow \mathbb{R}$ is the \emph{Lagrange} term. The former allows specifying the goal the system has to reach in the terminal time, while the latter defines the preferred path to reach such goal.

Finally, an optimal control problem minimizes Eq. \eqref{eq:cost} through a control policy $\bm{u}(\cdot)$ in the interval $\left[t_0, \, T\right]$, i.e. $\bm{u}\left[t_0, T\right]$. It can be written as:
\begin{IEEEeqnarray}{CL}
	\phantomsection \IEEEyesnumber \label{eq:ocp}
	\minimize_{\bm{u}\left[t_0, T\right]} & \mathcal{J}\left(\bm{x}_0, \bm{u}(\cdot), t\right) = \text{m}\left(\bm{x}(T)\right) + \int_{t_0}^T \ell \left(\bm{x}(\tau), \bm{u}(\tau)\right) \mathrm{d}\tau \IEEEyessubnumber \\
	\text{subject to: } & \nonumber \\
	& 	\dot{\bm{x}} = \bm{f}\left(\bm{x}(t), \bm{u}(t), t\right), \IEEEyessubnumber \label{eq:dynamics_ocp}\\[1pt]
	&   \bm{x}(t_0) = \bm{x}_0, \IEEEyessubnumber \\[1pt]
	&	{}_g\mathbf{l} \leq \bm{g}\left(\bm{x}(t), \bm{u}(t), t\right) \leq {}_g\mathbf{u}. \IEEEyessubnumber
\end{IEEEeqnarray}

Techniques for solving the optimal control problem described in Eq. \eqref{eq:ocp} can be classified as either \emph{direct} or \emph{indirect} methods.
\section{Indirect methods}\label{sec:indirect}

Indirect methods attempt to find a solution to Eq. \eqref{eq:ocp} by exploiting conditions for optimality. Historically, these have been developed considering no constraints, i.e. $n_g = 0$.

Denote the optimal \emph{cost-to-go} $\mathcal{J}^\star(\bm{x}(t), t)$ as the following:
\begin{equation}\label{eq:cost_to_go}
	\mathcal{J}^\star(\bm{x}(t), t) = \inf_{u[t, T]}  \text{m}\left(\bm{x}(T)\right) + \int_{t}^T \ell \left(\bm{x}(\tau), \bm{u}(\tau)\right) \mathrm{d}\tau.
\end{equation}
Notice that $\mathcal{J}^\star(\bm{x}(t), t)$ depends on $\bm{x}(t)$ but not on the state evolution up to state $t$. In this regard, let us introduce the \emph{Bellman principle of optimality} \citep[Chap. 3.3]{bellman1957dynamic}:
\begin{customQuote}{3.5em}
\textit{An optimal policy has the property that whatever the initial state and initial decision are, the remaining decisions must constitute an optimal policy with regard to the state resulting from the first decision.}
\end{customQuote}

Consider an instant $t_1: t \leq t_1 \leq T$. If the optimal control policy has been applied in the interval from $t_1$ to $T$, the optimal \emph{cost-to-go} starting from $t$ is obtained by minimizing the sum of the cost from $t$ to $t_1$, plus the optimal cost from $t_1$ to $T$. In fact, we can split Eq. \eqref{eq:cost_to_go} as follows: 
\begin{equation}
\begin{split}
	\mathcal{J}^\star(\bm{x}(t), t) = &\inf_{u[t, t_1]}\left\{ \int_{t}^{t_1} \ell \left(\bm{x}(\tau), \bm{u}(\tau)\right) \mathrm{d}\tau  + \right.\\
	+&\left.\inf_{u[t, t_1]}\left[\text{m}\left(\bm{x}(T)\right) + \int_{t_1}^T \ell \left(\bm{x}(\tau), \bm{u}(\tau)\right) \mathrm{d}\tau \right]\right\}.
\end{split}
\end{equation}
Then, the second part corresponds to the optimal cost-to-go starting from $t_1$ up to $T$. Thus, the Bellman principle of optimality implies:
\begin{equation}
	\mathcal{J}^\star(\bm{x}(t), t) = \inf_{u[t, t_1]}\left\{ \int_{t}^{t_1} \ell \left(\bm{x}(\tau), \bm{u}(\tau)\right) \mathrm{d}\tau  + \mathcal{J}^\star(\bm{x}(t_1), t_1) \right\}.
\end{equation}

Consider $t_1 = t + \mathrm{d}t$. Then, by employing the mean value theorem, there exists a constant $\alpha \in \left[0, 1\right]$ such that 
\begin{equation*}
	\mathcal{J}^\star(\bm{x}(t), t) {=} \inf_{u[t, t + \mathrm{d}t]}\left\{ \ell \left(\bm{x}(t + \alpha\mathrm{d}t), \bm{u}(t + \alpha\mathrm{d}t)\right)\mathrm{d}t {+} \mathcal{J}^\star(\bm{x}(t + \mathrm{d}t), t + \mathrm{d}t) \right\}.
\end{equation*}
Under regularity assumptions for the \emph{cost-to-go} function, we can obtain $ \mathcal{J}^\star(\bm{x}(t + \mathrm{d}t), t + \mathrm{d}t)$ through Taylor expansion:
\begin{equation}
\begin{split}
	 \mathcal{J}^\star(\bm{x}(t + \mathrm{d}t), t + \mathrm{d}t) =~  &\mathcal{J}^\star(\bm{x}(t), t) + \frac{\partial \mathcal{J}^\star(\bm{x}(t), t)}{\partial \bm{x}} \frac{\partial \bm{x}(t)}{\partial t}\mathrm{d}t + \\
	 &+ \frac{\partial \mathcal{J}^\star(\bm{x}(t), t)}{\partial t} \mathrm{d}t + O(\mathrm{d}t)^2.
\end{split}
\end{equation}
This relation can be substituted into the optimal \emph{cost-to-go} formula, obtaining
\begin{equation}
\begin{split}
	\mathcal{J}^\star(\bm{x}(t), t) = &\inf_{u[t, t + \mathrm{d}t]}\left\{\ell \left(\bm{x}(t + \alpha\mathrm{d}t), \bm{u}(t + \alpha\mathrm{d}t)\right)\mathrm{d}t + \mathcal{J}^\star(\bm{x}(t), t) + \right. \\
	& \left. + \frac{\partial \mathcal{J}^\star(\bm{x}(t), t)}{\partial \bm{x}} \frac{\partial \bm{x}(t)}{\partial t}\mathrm{d}t + \frac{\partial \mathcal{J}^\star(\bm{x}(t), t)}{\partial t} \mathrm{d}t + O(\mathrm{d}t)^2 \right\}.
\end{split}
\end{equation}
First we simplify and then we divide by $\mathrm{d}t$:
\begin{equation*}
\begin{split}
    0 = \inf_{u[t, t + \mathrm{d}t]}\bigg\{&\ell \left(\bm{x}(t + \alpha\mathrm{d}t), \bm{u}(t + \alpha\mathrm{d}t)\right) + \frac{\partial \mathcal{J}^\star(\bm{x}(t), t)}{\partial \bm{x}} \bm{f}\left(\bm{x}(t), \bm{u}(t), t\right) + \\
    &\left. +\frac{\partial \mathcal{J}^\star(\bm{x}(t), t)}{\partial t} + O(\mathrm{d}t) \right\}.
\end{split}
\end{equation*}
Notice that $\frac{\partial \mathcal{J}^\star(\bm{x}(t), t)}{\partial t}$ does not depend on the control input. In addition, the \emph{cost-to-go} in the terminal instant $T$ corresponds to the \emph{Mayer} term. Finally, by letting $\mathrm{d}t \rightarrow 0$, we obtain
\begin{equation} \label{eq:hjb}
\left\{
	\begin{IEEEeqnarraybox}[][c]{RCLL}
	\frac{\partial \mathcal{J}^\star(\bm{x}(t), t)}{\partial t} &=& -\inf_{u}\bigg\{&\ell \left(\bm{x}(t), \bm{u}(t)\right) + \\
	&&&+\frac{\partial \mathcal{J}^\star(\bm{x}(t), t)}{\partial \bm{x}} \bm{f}\left(\bm{x}(t), \bm{u}(t), t\right)\bigg\}, \\
	\mathcal{J}^\star(\bm{x}(T), T) &=& \IEEEeqnarraymulticol{2}{L}{\text{m}\left(\bm{x}(T)\right).}
	\end{IEEEeqnarraybox}
\right.
\end{equation}
Eq. \eqref{eq:hjb} is the \emph{Hamilton-Jacobi-Bellman}(HJB) equation of optimal control \citep[Sec. 3.11]{kirk2012optimal}. It represents a necessary condition for a control policy to be optimal. Under regularity assumption for $J^*$, it is also sufficient. 

Indirect methods attempts to find the optimal control policy starting from optimality conditions like the HJB. Hence, indirect methods require the user to solve the partial differential equation defined in Eq. \eqref{eq:hjb}, limiting the applicability of the controller to the specific system under control. In addition, the introduction of constraints (especially inequalities) is not straightforward. Finally, these methods may lack robustness \citep[Sec. 4.3]{betts2010practical}. Small changes to the boundary conditions may result in very different policies.

Notably, in case of unconstrained linear time-invariant systems subject to quadratic costs, Eq. \eqref{eq:hjb} can be solved analytically obtaining the \emph{linear-quadratic regulator}.

\subsection{Linear Quadratic Regulator} \label{sec:lqr}
The linear quadratic regulator applies to linear time invariant (LTI) systems, characterized by the following relation:
\begin{equation}\label{eq:lti}
\dot{\bm{x}} = \bm{A}\bm{x} + \bm{B}\bm{u}.
\end{equation}
The cost function to be minimized is quadratic, corresponding to:
\begin{equation}
	\mathcal{J} = \frac{1}{2}\bm{x}(T)^\top \bm{F} \bm{x}(T) + \frac{1}{2}\int_{t_0}^T \left(\bm{x}(\tau)^\top \bm{Q} \bm{x}(\tau) + \bm{u}(\tau)^\top \bm{R} \bm{u}(\tau)\right) \dif \tau,
\end{equation}
where $\bm{F}, \bm{Q} \in \mathbb{R}^{n\times n}$ are positive semi-definite matrices, while $\bm{R} \in \mathbb{R}^{m \times m}$ is a positive definite matrix. They are all user-defined gain matrices.
Through the HJB equations, we obtain this simple control law \citep[Sec. 3.12]{kirk2012optimal}:
\begin{equation}\label{eq:state_feedback}
	\bm{u}(t) = - \bm{K}(t)\bm{x}(t),
\end{equation}
where $\bm{K} \in \mathbb{R}^{m\times n}$ is a time-varying gain matrix given by:
\begin{equation}
	\bm{K}(t) = \bm{R}^{-1}\bm{B}^\top\bm{P}(t).
\end{equation}
$\bm{P}(t) \in \mathbb{R}^{n\times n}$ is obtained by solving the continuous time \emph{Riccati differential equation} \citep{bittanti2012riccati}:
\begin{equation}
	\dot{\bm{P}}(t) = -\bm{Q} + \bm{P}(t)\bm{B}\bm{R}^{-1}\bm{B}^\top\bm{P}(t) - \bm{P}(t)\bm{A} - \bm{A}^\top\bm{P}(t)
\end{equation}
with the boundary condition
\begin{equation}
	\bm{P}(T) = \bm{F}.
\end{equation}

Hence, the LQR provides the optimal gains of a state-feedback control law, obtained by solving a \emph{boundary value problem}. 

The LQR can be easily extended to the case where the horizon is infinite, corresponding to the following cost function:
\begin{equation}
\mathcal{J} = \frac{1}{2}\int_{0}^\infty \left(\bm{x}(\tau)^\top \bm{Q} \bm{x}(\tau) + \bm{u}(\tau)^\top \bm{R} \bm{u}(\tau)\right) \dif \tau.
\end{equation}
In this case, the optimal control law is equal to Eq. \eqref{eq:state_feedback}, with the difference that $\bm{P}(t) = \bar{\bm{P}}$ is constant, obtained by computing the stationary point of the Riccati equation:
\begin{equation}
0 = -\bm{Q} + \bar{\bm{P}}\bm{B}\bm{R}^{-1}\bm{B}^\top\bar{\bm{P}} - \bar{\bm{P}}\bm{A} - \bm{A}^\top\bar{\bm{P}}.
\end{equation}
This is called \emph{Algeraic Riccati equation}. The advantage of this approach is that the gain matrix is constant and can be computed off-line.
\section{Direct methods}\label{sec:direct}
Direct methods reformulate Eq. \eqref{eq:ocp} as an optimization problem, thus \emph{directly} attempting at finding the minimum of Eq. \eqref{eq:cost} subject to dynamics and algebraic constraints. A generic optimization problem is the following:
\begin{IEEEeqnarray}{CL}
	\phantomsection \IEEEyesnumber \phantomsection \label{eq:optimization}
	\minimize_{\bm{\chi}} & \Gamma(\bm{\chi}) \IEEEyessubnumber \\
	\text{subject to: } & \nonumber \\
	&	{}_h\mathbf{l} \leq \bm{h}\left(\bm{\chi}\right) \leq {}_h\mathbf{u}. \IEEEyessubnumber
\end{IEEEeqnarray}
Compared to Eq. \eqref{eq:ocp} there are some notable differences. First of all we have only one set of variables $\bm{\chi} \in \mathbb{R}^{n_\chi}$. In addition the dynamics (Eq. \eqref{eq:dynamics_ocp}) or the concept of time evolution of the system are absent. This gap is overcome by discretizing the continuous system dynamics using numerical integration techniques, at the cost of inserting approximation errors. The discretization techniques are the same used to simulate generic ODEs.

\subsection{Common integration methods}\label{sec:integration_methods}
Integration methods determine an approximated value for $\bm{x}(t + \mathrm{d}t)$ given the state evolution:
\begin{equation}
	\bm{x}(t + \mathrm{d}t) = \bm{\upsilon}\left(\bm{x}(t), \bm{u}(t), t\right).
\end{equation}
The integrator $\bm{\upsilon}: \mathbb{R}^{n} \times \mathbb{R}^m \times \mathbb{R}  \rightarrow \mathbb{R}^{n}$ is a function considering only one previous value, $\bm{x}(t)$. In case other previous values are also considered, e.g. $\bm{x}(t-\mathrm{d}t)$, $\bm{x}(t-2\mathrm{d}t)$, and so on, the integration method is called \emph{multi-step}. In this thesis we will focus on \emph{single-step} integration methods only.
The most common \emph{single-step} method is the \emph{forward Euler} integration scheme:
\begin{equation}\label{eq:forward_euler}
	\bm{x}(t + \mathrm{d}t) = \bm{x}(t)  + \mathrm{d}t \bm{f}\left(\bm{x}(t), \bm{u}(t), t\right),
\end{equation}
which can be seen as an approximation of the derivative operator, i.e.
\begin{equation}
	\dot{\bm{x}} \approx \frac{\bm{x}(t + \mathrm{d}t) - \bm{x}(t)}{\mathrm{d}t}.
\end{equation}
Interestingly, it can be shown that for sufficiently large $\mathrm{d}t$, this method can be subjected to numerical instability. In other words, depending on $\mathrm{d}t$, it may be possible that the integrated state evolution diverges from the equilibrium point even if the dynamical system has a globally asymptotically stable equilibrium. On the contrary, the \emph{backward Euler} method
\begin{equation} \label{eq:backward_euler}
	\bm{x}(t + \mathrm{d}t) = \bm{x}(t)  + \mathrm{d}t \bm{f}\left(\bm{x}(t + \mathrm{d}t), \bm{u}(t + \mathrm{d}t), t + \mathrm{d}t\right)
\end{equation}
is numerically stable independently from the chosen time step \citep[Sec. 3.5]{ascher1997computer}. This is a property of \emph{implicit single step} methods. The \emph{implicitness} is due to the presence of $\bm{x}(t + \mathrm{d}t)$ on both sides of Eq. \eqref{eq:backward_euler}, defining an \emph{algebraic loop}. When using this method to simulate a dynamical system, it would be necessary to solve such algebraic loop at every integration step, usually through a \emph{Newton's method} \citep[Sec. 1.5]{betts2010practical}, increasing consistently the computational burden.

Another implicit scheme is the trapezoidal method:
\begin{equation}
\begin{split}
	\bm{x}(t + \mathrm{d}t) = \bm{x}(t) + \frac{\mathrm{d}t}{2} \Big(&\bm{f}\left(\bm{x}(t), \bm{u}(t), t\right) + \\
	& + \bm{f}\left(\bm{x}(t + \mathrm{d}t), \bm{u}(t + \mathrm{d}t), t + \mathrm{d}t\right)\Big)
\end{split}
\end{equation}
which evaluates the dynamics at both ends of the integration step. For this reason, it is considered a \emph{multiple collocation} method, while Euler schemes consider a single collocation point, i.e. they evaluate the system dynamics only once in the integration step. Another difference with respect to Euler methods is its accuracy. Indeed, the trapezoidal method is of \emph{order} 2, while Euler methods are of \emph{order} 1. The order indicates the approximation error introduced by the method. Intuitively, a method of order $k$ corresponds to a Taylor expansion stopped at the $k^{th}$ order, without the need of computing all the partial derivatives. Hence, higher order methods allow obtaining more accurate results using the same integration step. 

All the these single step methods can be represented via a Butcher tableau \citep[Sec. 232]{butcher2016numerical}. It allows storing all the relevant coefficients in a condensed form. For example, the Butcher tables of Euler methods are
\begin{equation*}
\text{Forward Euler: }
\begin{array}{c|c}
0 & 0\\
\hline
& 1
\end{array}, \quad \text{Backward Euler: }
\begin{array}{c|c}
1 & 1 \\
\hline
& 1
\end{array}
\end{equation*}
while the implicit trapezoidal one is 
\begin{equation*}
\begin{array}{c|c c}
0 &  0  &  0\\
1 & 1/2 & 1/2 \\
\hline
& 1/2 & 1/2
\end{array}.
\end{equation*}

In its general form, the table has the following structure:
\begin{equation}
\begin{array}{ c | c c c c c }
c_1 & a_{1,1} & a_{1,2} & \dotsm & a_{1,s} \\
c_2 & a_{2,1} & a_{2,2} & \dotsm & a_{2,s}  \\
\vdots & \vdots & \vdots & \ddots & \vdots \\
c_s & a_{s,1} & a_{s,2} & \dotsm & a_{s,s} \\ \hline
& b_1 & b_2 & \dotsm & b_s
\end{array} =
\begin{array} {c | c}
\mathbf{c} & \bm{A} \\ \hline \\[-10pt]
& \mathbf{b}^T
\end{array}.
\end{equation} 
$\mathbf{c}$ contains the portions of time step at which \emph{stages} will be evaluated (i.e. stage $i$ will be evaluated at time $t + c_i \mathrm{d}t$). The $i$-th stage, defined here with the symbol ${}^i\bm{s}$ is an intermediate evaluation of function $\bm{f}$:
\begin{equation}\label{eq:stage}
{}^i\bm{s}\left(\bm{x}(t), \bm{u}(\cdot), t\right) = \bm{f}\bigg(\bm{x}(t) + \mathrm{d}t \sum_{j=1}^{n_s} a_{i,j}{}^j\bm{s}, \bm{u}(t + c_i \mathrm{d}t), t + c_i \mathrm{d}t\bigg),
\end{equation}
with $n_s$ the number of stages. Notice that if $a_{i,j} \neq 0$ for $j \geq i$ the method is implicit, because stage $i$ will depends on itself, and/or following stages that are yet to be computed. The evaluation of stage $i$ requires knowing the control values at a fraction of the time step, depending on the value of $c_i$. It is often practical to assume that $\bm{u}(t + c_i \mathrm{d}t) = \bm{u}(t)$, independently from $c_i$. In this case, we write ${}^i\bm{s}\left(\bm{x}(t), \bm{u}(t), t\right)$. Finally, $\mathbf{b}$ contains the coefficients used to sum up all the stages, retrieving the final solution:
\begin{equation}\label{RKconstraint}
\bm{x}(t + \mathrm{d}t) = \bm{x}(t) + \mathrm{d}t\sum_{i=1}^{n_s} b_i {}^i\bm{s},
\end{equation}
where we momentarily dropped part of the notation for simplicity. We can rewrite this equation in a more compact form:
\begin{equation}\label{RKconstraintCompact}
\bm{x}(t + \mathrm{d}t) = \bm{x}(t) + \mathrm{d}t \, \textbf{S}\left(\bm{x}(t), \bm{u}(\cdot), t\right) \bm{b}, \quad \textbf{S} = 
\begin{bmatrix}
{}^1\bm{s} & {}^2\bm{s} & \cdots & {}^{n_s}\bm{s}
\end{bmatrix}.
\end{equation}

The methods described by Eq.s \eqref{RKconstraintCompact} belongs to the Runge-Kutta integrator family. A famous example is the 4th order Runge-Kutta method \citep[Sec. 235]{butcher2016numerical} whose Butcher tableau is the following:
\begin{equation*}
	\begin{array}{c| c c c c}
 	 0  &  0  &  0  & 0 & 0 \\
	1/2 & 1/2 &  0  & 0 & 0 \\
	1/2 &  0  & 1/2 & 0 & 0 \\
	 1  &  0  &  0  & 1 & 0 \\
	\hline
	& 1/6 & 1/3 & 1/3 & 1/6
\end{array},
\end{equation*}
which is explicit. In case of explicit Runge-Kutta methods, it can be proven that the number of stages is greater or equal than the order of the method \citep[Sec. 324]{butcher2016numerical}. In particular, explicits methods for which the order is equal to number of stages have been found up to order 4. On the other hand, the method with the least amount of stages having order 8, is composed by 11 stages.

To conclude, Eq. \eqref{RKconstraintCompact} can be used as a constraint in place of Eq. \eqref{eq:dynamics_ocp}. While it may be tempting to use methods with a high number of stages to exploit their accuracy, they require composing function $\bm{f}$ multiple times. Hence, in the case of 
non-linear dynamics, this would make the optimal control problem difficult to solve. Conversely, the advantages of explicit methods are limited to the solution of \emph{initial value problems} (IVP). An optimal control problem resembles more a \emph{boundary value problem} (BVP) because it finds a path to a target state. In this case, the distinction between implicit and explicit methods becomes less important \citep{ascher1994numerical}. 

Eq. \eqref{RKconstraintCompact} can be used recursively to determine the state evolution from $t_0$ to $T$. This method is called \emph{shooting}.

\subsection{Shooting}\label{sec:shooting}
\subsubsection{Single shooting}
Let us consider a (LTI) system like the one in Eq. \eqref{eq:lti}.
If we discretize it with the forward Euler method, we obtain the following discrete system
\begin{equation}\label{eq:discrete_example}
	\bm{x}(t + \mathrm{d}t) = \bm{x}(t) + \mathrm{d}t\bm{A}\bm{x}(t) + \mathrm{d}t\bm{B}\bm{u}(t).
\end{equation}
By assuming $\mathrm{d}t = \frac{T-t_0}{N}$, with $N$ being the number of integration steps, it is possible to obtain $\bm{x}(t)$ as a function of the initial state and the control inputs only. In fact,
\begin{IEEEeqnarray*}{RCL}
	\bm{x}(t_0 + \mathrm{d}t) &=& \bm{x}_0 + \mathrm{d}t\bm{A}\bm{x}_0 + \mathrm{d}t\bm{B}\bm{u}(t_0), \\
	\bm{x}(t_0 + 2\mathrm{d}t) &=& \bm{x}(t_0 + \mathrm{d}t) + \mathrm{d}t\bm{A}\bm{x}(t_0+ \mathrm{d}t) + \mathrm{d}t\bm{B}\bm{u}(t_0+ \mathrm{d}t),\\
	&\vdots&\\
	\bm{x}(T) &=& \bm{x}(T - \mathrm{d}t) + \mathrm{d}t\bm{A}\bm{x}(T - \mathrm{d}t) + \mathrm{d}t\bm{B}\bm{u}(T - \mathrm{d}t),
\end{IEEEeqnarray*}
hence it is possible to remove all the intermediate state variables obtaining:
\begin{equation}\label{eq:single_shooting_LTI}
	\bm{x}(T) = \left(\mathds{1}_N + \mathrm{d}t\bm{A}\right)^N\bm{x}_0 + \bm{\mathds{C}} \begin{bmatrix}
	\bm{u}(t_0) \\ \vdots \\ \bm{u}(T - 2\mathrm{d}t) \\ \bm{u}(T - \mathrm{d}t)
	\end{bmatrix},	
\end{equation}
where $\bm{\mathds{C}} \in \mathbb{R}^{n \times Nm}$ is the controllability matrix
\begin{equation}
	\bm{\mathds{C}} = \begin{bmatrix}
	\left(\mathds{1}_N + \mathrm{d}t\bm{A}\right)^{N-1}\mathrm{d}t\bm{B} & \cdots & \left(\mathds{1}_N + \mathrm{d}t\bm{A}\right)\mathrm{d}t\bm{B} & \mathrm{d}t\bm{B}
	\end{bmatrix}.
\end{equation}
Henceforth, with a \emph{single shooting} method it is possible to obtain the terminal state starting from the initial one, without having to consider all the intermediate states. In an optimization framework, Eq. \eqref{eq:single_shooting_LTI} will correspond to the constraint
\begin{equation}
	{}_d\bm{h} \coloneqq \left(\mathds{1}_N + \mathrm{d}t\bm{A}\right)^N\bm{x}_0 + \bm{\mathds{C}} \begin{bmatrix}
	\bm{u}(t_0) \\ \vdots \\ \bm{u}(T - 2\mathrm{d}t) \\ \bm{u}(T - \mathrm{d}t)
	\end{bmatrix} - \bm{x}(T) = 0.
\end{equation}
The set of optimization variables $\bm{\chi}$ would consist of the control inputs applied at each step. As a consequence, the partial derivative of ${}_d\bm{h}$ with respect to $\bm{\chi}$, a.k.a. the constraint \emph{Jacobian}, corresponds to $\bm{\mathds{C}}$, which is a \emph{dense} matrix.
 
The applicability of this method is not limited to LTI systems, but it can be applied also to generic non-linear systems, using any explicit integration method, by applying Eq. \eqref{RKconstraintCompact} recursively.

\subsubsection{Multiple shooting}
If the system is nonlinear, the composition of function $\bm{f}$ $N$ times (as it has been done in Eq. \eqref{eq:single_shooting_LTI}) may result in a very complex expression. In addition, it is difficult to specify \emph{path} constraints, i.e. those involving intermediate states. These problems, typical of the method just introduced, are solved using a \emph{multiple shooting} approach \citep{Bock84,diehl2006fast}, where all the intermediate state variables are also optimization variables. It corresponds to fractioning the prediction horizon in multiple segments in which we apply an integration method. This has the clear disadvantage of including much more variables and constraints in the optimization problem. In fact, it is necessary to add a constraint similar to Eq. \eqref{RKconstraintCompact} for each pair of optimization variables. The dynamical constraints are the following,
\begin{equation} \label{eq:multiple_shooting}
	{}_d\bm{h} \coloneqq 	\left\{\begin{IEEEeqnarraybox}[][c]{RCL}
	0 &=& \bm{x}_0 + \mathrm{d}t \, \textbf{S}\left(\bm{x}_0, \bm{u}(t_0), t_0\right) \bm{b} - \bm{x}(t_0 + \mathrm{d}t),\\
	0 &=& \bm{x}(t_0+ \mathrm{d}t) + \mathrm{d}t \, \textbf{S}\left(\bm{x}(t_0+ \mathrm{d}t), \bm{u}(t_0+ \mathrm{d}t), t_0 + \mathrm{d}t\right) \bm{b}  + \\
	&&- \bm{x}(t_0 + 2\mathrm{d}t),\\
	&\vdots& \\
	0 &=& \bm{x}(T - \mathrm{d}t) + \mathrm{d}t \, \textbf{S}\left(\bm{x}(T - \mathrm{d}t), \bm{u}(T - \mathrm{d}t), T - \mathrm{d}t\right) \bm{b} + \\
	&&- \bm{x}(T),
	\end{IEEEeqnarraybox}
	\right.
\end{equation}
where each constraint depends only on the optimization variables related to the corresponding collocation points. For example, the second constraint in Eq. \eqref{eq:multiple_shooting}, would depend on $\bm{x}(t_0+ \mathrm{d}t)$, $\bm{u}(t_0+ \mathrm{d}t)$ and $\bm{x}(t_0 + 2\mathrm{d}t)$ only, while the next one on $\bm{x}(t_0+ 2\mathrm{d}t)$, $\bm{u}(t_0+ 2\mathrm{d}t)$ and $\bm{x}(t_0 + 3\mathrm{d}t)$. By defining the optimization variable $\bm{\chi}$ as
\begin{equation}\label{eq:variables_multiple_shooting}
	\bm{\chi} = \begin{bmatrix}
	\bm{u}(t_0) \\
	\bm{x}(t_0 + \mathrm{d}t) \\
	\bm{u}(t_0 + \mathrm{d}t) \\
	\bm{x}(t_0 + 2\mathrm{d}t) \\
	\vdots\\
	\bm{u}(T - \mathrm{d}t) \\
	\bm{x}(T) \\
	\end{bmatrix},
\end{equation}
the constraint Jacobian would assume a block-diagonal structure which can be exploited by modern solvers like in \citep{IPOpt2006}.

Since the system dynamics is discretized, so it is the cost function. In particular, Eq. \eqref{eq:cost} would become:
\begin{equation}
	\Gamma(\bm{\chi}) = \text{m}\left(\bm{x}(T)\right) + \sum_i^{N -1} \ell \left(\bm{x}(t_0 + i\mathrm{d}t), \bm{u}(t_0 + i\mathrm{d}t)\right) \mathrm{d}t.
\end{equation}

Similarly, if algebraic constraints are present, it is sufficient to ``repeat'' them across the time horizon, constraining only a portion of $\bm{\chi}$ depending on the time instant. To summarize, starting from the optimal control problem of Eq. \eqref{eq:ocp}, we obtain the following optimization problem:
\begin{IEEEeqnarray*}{CL}
	\minimize_{\bm{\chi}} & \text{m}\left(\bm{x}(T)\right) + \sum_i^{N -1} \ell \left(\bm{x}(t_0 + i\mathrm{d}t), \bm{u}(t_0 + i\mathrm{d}t)\right) \mathrm{d}t \\
	\text{subject to: } & \nonumber \\
	&	{}_g\mathbf{l} \leq \bm{g}\left(\bm{x}_0, \bm{u}(t_0), t_0\right) \leq {}_g\mathbf{u}, \\
	&0 = \bm{x}_0 + \mathrm{d}t \, \textbf{S}\left(\bm{x}_0, \bm{u}(t_0), t_0\right) \bm{b} - \bm{x}(t_0 + \mathrm{d}t),\\
	&	{}_g\mathbf{l} \leq \bm{g}\left(\bm{x}(t_0 + \mathrm{d}t), \bm{u}(t_0 + \mathrm{d}t), t_0 + \mathrm{d}t\right) \leq {}_g\mathbf{u}, \\
	&0 = \bm{x}(t_0+ \mathrm{d}t) + \mathrm{d}t \, \textbf{S}\left(\bm{x}(t_0+ \mathrm{d}t), \bm{u}(t_0+ \mathrm{d}t), t_0 + \mathrm{d}t\right) \bm{b}  + \\
	&\hphantom{0 = }- \bm{x}(t_0 + 2\mathrm{d}t),\\
	&\vdots \\
	&	{}_g\mathbf{l} \leq \bm{g}\left(\bm{x}(T - \mathrm{d}t), \bm{u}(T - \mathrm{d}t), T - \mathrm{d}t\right) \leq {}_g\mathbf{u}, \\
	&0 = \bm{x}(T - \mathrm{d}t) + \mathrm{d}t \, \textbf{S}\left(\bm{x}(T - \mathrm{d}t), \bm{u}(T - \mathrm{d}t), T - \mathrm{d}t\right) \bm{b} + \\
	&\hphantom{0 = }- \bm{x}(T),
\end{IEEEeqnarray*}
where the optimization variables $\bm{\chi}$ are defined as in Eq. \eqref{eq:variables_multiple_shooting}. This optimization problem can be solved with an appropriate solver.
\section{Receding horizon principle}\label{sec:receding_horizon}
When solving the optimal control problem described by Eq. \eqref{eq:ocp} using an indirect method, as presented in Sec. \ref{sec:indirect}, we obtain a control law which is function of the state. For example, a linear-quadratic regulator can be seen as a linear state-feedback control law with an optimal gain, as shown in Sec. \ref{sec:lqr}. On the contrary, when using a direct method, like those of Sec. \ref{sec:direct}, the output is not a control \emph{law}, but a series of control \emph{values}. In fact, given $\bm{x}_0$, we obtain $\bm{u}(t_0 + i\mathrm{d}t)$, with $i$ from $0$ to $N-1$. The application of all these control inputs would result in \emph{open-loop} control. Alternatively, we can apply $\bm{u}(t_0)$ only, discarding all the other control inputs. The system evolves and a new feedback $\bm{x}_0^\prime$ is retrieved. The time horizon is shifted of an amount equal to $\mathrm{d}t$ and the optimal control problem is solved again with a different initial condition. This method of discarding the control values except the first, and shifting the time horizon, is called \emph{receding horizon} principle \citep{Mayne90MPC, clarke1991constrained}. It provides the basis for the so-called \emph{model predictive control} (MPC) \citep{garcia1989model}. The stability of MPC can be demonstrated also in the constrained case \citep{zheng1995stability,Mayne2000Stability}.
\section{Basics of non-linear optimization}\label{sec:nlp}
Consider the following optimization problem:
\begin{IEEEeqnarray}{CL}
	\phantomsection \IEEEyesnumber\label{eq:optimization_separated}
	\minimize_{\bm{\chi}} & \Gamma(\bm{\chi}) \IEEEyessubnumber \\
	\text{subject to: } & \nonumber \\
	&	{}_e\bm{h}\left(\bm{\chi}\right) = 0, \IEEEyessubnumber \label{eq:equality} \\
	&	{}_i\bm{h}\left(\bm{\chi}\right) \leq 0. \IEEEyessubnumber \label{eq:inequality}
\end{IEEEeqnarray}
Compared to Eq. \eqref{eq:optimization}, we separated \emph{equality} constraints, ${}_e\bm{h}: \mathbb{R}^{n_\chi} \rightarrow \mathbb{R}^{n_e}$, from \emph{inequality} constraints, ${}_i\bm{h}: \mathbb{R}^{n_\chi} \rightarrow \mathbb{R}^{n_i}$, assuming to have only upper-bounds, without loss of generality. A point $\bm{\chi}$ is said to be \emph{feasible} if both the conditions \eqref{eq:equality}$-$\eqref{eq:inequality} are satisfied. A feasible point $\bm{\chi}$ is \emph{locally optimal}, if there exist a scalar $R > 0$ such that:
\begin{equation}\label{eq:local_optimality}
	\Gamma(\bm{\chi}) = \inf\{\Gamma(\bm{z})\, |\, {}_e\bm{h}\left(\bm{z}\right) = 0, \,  {}_i\bm{h}\left(\bm{z}\right) \leq 0, \, \|\bm{z} - \bm{\chi}\| < R \}.
\end{equation} 
In other words, $\bm{\chi}$ is locally optimal if all the feasible points $\bm{z}$ in a hypersphere of radius $R$ around it, provide a higher cost. If an inequality constraint $k \leq n_i$ is equal to 0, i.e. ${}_i\bm{h}\left(\bm{\chi}\right)_k = 0$, it is said to be \emph{active}.

\subsection{Convex optimization problems}
Let us start with few definitions.
A set $C$ is \emph{convex} if the line connecting any two points in $C$ is fully contained in $C$, i.e.
\begin{equation}
	C \text{ is convex } \iff \bm{x}, \bm{y} \in C : \theta\bm{x} + (1- \theta)\bm{y} \in C, \quad \forall \,\theta \in [0, 1].
\end{equation}
A generic function $\bm{f}_\text{cv}$ is \emph{convex} if its domain is a convex set and, for all $\bm{x}$ and $\bm{y}$ belonging to its domain, we have:
\begin{equation}
	\bm{f}_\text{cv}\left(\theta\bm{x} + (1- \theta)\bm{y}\right) \leq \theta\bm{f}_\text{cv}(\bm{x}) + (1- \theta)\bm{f}_\text{cv}(\bm{y}) \quad \forall \,\theta \in [0, 1].
\end{equation} 
Intuitively, this means that if we take two points on the surface representing $\bm{f}_\text{cv}$, the line connecting them is ``above'' the surface. As an example, quadratic functions like $\bm{\chi}^\top \bm{H} \bm{\chi}$, with $\bm{H}$ positive definite, are convex.
If in Eq. \eqref{eq:optimization_separated} the cost function and the inequality constraints are convex functions, while the equalities are affine, then the optimization problem is said to be \emph{convex}:
\begin{IEEEeqnarray}{CL}
	\phantomsection \IEEEyesnumber\label{eq:optimization_convex}
	\minimize_{\bm{\chi}} & \Gamma_\text{cv}(\bm{\chi}) \IEEEyessubnumber \\
	\text{subject to: } & \nonumber \\
	&	\bm{A}_e \bm{\chi}  = {}_e\bm{b}, \IEEEyessubnumber \\
	&	{}_i\bm{h}_\text{cv}\left(\bm{\chi}\right) \leq 0. \IEEEyessubnumber 
\end{IEEEeqnarray}
Notice that neither $\bm{A}_e \in \mathbb{R}^{n_e \times n}$, nor ${}_e\bm{b} \in \mathbb{R}^{n_e}$ depend on $\bm{\chi}$. For this particular problem, local optima are also global, i.e. condition \eqref{eq:local_optimality} holds for any $R$. The proof of this statement can be found in \citep[Sec. 4.2.2]{boyd2004convex}. \emph{Quadratic programming} (QP) problems, constituted by a quadratic cost function and affine constraints, are an example of convex optimization problems. 

\subsection{The Lagrangian}

Given the optimization problem defined in Eq. \eqref{eq:optimization_separated}, we define the \emph{Lagrangian} $\mathcal{L}: \mathbb{R}^{n_\chi}\times\mathbb{R}^{n_e}\times\mathbb{R}^{n_i} \rightarrow \mathbb{R}$ as:
\begin{equation}\label{eq:lagrangian}
	\mathcal{L}\left(\bm{\chi}, {}_e\bm{\lambda}, {}_i\bm{\lambda}\right) = \Gamma(\bm{\chi}) + {}_e\bm{\lambda}^\top {}_e\bm{h}(\bm{\chi})	+ {}_i\bm{\lambda}^\top {}_i\bm{h}(\bm{\chi}),	
\end{equation}
where ${}_e\bm{\lambda} \in \mathbb{R}^{n_e}$ and ${}_i\bm{\lambda} \in \mathbb{R}^{n_i}$ are called \emph{Lagrangian multipliers}. 

The \emph{dual function} $\mathpzc{g}: \mathbb{R}^{n_i}\times \mathbb{R}^{n_e} \rightarrow \mathbb{R}$ is obtained by taking the minimum of Eq. \eqref{eq:lagrangian} over $\bm{\chi}$:
\begin{equation}\label{eq:dual}
	\mathpzc{g}\left({}_e\bm{\lambda}, {}_i\bm{\lambda}\right) = \inf_{\bm{\chi}} \mathcal{L}\left(\bm{\chi}, {}_e\bm{\lambda}, {}_i\bm{\lambda}\right).
\end{equation}

Interestingly, the dual function provides a lower-bound to the optimal cost value $\Gamma^\star = \Gamma(\bm{\chi}^\star)$, with $\bm{\chi}^\star$ the optimal point of \eqref{eq:optimization_separated}, as we show in the following. 
For any ${}_e\bm{\lambda} \in \mathbb{R}^{n_e}$, we have that 
\begin{equation*}
{}_e\bm{\lambda}^\top {}_e\bm{h}(\bm{\chi}^\star) = 0
\end{equation*}
since $\bm{\chi}^\star$ is feasible. For the same reason, for  any ${}_i\bm{\lambda} \geq 0$, we have
\begin{equation*}
{}_i\bm{\lambda}^\top {}_i\bm{h}(\bm{\chi}^\star) \leq 0,
\end{equation*}
Hence, given the Lagrangian definition in Eq. \eqref{eq:lagrangian}, the following holds:
\begin{equation}
	\mathcal{L}\left(\bm{\chi}^\star, {}_e\bm{\lambda}, {}_i\bm{\lambda}\right) \leq \Gamma(\bm{\bm{\chi}^\star}).
\end{equation}
In addition, $\mathpzc{g}(\cdot)$ is the minimum of $\mathcal{L}\left(\cdot\right)$ over $\bm{\chi}$. We finally have
\begin{equation}
	\mathpzc{g}\left({}_e\bm{\lambda}, {}_i\bm{\lambda}\right) \leq \Gamma(\bm{\bm{\chi}^\star}), \quad \forall {}_e\bm{\lambda} \in \mathbb{R}^{n_e}\text{, and } {}_i\bm{\lambda} \geq 0.
\end{equation}

Intuitively, the most meaningful lower-bound is obtained by maximizing $\mathpzc{g}(\cdot)$ over ${}_e\bm{\lambda}$ and ${}_i\bm{\lambda}$, because it provides the highest lower-bound for the optimal cost. It can be obtained by solving the following optimization problem, called the \emph{dual problem}:
\begin{IEEEeqnarray}{CL}
	\phantomsection \IEEEyesnumber\label{eq:dual_optimization}
	\maximize_{{}_e\bm{\lambda}, {}_i\bm{\lambda}} & \mathpzc{g}\left({}_e\bm{\lambda}, {}_i\bm{\lambda}\right) \IEEEyessubnumber \\
	\text{subject to: } & \nonumber \\
	&	{}_i\bm{\lambda} \geq 0. \IEEEyessubnumber 
\end{IEEEeqnarray}

In case $\mathpzc{g}\left({}_e\bm{\lambda}^*, {}_i\bm{\lambda}^*\right) = \Gamma(\bm{\bm{\chi}^\star})$ holds, with ${}_e\bm{\lambda}^*$ and ${}_i\bm{\lambda}^*$ the optimal Lagrange multipliers, we have \emph{strong duality}. It is usually a property of convex optimization problems \citep[Sec. 5.3.2]{boyd2004convex}, but it applies also in the non-convex case under some conditions called \emph{constraint qualifications} \citep{giorgi2018guided}. One of these is the \emph{Linear Independence Constraint Qualification} (LICQ). It requires the gradient of all active constraints (equalities included) to be linear independent at the optimum. If we define ${}_{ai} \bm{h}$ as the set of active inequalities $\left(\text{i.e. }{}_{ai}\bm{h}(\bm{\chi}^\star) = 0\right)$, the matrix
\begin{equation}
\begin{bmatrix}
\nabla_{\bm{\chi}}\,{}_e\bm{h}(\bm{\chi}^\star) \\
\nabla_{\bm{\chi}}\,{}_{ai}\bm{h}(\bm{\chi}^\star)
\end{bmatrix}
\end{equation}
has to be full row rank.

\subsection{Necessary conditions for optimality}
Let us consider the following assumptions.
\begin{enumerate}	
	\item Functions $\Gamma(\cdot)$, ${}_e\bm{h}(\cdot)$ and ${}_i\bm{h}(\cdot)$ are continuously differentiable in $\bm{\chi}^\star$.
	\item $\bm{\chi}^\star$ is a locally optimal point, while ${}_e\bm{\lambda}^*$ and ${}_i\bm{\lambda}^*$ solve problem \eqref{eq:dual_optimization}.
	\item Strong duality holds. 
\end{enumerate} 
Then, for a point $\bm{\chi}^\star$ to be optimal, the following \emph{necessary} conditions have to be satisfied \citep{kuhn1951}:
\begin{IEEEeqnarray}{RCL}
	\IEEEyesnumber \phantomsection \label{eq:kkt}
	{}_e\bm{h}\left(\bm{\chi}^\star\right) &=& 0 \IEEEyessubnumber  \label{eq:kkt1}\\
    {}_i\bm{h}\left(\bm{\chi}^\star\right) &\leq& 0 \IEEEyessubnumber \label{eq:kkt2}\\
	{}_i\bm{\lambda}^\star &\geq& 0 \IEEEyessubnumber \label{eq:kkt3}\\
	{}_i\bm{\lambda}^{\star\top} {}_i\bm{h}(\bm{\chi}^\star) &=& 0 \IEEEyessubnumber \label{eq:kkt4}\\
	\nabla_{\bm{\chi}}\Gamma(\bm{\chi}^\star) + {}_e\bm{\lambda}^{\star\top} \nabla_{\bm{\chi}}\,{}_e\bm{h}(\bm{\chi}^\star) + {}_i\bm{\lambda}^{\star\top} \nabla_{\bm{\chi}}\,{}_i\bm{h}(\bm{\chi}^\star) &=& 0. \IEEEyessubnumber\label{eq:kkt5}
\end{IEEEeqnarray}

Eq. \eqref{eq:kkt} defines the \emph{Karush-Kuhn-Tucker} conditions (KKT). Eq.s \eqref{eq:kkt1}{-}\eqref{eq:kkt2} impose $\bm{\chi}^\star$ to be feasible. Similarly, Eq. \eqref{eq:kkt3} is the dual feasibility condition. Condition \eqref{eq:kkt4} is the so-called \emph{complementarity slackness}. Notice that, because of Eq. \eqref{eq:kkt2}-\eqref{eq:kkt3}, the complementarity slackness imposes every product ${}_i\lambda_k^\star \cdot {}_i h_k$ to be equal to zero. In other words, if an inequality is not active, i.e. ${}_i h_k < 0$, the corresponding Lagrange multiplier ${}_i\lambda_k^\star$ has to be null. Conversely, if the inequality is active, the multiplier can be different from zero. Finally, condition \eqref{eq:kkt5} ensures that the optimal point is also a stationary point of the Lagrangian. 

If the optimization problem is convex, KKT conditions are also sufficient \citep[Sec. 5.5.3]{boyd2004convex}. Otherwise, \emph{second order optimality conditions} \citep[Sec. 3.3]{Diehl2018} have to be employed, involving the \emph{Hessian} of the Lagrangian, i.e. its double derivative with respect to the optimization variables. This means that if a point $\bm{\chi}$ satisfies Eq. \eqref{eq:kkt} (plus other second order conditions in case of non-convex problems), then $\bm{\chi}$ is optimal. As a consequence, a classical method for finding the optimal solution consists in applying root-finding techniques, like the \emph{Newton method} \citep[Sec. 1.5]{betts2010practical}, to Eq. \eqref{eq:kkt5}. The complication here is represented by inequalities, because of constraints \eqref{eq:kkt3}$-$\eqref{eq:kkt4}. A possibility then, is to assume to know which inequality is active and consider it as an equality. In other words we guess the \emph{active set}. If after some iteration one of the constraints \eqref{eq:kkt2}$-$\eqref{eq:kkt3} is no more satisfied, we change our guess on the set of active constraints \citep[Sec. 1.9]{betts2010practical}.  The solver qpOASES~\citep{Ferreau2014} uses this technique to solve QP problems. Interestingly, if the guess on the active set is correct, the global optimum of QP problems can be found in a single step \citep[Sec. 1.10]{betts2010practical}. QP solvers can also be used to solve generic non-linear optimization problems by resorting to \emph{sequential} approximations of the problem. This method is called SQP (sequential quadratic programming) \citep[Sec. 1.13]{betts2010practical} and it is adopted in the ESA (European Space Agency) solver WORHP \citep{buskens2012esa}.

Alternatively, the \emph{interior point} methods exploit logarithmic \emph{barrier functions} to deal with inequalities \citep[Sec. 4.3]{Diehl2018}. In particular, they solve the following optimization problem:
\begin{IEEEeqnarray}{CL}
	\IEEEyesnumber \phantomsection
	\minimize_{\bm{\chi}} & \Gamma(\bm{\chi}) - \mu \sum_{k}^{n_i}\ln\left(-{}_ih_k(\bm{\chi})\right) \IEEEyessubnumber \\
	\text{subject to: } & {}_e\bm{h} = 0. \IEEEyessubnumber
\end{IEEEeqnarray}
When an inequality tends to zero, the cost tends to infinity hence implicitly preventing it to be violated. The multiplier $\mu$ is then lowered in an iterative manner to reduce the approximation introduced by the barrier function. An example of interior point solver is \texttt{Ipopt} \citep{IPOpt2006}, widely used in this thesis to solve generic non-linear optimization problems.

\chapter{Modeling of Floating-Base Robots}\label{chap:robot_model}
While Chap. \ref{chap:oc} considers a generic dynamic model, in this chapter we focus on the modeling of floating-base robots. We start defining rotation and transformation matrices in Sec. \ref{sec:modelling_basics}, while the velocity of a rigid body is discussed in Sec. \ref{sec:trivializatons}. We then extend these considerations in the multi-body case in Sec. \ref{sec:multibody_kin}.  Finally Sec. \ref{sec:modelling} discusses multi-body dynamics and Sec. \ref{sec:intro_momentum} presents a fundamental modeling tool widely used in the thesis, i.e the \emph{centroidal dynamics}. Sec. \ref{sec:simplified} presents some of the simplifications for the centroidal dynamics which are popular in the literature.
\section{Rotation and transformation matrices}\label{sec:modelling_basics}
Let us consider two frames in $\mathbb{R}^3$ called C and D, with a coincident origin. Here, and in the rest of the thesis, we assume the corresponding bases to be orthonormal. 
Define ${}^F\bm{d}_i \in \mathbb{R}^3$ as the $i$-th base of frame $D$ expressed in another frame $F$ (equivalently ${}^F\bm{c}_i$ for frame $C$). Trivially we have that ${}^D\bm{d}_i = \bm{e}_i$. At the same time, given a point $\bm{p}$ whose coordinates are defined in frame $D$, namely ${}^D\bm{p}$, we obtain the corresponding coordinates in frame $C$ as follows:
\begin{equation}
	{}^C\bm{p} = \begin{bmatrix}
	{}^C\bm{d}_1 & {}^C\bm{d}_2 & {}^C\bm{d}_3
	\end{bmatrix} {}^D\bm{p} = {}^C\bm{R}_D {}^D\bm{p}.
\end{equation}
${}^C\bm{R}_D$ is the rotation matrix and it belongs to $\SO(3)$, i.e. the set of $\mathbb{R}^{3\times 3}$ orthogonal matrices with determinant equal to 1.

Let us consider now another frame $F$ having a different origin. Define as ${}^C \bm{o}_F$ the position of the frame $F$ origin measured in frame $C$. Then, we can transform the coordinates of a point ${}^F\bm{q}$ into $C$ as follows:
\begin{equation}\label{eq:transformation_initial}
	{}^C\bm{q} = {}^C\bm{R}_F {}^F\bm{q} + {}^C \bm{o}_F,
\end{equation}
or in a more compact form
\begin{equation}
	\begin{bmatrix}
	{}^C\bm{q} \\
	1
	\end{bmatrix} = \begin{bmatrix}
	{}^C\bm{R}_F & {}^C \bm{o}_F \\
	\bm{0}_{1 \times 3} & 1
	\end{bmatrix} \begin{bmatrix}
	{}^F\bm{q} \\
	1
	\end{bmatrix}.
\end{equation}
In the following, with a slight abuse of notation, we simply write
\begin{equation}
	{}^C\bm{q} = {}^C\bm{H}_F {}^F\bm{q},
\end{equation}
where the transformation matrix ${}^C\bm{H}_F$ belongs to the set of $\mathbb{R}^{4\times 4}$ called $\SE(3)$. If we assume these frames to be rigidly attached to rigid bodies, transformation matrices can be used to retrieve their relative pose.

The transformation matrix ${}^C\bm{H}_F$, also called \emph{homogeneous transformation}, can be easily inverted starting from Eq. \eqref{eq:transformation_initial}:
\begin{equation}
	{}^C\bm{H}_F^{-1} = \begin{bmatrix}
	{}^C\bm{R}_F^{\top} & -{}^C\bm{R}_F^{\top}{}^C \bm{o}_F \\
	\bm{0}_{1 \times 3} & 1
	\end{bmatrix},
\end{equation}
where we exploited the orthogonality properties of the rotation matrix. 

\section{Rigid body velocity}\label{sec:trivializatons}

It is possible to define the 6D velocity of a rigid body in multiple ways. Throughout the thesis we make use of different representations depending on the specific case. These \emph{trivializations}, as they are also called, arise from the fact that the relative velocity between two frames can be \emph{represented} with different coordinate vectors depending on the frame used to measure it.

Defining $C$ and $D$ as two frames moving with respect to each other, we define the time derivative of the relative transformation as:
\begin{equation}
	{}^C\dot{\bm{H}}_D \coloneqq \frac{\dif}{\dif t}\left({}^C{\bm{H}}_D\right) = \begin{bmatrix}
	{}^C\dot{\bm{R}}_D & {}^C\dot{\bm{o}}_D \\
	\bm{0}_{1\times 3} & 0
	\end{bmatrix}.
\end{equation}
A more compact representation can be obtained by multiplying it on the \emph{left} with the inverse of the transformation matrix:
\begin{equation}\label{eq:left_trivialization_adjoint}
\begin{split}
	{}^C{\bm{H}}^{-1}_D{}^C\dot{\bm{H}}_D &= \begin{bmatrix}
	{}^C{\bm{R}}^\top_D & -{}^C{\bm{R}}^\top_D{}^C{\bm{o}}_D \\
	\bm{0}_{1\times 3} & 1
	\end{bmatrix}\begin{bmatrix}
	{}^C\dot{\bm{R}}_D & {}^C\dot{\bm{o}}_D \\
	\bm{0}_{1\times 3} & 0
	\end{bmatrix} \\
	&= \begin{bmatrix}
	{}^C{\bm{R}}^\top_D {}^C\dot{\bm{R}}_D & {}^C{\bm{R}}^\top_D{}^C\dot{\bm{o}}_D \\
	\bm{0}_{1\times 3} & 0
	\end{bmatrix}.
\end{split}
\end{equation}
Note that ${}^C{\bm{R}}^\top_D {}^C\dot{\bm{R}}_D$ is skew-symmetric. In fact, starting from the equality
\begin{equation}\label{eq:rotation_equality}
	{}^C{\bm{R}}^\top_D {}^C{\bm{R}}_D = \mathds{1}_3,
\end{equation}
by differentiation we have
\begin{equation}
{}^C\dot{\bm{R}}^\top_D {}^C{\bm{R}}_D  + {}^C{\bm{R}}^\top_D {}^C\dot{\bm{R}}_D= \bm{0}_{3\times 3}.
\end{equation}
Hence, we have
\begin{equation}
{}^C\dot{\bm{R}}^\top_D {}^C{\bm{R}}_D  = - \left({}^C\dot{\bm{R}}^\top_D {}^C{\bm{R}}_D\right)^\top,
\end{equation}
which is the definition of a skew-symmetric matrix.

The information contained in Eq. \eqref{eq:left_trivialization_adjoint} can be summarized by two $3D$ vectors, the linear and angular velocity, as follows
\begin{IEEEeqnarray}{RCL}
	\IEEEyesnumber \phantomsection
	{}^D\bm{v}_{C, D} &=&  {}^C{\bm{R}}^\top_D{}^C\dot{\bm{o}}_D \IEEEyessubnumber \\
	{}^D\bm{\omega}_{C, D} &=& \left({}^C{\bm{R}}^\top_D {}^C\dot{\bm{R}}_D\right)^\vee \IEEEyessubnumber
\end{IEEEeqnarray}
composing the \emph{left-trivialized} 6D velocity:
\begin{equation}
	{}^D\bm{V}_{C, D} = \begin{bmatrix}
	{}^D\bm{v}_{C, D} \\
	{}^D\bm{\omega}_{C, D}
	\end{bmatrix}.
\end{equation}
It is also called \emph{body velocity}. In fact, assuming $D$ to be attached to a moving rigid body and $C$ an inertial frame, the superscript indicates that the velocity coordinates are expressed in the moving $D$ frame. With an abuse of notation, we can express with the operator $\vee$ (and $\wedge$ for its inverse) the ``extraction'' of a $6D$ velocity from the matrix defined in Eq. \eqref{eq:left_trivialization_adjoint}:
\begin{equation}
	{}^D\bm{V}_{C, D} = \begin{bmatrix}
	{}^C{\bm{R}}^\top_D {}^C\dot{\bm{R}}_D & {}^C{\bm{R}}^\top_D{}^C\dot{\bm{o}}_D \\
	\bm{0}_{1\times 3} & 0
	\end{bmatrix}^\vee. 
\end{equation}

Alternatively, we can multiply ${}^C\dot{\bm{H}}_D$ on the \emph{right} side by the inverse of the transformation matrix:
\begin{equation}
\begin{split}
{}^C\dot{\bm{H}}_D{}^C{\bm{H}}^{-1}_D &= \begin{bmatrix}
{}^C\dot{\bm{R}}_D & {}^C\dot{\bm{o}}_D \\
\bm{0}_{1\times 3} & 0
\end{bmatrix} \begin{bmatrix}
{}^C{\bm{R}}^\top_D & -{}^C{\bm{R}}^\top_D{}^C{\bm{o}}_D \\
\bm{0}_{1\times 3} & 1
\end{bmatrix}\\
&= \begin{bmatrix}
{}^C\dot{\bm{R}}_D{}^C{\bm{R}}^\top_D & {}^C\dot{\bm{o}}_D-{}^C\dot{\bm{R}}_D{}^C{\bm{R}}^\top_D{}^C{\bm{o}}_D \\
\bm{0}_{1\times 3} & 0
\end{bmatrix}.
\end{split}
\end{equation}
It is trivial to show that ${}^C\dot{\bm{R}}_D{}^C{\bm{R}}^\top_D$ is skew-symmetric, similarly to what has been done starting from Eq. \eqref{eq:rotation_equality}. Then, we can construct the \emph{right-trivialized} velocity
\begin{equation}
		{}^C\bm{V}_{C, D} = \begin{bmatrix}
		{}^C\bm{v}_{C, D} \\
		{}^C\bm{\omega}_{C, D}
		\end{bmatrix}.
\end{equation}
It is composed as follows:
\begin{IEEEeqnarray}{RCL}
	\IEEEyesnumber \phantomsection
	{}^C\bm{v}_{C, D} &=&  {}^C\dot{\bm{o}}_D-{}^C\dot{\bm{R}}_D{}^C{\bm{R}}^\top_D{}^C{\bm{o}}_D \IEEEyessubnumber \\
	{}^C\bm{\omega}_{C, D} &=& \left({}^C\dot{\bm{R}}_D{}^C{\bm{R}}^\top_D\right)^\vee. \IEEEyessubnumber
\end{IEEEeqnarray}
This is also called \emph{inertial} velocity since it is measured on the $C$ frame. Notice that the linear velocity can be alternatively written as ${}^C\bm{v}_{C, D} =  {}^C\dot{\bm{o}}_D+{}^C{\bm{o}}_D^\wedge{}^C\bm{\omega}_{C, D}$.

In some applications, it may be helpful to define the 6D velocity as
\begin{equation}\label{eq:vel_mixed_unnamed}
	\begin{bmatrix}
	{}^C\dot{\bm{o}}_D \\
	{}^C\bm{\omega}_{C, D}
	\end{bmatrix},
\end{equation}
thus having the linear velocity corresponding to the time derivative of the relative position of the two frames. This is particularly helpful when we use one of the integration techniques introduced in Sec. \ref{sec:integration_methods} to retrieve the corresponding position over time. Interestingly, the quantity in Eq. \eqref{eq:vel_mixed_unnamed} corresponds to the velocity expressed in a frame having the same origin of $D$, but oriented as $C$. Such a frame is indicated with the notation $D[C]$. It is called \emph{hybrid} or \emph{mixed} velocity representation  \citep{traversaro2017thesis}:
\begin{equation}
\begin{split}
	{}^{D[C]} \bm{V}_{C,D} = {}^{D[C]} \bm{X}_{D} {}^{D} \bm{V}_{C,D} &= \begin{bmatrix}
	{}^C\bm{R}_D & \bm{0}_{3\times 3} \\
	\bm{0}_{3\times 3} & {}^C\bm{R}_D
	\end{bmatrix} \begin{bmatrix}
	{}^C{\bm{R}}^\top_D{}^C\dot{\bm{o}}_D \\
	{}^D\bm{\omega}_{C, D}
	\end{bmatrix} \\
	&= 	\begin{bmatrix}
	{}^C\dot{\bm{o}}_D \\
	{}^C\bm{\omega}_{C, D}
	\end{bmatrix}.
\end{split}
\end{equation}
The equality ${}^C\bm{R}_D{}^D\bm{\omega}_{C, D} = {}^C\bm{\omega}_{C, D}$ is proven as follows:
\begin{IEEEeqnarray}{RCL}
	\IEEEyesnumber \phantomsection
	\left({}^C\bm{R}_D{}^D\bm{\omega}_{C, D}\right)^\wedge &=&  {}^C\bm{R}_D\left({}^D\bm{\omega}_{C, D}\right)^\wedge {}^C\bm{R}_D^\top \IEEEyessubnumber \label{eq:change_base_angular1}\\
	&=& {}^C\bm{R}_D\left(\left({}^C{\bm{R}}^\top_D {}^C\dot{\bm{R}}_D\right)^\vee\right)^\wedge {}^C\bm{R}_D^\top \IEEEyessubnumber \\
	&=& {}^C\bm{R}_D{}^C{\bm{R}}^\top_D {}^C\dot{\bm{R}}_D {}^C\bm{R}_D^\top \IEEEyessubnumber \\
	&=& {}^C\dot{\bm{R}}_D {}^C\bm{R}_D^\top \IEEEyessubnumber \\
	&=& \left({}^C\bm{\omega}_{C, D}\right)^\wedge \IEEEyessubnumber \\
	{}^C\bm{R}_D{}^D\bm{\omega}_{C, D} &=& {}^C\bm{\omega}_{C, D}, \IEEEyessubnumber
\end{IEEEeqnarray}
where in Eq. \eqref{eq:change_base_angular1} we used the \emph{rotational equivalence} of cross products. 

Similarly, the adjoint matrix ${}^C \bm{X}_D$ allows obtaining a right-trivialized velocity starting from the left-trivialization:
\begin{equation} \label{eq:from_left_to_right_trivialization}
	{}^C\bm{V}_{C, D} = {}^C\bm{X}_{D}{}^D\bm{V}_{C, D} = \begin{bmatrix}
	{}^C\bm{R}_D & {}^C{\bm{o}}_D^\wedge {}^C\bm{R}_D  \\[1pt]
	\bm{0}_{3\times 3} & {}^C\bm{R}_D
	\end{bmatrix}{}^D\bm{V}_{C, D}.
\end{equation}
This highlights that the relative velocity between the two frames, intended as a physical entity, is unique. Nevertheless, it assumes different \emph{coordinates} according to the frame in which it is expressed. Thus, without loss of generality, in the following we use mainly the \emph{left-trivialized} version.

In general, the adjoint matrix can be used to express a velocity in a different frame. For example, let us express the velocity ${}^D\bm{V}_{C, D}$ in another frame $F$ at a constant relative pose from $D$, ${}^F\bm{H}_D$. It can be shown that ${}^F\bm{V}_{C, F} = {}^F\bm{V}_{C, D}$ (there is no relative velocity between $D$ and $F$). Hence,
\begin{IEEEeqnarray}{RCL}
	\IEEEyesnumber \phantomsection \label{eq:adjoint_chang_coordinates_full}
	{}^F\bm{V}_{C, D} = {}^F\bm{V}_{C, F} &=& \left({}^C{\bm{H}}^{-1}_F \frac{\dif}{\dif t}\left({}^C{\bm{H}}_D{}^D{\bm{H}}_F\right)\right)^\vee \IEEEyessubnumber \\
	&=& \left({}^D{\bm{H}}_F^{-1} {}^C{\bm{H}}_D^{-1} {}^C\dot{\bm{H}}_D{}^D{\bm{H}}_F\right)^\vee \IEEEyessubnumber \label{eq:trivialization_inversion}\\
	&=& \left({}^D{\bm{H}}_F^{-1}{}^D\bm{V}_{C, D}^\wedge {}^D{\bm{H}}_F\right)^\vee \IEEEyessubnumber \label{eq:adjoint_change_preliminary}\\
	&=& {}^F\bm{X}_{D} {}^D\bm{V}_{C, D}. \IEEEyessubnumber \label{eq:adjoint_chang_coordinates}
\end{IEEEeqnarray}
The passage from Eq. \eqref{eq:adjoint_change_preliminary} to \eqref{eq:adjoint_chang_coordinates} is similar to the \emph{rotational equivalence} of cross products used previously, and it can be verified by hand calculations. 

Let us drop the assumption of constant relative pose for the frame $F$. Its velocity with respect to $C$ is the following:
\begin{IEEEeqnarray}{RCL}
	\IEEEyesnumber \phantomsection \label{eq:velocity_composition}
	{}^F\bm{V}_{C, F} &=& \left({}^C{\bm{H}}^{-1}_F \frac{\dif}{\dif t}\left({}^C{\bm{H}}_D{}^D{\bm{H}}_F\right)\right)^\vee \IEEEyessubnumber \\
	&=&\left({}^D{\bm{H}}_F^{-1} {}^C{\bm{H}}_D^{-1} {}^C\dot{\bm{H}}_D{}^D{\bm{H}}_F + {}^C{\bm{H}}^{-1}_F{}^C{\bm{H}}_D{}^D\dot{\bm{H}}_F\right)^\vee \IEEEyessubnumber \\
	&=& \left({}^D{\bm{H}}_F^{-1}{}^D\bm{V}_{C, D}^\wedge {}^D{\bm{H}}_F + {}^D{\bm{H}}^{-1}_F{}^D\dot{\bm{H}}_F\right)^\vee \IEEEyessubnumber \\
	&=& {}^F\bm{X}_{D} {}^D\bm{V}_{C, D} + {}^F\bm{V}_{D, F}, \IEEEyessubnumber
\end{IEEEeqnarray}
which provides an expression for the composition of velocities.
\section{Multi-body kinematics}\label{sec:multibody_kin}
\begin{figure}[tpb]
	\def\svgwidth{0.9\columnwidth}
	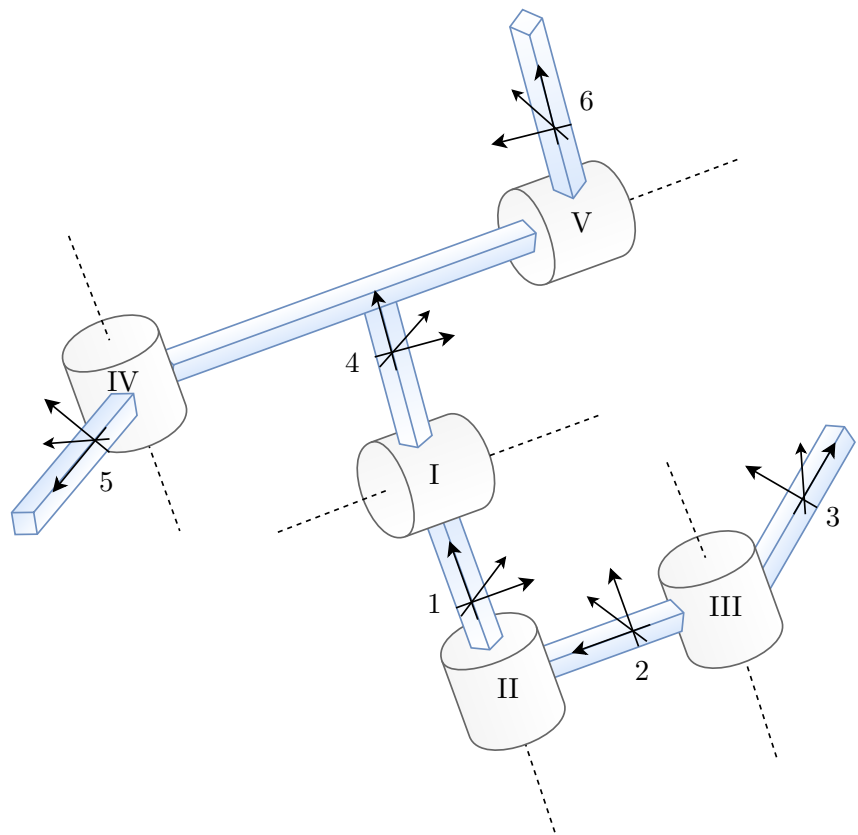
	\caption{Schematic representation of a multi-body structure. Links are in blue, while joints are in grey. These are assumed to be revolute with a single degree of freedom. The corresponding axis of rotation is indicated with a dotted line. Each joint is associated with a roman numeral, while each link has a frame attached, named with a number.}
	\label{fig:multibody}
\end{figure}
Let us analyze a robot composed of $n_j+1$ rigid bodies, called links, connected by $n_j$ joints with one degree of freedom each. An example of this type of kinematic structure is presented in Fig. \ref{fig:multibody}. Each link has an associated frame. We also assume that none of the links has an \emph{a priori} constant pose with respect to an inertial frame, i.e. the system is \emph{free floating}.

We can arbitrarily decide one of the rigid bodies to be the base link $B$, whose transformation with respect to the inertial frame is defined as ${}^\mathcal{I}\bm{H}_B$, and composed as follows:
\begin{equation}
{}^\mathcal{I}\bm{H}_B = \begin{bmatrix}
{}^\mathcal{I}\bm{R}_B & {}^\mathcal{I}\bm{p}_B \\
\bm{0}_{1 \times 3} & 1
\end{bmatrix}.
\end{equation}

We name $\pi_B(i)$ as the set of joints connecting link $i$ to the base. In addition, we assume the robot to have no closed kinematic chains. The joints in $\pi_B(i)$ are ordered such that the ($k + 1$)-th joint is not in the path from $B$ to the $k$-th joint. Hence, for a given choice of $B$, the set $\pi_B(i)$ is unique. For example, given Fig. \ref{fig:multibody}, if we define link 1 as our base, we have that $\pi_\text{1}(6) = $\{I, V\}, i.e. the path from link 1 to link 6 includes joints I and V. If we change base, e.g. link 5, we have $\pi_\text{5}(1) = $\{IV, I\} and $\pi_\text{5}(6) = $\{IV, V\}.

We define $\lambda_B(k)$ and $m_B(k)$ as the parent and child link of joint $k$, respectively. Each joint has a single parent and child link, but a rigid body can be connected to multiple joints, hence obtaining a tree structure. For example, referring again to Fig. \ref{fig:multibody}, we have $\lambda_1$(III) = 2, i.e. if we choose link 1 as base, link 2 appear to be the parent of joint III, while  $m_1$(III) = 3. If we choose link 3 as base, these two would be inverted.

We can finally reconstruct the position of a link $L$ as follows:
\begin{equation}\label{eq:fk_explicit}
	{}^\mathcal{I}\bm{H}_L = {}^\mathcal{I}\bm{H}_B \prod_{k \in \pi_B(L)} {}^{\lambda_B(k)}\bm{H}_{m_B(k)}.
\end{equation}

The transformation matrix ${}^{\lambda_B(k)}\bm{H}_{m_B(k)}$ depends upon the position of joint $k$, i.e. ${}^{\lambda_B(k)}\bm{H}_{m_B(k)}(s_k)$.
In this thesis we consider only \emph{revolute joints}, characterized by an axis ${}_k\bm{a} \in \mathbb{R}^3$ with unitary modulus. Let us consider an additional two frames $\bar{\lambda}_B(k)$ and $\bar{m}_B(k)$ belonging to the parent and child link respectively. They are placed such that ${}^{\bar{\lambda}_B(k)}\bm{H}_{\bar{m}_B(k)}(0) = \mathds{1}_4$. Thanks to this assumption, ${}_k\bm{a}$ has the same (constant) coordinates when defined either in $\bar{\lambda}_B(k)$ or $\bar{m}_B(k)$. Then, the joint transform is obtained thanks to the \emph{axis-angle} formalism:
\begin{equation}
\begin{split}
	{}^{\bar{\lambda}_B(k)}\bm{H}_{\bar{m}_B(k)}(s_k) &= \begin{bmatrix}
	{}^{\bar{\lambda}_B(k)}\bm{R}_{\bar{m}_B(k)}(s_k) & \bm{0}_{3 \times 1} \\
	\bm{0}_{1 \times 3} & 1
	\end{bmatrix} \\
	{}^{\bar{\lambda}_B(k)}\bm{R}_{\bar{m}_B(k)}(s_k) &= \mathds{1}_3 + \cos(s_k)({}_k\bm{a})^\wedge + \sin(s_k)\left(({}_k\bm{a})^\wedge\right)^2,
\end{split}
\end{equation}
obtaining the final relation
\begin{equation}
	{}^{\lambda_B(k)}\bm{H}_{m_B(k)}(s_k) = {}^{{\lambda}_B(k)}\bm{H}_{\bar{\lambda}_B(k)} \, {}^{\bar{\lambda}_B(k)}\bm{H}_{\bar{m}_B(k)}(s_k) \; {}^{\bar{m}_B(k)}\bm{H}_{{m}_B(k)}.
\end{equation}

Eq. \eqref{eq:fk_explicit} defines the \emph{forward kinematics} function. It returns the absolute pose of a link given the base pose and the joint values $\bm{s} \in \mathbb{R}^{n_j}$, i,e. $FK : \SE(3) \times \mathbb{R}^{n_j} \rightarrow \SE(3)$:
\begin{equation}
	{}^\mathcal{I}\bm{H}_L = FK\left({}^\mathcal{I}\bm{H}_B, \bm{s}\right).
\end{equation}

Let us now define the velocity of a link. We can express Eq. \eqref{eq:fk_explicit} by separating the part depending on the base pose from the joints dependency:
\begin{equation}
	{}^\mathcal{I}\bm{H}_L = {}^\mathcal{I}\bm{H}_B {}^B\bm{H}_L,
\end{equation}
thus, by applying Eq. \eqref{eq:velocity_composition}, we obtain
\begin{equation}\label{eq:velocity_inertial}
	{}^L\bm{V}_{\mathcal{I}, L} = {}^L\bm{X}_B{}^B\bm{V}_{\mathcal{I}, B} + {}^L\bm{V}_{B, L}.
\end{equation}
While ${}^B\bm{V}_{\mathcal{I}, B}$ is assumed to be known, ${}^L\bm{V}_{B, L}$ is a function of joints velocity $\dot{\bm{s}} \in \mathbb{R}^{n_j}$. In fact,
\begin{equation}
	{}^B\dot{\bm{H}}_{L} = \sum_{k \in \pi_B(L)} {}^B{\bm{H}}_{\lambda_B(k)}  \frac{\partial}{\partial s_k}\left({}^{\lambda_B(k)}\bm{H}_{m_B(k)}\right)  {}^{m_B(k)}{\bm{H}}_{L} \dot{s}_k
\end{equation}
and, by using the same derivation of Eq. \eqref{eq:adjoint_chang_coordinates_full}, we obtain
\begin{IEEEeqnarray}{RCL}
\phantomsection \IEEEyesnumber \label{eq:velocity_expanded}
	{}^L\bm{V}_{B, L} &=& \left({}^{B}\bm{H}_{L}^{-1} {}^B\dot{\bm{H}}_{L} \right)^\vee \nonumber \\
	&=& \sum_{k \in \pi_B(L)}\left({}^L{\bm{H}}_{\lambda_B(k)}  \frac{\partial}{\partial s_k}\left({}^{\lambda_B(k)}\bm{H}_{m_B(k)}\right)  {}^{m_B(k)}{\bm{H}}_{L}\right)^\vee \dot{s}_k \nonumber\\
	&=& \sum_{k \in \pi_B(L)}\left({}^L{\bm{H}}_{m_B(k)} {}^{\lambda_B(k)}{\bm{H}}_{m_B(k)}^{-1} \frac{\partial}{\partial s_k}\left({}^{\lambda_B(k)}\bm{H}_{m_B(k)}\right)  {}^{m_B(k)}{\bm{H}}_{L}\right)^\vee \dot{s}_k \nonumber\\
	&=& \sum_{k \in \pi_B(L)} {}^L\bm{X}_{m_B(k)} \left({}^{\lambda_B(k)}\bm{H}_{m_B(k)}^{-1} \frac{\partial}{\partial s_k}\left({}^{\lambda_B(k)}\bm{H}_{m_B(k)}\right) \right)^\vee \dot{\bm{s}}_k \nonumber \\
	{}^L\bm{V}_{B, L} &=& \sum_{k \in \pi_B(L)} {}^L\bm{X}_{m_B(k)}\,{}^{m_B(k)}\textbf{s}\, \dot{\bm{s}}_k. 
\end{IEEEeqnarray}
${}^{m_B(k)}\textbf{s}$ (more verbosely ${}^{m_B(k)}\textbf{s}_{\lambda_B(k), m_B(k)}$) is the \emph{left-trivialized joint motion subspace}. It is constant and it can be proven that, for a revolute joint, this depends on the rotation axis ${}_k\bm{a}$, \citep{traversaro2017thesis}:
\begin{equation}
	{}^{m_B(k)}\textbf{s}_{\lambda_B(k), m_B(k)} = {}^{{\lambda}_B(k)}\bm{X}_{\bar{\lambda}_B(k)} \, \begin{bmatrix}
	\bm{0}_{3\times 1} \\
	{}_k\bm{a}
	\end{bmatrix}.
\end{equation}

To summarize, ${}^L\bm{V}_{B, L}$ is an affine function of joints velocity $\dot{\bm{s}} \in \mathbb{R}^{n_j}$, allowing us to write:
\begin{equation}
	{}^L\bm{V}_{B, L} = {}^L\bm{J}_{B, L}^{\dot{s}}  \dot{\bm{s}},
\end{equation}
and including also the base velocity,
\begin{equation}
	{}^L\bm{V}_{\mathcal{I}, L} = \begin{bmatrix}
	{}^L\bm{X}_B & {}^L\bm{J}_{B, L}^{\dot{s}}
	\end{bmatrix} \begin{bmatrix}
	{}^B\bm{V}_{\mathcal{I}, B} \\
	\dot{\bm{s}}
	\end{bmatrix} = {}^L\bm{J}_{\mathcal{I}, L} \, \bm{\nu}.
\end{equation}
${}^L\bm{J}_{\mathcal{I}, L} \in \mathbb{R}^{6\times (6+n_j)}$ is the \emph{left-trivialized Jacobian} of link $L$, while $\bm{\nu} \in \mathbb{R}^{6+n_j}$ is the multi-body system velocity. The $k$-th column of ${}^L\bm{J}_{B, L}^{\dot{s}}$ is equal to ${}^L\bm{X}_{m_B(k)}\,{}^{m_B(k)}\textbf{s}$ if $k \in \pi_B(L)$, the zero vector otherwise.

\section{Multi-body dynamics}
\label{sec:modelling}
The robot configuration space is characterized by the \emph{position} and the \emph{orientation} of the base frame $B$, and the joint configurations. Thus, it corresponds to the group $\mathbb{Q} = \mathbb{R}^3 \times SO{(3)} \times \mathbb{R}^n$ and an element $\bm{q} \in \mathbb{Q}$ can be defined as the following triplet: $\bm{q} = ({}^{\mathcal{I}}\bm{p}_B, {}^{\mathcal{I}}\bm{R}_{B}, \bm{s})$.

The \emph{velocity} of the multi-body system can be characterized by the \emph{algebra} $\mathbb{V}$ of $\mathbb{Q}$ defined by: $\mathbb{V} = \mathbb{R}^3 \times \mathbb{R}^3 \times \mathbb{R}^n$.
An element of $\mathbb{V}$ corresponds to $\bm{\nu}$.

We also assume that the robot is interacting with the environment exchanging $n_c$ distinct wrenches\footnote{As an abuse of notation, we define as \emph{wrench} a quantity that is not the dual of a 
\emph{twist}, but a 6D force/moment vector.}. Due to the fact that the configuration space is not vectorial, we cannot apply the classical Euler-Lagrange equations. This is solved by employing the Euler-Poincar\'e formalism \citep[Ch. 13.5]{Marsden2010}, obtaining as a final result:
\begin{equation}
\label{eq:system_initial}
\bm{M}(\bm{q})\dot{\bm{\nu}} + \bm{C}(\bm{q}, \bm{\nu})\bm{\nu} + \bm{G}(\bm{q}) =  \begin{bmatrix}
\bm{0}_{6\times n} \\ \mathds{1}_n
\end{bmatrix}\bm{\tau}_s + \sum_{k = 1}^{n_c} \bm{J}^\top_{\mathcal{C}_k} {}_k\textbf{f}
\end{equation}
where $\bm{M} \in \mathbb{R}^{n+6 \times n+6}$ is the mass matrix, $\bm{C} \in \mathbb{R}^{(n+6) \times (n+6)}$ accounts for Coriolis and centrifugal effects, $\bm{G} \in \mathbb{R}^{n+6}$ is the gravity term. $\bm{\tau}_s \in \mathbb{R}^{n}$ is a vector representing the internal actuation torques, while ${}_k\textbf{f} \in \mathbb{R}^{6}$ denotes the $k$-th external wrench applied by the environment on the robot. In particular, it is composed as follows:
\begin{equation}
	{}_k\textbf{f} = \begin{bmatrix}
	{}_k\bm{f} \\
	{}_k\bm{\tau}
	\end{bmatrix}
\end{equation} 
with ${}_k\bm{f},\, {}_k\bm{\tau} \in \mathbb{R}^3$ being the contact force and torque respectively.
The Jacobian $\bm{J}_{\mathcal{C}_k} = \bm{J}_{\mathcal{C}_k}(\bm{q})$ is defined in different trivializations depending on the choice of frame used to measured the external wrench. In most of the applications described in this thesis, we assume the wrenches to be measured in a frame located on the contact link origin and oriented as $\mathcal{I}$. Hence, $\bm{J}_{\mathcal{C}_k}$ is expressed in \emph{mixed} representation.
By stacking all the Jacobians and contact wrenches, we can rewrite Eq. \eqref{eq:system_initial} as follows:
\begin{equation}
\label{eq:system}
\bm{M}(\bm{q})\dot{\bm{\nu}} + \bm{C}(\bm{q}, \bm{\nu})\bm{\nu} + \bm{G}(\bm{q}) =  \begin{bmatrix}
\bm{0}_{6\times n} \\ \mathds{1}_n
\end{bmatrix}\bm{\tau}_s + \bm{J}_{\mathcal{C}}^\top \textbf{f}
\end{equation}
where
\begin{equation}
	\bm{J}_{\mathcal{C}}(\bm{q}) = 
	\begin{bmatrix}\bm{J}_{\mathcal{C}_1}(\bm{q}) \\ \vdots \\ \bm{J}_{\mathcal{C}_{n_c}}(\bm{q})  \end{bmatrix}, \quad
	\textbf{f} = \begin{bmatrix}
		{}_1\textbf{f} \\
		\vdots\\
		{}_k\textbf{f}
	\end{bmatrix}.
\end{equation}

Lastly, it is assumed that a set of holonomic constraints acts on System \eqref{eq:system}. These are of the form $\bm{c}(\bm{q}) = 0$.
The holonomic constraints associated with all the rigid contacts can be represented as
\begin{equation}\label{eqn:constraintsAll}
\bm{J}_{\mathcal{C}}(\bm{q}) \bm{\nu} = 0,
\end{equation}
indicating that the velocity of the associated links is supposed to be zero. Eq. \eqref{eqn:constraintsAll} can be differentiated
\begin{equation}\label{eq:holonomic_constraint}
	\bm{J}_{\mathcal{C}} \dot{\bm{\nu}} + \dot{\bm{J}}_{\mathcal{C}} \bm{\nu} = 0 
\end{equation}
obtaining a dependency on $\dot{\bm{\nu}}$. Eq.s \eqref{eq:system} and \eqref{eq:holonomic_constraint} together are the dynamical equations describing the motion of a floating-base system instantiating contacts with the environment.

As described in \citep[Sec. 5]{traversaro2017thesis}, it is possible to apply a coordinate transformation in the state space $(q,{\nu})$ that transforms the system dynamics~\eqref{eq:system} into a new form where
the mass matrix is block diagonal, thus decoupling joint and base frame accelerations. Also, in this new set of coordinates,  the first six rows of Eq. \eqref{eq:system} corresponds to the \emph{centroidal dynamics}. In the specialized literature, this term is used to indicate the rate of change of the robot's momentum expressed at the center of mass, which then equals the summation of all external wrenches acting on the multi-body system \citep{orin2013centroidal, wensing2016improved}.

\section{Centroidal dynamics}\label{sec:intro_momentum}
By definition, the center of mass (CoM) $\bm{x}_\text{CoM}$ corresponds to the weighted average of all the links CoM position:
 \begin{equation}\label{eq:com_definition}
 \bm{x}_\text{CoM} = {}^\mathcal{I} \bm{H}_B\sum_i \frac{m_i}{m}{}^{B} \bm{H}_i  ~ {}^i\textbf{p}_{\text{CoM}},
 \end{equation}
where ${}^i\textbf{p}_{\text{CoM}} \in \mathbb{R}^3$ is the (constant) CoM position of link $i$ expressed in $i$ coordinates. $m, m_i \in \mathbb{R}$ are respectively the robot total mass and the $i$-th link mass. 
In order to introduce the \emph{centroidal dynamics}, it is convenient to define a frame, called $\bar{G}$, whose origin is located on the CoM, while the orientation is parallel to the inertial frame $\mathcal{I}$.
We introduce ${}_{\bar{G}} \bm{h} \in \mathbb{R}^6$ as the robot total momentum expressed in this frame, such that
\begin{equation}
{}_{\bar{G}} \bm{h} = \begin{bmatrix}
{}_{\bar{G}} \bm{h}^p \\
{}_{\bar{G}} \bm{h}^\omega
\end{bmatrix},
\end{equation}
where ${}_{\bar{G}} \bm{h}^p \in \mathbb{R}^3$ and ${}_{\bar{G}} \bm{h}^\omega \in \mathbb{R}^3$ are respectively the linear and angular momentum. In addition, the following holds:
\begin{equation}\label{eq:com_from_momentum}
	\dot{\bm{x}}_\text{CoM} = \frac{1}{m}{}_{\bar{G}} \bm{h}^p.
\end{equation}

The robot total momentum corresponds to the summation of all the links momenta, projected on the same frame $\bar{G}$:
\begin{equation}
{}_{\bar{G}}\bm{h} = \sum_i {}_{\bar{G}} \bm{X}^B {}_B\bm{h}.
\end{equation}
Notice that the adjoint matrix is the one used to transform wrenches. They are the dual of the adjoint matrices presented in Sec. \ref{sec:trivializatons}. In particular ${}_P\bm{X}^C := {}^C\bm{X}_P^\top$. We can expand all the momenta and write:
\begin{equation}\label{eq:momentumExpanded}
{}_{\bar{G}} \bm{h} = {}_{\bar{G}} \bm{X}^B \sum_i {}_{B}\bm{X}^i \bm{I}_i {}^i \bm{V}_{A,i}
\end{equation}
with $\bm{I}_i \in \mathbb{R}^{6\times 6}$ being the (constant) link inertia expressed in link frame. Hence, it is a function of the robot velocity $\bm{\nu}$:
\begin{equation}\label{eq:cmm_intro}
	{}_{\bar{G}} \bm{h} = \bm{J}_\text{CMM}\bm{\nu}
\end{equation}
where $\bm{J}_\text{CMM} \in \mathbb{R}^{6\times n}$ is the \emph{Centroidal Momentum Matrix} (CMM) \citep{orin08}.

The centroidal momentum rate of change balances the external wrenches applied to the robot:
\begin{equation}\label{eq:centroidal_momentum_dynamics}
\begin{split}
	{}_{\bar{G}} \dot{\bm{h}} &= \sum_{k = 1}^{n_c} {}_{\bar{G}}\bm{X}^k {}_k\textbf{f} + m \bar{\bm{g}} \\
	&= \sum_{k = 1}^{n_c} \begin{bmatrix}
	{}^{\mathcal{I}}\bm{R}_k & \bm{0}_{3\times 3} \\
	({}^{\mathcal{I}}\bm{o}_k - \bm{x}_\text{CoM})^\wedge\,{}^{\mathcal{I}}\bm{R}_k & {}^{\mathcal{I}}\bm{R}_k
	\end{bmatrix} {}_k\textbf{f} + m \bar{\bm{g}} 
\end{split}
\end{equation}
The adjoint matrix ${}_{\bar{G}}\bm{X}^k \in \mathbb{R}^{6 \times 6}$ transforms the contact wrench from the application frame (located in ${}^{\mathcal{I}}\bm{o}_k$ with orientation ${}^{\mathcal{I}}\bm{R}_k$) to $\bar{G}$. Finally, $\bar{\bm{g}} = \left[\begin{smallmatrix} 0 & 0 & -g & 0 & 0 & 0\end{smallmatrix}\right]^\top$ is the 6D gravity acceleration vector.

Alternatively, the centroidal momentum dynamics can be obtained by differentiating Eq. \eqref{eq:cmm_intro}:
\begin{equation}\label{eq:momentum_derivative_cmm}
	{}_{\bar{G}} \dot{\bm{h}} = \bm{J}_\text{CMM}\dot{\bm{\nu}} +  \dot{\bm{J}}_\text{CMM}\bm{\nu},
\end{equation}
thus highlighting its dependency on $\dot{\bm{\nu}}$.
\section{Simplified models}\label{sec:simplified}
In this short section we introduce how the robot dynamics is simplified obtaining two linear models which are widely used in the legged locomotion literature: the linear inverted pendulum model and the Capture Point.
\subsection{The linear inverted pendulum}\label{sec:lip}
The linear inverted pendulum model (in short LIP) approximates a legged robot by considering only the center of mass, represented as a point mass on the top of a pendulum with no inertia. The other edge is the ``foot", set in rigid contact with the ground. ${}_\text{p}\bm{x} \in \mathbb{R}^3$ is its position in an inertial frame. 

The LIP dynamics is obtained starting from Eq. \eqref{eq:centroidal_momentum_dynamics} considering only the linear momentum. We assume a single external force to be present, called ${}_\text{p}\bm{f} \in \mathbb{R}^3$. This is measured in an inertial frame, applied on ${}_\text{p}\bm{x}$, and parallel to the pendulum. Thus, the equations describing the LIP dynamics are:
\begin{IEEEeqnarray}{c} 
	\IEEEyesnumber \phantomsection \label{eq:lip_dynamics_full}	
	m\ddot{\bm{x}}_\text{CoM} = {}_\text{p}\bm{f} + m\bar{\bm{g}},  \IEEEyessubnumber \\
	\left({}_\text{p}\bm{x}-\bm{x}_\text{CoM}\right)^\wedge~ {}_\text{p}\bm{f} = 0.  \IEEEyessubnumber \label{eq:lip_force_constraint}    
\end{IEEEeqnarray}	
Notice that, thanks to Eq. \eqref{eq:lip_force_constraint}, the angular momentum is constant, i.e.:
\begin{equation}
	{}_{\bar{G}} \dot{\bm{h}}^\omega = 0.
\end{equation} 

Let us assume now the point mass to be always at a constant height, i.e.
\begin{equation}\label{eq:constant_height}
	\bm{e}_3^\top \left({}_\text{p}\bm{x}-\bm{x}_\text{CoM}\right) = \bar{x}_{\text{CoM}, z}.
\end{equation}
This implies $m\ddot{\bm{x}}_{\text{CoM},z} = 0$, which corresponds to 
\begin{equation}\label{eq:lip_normal_force}
{}_\text{p}f_z = -m|g|. 
\end{equation}    
The remaining components of ${}_\text{p}\bm{f}$ can be computed starting from Eq. \eqref{eq:lip_force_constraint} and Eq. \eqref{eq:constant_height}: 
\begin{IEEEeqnarray}{RCL}
	\IEEEyesnumber \phantomsection \label{eq:lip_tang_forces}
	{}_pf_x &=& m\omega^2\bm{e}_1^\top \left(\bm{x}_\text{CoM}- {}_\text{p}\bm{x}\right), \IEEEyessubnumber\\
	{}_pf_y &=& m\omega^2\bm{e}_2^\top \left(\bm{x}_\text{CoM} - {}_\text{p}\bm{x}\right), \IEEEyessubnumber
\end{IEEEeqnarray}    
where $\omega = \sqrt{\frac{|g|}{\bar{x}_{\text{CoM}, z}}}$ is the dominant time constant of the LIP. 

By substituting Eq.s \eqref{eq:lip_normal_force}-\eqref{eq:lip_tang_forces} into Eq. \eqref{eq:lip_dynamics_full}, we obtain:
\begin{equation*}
\ddot{\bm{x}}_\text{CoM} = \begin{bmatrix}
\bm{e}_1^\top \\
\bm{e}_2^\top \\
\bm{0}_{3\times 1}
\end{bmatrix} \omega^2\left(\bm{x}_\text{CoM} - {}_\text{p}\bm{x}\right).
\end{equation*}
Since the dynamics is different from zero only in the planar coordinates, we can define
\begin{equation}
	\bm{x}_\text{LIP} = \begin{bmatrix}
	\bm{e}_1^\top \\
	\bm{e}_2^\top
	\end{bmatrix}\bm{x}_\text{CoM}, \quad \bm{x}_\text{LIP} \in \mathbb{R}^2,
\end{equation}
and, without loss of generality, we consider ${}_\text{p}\bm{x} \in \mathbb{R}^2$. We then obtain the LIP dynamic equation,
\begin{equation}\label{eq:lip_dynamics}
	\ddot{\bm{x}}_\text{LIP} = \omega^2\left(\bm{x}_\text{LIP} - {}_\text{p}\bm{x}\right),
\end{equation} 
which is linear. It possess a single equilibrium point, namely
\begin{equation}\label{eq:lip_equilibrium}
	\begin{bmatrix}
	{\bm{x}}_\text{LIP}\\
	{\dot{\bm{x}}}_\text{LIP}
	\end{bmatrix}
	 := \begin{bmatrix}
	 		 {}_\text{p}\bm{x}\\
	 	\bm{0}
	 	\end{bmatrix}.
\end{equation}

This model can be enriched considering also a finite sized-foot \citep{Kajita2001,kajita20013d,koolen2012capturability}. In this case, it can be shown that ${}_\text{p}\bm{x}$ corresponds to the Zero Moment Point position $\bm{x}_\text{ZMP}$ \citep{vukobratovic2004zero}, instead of the foot position.
\subsection{The Capture Point}\label{sec:capture_point}
\begin{figure}[tpb]
	\def\svgwidth{0.9\columnwidth}
	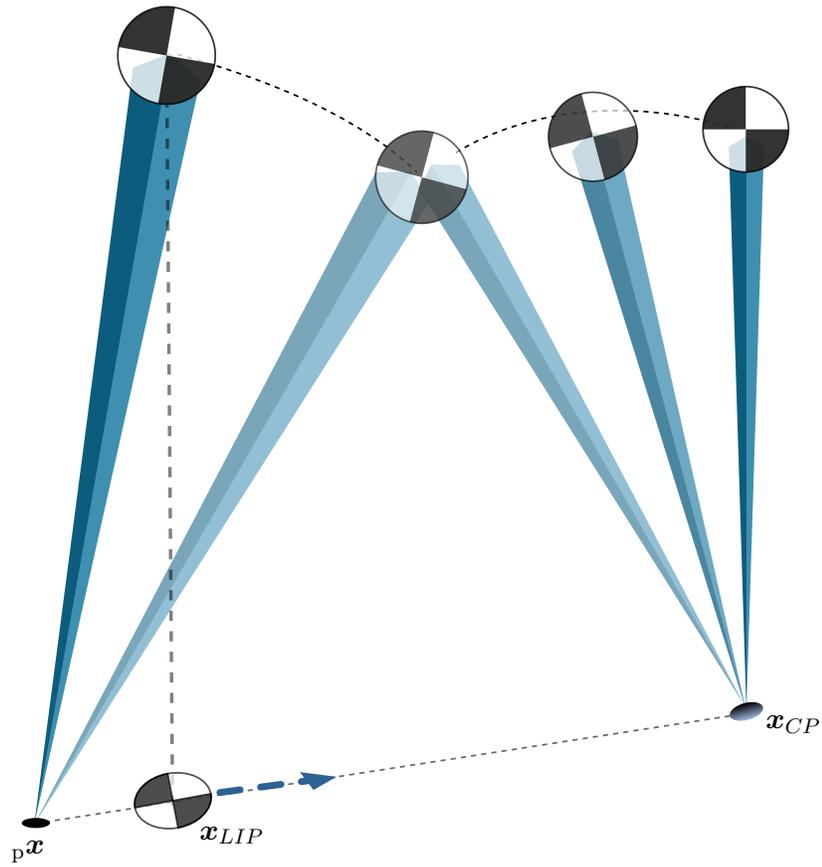
	\caption{Illustration of the Capture Point concept. It is the point over which the pendulum ``steps'' in order to stop in the upright position. This point depends on the $\bm{x}_{LIP}$ position and velocity. }
	\label{fig:capture_point}
\end{figure}

The Capture Point \citep{Pratt2006} is defined as the point on the ground where the robot has to step in order to come to a full stop, as shown in Fig. \ref{fig:capture_point}. Eq. \eqref{eq:lip_dynamics} can be rewritten as follows\footnote{ With an abuse of notation we write the dynamical system matrices as if $\bm{x}_\text{LIP}$ and its velocity were 1-dimensional, exploiting the fact that planar directions are decoupled.}:
\begin{equation}\label{eq:lip_matrix}
	\dot{\bm{\sigma}} = \begin{bmatrix}
	0 & 1\\
	\omega^2 & 0
	\end{bmatrix}\bm{\sigma} + \begin{bmatrix}
	0 \\
	-\omega^2
	\end{bmatrix}{}_\text{p}\bm{x}
\end{equation}
with $\bm{\sigma} = \left[\bm{x}_\text{LIP}^\top,~ \dot{\bm{x}}_\text{LIP}^\top \right]^\top$. The dynamical system of Eq. \eqref{eq:lip_matrix} can be explicitly solved \citep{englsberger2011bipedal}, obtaining the following relation:
\begin{equation}\label{eq:lip_in_time}
	\bm{\sigma}(t) = \begin{bmatrix}
	\cosh(\omega t) & \frac{1}{\omega}\sinh(\omega t) \\
	\omega\sinh(\omega t) & \cosh(\omega t)
	\end{bmatrix}\bm{\sigma}(0) + \begin{bmatrix}
	1 - \cosh(\omega t) \\
	-\omega \sinh(\omega t)
	\end{bmatrix}{}_\text{p}\bm{x}.
\end{equation}
Eq. \eqref{eq:lip_in_time} can be used to find an expression for the Capture Point starting from its definition. In order for the pendulum to come to a complete stop, its asymptotic position needs to converge to the equilibrium point. Hence, let us substitute $\bm{x}_\text{LIP}$ with ${}_\text{p}\bm{x}$, obtaining
\begin{equation}
	\cosh(\omega t)~\bm{x}_\text{LIP}(0) + \frac{1}{\omega}\sinh(\omega t)~\dot{\bm{x}}_\text{LIP}(0) - \cosh(\omega t)~{}_\text{p}\bm{x} = 0,
\end{equation}
from which we get
\begin{equation}
	{}_\text{p}\bm{x} = \bm{x}_\text{LIP}(0) + \frac{1}{\omega}\tanh(\omega t)~\dot{\bm{x}}_\text{LIP}(0).
\end{equation}
Since we are interested in the asymptotic solution, we let $t\rightarrow\infty$ which results in:
\begin{equation}
	{}_\text{p}\bm{x} = \bm{x}_\text{LIP}(0) + \frac{1}{\omega}\dot{\bm{x}}_\text{LIP}(0).
\end{equation}
Hence, given a generic state $\bm{\sigma}$, the Capture Point $\bm{x}_{CP} \in \mathbb{R}^2$ is defined as:
\begin{equation}\label{eq:capture_point}
	\bm{x}_{CP} = \bm{x}_\text{LIP} + \frac{1}{\omega}\dot{\bm{x}}_\text{LIP}.
\end{equation}

In order to find the Capture Point dynamics, let us analyze again the system of Eq. \eqref{eq:lip_matrix}. The associated eigenvalues are:
\begin{equation}
\lambda_{1,2} = \pm\omega
\end{equation}
which indicate a neutral saddle point (an equilibrium with two real eigenvalues symmetric with respect to the imaginary axis \cite[Section 2.1]{khalil2002}). The corresponding eigenvectors are:
\begin{equation}\label{eq:lip_eigenvectors}
\bm{v}=(\bm{v}_1\:\bm{v}_2)=\begin{bmatrix}
1 & 1\\
\omega_0 & -\omega_0
\end{bmatrix}.
\end{equation}	
We diagonalize the dynamical system of Eq. \eqref{eq:lip_matrix} through the relation: 
\begin{equation}
\bm{\sigma} = \bm{T}
\begin{bmatrix}
\bm{\xi}\\
\bm{\beta}
\end{bmatrix}, \quad \bm{\xi} \in \mathbb{R}^2,~\bm{\beta} \in \mathbb{R}^2,
\end{equation}
where the transformation matrix $\bm{T}$ and its inverse $\bm{T}^{-1}$ are defined as
\begin{equation}
	\bm{T} = \begin{bmatrix}
	1/2 & \hphantom{-}1/2\\
	\omega/2 & -\omega/2 
	\end{bmatrix}, \quad 	\bm{T}^{-1} = 
	\begin{bmatrix}
	1 & \hphantom{-}1/\omega\\
	1 & -1/\omega 
	\end{bmatrix}.
\end{equation}
The matrix $\bm{T}$ is obtained by multiplying the eigenvectors matrix by $1/2$. It is possible to notice that, using this transformation matrix, $\bm{\xi}$ is identical to $\bm{x}_{CP}$. Then, the corresponding diagonal system is:
\begin{equation}
\begin{bmatrix}
\dot{\bm{\xi}}\\
\dot{\bm{\beta}}
\end{bmatrix}=
\begin{bmatrix}
\dot{\bm{x}}_{CP}\\
\dot{\bm{\beta}}
\end{bmatrix}=
	\begin{bmatrix}
	\omega & 0 \\
	0 & -\omega
	\end{bmatrix}
\begin{bmatrix}
\bm{x}_{CP}\\
\bm{\beta}
\end{bmatrix}+
	\begin{bmatrix}
	-\omega\\
	\hphantom{-}\omega
    \end{bmatrix} {}_\text{p}\bm{x}.
\end{equation}       
Hence, the Capture Point dynamics
\begin{equation}\label{eq:capture_point_dynamics}
	\dot{\bm{x}}_{CP} = \omega\left(\bm{x}_{CP} - {}_\text{p}\bm{x}\right)
\end{equation}
corresponds to the unstable part of the LIP dynamics (positive eigenvalue).

\chapter{State of the Art and Thesis Context}\label{chap:soa_context}
This short chapter presents the state of the art, focusing on the different components of a layered architecture. Sec. \ref{sec:context} contextualize the thesis.
\section{State of the art}\label{sec:generic_soa}
Humanoid robots are underactuated \citep{Spong1998} which, in essence, means that by exploiting all theirs actuated degrees-of-freedom they can control their internal configuration, but they cannot affect their global pose directly.
Nevertheless, it is possible to circumvent this limitation by exploiting contacts with the environment and the robot ability to change ``shape''.
Additional difficulties arise from the fact that contacts may change over time. This results in a different evolution of the constrained dynamical system making the overall system hybrid \citep{lygeros1999hybrid}, i.e. it possesses both a continuous and  discrete time dynamics.

A recent approach for bipedal locomotion control that became popular during the DARPA Robotics Challenge,
consists in defining a hierarchical control architecture \citep{feng2015optimization}. Each layer of this architecture receives inputs from the robot and its surrounding environment, and provides references to the layer below. The lower the layer, the shorter the time horizon that is used to evaluate the outputs. Also, lower layers usually employ more complex models to evaluate the outputs, but the shorter time horizon often results in faster computations for obtaining these outputs.

From higher to lower layers, \emph{trajectory optimization for footstep planning}, \emph{simplified model control}, and \emph{whole-body control} represent a common stratification of architectures for bipedal locomotion control \citep{carpentier2017multi, romualdi2018benchmarking}.

\subsection{Trajectory optimization for footstep planning} \label{sec:soa_trajectory}

This first layer is in charge of finding a sequence of robot footsteps, which is also crucial for maintaining the robot balance. A fundamental ability of humanoid robots consists, for example, in moving over steps and stairs, where they can exploit the legged configuration. A possible approach for the generation of humanoid motions over stairs is to tackle the problem as an extension of the planar walking motion generation \citep{hirai1998development,hu2016walking, caron2019stair}. More in general, when planning locomotion trajectories, the definition of contacts plays a central role. Several strategies are available in literature, here summarized in four categories. In particular, we distinguish them depending on how much information on the contact is supposed to be provided by the designer.
\begin{enumerate}
	\item Everything may be set by the designer, including contact sequence, location and timing. In other words, it is specified beforehand where, when and how the robot is supposed to step. Planning is performed on the remaining quantities, typically CoM state and body postures.
	\item With respect to the previous case, the designer may specify only the sequence of contacts. The actual contact timing and location are an output of the trajectory planner.
	\item In an increased level of detail, it is possible to model the contact activation and the deactivation through integer variables.
	\item Lastly, it is possible to model contacts as part of the dynamical system describing the robot behavior. With such a complete modeling, all the contact related quantities are an output of the planner. In this particular category, we focus on \emph{complementarity-free} methods which do not rely on classical complementarity conditions. The reason behind this choice is addressed in Chapter \ref{chap:modeling_dp}.
\end{enumerate} 
In the following, we expand these four categories.

\paragraph{Fixed contact sequence, timing and location.} The assumption of knowing in advance where the contacts will be established and in which instant simplifies the planning problem. Planning then focuses on the CoM state.
In \citep{caron2016multi} only the CoM linear acceleration is taken into consideration, while in \citep{daiplanning} the authors propose a convex upper bound of the angular momentum to be minimized. In \citep{ponton2016convex} the derivative of the angular momentum is approximated by using quadratic constraints together with slack variables necessary to keep the approximation error low. Nevertheless, in this approach, it is not possible to directly penalize the use of the angular momentum, while it introduces many additional variables into the formulation. Similarly, in \citep{fernbach2019c} the CoM trajectory is forced to be polynomial with only one free variable. The resulting optimization problem is convex with the drawback of limiting the set of trajectories that can be generated depending on the choice of polynomial function.

Body postures can also be planned together with the centroidal quantities, thus considering also the robot kinematic structure, \citep{de2010feature,herzog2015trajectory,kudruss2015optimal, posa2016optimization, serra2016newton, fernbach2018croc}.

These methods need to rely on external contact planners. In some applications, where multiple contacts can be established in several regions, this approach may be the most viable solution. In this case, search algorithms can be used to determine suitable contact locations \citep{chestnutt2003planning, chestnutt2005footstep, perrin2011fast, stumpf2014supervised, karkowski2016real, griffin2019footstep, lin2019efficient}.

Under the same simplifications, authors of \citep{feng20133d, budhiraja2018differential, giraud2020motion} adopted Differential Dynamic Programming to generate whole-body motions, thus solving the optimization problem using the Bellman's principle of optimality presented in Sec. \ref{sec:indirect}. These are some of the very few applications of indirect methods in humanoid robotics.

\paragraph{Predefined contact sequence.} 
During locomotion, it can be assumed to know in advance the contact sequence. As an example, for a biped robot, it can be assumed that a contact with the left foot will be followed by another one with the right foot. Recently \citep{carpentier2016versatile,caron2017make,winkler18}, authors employ a similar method where contact phases are predefined. By specifying a different set of equations depending on the contact state, the hybrid nature arising from the establishment of contacts is easily modeled. The time spent in each phase can be turned into an optimization variable. Nevertheless, in case of several point contacts, the definition of the various phases could become intractable. 

\paragraph{Mixed integer programming.} 
Instead of receiving the contact sequence as input, it is possible to use integer variables to determine where a contact should be established \citep{deits2014footstep, mason2018mpc, mirjalili2018whole} and in which instant \citep{ibanez2014emergence, aceituno2018simultaneous}. These approaches require \emph{Mixed Integer Programming} tools. While providing enhanced modeling capabilities, the exploitation of integer variables strongly affects the computational performances, especially in case several contacts can be established. Furthermore, the availability of specialized solvers is limited.

\paragraph{Complementarity-free contact modeling.}
The \emph{complementarity-free} approach allows simulating multi-body systems subject to contacts, without enforcing complementarity conditions directly \citep{todorov2011convex, drumwright2010modeling}. Equivalently accurate results are obtained by maximizing the rate of energy dissipation.
These methods consider contacts explicitly inside the planner. Hence, contact location, timing and sequence are decided directly by the planner, allowing to generate complex motions \citep{mordatch2012discovery, tassa2012synthesis, erez2013integrated}. 

\subsection{Simplified model control}

This layer is in charge of finding feasible trajectories for the robot center of mass along the walking path. The computational burden to find feasibility regions, however, usually calls for simplified models to characterize the robot dynamics.
Indeed, when restricting the robot center of mass (CoM) on a plane at a fixed height and assuming constant angular momentum, it is possible to derive simple and effective control laws based on the well known Linear Inverted Pendulum (LIP) \citep{Kajita2001, kajita20013d} and the Capture Point \citep{Pratt2006}, as also introduced in Sec. \ref{sec:simplified}. The \emph{Divergent Component of Motion} (DCM) is a similar concept \citep{takenaka2009real}. Initially conceived in 2D , it has been extended in the 3D case too \citep{englsberger2013three}. The Capture Point and the DCM can be adopted to draw stability properties of walking motions \citep{englsberger2011bipedal, koolen2012capturability,krause2012stabilization}. 
These simplified models have a widespread diffusion, also amongst torque controlled robots \citep{stephens2010pushforce, pratt2012capturability, dafarra2016torque, griffin2017walking, englsberger2018torque}. Thus, they present high level of robustness, also in case of strong perturbations as it is usually the case when controlling a humanoid robot in torque mode.

The linearity of the model is based on the assumption of constant CoM height, resulting in a fixed pendulum constant $\omega$. Instead, when considering $\omega$ as a tuning parameter, the model linearity can be preserved even in case of varying CoM height  \citep{englsberger2013three}. If it is left unconstrained, it is possible to generate walking trajectories with a straight knee configuration \citep{griffin2018straight}.
In other works, balance is maintained by varying only the CoM height \citep{koolen2016balance}, hence completely removing the assumption for it to be constant. The robustness properties of this strategy have been further studied through the use of Sums-of-Squares optimization \citep{posa2017balancing}. A similar model has been used also in a multi-contact scenario \citep{perrin2018effective} and extended considering zero-moment point (ZMP) \citep{vukobratovic2004zero} motions to determine stability conditions \citep{caron2017dynamic,caron2018balance}. These approaches have been used to control a humanoid robot considering only single support phases.

These simplified linear models have allowed on-line model predictive control \citep{wieber2006trajectory,diedam2008online,missura2014balanced, naveau2016reactive, griffin2016model, bombile2017capture, joe2018balance}, also providing  references for the footstep locations in the form of deltas with respect to desired values. In addition, it is possible to derive MPC schemes with the guarantee of producing ``stable'' CoM trajectories using the DCM model \citep{scianca2016intrinsically}.

Model predictive controllers are appealing also for controlling hybrid systems \citep{bemporad2002hybrid,lazar2006stabilizing} since the full hybrid model can be exploited. 
Indeed, thanks to the prediction capabilities, it is possible to include inside the formulation both time- and state-dependent switching, performing anticipatory actions for the imminent variation in the dynamics. However MPC does not solve those problems related to the numerical integration of hybrid systems, which indeed are an open research problem \citep{olejnik2017modeling, azhmyakov2019relaxation,  rijnen2019sensitivity}.

When complete robot models are combined with impact dynamics, the control problem increases in complexity. Furthermore, ensuring stability and convergence properties of the underlying system requires employing control frameworks developed for hybrid systems \citep{engell2003modelling}. A possibility to ensure these properties is to use the control approach based on virtual constraints and hybrid zero dynamics \citep{grizzle2001asymptotically,westervelt2003hybrid,grizzle2014models,ames2014rapidly}, which has also been proved in real experiments performed using humanoid robots with point feet \citep{chevallereau2003rabbit} or passive compliant ankles \citep{reher2016realizing}.

\subsection{Whole-body controllers} \label{sec:whole-body-controllers}
The whole-body controllers are instantaneous algorithms that usually find (desired) joint torques and contact forces achieving some desired robot accelerations. In this framework, the generated joint-torques and contact forces can satisfy inequality constraints, which allow imposing friction constraints. The outputs of these controllers shall ensure the tracking of reference positions coming from the simplified model control layer. Although the reference positions may be stabilized  by a joint-position control loop, joint-torque based controllers have shown to ensure a degree of compliance, which also allows safe interactions with the environment \citep{Ott2011, Saab2013}. From the modeling point of view, full-body \emph{floating-base} models are usually employed in QP controllers. 

These controllers are often composed of several tasks, organised with strict or weighted hierarchies. In the former case, null-space projections are performed, making sure that a low priority task does not interfere with those at higher priority \citep{park2006contact, wensing2013generation, Herzog2014,nava2016stability, pucci2016highly, koolen2016design, padois2017whole}.  Instead, in the latter case, all the tasks are usually included in a cost function, regulating the relative priority through a set of weights~\citep{lee2012, bouyarmane2017weight}. With this approach, even the highest priority task is affected by other tasks depending on the choice of weights. Hence, this approach may require more tuning. On the other hand, there is less chance of \emph{task unfeasibility}. This is the case when high priority tasks require an unfeasible effort from the robot. At the same time, even if all tasks are feasible, ill-defined tasks may still cause unwanted motions. Hence, in the weighted case, all tasks act as regularization, avoiding to pick solutions which are good for one task but inefficient for all the others.
In \citep{Stephens2010} multiple tasks are achieved by introducing fictitious desired wrenches in the controller formulation.

Interestingly, in \citep{henze2014posture,koenemann2015whole, neunert2018whole, giftthaler2018family}, an MPC controller is applied for whole-body balance and tracking of walking trajectories.

\section{Thesis context}\label{sec:context}

\subsection[Part \ref{part:applications}]{Part \ref{part:applications}: \\ \nameref{part:applications}} 

This part presents three applications of optimal control and trajectory planning to humanoid locomotion. Here follows the context for each chapter. 

\subsubsection*{Chapter \ref{chap:steprecovery}: \nameref{chap:steprecovery}}
We present a momentum-based whole-body torque controller based on a MPC formulation. 
In particular, as in \citep{caron2016multi,daiplanning,ponton2016convex} the dynamic evolutions of the robot linear and angular momentum are taken into account. We thus make use of a \emph{reduced} model: differently from simplified models used in literature, the robot momentum is an exact model that captures the \emph{global} behaviour of the robot. We deal with the complications introduced by the derivative of the angular momentum by resorting to a Taylor expansion. Indeed, planners which resort to simplified models usually neglect angular momentum, supposing it to be constant. Exceptionally, in \citep{seyde2018inclusion} the angular momentum is considered while defining a desired Capture Point trajectory.

 The presented controller allows dealing directly with the intrinsic hybrid dynamics of the system by considering time-varying constraints. 
The computed control inputs, i.e. the contact wrenches, are directly applied to the robot, rather than being used to define a joint reference trajectory, thus representing a particularity of our approach. 

The controller architecture inherits its structure from the momentum-based whole-body torque controller \citep{Frontiers2015,pucci2016highly,nava16} implemented on iCub. Torque control is particularly suitable for our application given that it permits absorbing the impacts efficiently, maintaining the balance also in case of robot positioning errors.
We tested this approach on the simulated iCub humanoid robot while performing a step recovery strategy.
To summarize, the application presented in Chapter \ref{chap:steprecovery} is placed amongst the trajectory generators which relies on external contact planners. Nevertheless, it interfaces directly to a whole-body controller without the need of a simplified model controller.

\subsubsection*{Chapter \ref{chap:iros_walking}: \nameref{chap:iros_walking}}
This chapter presents an on-line predictive kinematic planner for foot-step positioning and center-of-mass (CoM) trajectories. This planner is also integrated with a stack-of-task torque controller, which ensures a degree of \emph{robot compliance}. The entire architecture uses on-line feedback from the robot and user-desired quantities. It implements the above three layers, and can be launched on both position and torque controlled robots. 

The \emph{trajectory optimization for foot-step planning} is achieved by a planning module that uses a simplified kinematic robot model, namely, the unicycle model. Foot positions are updated depending on the robot state and on a desired direction coming from a joypad, which gives a human user teleoperating the robot the possibility of defining  desired walking paths.   Differently from~\citep{faragasso2013vision}, we do not fix a desired robot velocity. Compared to \citep{diedam2008online,missura2014balanced,bombile2017capture}, we do not assume the robot to be always in single support. As a consequence, the robot  avoids to step in place continuously if the desired robot position does not change, or changes slowly. Another drawback of these approaches is that foot rotations are planned separately from linear positions, and this drawback is overcome by our approach. 

Once footsteps are defined, a MPC module generates kinematically feasible trajectories for the robot center of mass and joint trajectories by using the LIP model \citep{Kajita2003} and  whole-body inverse kinematics. 

The CoM and foot trajectories are then streamed as desired values to either a joint-position control loop, or to a \emph{whole-body quadratic-programming (QP) controller}. This latter controller generates desired joint torques, ensuring a degree of robustness and robot compliance. The desired joint torques are then stabilized by a low-level joint torque controller. Experimental validations of the proposed approach are carried out on the iCub humanoid robot, with both position and joint torque control experiments, highlighting the differences between these two modes.

\subsubsection*{Chapter \ref{chap:stepups}: \nameref{chap:stepups}}

In this chapter we leverage trajectory optimization techniques to generate motions which allow a humanoid robot to perform large step-ups. In particular, we exploit a reduced model of the centroidal dynamics, where the forces on both feet are considered. The angular momentum is assumed constant, thus extending \citep{koolen2016balance}. By adopting a \emph{phase-based trajectory optimization} technique \citep{winkler18}, we plan over a fixed series of contacts. Through a particular heuristic we avoid generating trajectories which would require a high torque expenditure. Experiments are performed on the Atlas humanoid robot, showing a reduction of the maximal knee torque up to 30\% in simulation and up to 20\% on the real hardware.

\subsection[Part \ref{part:dynamic_planner}]{Part \ref{part:dynamic_planner}: \\ \nameref{part:dynamic_planner}}

In this part, we merge the first two layers, generating locomotion trajectories adopting the full kinematics of the robot and the centroidal dynamics.
 
\subsubsection*{Chapter \ref{chap:modeling_dp}: \nameref{chap:modeling_dp}}

In this chapter we model the robot in contact with the environment similarly to \citep{dai2014whole}. In particular, we use the full kinematics of the robot and the centroidal dynamics but, in addition, we introduce new methods to define the complementarity constraints, i.e. those which link the velocity of a contact point and the force applied to it. In addition, we present a novel method to constrain their planar motion. This method, similarly to \citep{todorov2011convex, drumwright2010modeling, dai2014whole}, does not consider a predefined contact sequence and does not need integer variables.

Finally, the robot in contact with environment is modeled as a single, non-linear, dynamical system subject to equality and inequality constraints. This allows applying the optimal control techniques presented in Chapter \ref{chap:oc}.

\subsubsection*{Chapter \ref{chap:tasks}: \nameref{chap:tasks}}

We present here the set of tasks and constraints which can generate walking motions. Compared to \citep{dai2014whole} we do not assume to have a joint trajectory to be followed. On the contrary, such reference is fixed, while walking motions are generated by defining a constant desired CoM velocity and a moving reference on the ground. Given the high dimensionality of the system under control, several other tasks are included to ``regularize'' the generated walking motion. 

This chapter complements Chap. \ref{chap:modeling_dp} in defining an optimal control problem. In particular, this chapter presents those tasks and constraints which are specific for the walking motions, thus not related to physical principles.

\subsubsection*{Chapter \ref{chap:experiments}: \nameref{chap:experiments}}
In this chapter we validate the model and the tasks presented in the previous chapters. In particular, we first show an example of generated trajectory. Secondly, we evaluate the performances of the planner when changing the method for defining the complementarity constraints. 

Finally, similarly to \citep{carpentier2016versatile}, we generate off-line a walking trajectory including the desired joints value using the method described in the previous chapters. These are first interpolated and then fed to the robot by means of a joint position controller.
\epigraphhead[500]{\begin{quotation}
		{\footnotesize
			\noindent\emph{Keep on the lookout for novel ideas that others have used successfully.\\ Your idea has to be original only in its adaptation to the problem you're working on.}
			\begin{flushright}
				Thomas Edison
			\end{flushright}
		}
	\end{quotation}}
\part{Predictive Control Based on Simplified Models}\label{part:applications}
\chapter{A Push Recovery Strategy for the iCub Humanoid Robot}\label{chap:steprecovery}
Balancing and reacting to strong and unexpected pushes is a critical requirement for humanoid robots. 
In this chapter, we use a predictive scheme for step recovery connected to a momentum-based whole-body torque controller, presented in Sec. \ref{sec:momentum}. We use prediction to consider the switching of contact constraints induced by the step. Using this strategy, the robot is able to stand and maintain the upright position even after pushes of intensity comparable to a third of the robot weight. Sec. \ref{sec:model} presents the modeling tools used to achieve such a goal, while Sec. \ref{sec:steprecovery_oc} defines the corresponding optimal control problem. The definition of references is introduced in Sec. \ref{sec:sr_references}. The prediction capabilities of the controller are also exploited to determine when the robot is about to fall and it is necessary to perform a step. This concept is presented in Sec. \ref{trigger}.
Experiments in simulation, shown in Sec. \ref{sec:simulation}, validate the proposed approach. Sec. \ref{sec:conclusion} presents concluding remarks introducing pros and cons of the approach.

\section{Background on momentum-based whole-body torque control} \label{sec:momentum}

The push recovery strategy proposed in this chapter interfaces with the momentum-based whole-body torque control implemented for the iCub humanoid robot.
In this section, we briefly outline it and we refer the reader to \citep{Frontiers2015,nava16} for additional details.

The momentum-based balancing controller is a hierarchical controller composed of two control objectives.
The first, and highest priority objective, is the tracking of a desired centroidal momentum while the second is the stabilization of the zero dynamics.
By assuming as virtual control inputs the contact wrenches $\textbf{f} = [{}_1\textbf{f}^\top, \cdots, {}_{n_c}\textbf{f}^\top]^\top$, it is possible to control the robot momentum by solving the following minimization problem:
\begin{IEEEeqnarray}{RL}
\phantomsection \IEEEyesnumber \label{eq:mom_min}
\minimize_\textbf{f} ~& \norm{{}_{\bar{G}} \dot{\bm{h}} - {}_{\bar{G}} \dot{\bm{h}}^*}^2 \nonumber\\
\text{subject to:}~ & \\
& {}_{\bar{G}} \dot{\bm{h}} :=  {}_{\bar{G}}\bm{X} \textbf{ f} + m \bar{\bm{g}}, \nonumber\\
& \textbf{\textit{A}}\,\textbf{f} \leq \textbf{\textit{b}}, \nonumber
\end{IEEEeqnarray}
where the matrix ${}_{\bar{G}}\bm{X}$ represents the column concatenation of all the transformation matrices ${}_{\bar{G}}\bm{X}^k$ of Eq. \eqref{eq:centroidal_momentum_dynamics}. The inequality constraint $\textbf{\textit{A}}\,\textbf{f} \leq \textbf{\textit{b}}$ represents friction cone, center of pressure and other constraints on the wrenches.
The desired momentum rate of change is obtained by mean of a PI control law plus a feed-forward action \citep{nava16}. In particular, it allows tracking a desired linear CoM position and velocity. The angular momentum is controlled to zero, including some custom-made integral terms to avoid instability of the zero-dynamics. 

The second objective is responsible for constraining the joint variables and avoid internal divergent behaviors.
As before, we can specify a minimization problem also for this second task, i.e.
\begin{IEEEeqnarray}{RCL}
	\IEEEyesnumber \phantomsection \label{eq:zero_stab_min}
	\minimize_{\bm{\tau}}&~ & \norm{\bm{\tau} - \bm{\psi}}^2  \IEEEyessubnumber \label{eq:zero_stab_min_cost}\\
	\text{subject to:}&& \bm{M}(\bm{q})\dot{\bm{\nu}} + \bm{C}(\bm{q}, \bm{\nu})\bm{\nu} + \bm{G}(\bm{q}) - \bm{J}^\top_{\mathcal{C}} \textbf{f} =  \begin{bmatrix}
		\bm{0}_{6\times n} \\ \mathds{1}_n
	\end{bmatrix}\bm{\tau}_s,  \IEEEyessubnumber \label{eq:zero_stab_min_dyn}\\
	&& \bm{J}_{\mathcal{C}} \dot{\bm{\nu}} + \dot{\bm{J}}_{\mathcal{C}} \bm{\nu} = 0,   \IEEEyessubnumber \label{eq:zero_stab_min_constr}\\
	&& \norm{{}_{\bar{G}} \dot{\bm{h}} - {}_{\bar{G}} \dot{\bm{h}}^*}^2 = \text{solution of }\eqref{eq:mom_min}\IEEEyessubnumber \label{eq:zero_stab_min_hier}.
\end{IEEEeqnarray}
Eq. \eqref{eq:zero_stab_min_dyn} corresponds to the free-floating dynamics of the mechanical system described in Sec. \ref{sec:modelling}.
Eq. \eqref{eq:zero_stab_min_constr} is the constraint equation describing the kinematic constraints associated with the contacts. $\bm{\psi}$ represents a joint torque reference computed as in a PD plus gravity and contact wrenches compensation controller. More details on this reference can be found in \citep{nava16}.
Eq. \eqref{eq:zero_stab_min_hier} is the hierarchical constraint, i.e. it prevents  the solution of this second problem from changing the optimum of Eq. \eqref{eq:mom_min}.

Equations \eqref{eq:zero_stab_min_dyn} and \eqref{eq:zero_stab_min_constr} together describes the dynamics of the constrained dynamical system, as described in Sec. \ref{sec:modelling}. Summarizing, from a functional point of view, the momentum-based controller takes as reference a desired momentum trajectory, a desired joints configuration and the set of contact constraints. The generated torques are then applied directly as references for the low-level torque control.

It is worth noting that when the constraint set changes, e.g. when the robot goes from two feet to one foot or vice versa, the constrained dynamics changes.
The overall system is thus a hybrid system and the discrete state transitions should be handled accordingly. Indeed, the strategy presented in this chapter deals with this problem.

\section{Modeling for push recovery} \label{sec:model}
Let us consider the case where no contacts are available except the two applied at the feet. Then, we can rewrite Eq. \ref{eq:centroidal_momentum_dynamics} as:
\begin{equation}\label{c6eq:model_2f}
\begin{bmatrix}
\ddot{\bm{x}}_\text{CoM}\\
{}_{\bar{G}} \dot{\bm{h}}^\omega
\end{bmatrix} {=} \begin{bmatrix}
				m^{-1}\mathds{1}_3 & \bm{0}_{3\times3}\\
				\bm{0}_{3\times3} & \mathds{1}_3
				\end{bmatrix}\left[
				{}_{\bar{G}}\bm{X}^{l} \: {}_{\bar{G}}\bm{X}^{r}
				\right]\begin{bmatrix}
									{}_l\textbf{f}\\{}_r\textbf{f}
							\end{bmatrix} {+} \bar{g},
\end{equation}
where the indexes $l$ and $r$ are relative to the left and right foot respectively. 
 
In order to model the step in a model predictive framework, we can assume to know the instant in which the swing foot touches the ground.
In fact, the considered model does not contain any information about the posture of the robot and therefore it is not possible to define a ``transition function''. For example, the distance of the foot from the ground can be used to predict when the step is going to take place, but this would require injecting also kinematic information in the planner. 
In other words, the controller is aware that the impact takes place in a precise instant in the future, but it does not know whether the quantities involved in the model affects this instant or not.
Then, the most viable choice is to consider this instant constant and known in advance, named $t_{impact}$ and equal to the time needed to perform a step. This quantity is set by the user, also depending on the dynamic capabilities of the robot.

For the same reason, the position that the swing foot takes after the step, is assumed to be known and constant too.
This takes particular importance since ${}_{\bar{G}}\bm{X}^{s}$ (where $s$ refers to the swing foot) directly depends on this position. In Sec. \ref{sec:sr_references} we describe how this position is obtained. 
Summarizing, the characteristics of the step, i.e. the duration and the target position, are assumed to be known and constant. 

\subsection{The angular momentum}\label{sec:ang_mom}
We focus now on the matrix ${}_{\bar{G}}\bm{X}^{i}$ (here $i$ is a placeholder for subscript $l$ or $r$), whose structure depends on the choice of the frame to describe the robot momentum. 
The matrix has the following form:
\begin{equation}
{}_{\bar{G}}\bm{X}^{i} = \begin{bmatrix}
								\mathds{1}_3 & \bm{0}_{3\times 3}\\
								({}_{i}\bm{p}-\bm{x}_{\text{CoM}})^{\wedge} & \mathds{1}_3
						\end{bmatrix},
\end{equation} 
with ${}_{i}\bm{p} \in \mathbb{R}^3$ being the foot position. In fact, we assume contact wrenches to be measured in a frame located in ${}_{i}\bm{p}$ and oriented as the inertial frame $\mathcal{I}$, thus parallel to $\bar{G}$. Hence, the time derivative of the angular momentum corresponds to:
\begin{equation}\label{eq:angular_momentum_def}
	{}_{\bar{G}} \dot{\bm{h}}^\omega = \sum_{i}^{l,r} ({}_{i}\bm{p}-\bm{x}_{\text{CoM}})^{\wedge} \: {}_i\bm{f} + {}_i\bm{\tau}.
\end{equation}
The product between $\bm{x}_\text{CoM}$ and ${}_i\bm{f}$ makes Eq. \eqref{eq:angular_momentum_def} bilinear (hence non-convex) with respect to these two variables.
In the literature this problem is addressed by minimizing an upper-bound of the angular momentum \citep{daiplanning}, or by approximating it through quadratic constraints \citep{ponton2016convex}. 
In the application presented in this chapter, angular momentum is mainly needed to bound the usage of contact wrenches, rather than being precisely controlled to zero. Thus, we can accept a more coarse approximation, relying on its Taylor expansion. It is computed around the values obtained from the latest available feedback, indicated with the superscript ${}^0$. Such approximation has the following expression:
\begin{IEEEeqnarray*}{RCL}
{}_{\bar{G}} \dot{\bm{h}}^\omega &\approx& {}_{\bar{G}} \dot{\bm{h}}^{\omega, 0} + \sum_{i}^{\{l,r\}} 
 \frac{\partial {}_{\bar{G}} \dot{\bm{h}}^\omega}{\partial {}_i\bm{f}} \left({}_i\bm{f} - {}_i\bm{f}^0\right)+ \frac{\partial {}_{\bar{G}} \dot{\bm{h}}^\omega}{\partial \bm{x}_\text{CoM}} \left(\bm{x}_\text{CoM} - \bm{x}_{\text{CoM}}^0\right), \nonumber\\
{}_{\bar{G}} \dot{\bm{h}}^\omega &\approx& \sum_{i}^{\{l,r\}} {}_i\bm{\tau} + \left({}_i\bm{p}-\bm{x}_{\text{CoM}}^0\right)^\wedge \: {}_i\bm{f}^0+  \left({}_{i}\bm{p}-\bm{x}_{\text{CoM}}^0\right)^\wedge \: \left({}_i\bm{f}-{}_i\bm{f}^0\right) +  \nonumber \\
&+& \left({}_i\bm{f}^0\right)^\wedge \: \left(\bm{x}_{\text{CoM}}-\bm{x}_{\text{CoM}}^0\right)  \label{eq:ang_mom_last} \nonumber\\
\end{IEEEeqnarray*}
where we exploited the anticommutative property of the cross product, i.e.  $\bm{x}^\wedge \bm{y} = -\bm{y}^\wedge  \bm{x}$. By simplifying, we finally obtain
\begin{equation}\label{eq:simplified_angular_momentum}
	{}_{\bar{G}} \dot{\bm{h}}^\omega \approx \sum_{i}^{\{l,r\}} {}_i\bm{\tau} + \left({}_i\bm{p}-\bm{x}_{\text{CoM}}^0\right)^\wedge \: {}_i\bm{f} + \left({}_i\bm{f}^0\right)^\wedge \: \left(\bm{x}_{\text{CoM}}-\bm{x}_{\text{CoM}}^0\right). \IEEEyesnumber
\end{equation}

The approximation introduced with the truncation to the first order is  $o\left(\left({}_i\bm{f}-{}_i\bm{f}^0\right)\left(\bm{x}_{\text{CoM}}-\bm{x}_{\text{CoM}}^0\right) \right)$. In fact, there are no other quadratic or higher order terms.

\section{Optimal control problem definition}\label{sec:steprecovery_oc}
First, let us define the state variables $\bm{\chi}$ and the control input variable $\textbf{f}$:
\begin{equation}
	\bm{\chi} \coloneqq \begin{bmatrix}
	\bm{x}_\text{CoM} \\
	\dot{\bm{x}}_\text{CoM} \\ 
	{}_{\bar{G}} {\bm{h}}^\omega
	\end{bmatrix},\quad
	\textbf{f} \coloneqq \begin{bmatrix}
	{}_l\textbf{f} \\
	{}_r\textbf{f}
	\end{bmatrix}, \quad \bm{\chi} \in \mathbb{R}^9\text{, }\textbf{f} \in \mathbb{R}^{12}.
\end{equation}
We can rewrite Eq. \eqref{c6eq:model_2f} using \eqref{eq:simplified_angular_momentum}, obtaining:
\begin{equation}\label{c6eq:model}
\dot{\bm{\chi}} = \tilde{\bm{E}}_s \bm{\chi} + \tilde{\bm{F}}_s \textbf{f} + \tilde{\bm{G}}_s + \tilde{\bm{K}}_s^0,
\end{equation}
\begin{IEEEeqnarray*}{RCL}
	\tilde{\bm{E}}_s &=& \begin{bmatrix}
		\bm{0}_{3\times3} & \mathds{1}_3 & \bm{0}_{3\times3}\\
		\bm{0}_{3\times3} & \bm{0}_{3\times3}  & \bm{0}_{3\times3}\\
		\left({}_l\bm{f}^0+{}_r\bm{f}^0\right)^\wedge & \bm{0}_{3\times3}  & \bm{0}_{3\times3}\\
	\end{bmatrix},\\
	\tilde{\bm{F}}_s &=& \begin{bmatrix}
		\bm{0}_{3\times 3} &\bm{0}_{3\times 3} &\bm{0}_{3\times 3} &\bm{0}_{3\times 3} \\
		m^{-1}\mathds{1}_3 & \bm{0}_{3\times 3} & m^{-1}\mathds{1}_3 & \bm{0}_{3\times 3}\\
		\left({}_{l}\bm{p}-\bm{x}_{\text{CoM}}^0\right)^\wedge & \mathds{1}_3 & \left({}_{r}\bm{p}-\bm{x}_{\text{CoM}}^0\right)^\wedge & \mathds{1}_3					           		
	\end{bmatrix},\\
	\tilde{\bm{G}}_s &=& \begin{bmatrix}
		\bm{0}_3\\
		-\bar{\bm{g}}
	\end{bmatrix},\quad \tilde{\bm{K}}^0_s = \begin{bmatrix}
	\bm{0}_6\\
	-\left({}_l\bm{f}^0+{}_r\bm{f}^0\right)^\wedge \bm{x}_{\text{CoM}}^0
\end{bmatrix}.
\end{IEEEeqnarray*}
Here, $\tilde{\bm{K}}^0_s$ introduces the constant terms resulting from the Taylor approximation. Thus, it is dependent from the latest available feedback coming from the robot.
Under the above considerations, the model described by Eq.\eqref{c6eq:model} is affine and can be easily inserted into a QP formulation, as described with more details in Sec. \ref{sec:qp}.

\subsection{Contact constraints}\label{c6sec:constraints}

The constraints to be enforced are mainly related to the feasibility of the exerted wrenches.
Furthermore the wrench acting on the swing foot should be null for every instant before the impact, i.e. $t_{impact}$.

We start by examining the constraints acting on the stance foot (defined with the subscript ${}_{st}$):
\begin{equation}\label{c6eq:wrench_constr}
\bm{A}_{st}\,{}_{st}\textbf{f} \leq {}_{st}\bm{b} \quad \forall t: t \leq T,
\end{equation}
with $T$ being the prediction horizon.
Eq. \eqref{c6eq:wrench_constr} encompasses all the considered inequalities constraints, namely: 
\begin{inparaenum}[i)]
    \item friction cone with linear approximation,
    \item center of pressure,
    \item positivity of the normal contact force.
\end{inparaenum}

The constraints on the swing foot instead, defined by the subscript ${}_{sw}$, introduce the hybrid nature of the system. In particular, the wrench must be null before the impact and feasible after it. Formally:
\begin{equation}\label{c6eq:fr_cont_costr}
\begin{cases}
	{}_{sw}\textbf{f} = \bm{0}_{6\times 1} &\forall t:\; t < t_{impact}\\
	\bm{A}{}_{sw}\,{}_{sw}\textbf{f} \leq {}_{sw}\bm{b}  &\forall t:\; t_{impact} \leq t \leq T
\end{cases}
\end{equation}
where $\bm{A}{}_{sw}$ and ${}_{sw}\bm{b}$ describe the same feasibility constraints as $\bm{A}_{st}$ and ${}_{st}\bm{b}$.
The above equation assumes that $t_{impact} \leq T$. If the impact does not occur inside the control horizon, then the wrench exerted by the right foot is forced to be null throughout the whole horizon.

We also included an additional constraint to enforce that balancing is kept after the step. In particular, after this instant, we can constrain the Capture Point $\bm{x}_{CP}$ (introduced in Sec. \ref{sec:capture_point}) to be inside the convex hull of the two feet, which can be predicted by knowing the future position of the swing foot.
Thus, we can define a set of linear inequalities such that if
\begin{equation}
	\bm{A}_{ch}\bm{x}_{CP} \leq {}_{ch}\bm{b}, \quad \forall t: t \geq t_{impact},
\end{equation}
is satisfied, than the Capture Point is inside the convex hull.
Hence, we force it to be in a \emph{capturable} state after the step is performed \citep{Pratt2006, koolen2012capturability}. In fact, the convex hull represents the region where it is possible to stabilize the Capture Point dynamics without additional steps. 

\subsection{Cost function definition}\label{c6sec:cost}
We define here the cost function applied within the MPC controller, defined by $\mathcal{J}$.
Note that different terms of the cost function act only after the step is taken. It has the following expression:
\begin{IEEEeqnarray}{rCl}
	\IEEEyesnumber \phantomsection \label{c6eq:cost}
	\mathcal{J} =& \frac{1}{2} & \left(  \int_{0}^{T} \norm{ \bm{\chi}(\tau) - \bm{\chi}^*(\tau)}^2_{\bm{K}_\chi}  \mathrm{d}\tau + \int_{0}^{T}  \norm{\textbf{f}(\tau)}^2_{\bm{K}_\text{f}} \dif\tau +\right. \IEEEyessubnumber \\
	&+&\int_{\bar{t}_{imp}}^{T} \norm{\bm{\chi}(\tau) - \bm{\chi}^*(\tau)}^2_{\bm{K}_\chi^{imp}}  \dif\tau + \IEEEyessubnumber \label{c6eq:cost_gamma2}\\
	&+& \norm{\bm{\chi}(T) - \bm{\chi}^*(T)}^2_{\bm{K}_\chi^{imp}}\Bigg). \IEEEyessubnumber \label{c6eq:cost_gamma_ter}
\end{IEEEeqnarray}
Since the impact may occur after the end of the prediction horizon, $\bar{t}_{imp}$ is the minimum between $t_{impact}$ and $T$. This allows us to weight the state $\bm{\chi}$ in a different way before ($\bm{K}_\chi$) and after ($\bm{K}_\chi^{imp}$) the step. 
For the sake of simplicity, the initial time instant is set to zero.

Finally, a terminal cost term is inserted (using the matrix $\bm{K}_\chi^{imp}$ introduced before) to model the finiteness of the control horizon. This is also useful in case $\bar{t}_{imp} = T$.

Consider now the final objective of having the center of mass over the centroid of the support polygon. We decide to weight only the normal component of the CoM error throughout the whole horizon, while weighting the transverse components (i.e. $x$ and $y$) only in the weight matrix $\bm{K}_\chi^{imp}$. Hence, they are considered only in the terminal cost, Eq.\eqref{c6eq:cost_gamma_ter}, and after the step is made, i.e in the cost of Eq.\eqref{c6eq:cost_gamma2}. 

Finally, the requested reaction forces are mildly weighted too. This term acts as a regularization, to prefer solutions requiring less stress to the robot.

The outputs of the MPC strategy are used as references for the whole-body torque controller described in Section \ref{sec:momentum}. In particular, we remove Eq. \eqref{eq:zero_stab_min_hier} since the MPC takes care of computing feasible wrenches $\textbf{f}$. These are directly included in Eq. \eqref{eq:zero_stab_min_dyn}. Desired joint positions are obtained by means of an inverse kinematics problem, i.e. we impose a desired pose for the swing foot, as defined in Sec. \ref{sec:sr_references}, following as close as possible the center of mass trajectory defined by the MPC controller. 
\subsection{Quadratic programming transcription} \label{sec:qp}

We solve the finite-horizon optimal control problem of Sec. \ref{sec:steprecovery_oc} with the direct multiple shooting method presented in Sec. \ref{sec:shooting}, i.e. we consider all the intermediate state variables. It is converted into a QP problem to be solved at each time step. 

We first discretize the model using a forward Euler scheme. Different approaches may have been chosen, as described in Sec. \ref{sec:integration_methods}, but since we already adopted a first order Taylor approximation in the dynamics (see Sec. \ref{sec:ang_mom}), it would be inconvenient to use higher order methods. In addition, we use a fixed integration step $\dif t$ of $10$ milliseconds, which is small enough (given the system into consideration) to avoid numerical instability.
By discretizing Eq.\eqref{c6eq:model}, we obtain:
\begin{equation}\label{c6eq:discrmodel}
\bm{\chi}(k+1) = \bm{E}_s \bm{\chi}(k) + \bm{F}_s \textbf{f}(k) +\bm{G}_s + \bm{K}^0_s,
\end{equation}
where
$
\bm{E}_s = \mathds{1}_9 + \mathrm{d}t\tilde{\bm{E}}_s,\, \bm{F}_s = \mathrm{d}t\tilde{\bm{F}}_s, \, \bm{G}_s  = \mathrm{d}t\tilde{\bm{G}}_s , \, \bm{K}^0_s = \mathrm{d}t\tilde{\bm{K}^0_s},
$
while $k \in \mathbb{N}$, $k = 0,\cdots\, , \, N-1$ is the discrete time, $N=\ceil{t_f/\mathrm{d}t}$.

According to the receding horizon principle, as presented in Sec. \ref{sec:receding_horizon}, at each time step, $\textbf{f}(0)$ is applied to the system through a modified version of the balancing controller implemented on iCub, as anticipated in Sec. \ref{c6sec:cost}. 

The definition of the time instant at which the impact occurs is a crucial point to be considered.
We assume the impact to be impulsive and taking place at the beginning of a time step. We denote this instant with $k_{impact}$.
We further assume that the control input $\textbf{f}(k)$ is applied piecewise constantly, i.e. constant from instant $k$ to $k+1$. 
Note that, in view of the above two assumptions, setting $k_{impact} = 0$ implies that both feet are in contact already at the initial time $k = 0$.

As mentioned previously, we do not directly model the distance between the swing foot and the ground, but we compute the position and the expected impact time at the beginning of the push strategy.
The following procedure describes how we update the expected impact instance throughout the proposed MPC controller.
Assuming the initial time instant to be always set to zero, we start with a value of $k_{impact}$ equal to $\lceil t_{impact}/\mathrm{d}t \rceil$. At each controller execution, we decrease it by one, i.e. $k_{impact} := k_{impact} - 1$. If the impact has not occurred yet, we saturate $k_{impact}$ to $1$, i.e. we expect the impact to occur at the second time step. This avoid requiring wrenches on a swinging limb.
If we \emph{measure} a certain amount of force applied at the swing leg, we assume the impact to have occurred. In this case, $k_{impact}$ is set constant to $0$.

\section{Definition of the swing foot reference position}\label{sec:sr_references}

The output of the optimal control problem presented in Sec. \ref{sec:steprecovery_oc} does not define the target position the swing foot should achieve. Nevertheless, such controller strongly depends on it. In this section, we present a simple heuristic we adopted to determine where the robot has to step \citep{dafarra2016torque}.

One key characteristic is that the foot position is not planned a priori, but it depends on the status of the robot when the push-recovery controller starts. This allows the robot to take different steps depending on the direction and intensity of the push.

The heuristic we use is based on the Capture Point concept described in Sec. \ref{sec:capture_point}. The solution to the differential equation of Eq. \eqref{eq:capture_point_dynamics}, with initial conditions in $t = 0$ is given by the following:
\begin{equation}
	\bm{x}_{CP}(t) = e^{\omega t}\left(\bm{x}_{CP}(0)- {}_\text{p}\bm{x}\right) + {}_\text{p}\bm{x},
\end{equation}
where ${}_\text{p}\bm{x}$ is the origin of the frame attached to the stance foot.
This equation describes the time evolution of the Capture Point given its initial condition $\bm{x}_{CP}$ and the (constant) stance foot position $x_\text{foot}$.
Knowing beforehand the time to perform the step $t_\text{step}$, the desired swing foot position ${}_{sw}\bm{x}^* \in \mathbb{R}^3$ is:
\begin{equation}
\label{eq:capture_point_predicted_lipm}
{}_{sw}\bm{x}^* = \begin{bmatrix}
e^{\omega t_\text{step}}\left(\bm{x}_{CP}(0)- {}_\text{p}\bm{x}\right) + {}_\text{p}\bm{x} \\
0
\end{bmatrix}.
\end{equation}
This quantity is kept constant for the whole prediction horizon. The orientation of the foot is defined by the user instead. In other words, we land the swing foot on the Capture Point position we predict at end of the step.

From the definition of the desired swing foot pose, it is also possible to compute a CoM position reference. In particular, for what concerns the $x$ and $y$ components, we choose it as the center of the convex hull given by the two feet. The desired height is kept equal to the initial one. The controller follows this reference only after the step is performed. During the swing instead, the controller does not follow any reference CoM trajectory, but it reaches this point thanks to its prediction capabilities.
\section{MPC as a step trigger}\label{trigger}
Let us consider the scenario where the robot has to perform a step because of an external perturbation. In principle we could leverage the presented MPC formulation to define the moment in which to start the motion, up to now considered as a datum. This moment can be defined as the instant where the controller is not able to bring the CoM back to a desired equilibrium point, given the present support configuration. In other words, the constraints related to the balancing configuration prevent the controller to recover from the push. Thus, it is necessary to step and change balancing configuration in order to avoid a fall. The availability of a prediction horizon particularly suits this idea. Hypothetically, if $T = \infty$, as soon as the robot is pushed, we could predict whether or not the controller will be able to absorb the disturbance completely. In practice this is not true. The finiteness of the prediction horizon hides the actual recovery capabilities of the controller, thus it is still necessary to define a heuristic.
By considering the very last predicted state, we can set the following condition:
\begin{equation}\label{eq:step_condition}
\norm{\bm{x}_{CoM}(T) - \bm{x}_{CoM}^*} + k_v \norm{\dot{\bm{x}}_{CoM}(T) - \dot{\bm{x}}_{CoM}^*} < \bar{d}.
\end{equation} 
When Eq.\eqref{eq:step_condition} is violated, the robot performs the step. Both $k_v$ and $\bar{d} \in \amsmathbb{R}$ are user-defined parameters affecting the sensitivity to pushes. The smaller $\bar{d}$ the more the robot will be inclined to take a step, while $k_v$ allows to regulate the relative importance of the two errors. 
Notice that Eq.\eqref{eq:step_condition} does not depend directly on the foot dimensions. Nonetheless, the heuristic depends on this quantity implicitly. A smaller foot would limit the capabilities of the MPC controller to steer the state towards its reference.

In order to be effective, this heuristic needs the MPC strategy to be in charge of sending references to the robot, even in case there are no disturbances. 
However, the goal is different from what has been presented in Sec. \ref{c6sec:cost}. The robot should do its best to avoid a step. As a consequence, $\bar{t}_{imp}$ will be set equal to $T$ (i.e. the step will not occur at all), while $\bm{K}_\chi$ should be equal to $\bm{K}_\chi^{imp}$, to maintain the robot in its current position.

\section{Validation and simulation results} \label{sec:simulation}

\begin{figure*}[tpb]
	\centering
	\subfloat[$t=t_0$] {\includegraphics[width=.23\textwidth]{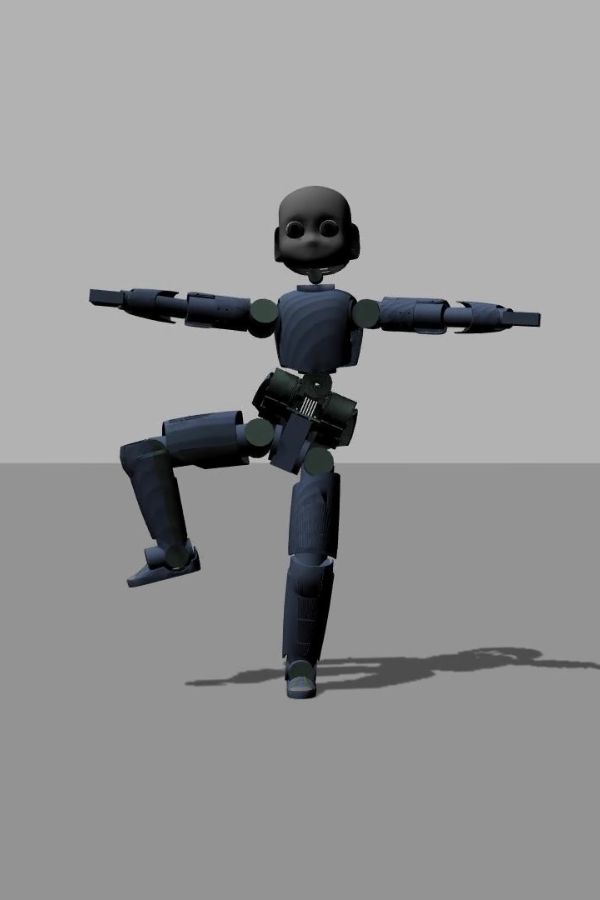}}
	\hspace{.01\textwidth}
	\subfloat[$t=t_0 + 1s$] {\includegraphics[width=.23\textwidth]{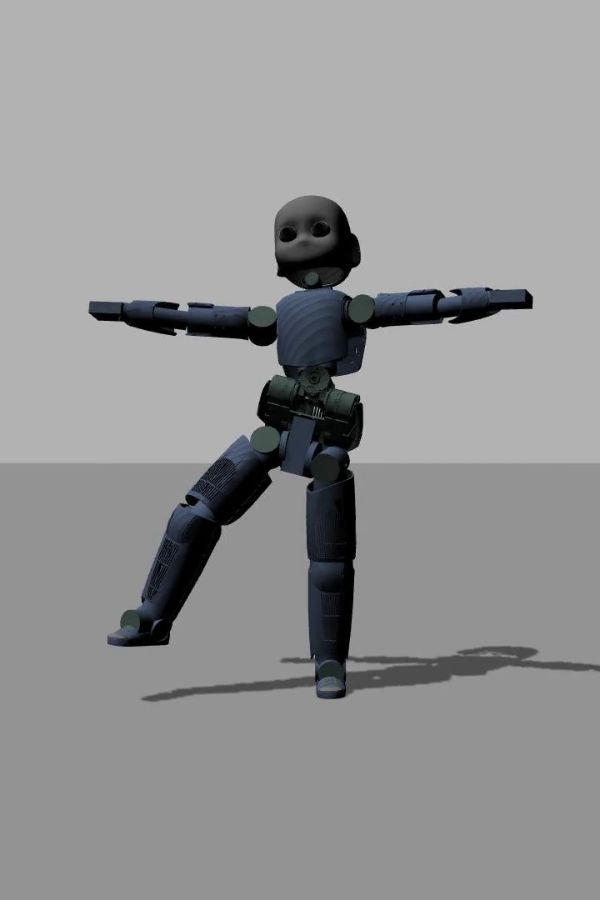}}
	\hspace{.01\textwidth}
	\subfloat[$t=t_0 + 2s$] {\includegraphics[width=.23\textwidth]{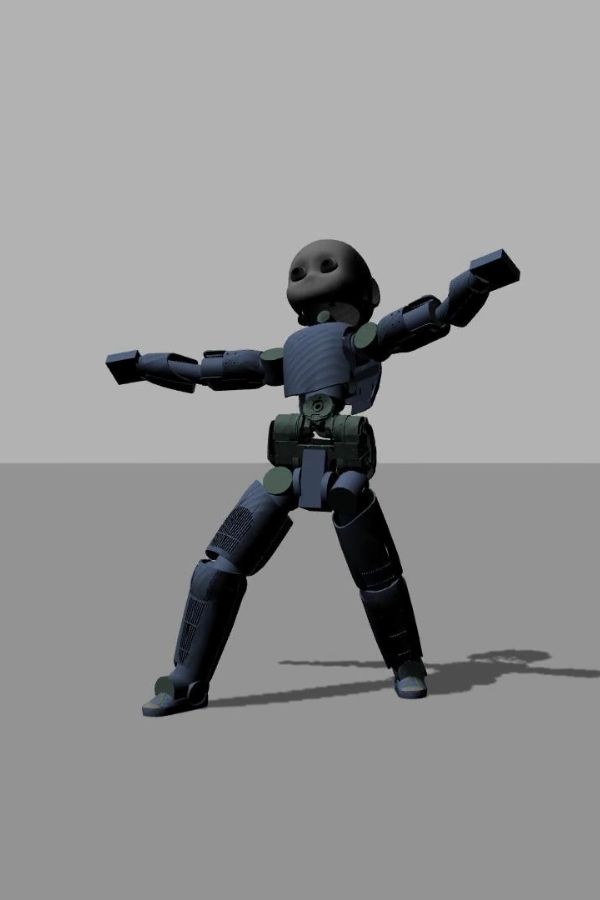}}
	\hspace{.01\textwidth}
	\subfloat[$t=t_0 + 2.5s$] {\includegraphics[width=.23\textwidth]{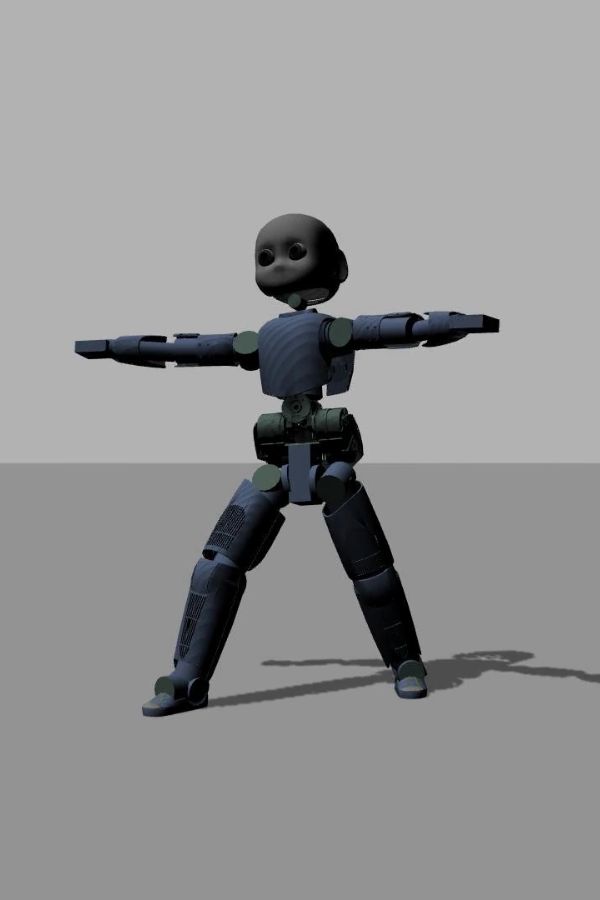}}
	\caption{Snapshots\protect\footnotemark ~of the step recovery motion when the robot is pushed at $45\degree$ with respect to the lateral axis.}
	   
	\label{fig_sr:snapshots}
\end{figure*}
\footnotetext{\url{https://www.youtube.com/playlist?list=PLBOchT-u69w9hJ6BmqPf06r0zWmungOrc}}

\begin{figure*}[tpb]
	\centering
	\subfloat[Side step] {\includegraphics[width=.9\textwidth]{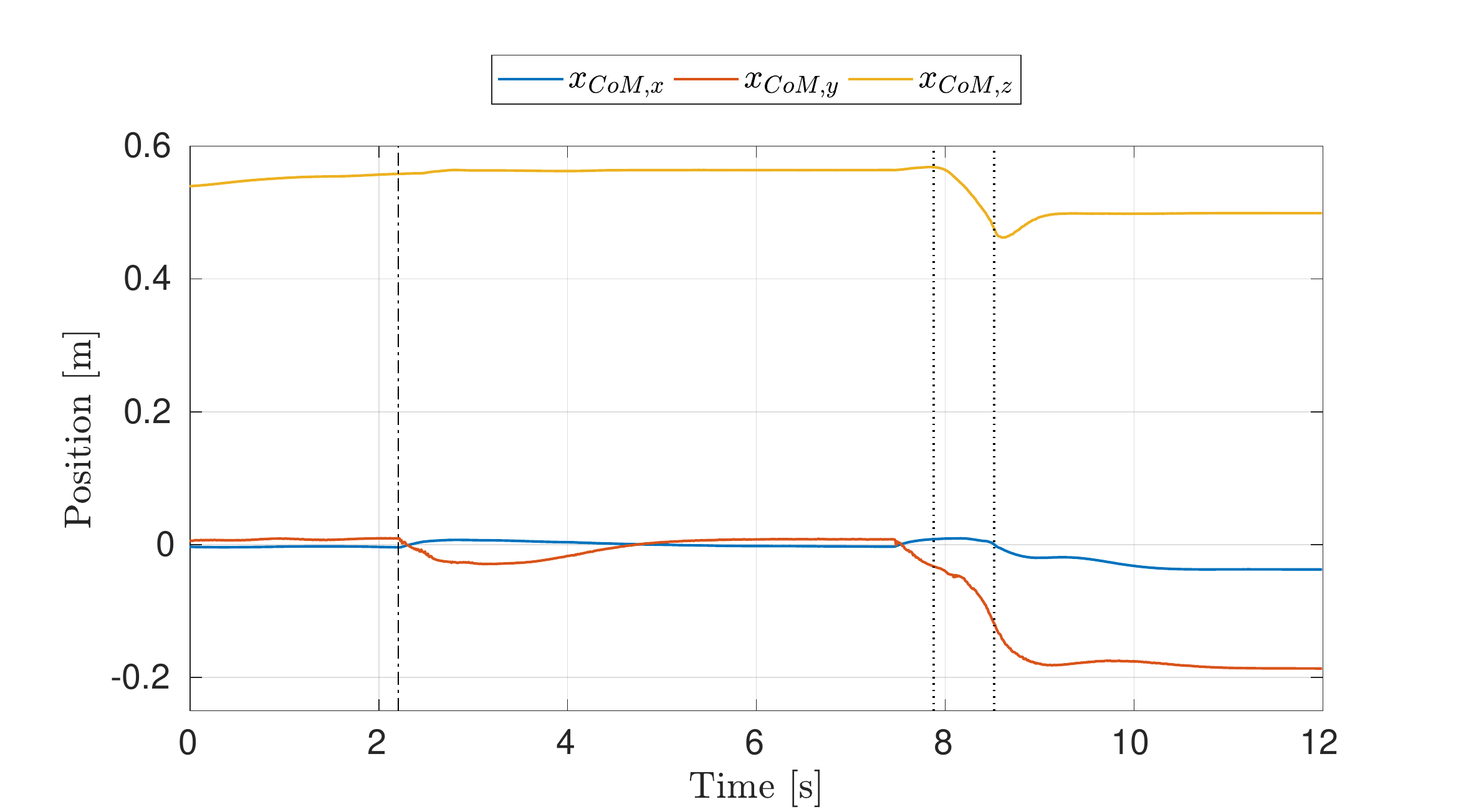}\label{fig:side}}
	
	\subfloat[Back step] {\includegraphics[width=.9\textwidth]{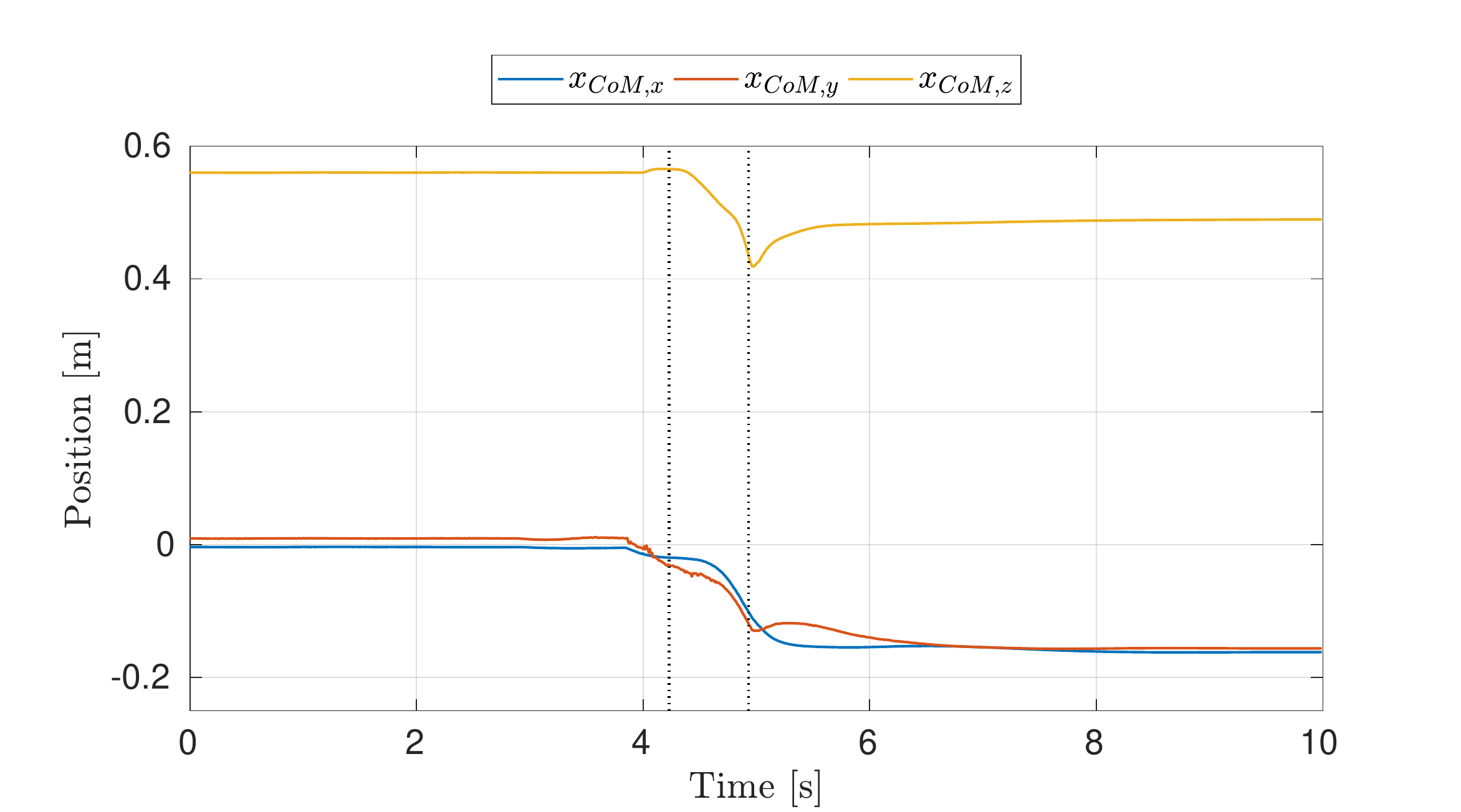}\label{fig:back}}
	\caption{CoM evolution during three different experiments of push recovery. The single dashed lines in \protect\subref{fig:side} and \protect\subref{fig:front} show a not-enough-strong push to violate the condition in Eq. \eqref{eq:step_condition}. The other two vertical dashed lines indicate the beginning and the end of the step (i.e. when the swing foot hits the ground). The external push force has been applied, with respect to the lateral axis, at $20\degree$, $-20\degree$ and $45\degree$ for the cases \protect\subref{fig:side}, \protect\subref{fig:back} and \protect\subref{fig:front} respectively.}
\end{figure*}
\begin{figure*}[tpb]\ContinuedFloat	
	\subfloat[Forward step] {\includegraphics[width=.9\textwidth]{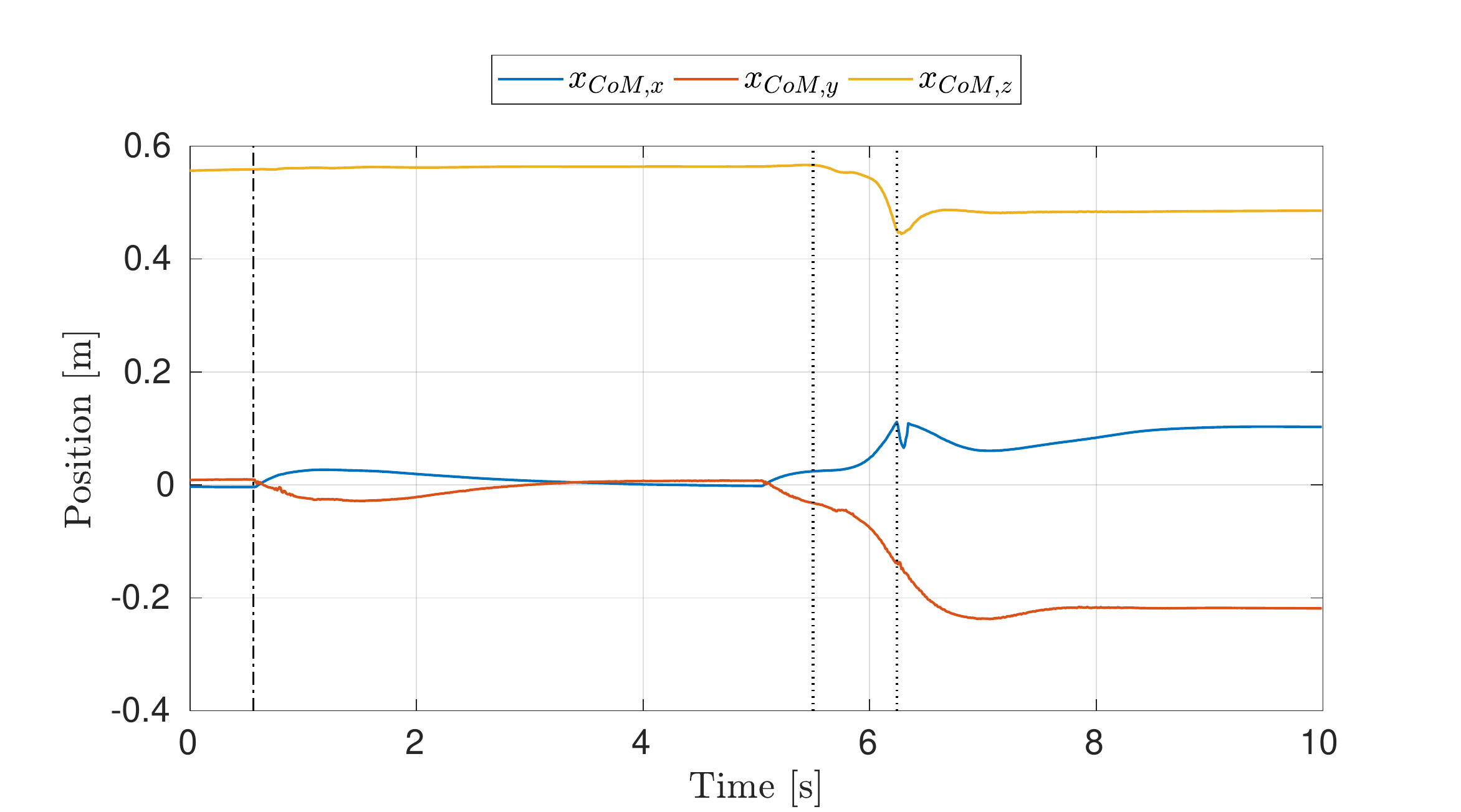}\label{fig:front}}
	\caption{(cont.) CoM evolution during three different experiments of push recovery. The single dashed lines in \protect\subref{fig:side} and \protect\subref{fig:front} show a not-enough-strong push to violate the condition in Eq. \eqref{eq:step_condition}. The other two vertical dashed lines indicate the beginning and the end of the step (i.e. when the swing foot hits the ground). The external push force has been applied, with respect to the lateral axis, at $20\degree$, $-20\degree$ and $45\degree$ for the cases \protect\subref{fig:side}, \protect\subref{fig:back} and \protect\subref{fig:front} respectively.}
	\label{fig:exp_com}
\end{figure*}
\begin{figure*}[tpb]	
	\subfloat[Side step] {\includegraphics[width=.9\textwidth]{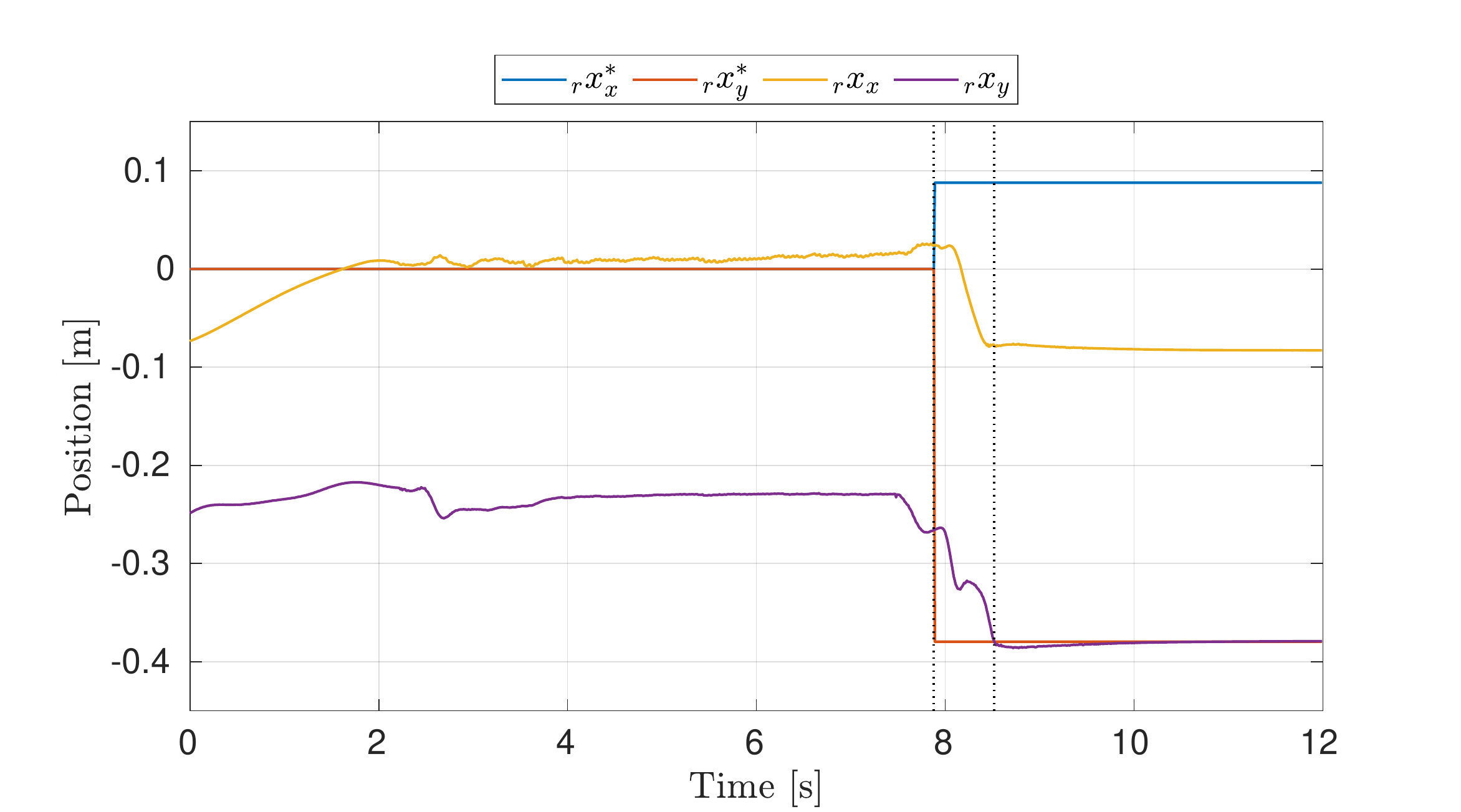}\label{fig:r_sole_side}}
	
	\subfloat[Back step] {\includegraphics[width=.9\textwidth]{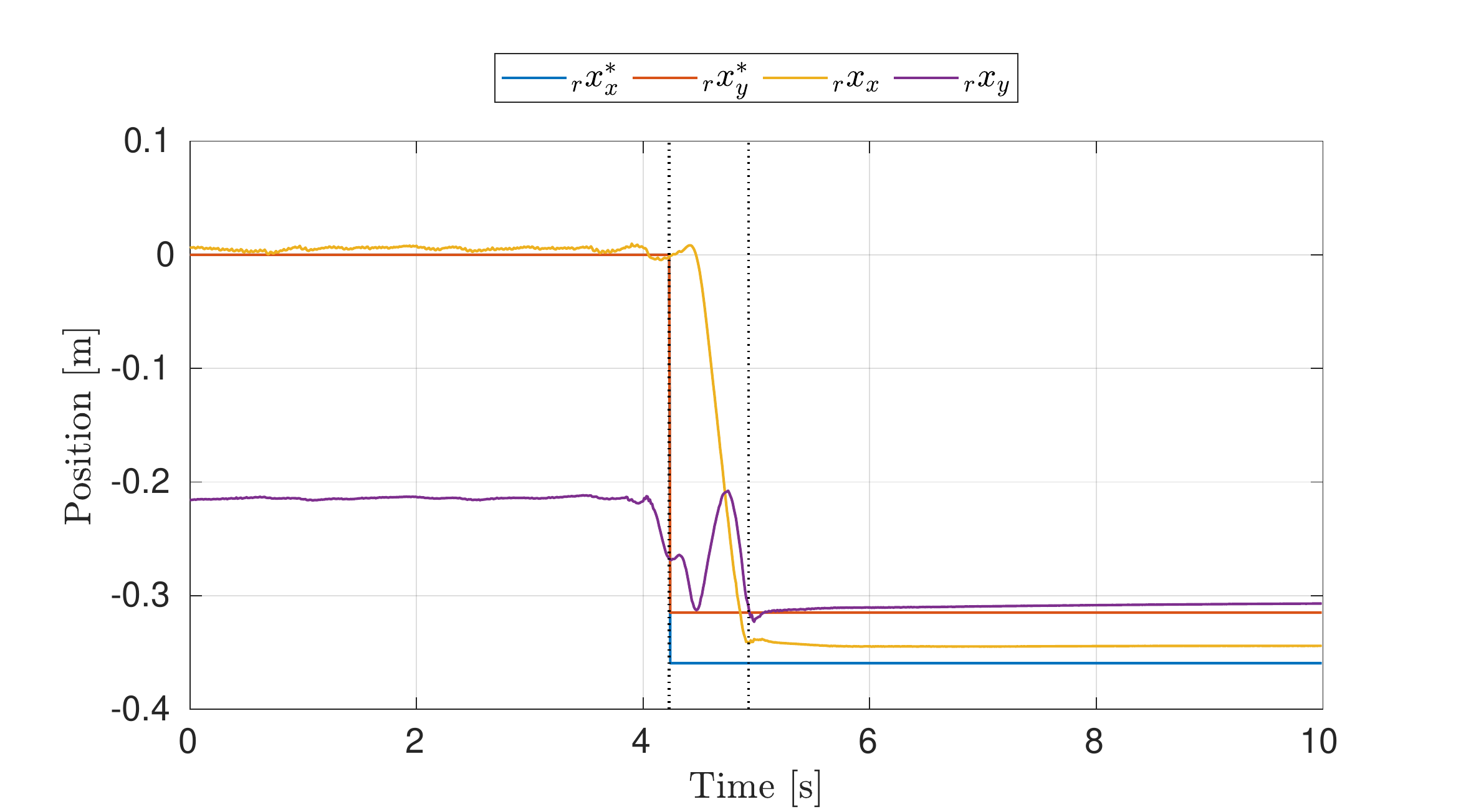}\label{fig:r_sole_back}}
	\caption{Tracking of the desired right foot position provided by the momentum controller introduced in Sec. \ref{sec:momentum}. The vertical dotted lines have the same meaning of Fig. \ref{fig:exp_com}, namely they represent the beginning and the end of the step. The reference position is computed with the method presented in Sec. \ref{sec:sr_references}. It is computed when the step starts and it is tracked only after this moment. The tracking is satisfactory in \protect\subref{fig:r_sole_back}, while it presents a large error on the $x-$direction in \protect\subref{fig:r_sole_side}. In \protect\subref{fig:r_sole_front} there are non-negligible errors for both the $x-$ and $y-$ direction, probably because of a too far reference, that is hard to track during such dynamic motion.}
\end{figure*}
\begin{figure*}[tpb]\ContinuedFloat	
	\subfloat[Forward step] {\includegraphics[width=.9\textwidth]{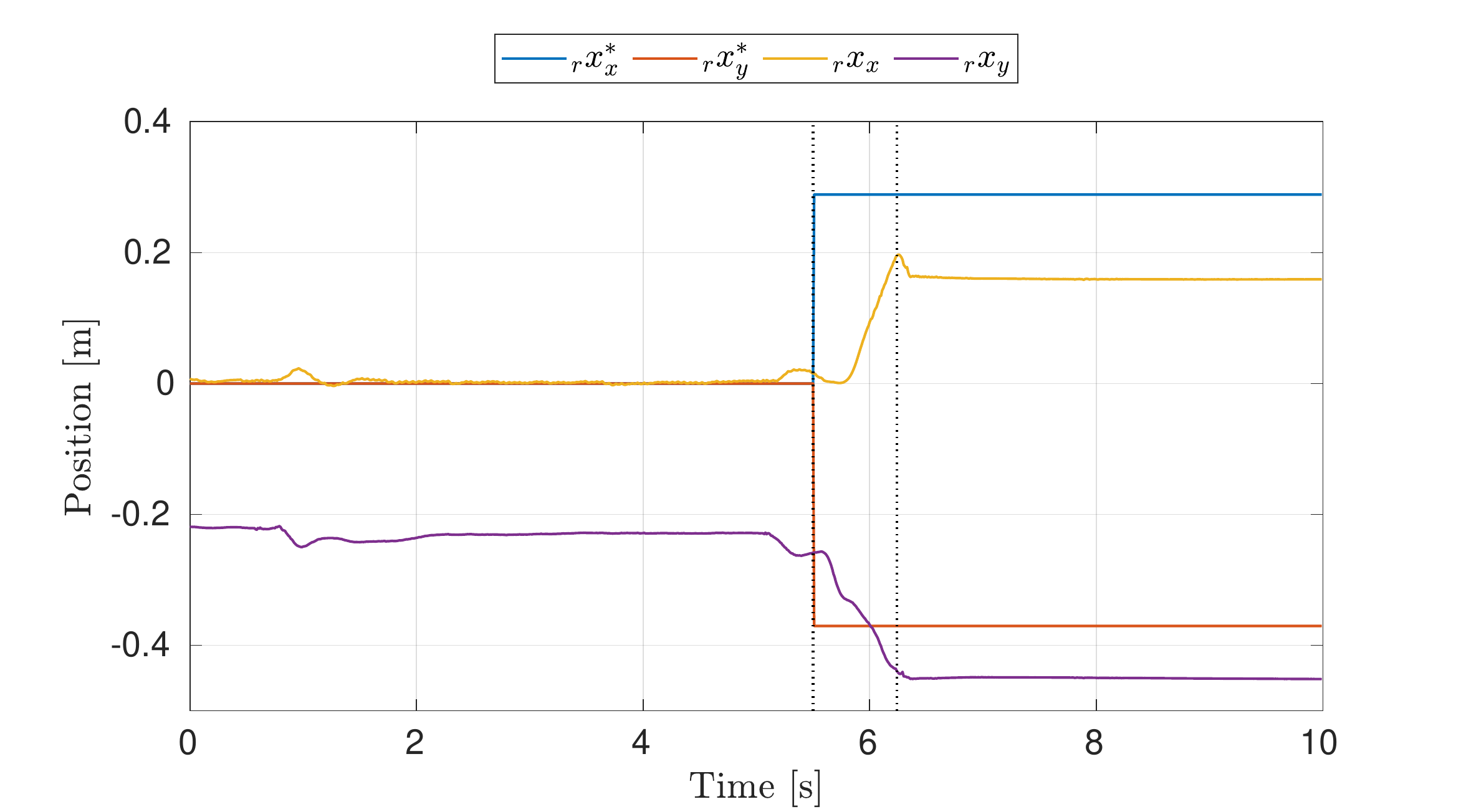}\label{fig:r_sole_front}}
	
	\caption{(cont.) Tracking of the desired right foot position provided by the momentum controller introduced in Sec. \ref{sec:momentum}. The vertical dotted lines have the same meaning of Fig. \ref{fig:exp_com}, namely they represent the beginning and the end of the step. The reference position is computed with the method presented in Sec. \ref{sec:sr_references}. It is computed when the step starts and it is tracked only after this moment. The tracking is satisfactory in \protect\subref{fig:r_sole_back}, while it presents a large error on the $x-$direction in \protect\subref{fig:r_sole_side}. In \protect\subref{fig:r_sole_front} there are non-negligible errors for both the $x-$ and $y-$ direction, probably because of a too far reference, that is hard to track during such dynamic motion.}
	\label{fig:r_sole_tracking}
\end{figure*}

\begin{figure}[tpb]
	\centering
	\subfloat[]{\includegraphics[width=.9\columnwidth]{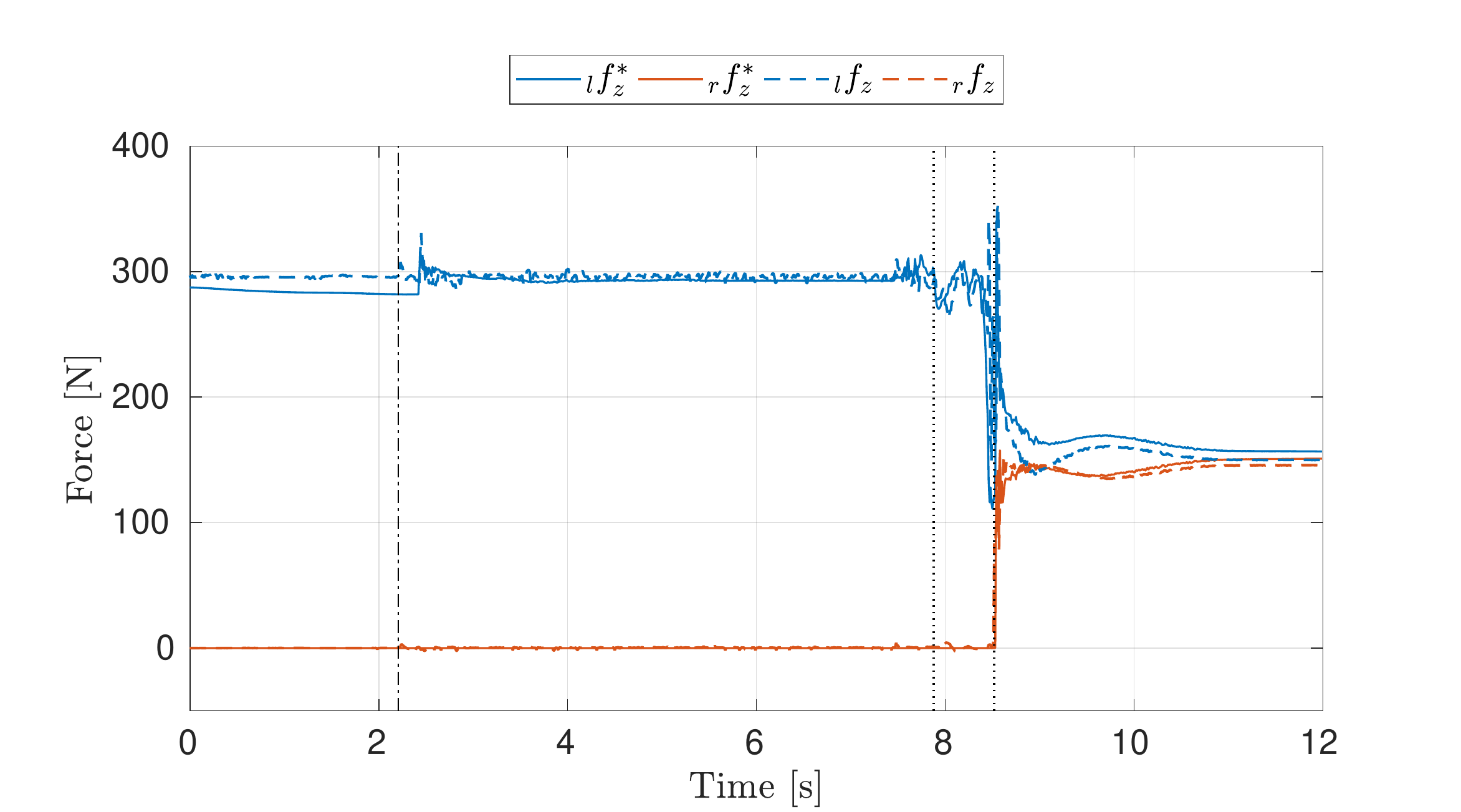} \label{fig:force}}
	
	\subfloat[]{\includegraphics[width=.9\columnwidth]{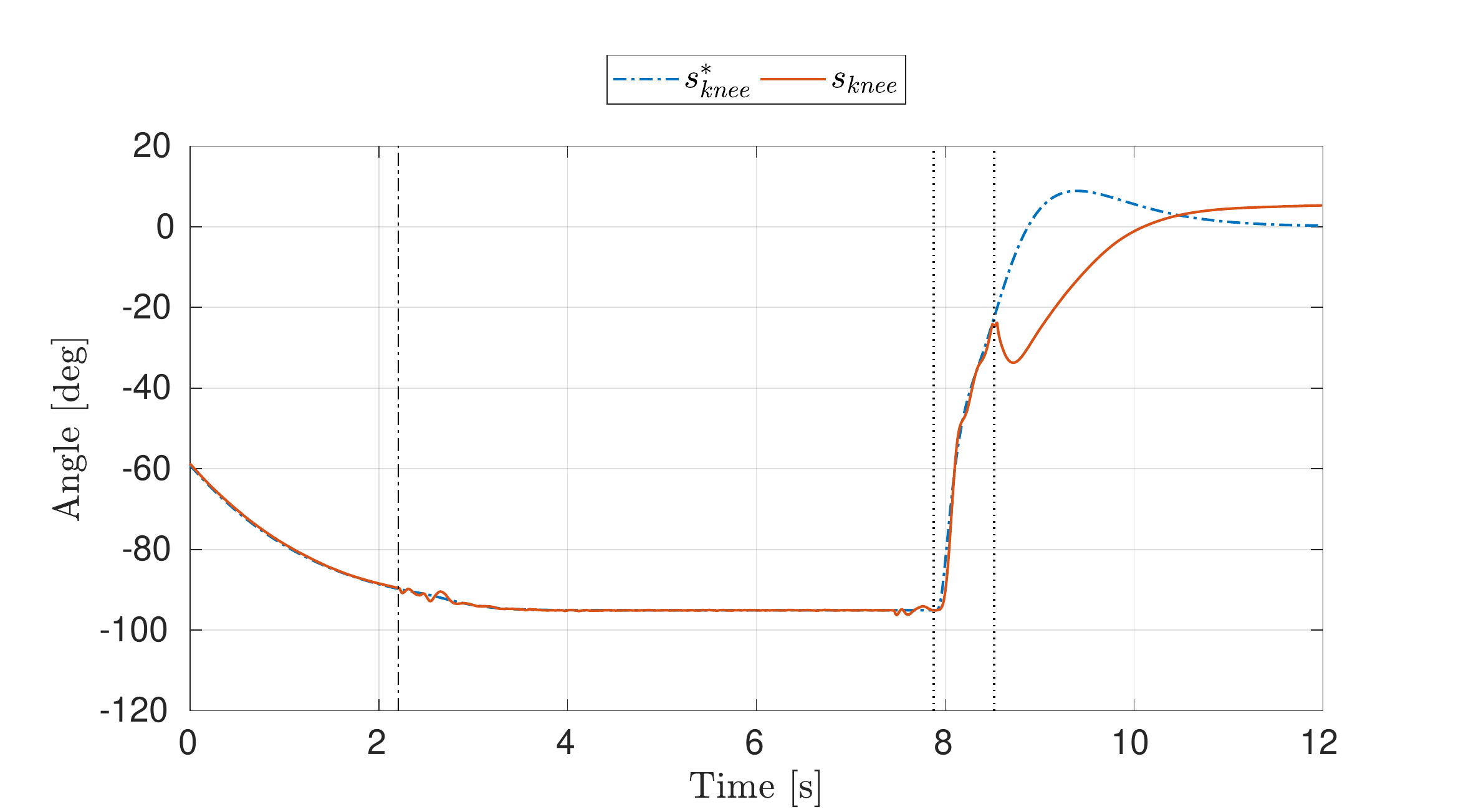}\label{fig:knee}}
	\caption{(a) Vertical components of the contact forces during the side-step experiment. Measured forces are plotted with dashed lines, while desired forces with solid lines. (b) Position tracking error of the right knee joint. The knee absorbs the impact with the ground, as it can be noticed by the peak in the error right after the impact.
	In both the plots, the first vertical dashed line shows a not-enough-strong push to violate the condition in Eq. \eqref{eq:step_condition}. The other two vertical dashed lines indicate the beginning and the end of the step.}	
\end{figure}

The presented MPC approach has been tested in the Gazebo simulator \citep{Koenig04} by using the iCub model.
The iCub humanoid robot, presented in Sec. \ref{sec:icub}, possesses 53 actuated joints, but only a total of 23 degrees of freedom (DoF) are used for balancing tasks. In fact, we do not consider those located in the eyes and in the hands.

Driven by the need of fast prototyping, the presented controller has been developed using the MATLAB\textsuperscript{\textregistered}/Simulink\textsuperscript{\textregistered}  environment.
For the presented tests, we use a machine running Ubuntu 16.04. The PC is equipped with a quad-core Intel\textsuperscript{\textregistered} Core i5@2.30GHz and 16GB of RAM. MOSEK\textsuperscript{\textregistered} is the selected solver, accessed through the MATLAB\textsuperscript{\textregistered} interface \texttt{CVX} \citep{cvx}.
In order to test the presented MPC scheme, we set a simple stepping scenario, where the robot, balancing on its left foot, uses the right foot to take a step.
This simple scenario allows us to test the performance of the proposed controller with a single contact activation. 

We present the results of different pushes, applied on the traverse plane, with an angle with respect to the lateral axis (pointing to the right of the robot), of $20\degree$ (Fig. \ref{fig:side}), $-20\degree$ (Fig. \ref{fig:back}) and $45\degree$ (Fig. \ref{fig:front}).

We choose a time step of $10\mathrm{ms}$, coincident with the rate of the whole-body torque controller, and a controller horizon of $N=25$, totaling $250\mathrm{ms}$.
We noticed that the chosen value of $N$ is sufficient to allow the effectiveness of the strategy when the robot is pushed from different directions. 
The push is nearly impulsive, applied on the chest with a magnitude of around $100\mathrm{N}$, which is about one third of the robot weight force. Fig. \ref{fig_sr:snapshots} presents a series of snapshots showing the robot while performing a large step.

Figure \ref{fig:exp_com} shows the CoM evolution for the three experiments, i.e. with different directions of pushes.
It can be noticed in both Figures \ref{fig:side} and \ref{fig:front} that two pushes occur. The first one does not violate the condition in Eq. \eqref{eq:step_condition}, thus it does not force the robot to take a step.
The second one, instead, triggers a change in the support foot configurations, and as a consequence, a new desired configuration for the CoM. The CoM height does not reach its initial value due to the kinematic limitations of the robot. Indeed, as it can be noticed from Figure \ref{fig_sr:snapshots}, legs are nearly fully extended.

The desired right foot positions for each experiment are depicted in Fig. \ref{fig:r_sole_tracking}, together with the corresponding estimated planar coordinates. As introduced in Sec. \ref{c6sec:cost}, the desired right foot position is turned into a joint reference by means of inverse kinematics. Such reference is given as postural task to the momentum controller introduce in Sec. \ref{sec:momentum}, which is considered a low priority objective. This is one of the main reasons behind the poor tracking performances. On the other hand, this also highlights the robustness properties of the presented architecture, which is able to maintain balance also in case of strong disturbances in the swing foot placement. Nevertheless, after the step is completed, the MPC strategy is fed with the measured foot position to be consistent with the new contact configuration. When the robot is pushed at an angle of $45\degree$, the reference is very far for the robot to be achieved. As it can be seen in Fig. \ref{fig:r_sole_front}, the foot does not reach its reference on the $x$-direction. Nevertheless, the controller is able to land the foot in a position which still allows it to maintain the balance.

Figure \ref{fig:force} represents the tracking of the desired vertical forces output by the MPC controller for the side-push experiment.
Remarkably, the normal force on the right foot appears to be tracked also across the step. 
Figure \ref{fig:knee} shows one of the benefits of torque control. 
The tracking of the joint position reference on the right knee undergoes a strong perturbation after the step. When hitting the ground, the intrinsic compliance introduced by torque control allows to absorb the impact, especially on this joint. This induce a peak of $30\degree$ of tracking error but the robot is still able to balance. 
In addition, torque control helps avoiding problems related to a not perfect placement of the swing foot before the impact. The compliance at the ankles helps absorbing the disturbances induced by an impact with a foot not perfectly parallel to the ground.


Summarizing, the presented controller allows the robot to recover from pushes of various intensity and directions, while remaining able to perform involved step movements. 

\section{Conclusions} \label{sec:conclusion}
The proposed controller adopts an approximated model of the robot linear and angular momentum dynamics in a predictive framework. It allows taking into account step movements by varying the structure of constraints and cost functions across the change of contact configuration. The uncertainty on its actual time instant is considered by a ``shift" in the prediction window. A heuristic is also employed as a condition for stepping. The contact wrenches are assumed to be control inputs and realized through a modified version of the iCub momentum-based whole-body torque controller. 

This approach avoids the definition of a CoM trajectory along the step, leaving such responsibility to the optimizer rather than to the designer. 
Currently, it takes almost $0.1$ seconds to solve an instance of the presented formulation. Unfortunately, this prevented the application on the real robot. In the following chapter, this problem is solved by using more simplified models. Nevertheless, the interest on developing a planner able to consider both the dynamics and the kinematics of the robot has not vanished, leading to the results of Part \ref{part:dynamic_planner}.

In particular, in the following, we focus on the generation and control of walking trajectories. Hence, the foot position tracking is fundamental and it requires the development of a new whole-body controller. Results have also shown the importance of generating references which can be kinetically achieved by the robot. This topic is analyzed in more details in Part \ref{part:dynamic_planner}.

\chapter{On-line Predictive Planning for Walking of the iCub Robot}
\label{chap:iros_walking}
We present here an architecture composed of three nested control loops. Sec. \ref{sec:iroswalking_background} introduces the background concepts exploited in the control architecture presented in Sec. \ref{sec:iroswalking_architecture}. The outer loop exploits a robot kinematic model to plan the footstep positions. In the mid layer, a predictive controller generates a CoM trajectory according to the well-known table-cart model. Through a whole-body inverse kinematics algorithm, we can define joint references for position controlled walking.
The outcomes of these two loops are then interpreted as inputs of a stack-of-task QP-based torque controller, which represents the inner loop of the presented control architecture. This resulting architecture allows the robot to walk also in torque control, guaranteeing higher level of \emph{compliance}. 
Real world experiments have been carried out on the humanoid robot iCub and they are shown in Sec. \ref{sec:iroswalking_results}. Finally, Sec. \ref{iroswalking_conclusions} concludes the chapter. 
\section{Background}\label{sec:iroswalking_background}
For a certain number of applications, the terrain can be considered flat.
In these cases, it is known that the human upper body is usually kept tangent to the \emph{walking path}~\citep{flavigne2010reactive,mombaur2010human} all the more so because stepping aside, i.e. perpendicular to the path, is energetically inefficient~\citep{handford2014sideways}. All these considerations suggest using a simple kinematic model to generate the walking trajectories: the unicycle model (see, e.g., \citep{PascalHandbook}). This model can be used to  plan footsteps in a corridor with turns and junctions using cameras~\citep{faragasso2013vision}, or to perform \emph{evasive} robot motions~\citep{cognetti2016real}. In all these cases, however,  the unicycle velocity is kept to a constant value.

\begin{figure}[tpb]
	\centering
	\def\svgwidth{0.85\columnwidth}
	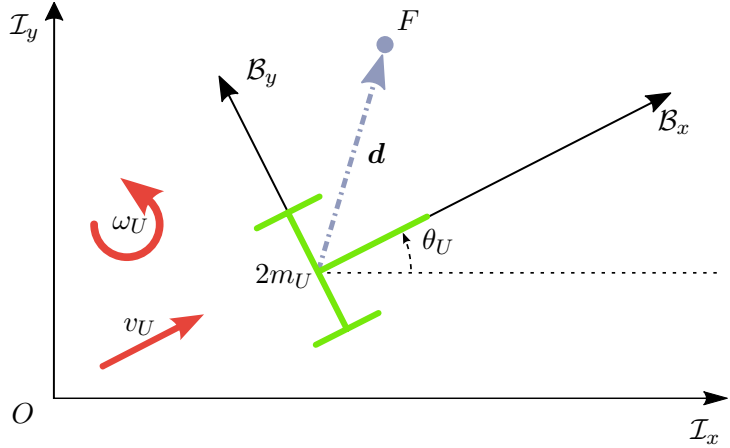
	\caption{Notation. The unicycle model is a planar model of a robot having two wheels placed at a distance $2m_U$, $m_U \in \mathbb{R}$ with a coinciding rotation axis.  Hence, this mobile robot cannot move sideways, i.e. along the wheel axis, but it can turn by moving the wheel at different velocity. $\mathcal{B}$ is a frame attached to the robot whose origin is located in the middle of the wheels axis. Point $F$ is attached to the robot. Its position expressed in $\mathcal{B}$ is given by $\bm{d}\in \mathbb{R}^2$, a constant vector.}
	\label{fig:notation}
\end{figure}

\subsection[The unicycle model]{Background on the unicycle model}
\label{sec:unicycleModel}
The unicycle model, represented in Fig. \ref{fig:notation} is described by the following model equations \citep{pucci2013nonlinear}:
\begin{IEEEeqnarray}{RCL}
	\IEEEyesnumber \phantomsection \label{unicycleDynamics}
	\dot{\bm{x}}_U & = & v_U R_2(\theta_U) \bm{e}_1, \IEEEyessubnumber \label{honolonimicCon} \\
	\dot{\theta}_U & = & \omega_U,       \IEEEyessubnumber
\end{IEEEeqnarray}
with $v_U \in \mathbb{R}$ and $\omega_U \in \mathbb{R}$ the robot's rolling and rotational speed, respectively. $\bm{x}_U \in \mathbb{R}^2$ is the unicycle position in the $xy$ plane of the inertial frame $\mathcal{I}$. $\theta_U \in \mathbb{R}$ represents the angle around the $z-$axis of $\mathcal{I}$ which aligns the inertial reference frame with a unicycle fixed frame.
The variables $v_U$ and $\omega_U$ are considered 
as kinematic control inputs.

A reasonable control objective for this kind of model is to asymptotically stabilize the point $F$ about a desired point $F^*$ whose position is defined as $\bm{x}_F^*$. For this purpose, we define the error $\tilde{\bm{x}}$ as 
\begin{equation}
\tilde{\bm{x}}  := \bm{x}_F - \bm{x}_F^*, \label{eq:xtilde}
\end{equation}
so that the control objective is equivalent to the asymptotic stabilization of $\tilde{\bm{x}}$ to zero.
Since we define $\bm{x}_F$ as 
\begin{equation*}
	\bm{x}_F = \bm{x}_U + \bm{R}_2(\theta_U)\bm{d},
\end{equation*}
then by differentiation it yields the dynamics
\begin{equation}
\label{errorDynamics}
	\dot{\bm{x}}_F = \dot{\bm{x}}_U + \omega_U \bm{R}_2(\theta_U) \bm{S}_2 \bm{d}. 
\end{equation}

Eq. \eqref{errorDynamics} describes the output dynamics. Substituting Eq. \eqref{unicycleDynamics} into Eq. \eqref{errorDynamics}, we can rewrite the output dynamics as:
\begin{IEEEeqnarray}{RCL}
	\IEEEyesnumber
\dot{\bm{x}}_F = \bm{B}_U(\theta_U)\bm{c}_U &=& \begin{bmatrix}\bm{R}_2(\theta_U) \bm{e}_1 &  \bm{R}_2(\theta_U) \bm{S}_2 \bm{d}\end{bmatrix} \bm{c}_U \IEEEyessubnumber\\
&=& \bm{R}_2(\theta_U) \begin{bmatrix}1 &  -d_2 \\0 & \hphantom{-}d_1 \end{bmatrix} \bm{c}_U \IEEEyessubnumber
\end{IEEEeqnarray}
where $\bm{c}_U = \left[v_U ~~ \omega_U\right]^\top$ is the vector of control inputs.
It can be noticed that $\det\left[ \bm{B}_U(\theta_U)\right] = d_1$, which means that when the control point $F$ is not located on the wheels' axis, its stabilization to an arbitrary reference position $\bm{x}_F^*$ can be achieved by the use of simple feedback laws. For example, if we define the following control law
\begin{equation}
\bm{c}_U = \bm{B}_U(\theta_U)^{-1}(\dot{\bm{x}}^*_F - \bm{K}_U\tilde{\bm{x}})
\end{equation}  
with $\bm{K}_U$ a positive definite matrix, then we have our error dynamics
\begin{equation} 
\dot{\tilde{\bm{x}}} = -\bm{K}\tilde{\bm{x}}.
\end{equation}
Thus, the origin of the error dynamics is an asymptotically stable equilibrium.
Notice that this control law is not defined when $d_1=0$.

\subsection[Zero moment point preview control]{Background on ZMP preview control}
\label{sec:zmppreview}
The MPC controller implemented in this work has been derived from the Zero Moment Point (ZMP) preview control described in \citep{Kajita2003}, here introduced.
This algorithm adopts a simplified model, i.e the Linear Inverted Pendulum (LIP) model, introduced in Sec. \ref{sec:lip}. In particular, we adopt a slightly modified version, where the ${}_p\bm{x}$ is to the ZMP position \citep{vukobratovic2004zero}. Nevertheless, the dynamic equations remain the same.

The ZMP can be related to the CoM projection, called $\bm{x}_{LIP}$ as in Sec. \ref{sec:lip}, through the following equation:
\begin{equation}
\label{eq:cop}
\bm{x}_{\text{ZMP}} = \bm{x}_{LIP} - \omega^2\ddot{\bm{x}}_{LIP}.
\end{equation}

Assuming we have a desired ZMP trajectory $\bm{x}_\text{ZMP}^*(t)$, we want to track this signal at every time instant.
One possibility is to consider the ZMP as an output of the following dynamical system:
\begin{equation}
\label{eq:cart_table_dyn_ss}
\begin{split}
\dot{\bm{\chi}} &= \bm{A}_{\text{ZMP}}\,\bm{\chi} + \bm{B}_{\text{ZMP}}\,\bm{u}, \\
 \bm{y} &= \bm{C}_{\text{ZMP}}\,\bm{\chi}
\end{split}
\end{equation}
where the new state variable $\bm{\chi}$ and control $\bm{u}$ are defined as
\begin{equation}
\label{eq:cart_table_dyn_ss_statedef}
\bm{\chi} := \begin{bmatrix}
\bm{x}_{LIP} \\
\dot{\bm{x}}_{LIP} \\
\ddot{\bm{x}}_{LIP}
\end{bmatrix} \in \mathbb{R}^6, \quad
\bm{u} := \begin{bmatrix}
\dddot{\bm{x}}_{LIP}
\end{bmatrix} \in \mathbb{R}^2.
\end{equation}
The system matrices are defined as in the following:
\begin{equation}
\label{eq:cart_table_dyn_ss_matrices}
\begin{aligned}
\bm{A}_{\text{ZMP}} & = \begin{bmatrix}
\bm{0}_{2\times 2} & \mathds{1}_2 & \bm{0}_{2\times 2}\\
\bm{0}_{2\times 2} & \bm{0}_{2\times 2} & \mathds{1}_2\\
\bm{0}_{2\times 2} & \bm{0}_{2\times 2} & \bm{0}_{2\times 2}
\end{bmatrix}, \quad
\bm{B}_{\text{ZMP}} = \begin{bmatrix}
\bm{0}_{4\times 2} \\
\mathds{1}_2
\end{bmatrix} \\
\bm{C}_{\text{ZMP}}& = \begin{bmatrix}
\mathds{1}_2 & \bm{0}_{2\times 2} & -\omega^2\mathds{1}_2
\end{bmatrix}.
\end{aligned}
\end{equation}

We define a cost function $\mathcal{J}$ as follows:
\begin{equation}
\label{eq:cost_linear}
\mathcal{J} = \int_0^{T} \norm{\bm{x}_\text{ZMP}^*- \bm{C}_{\text{ZMP}}\,\bm{\chi}}_{\bm{Q}}^2  + \norm{\bm{u}}^2_{\bm{R}} \dif \tau .
\end{equation}
$\mathcal{J}$ penalizes the output tracking error plus a regularization term on the effort. The system described by Eq. \eqref{eq:cart_table_dyn_ss} is linear, while the cost $\mathcal{J}$ is quadratic. This formulation provides the basis for the constrained MPC controller we present in Sec. \ref{sec:zmp_mpc_constrained}.

\section{Control architecture}\label{sec:iroswalking_architecture}
In this section we summarise the components constituting the presented architecture, namely:
%
%
\begin{itemize}
	\item the footstep planner,
	\item the MPC controller,
	\item the stack-of-tasks balancing controller.
\end{itemize}

These components share a lot of commonalities with other state of the art approaches. Nevertheless, this section presents how these three components interconnect to define a walking architecture. In particular, we focus on all the aspects necessary for a simple walking task, starting from the user inputs, up to the references to the low-level robot controller, while dealing with its balance. Such architecture defines the main contribution of this chapter. 

\subsection{The footstep planner}
\label{sec:footstepPlanner}
\begin{figure}[tpb]
	\centering
	\def\svgwidth{0.7\columnwidth}
	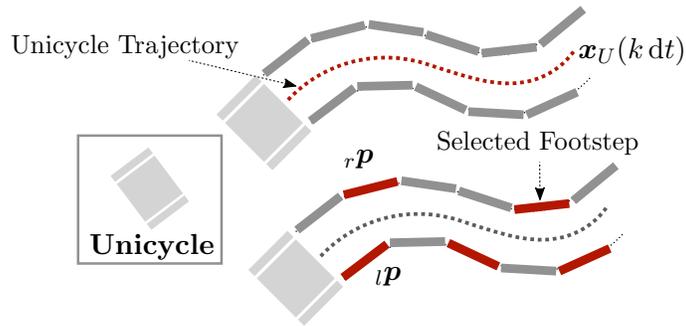
	\caption{Representation of the sampling algorithm. The unicycle trajectory is initially discretized to obtain $\bm{x}_U(k\dif t)$ (red dots), then the foot poses are sampled, obtaining ${}_l\bm{p}$ and ${}_r\bm{p}$.}
	\label{fig:sampler}
\end{figure}
Consider the unicycle model presented in Section \ref{sec:unicycleModel}. Assuming we know the reference trajectory for point $F$ up to time $T$, we can integrate the closed-loop system described in Section \ref{sec:unicycleModel} to obtain the trajectory spanned by the unicycle. The next step consists in discretizing the unicycle trajectories at a fixed rate $1/\mathrm{d}t$. This passage allows us to search for the best foot placement in a smaller space, constituted by the set of points obtained by discretization from the original unicycle trajectories. 

To determine the foot placements, we assume them to be coincident with the corresponding unicycle's wheel, as shown in Fig. \ref{fig:sampler}. Hence, the desired planar position of the left and right foot\footnote{With an abuse of notation, here we assume ${}_l\bm{p}, {}_r\bm{p} \in \mathbb{R}^2$ since we are just interested in the planar position.}, ${}_l\bm{p}$ and ${}_r\bm{p}$, can be obtained as:
\begin{IEEEeqnarray}{C}
	\IEEEyesnumber \phantomsection \label{sampledPositions}
		{}_l\bm{p} = \bm{x}_U + \bm{R}_2(\theta_U)\begin{bmatrix} 0 \\ m_U \end{bmatrix}, \IEEEyessubnumber\\
		 {}_r\bm{p} = \bm{x}_U + \bm{R}_2(\theta_U)\begin{bmatrix} 0 \\ -m_U \end{bmatrix}, \IEEEyessubnumber
\end{IEEEeqnarray}
where $m_U$ is the distance of a wheel from the center of the unicycle, see Fig. \ref{fig:notation}.
The foot orientations in the $xy$ plane, ${}_l\theta$ and ${}_r\theta$, coincide with $\theta_U$.

A step contains two phases: double support and single support. During double support, both robot feet are on the ground and, depending on the foot, we can distinguish two different states: \textit{switch-in} if the foot is being loaded, \textit{switch-out} otherwise. Instead, in single support, a foot is in a \textit{swing} state if it is moving, \textit{stance} otherwise. The instant in which a foot lands on the ground is called \textit{impact time}, $t_\text{impact}^f$ ($f$ is a placeholder for either $l$ or $r$). After this event, the foot will experience the following sequence of states: 
$
\textit{switch-in} \rightarrow \textit{stance} \rightarrow \textit{switch-out} \rightarrow \textit{swing}.
$
This sequence is terminated by another \textit{impact time}. At the beginning of the \textit{switch-out} phase, the \emph{other} foot has landed on the ground with an impact time $t_\text{impact}^{\sim f}$. The step duration, $\Delta_t$ is then defined as:
\begin{equation*}
\Delta_t = t_\text{impact}^{\sim f} - t_\text{impact}^f.
\end{equation*}
We define additional quantities relating the two feet when in double support (this phase is indicated with the $ds$ subscript):
\begin{itemize}
	\item The orientation difference $\Delta_\theta = \norm{{}_l\theta_{ds} - {}_r\theta_{ds}}$;
	\item The feet distance $\Delta_p = \|{}_l\bm{p}_{ds} - {}_r\bm{p}_{ds}\|$;
	\item The position of the left foot expressed on the frame rigidly attached to the right foot, ${}^r\bm{o}_l$.
\end{itemize}
These quantities will be used to determine the foot trajectories starting from the unicycle ones.

Given the discretized unicycle trajectories, a possible policy to define footsteps consists in fixing the duration of a step $\Delta_t$, or in fixing its length, $\Delta_p$. Both these two strategies are not desirable. In the former case the robot take always steps which may be too short at maximum speed. In the latter, if the unicycle is advancing slowly (because of slow moving references), the robot will take steps always at maximum length sacrificing the walking speed. In view of these considerations, we would like the planner to modify step length and speed depending on the reference trajectory. To avoid fixing any variable, it is necessary to define a cost function.
Our choice is composed of two parts. The first part weights the squared inverse of $\Delta_t$, thus penalizing fast steps, while the second penalize the squared 2-norm of $\Delta_p$, avoiding to take long steps. Thus, the cost function $\mathcal{J}$ can be written as:
\begin{equation} \label{cost_fcn}
\mathcal{J} = k_t \frac{1}{\Delta_t^2} + k_p \|\Delta_p\|^2,
\end{equation}
where $k_t$ and $k_p$ are two positive numbers. Depending on their ratio, the robot will take either long-and-slow or short-and-fast steps.
Notice that $\mathcal{J}$ is not defined when $\Delta_t = 0$, but the robot would not be able to take steps so fast. Thus, we need to bound $\Delta_t$:
\begin{equation}
	t_{min} \leq \Delta_t \leq t_{max},
\end{equation}
where the upper bound avoids a step to be too slow.

$\Delta_p$ needs to be lower than an upper-bound $d_{max}$, bounding the swinging foot into a circular area drawn around the stance foot.
Another constraint to be considered is the relative rotation of the two feet. In particular the absolute value of $\Delta_\theta$ must be lower than $\theta_{max}$.

Finally, to avoid the robot twisting its legs, we simply resort to a bound on the $y-$component of ${}^r\bm{o}_l$ vector. By applying a lower-bound ($w_{min}$) on it, we avoid the left foot to be on the right of the other foot.

Finally we can write the constraints defined above, together with $\mathcal{J}$, as an optimization problem:
\begin{IEEEeqnarray}{LRCCCL}	
	\IEEEyesnumber \phantomsection \label{unicycle_optimization}
	\minimize_{t_{impact}}    & \IEEEeqnarraymulticol{5}{C}{ K_t \frac{1}{\Delta_t^2}  +  K_p \|\Delta_p\|^2 }\nonumber\\
	\text{subject to}: \\
	& t_{min} & \leq & \Delta_t &\leq& t_{max}, \nonumber \\
	&&& \Delta_p &\leq & d_{max}, \nonumber\\
	&&& \|\Delta_\theta\| &\leq&  \theta_{max}, \nonumber \\
	& w_{min} &\leq& {}^r\bm{o}_{l,y}.&& \nonumber
\end{IEEEeqnarray} 
The decision variables are the \textit{impact times}. If we select an instant $k$ to be the \textit{impact time} for a foot, then we can obtain the corresponding foot pose, depending on the wheel position, see Eq. \eqref{sampledPositions}, and the corresponding unicycle orientation. 
The foot will keep this configuration until the beginning of the \textit{swing} phase, where it moves to a new configuration. 

The optimization problem solution is obtained through a simple iterative algorithm. The initialization can be done by using the measured position of the feet. Starting from this configuration we can integrate the unicycle trajectories assuming to know the reference trajectories. Once we discretize them, we can iterate over the discrete instant $k$ until we find a set of points which satisfy the conditions defined above. Among the remaining points we can evaluate $\mathcal{J}$ to find our optimum. This point corresponds to a new footstep, providing a new initialization for the iterative algorithm. An outer loop repeats this algorithm to obtain a series of footsteps.

The algorithm described here is flexible enough to include some features which are useful when commanding a humanoid robot. For example, it is possible to avoid steps which would be too small, pausing the motion until a larger step can be made. Additionally, we can impose the robot to stop always in double support.

Once footsteps are planned, we can proceed in interpolating the foot trajectories and other relevant quantities for the walking motion, e.g. the Zero Moment Point \citep{vukobratovic2004zero}. This passage is fundamental to obtain trajectories that can be tracked by the following control layers.

\begin{figure}[tpb]
	\centering
	\def\svgwidth{0.9\columnwidth}
	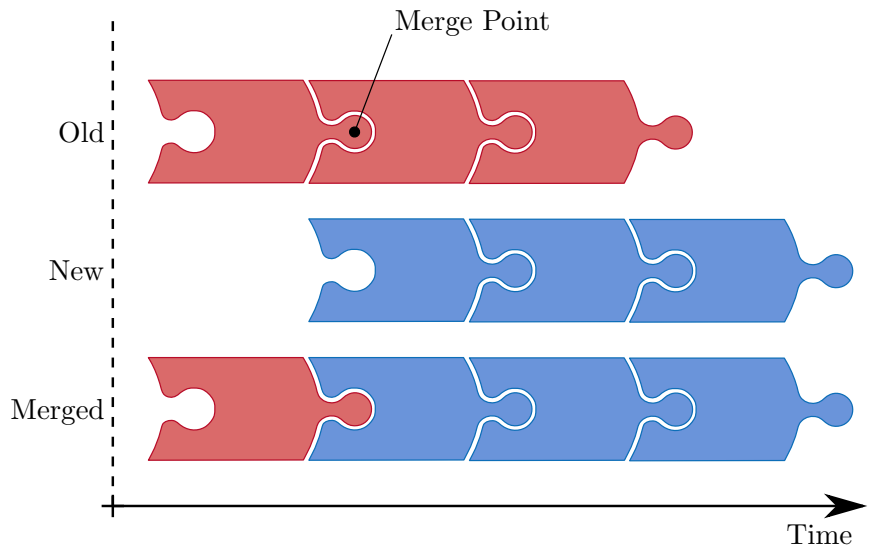
	\caption{Schematic representation of the merge point mechanism. A merge point represents the moment where it is possible to interconnect a new trajectory discarding the remaining part of the old trajectory. In particular, this correspond to the mid point of double support phases. }
	\label{fig:merge_mechanism}
\end{figure}

In addition, while walking, references may change and it is not desirable to wait the end of the planned trajectories for updating them. Thus, the idea is to \emph{merge} two trajectories instead of serializing them. When generating a new trajectory, the robot is supposed to be standing on two feet and the first \textit{switch} time needs to last half of its nominal duration. In view of these considerations, the middle instant of the double support phase is a suitable point to merge two trajectories. This choice eases the merging process since the feet are not supposed to move in that instant. Hence, the trajectories' initialization is trivial. Notice that there may be different \textit{merge points} along the trajectories, depending on the number of (full) double support phases. Two trivial \textit{merge points} are the very first instant (full replacement of trajectories) and the final point (serialization case). From an intuitive point of view, this process may be seen as the interlocking of two jigsaw puzzles, as in Fig. \ref{fig:merge_mechanism}. Each step represents a piece and different trajectories can be merged by interlocking them in the merge point, corresponding to the mid point of a double support phase.

This method is similar to the \emph{Receding Horizon Principle} described in Sec. \ref{sec:receding_horizon}. In fact, we plan trajectories for a horizon $T$ but only a portion of them is actually used, namely the first generated step. This simple strategy allows us to change the unicycle reference trajectory on-line (through the use of a joystick for example), directly affecting the robot motion. In addition, we could correct the new trajectories with the \emph{actual} position of the feet. This is particularly suitable when the foot placement is not perfect, as in torque-controlled walking.

\subsection{The model predictive controller}\label{sec:zmp_mpc_constrained}
The receding horizon controller used in our architecture inherits from the basic formulation described in Sec. \ref{sec:zmppreview}. Nevertheless, differently from \citep{Kajita2003, wieber2006trajectory, diedam2008online}, we have added time-varying constraints to bound the ZMP inside the convex hull, both during single and double support phases. In this way we increase the robustness of the controller. 
These constraints are modeled as linear inequalities acting on the state variables, i.e.
\begin{align}
\label{eq:constraints_linear}
\bm{Z}(t) \bm{\chi}  \leq \bm{z}(t).
\end{align}
The constraint matrix $\bm{Z}$ and bound $\bm{z}$ depend on time. In fact, we can exploit the knowledge on the desired foot positions in time to constrain the ZMP throughout the \emph{whole} horizon, while retaining linearity. In other words, we consider the convex hull as depending only on time, hence changing the constraints accordingly. This approach is similar to what presented in Chapter \ref{chap:steprecovery}. In fact, we exploit the controller prediction capabilities to take into account changes in the support polygon before they are happening.

To summarize, the full optimal control problem can be represented by the following minimization problem:
\begin{IEEEeqnarray}{RLL}
\IEEEyesnumber \phantomsection\label{eq:mpc_linear}
\minimize_{\bm{\chi} ,\bm{u}} & \IEEEeqnarraymulticol{2}{l}{\int_0^{T} \norm{\bm{x}_\text{ZMP}^*- \bm{C}_{\text{ZMP}} \bm{\chi}}_{\bm{Q}}^2  + \norm{\bm{u}}^2_{\bm{R}} \dif \tau} \nonumber\\
\text{subject to:} \\
&~ \dot{\bm{\chi}} = \bm{A}_{\text{ZMP}}\,\bm{\chi} + \bm{B}_{\text{ZMP}}\,\bm{u},  \nonumber\\
&~ \bm{Z}(t) \bm{\chi}  \leq \bm{z}(t), &\forall t \in [0, T),\nonumber\\
&~ \bm{\chi}(0) = \bm{\chi}_0.\nonumber
\end{IEEEeqnarray}

The problem in Eq. \eqref{eq:mpc_linear} is solved at each control sampling time by means of a Direct Multiple Shooting method (as presented in Section \ref{sec:shooting}), thus differently from \citep{dimitrov2008implementation}. This choice has been driven by the necessity of formulating state constraints, i.e. Eq. \eqref{eq:constraints_linear}, throughout the whole horizon. 
We apply the Receding Horizon Principle described in Sec. \ref{sec:receding_horizon}, thus only the first output is used. Since the control input is the CoM jerk, we integrate it in order to obtain a desired CoM velocity and position. The latter is sent to the inverse kinematics (IK) library \citep[InverseKinematics sub-library]{libiDynTree}, together with the desired foot positions to retrieve the desired joint coordinates. Both the MPC an the IK modules use \texttt{Ipopt} \citep{IPOpt2006} to solve the corresponding optimization problems.

\subsection{The stack of tasks balancing controller}
\label{sec:sot}
The balancing controller stabilizes the center of mass position, the left and right foot positions and orientations by defining joint torques. The velocities associated to these tasks are stacked together into $\bm{\Upsilon}$:
\begin{equation}
\label{eq:tasks}
\bm{\Upsilon} = \begin{bmatrix}
\dot{\bm{x}}_{\text{CoM}}^\top & {}^l\bm{V}_{\mathcal{I}, l}^\top & {}^r\bm{V}_{\mathcal{I}, r}^\top
\end{bmatrix}^\top.
\end{equation}
${}^l\bm{V}_{\mathcal{I}, l}, {}^r\bm{V}_{\mathcal{I}, r} \in \mathbb{R}^6$ are the linear and angular foot velocities.

Letting $\bm{J}_\text{CoM}$, $\bm{J}_l$ and $\bm{J}_r$ denote respectively the Jacobians of the center of mass position, left and right foot configurations, $\bm{J}_{\Upsilon}$ can be defined as a stack of the Jacobians associated to each task:
\begin{equation}
\bm{J}_{\Upsilon} = 
\begin{bmatrix}
\bm{J}_{\text{CoM}}^\top & \bm{J}_{l}^\top & \bm{J}_{r}^\top
\end{bmatrix}^\top
\end{equation}

Furthermore, the task velocities $\bm{\Upsilon}$ can be computed from the robot state $\bm{\nu}$ using $\bm{\Upsilon} = \bm{J}_{\Upsilon} \bm{\nu}$. By deriving this expression, the task acceleration is
\begin{equation}
\label{eq:task_acceleration}
\dot{\bm{\Upsilon}} = \dot{\bm{J}}_{\Upsilon} \bm{\nu} + \bm{J}_{\Upsilon} \dot{\bm{\nu}}
\end{equation}

In view of~\eqref{eq:system} and~\eqref{eq:task_acceleration}, the task accelerations $\dot{\bm{\Upsilon}}$ can be formulated as a function of joint torques $\bm{\tau}_s$ and contact wrenches $\textbf{f}$, supposed to be control inputs:
\begin{equation}
\label{eq:task_acceleration_from_U}
\dot{\bm{\Upsilon}}(\bm{\tau}_s, \textbf{f}) = \dot{\bm{J}}_{\Upsilon} \bm{\nu} + \bm{J}_{\Upsilon} \bm{M}^{-1} \left(\begin{bmatrix}
\bm{0}_{6\times n} \\ \mathds{1}_n
\end{bmatrix}\bm{\tau}_s + \bm{J}_{\mathcal{C}}^\top \textbf{f} - \bm{C}\bm{\nu} - \bm{G}\right).
\end{equation}
In our formulation we want $\dot{\bm{\Upsilon}}(\bm{\tau}_s, \textbf{f})$ to track a specified task acceleration $\dot{\bm{\Upsilon}}^*$, namely:
\begin{equation}
\label{eq:task_constraint}
	\dot{\bm{\Upsilon}}(\bm{\tau}_s, \textbf{f}) = \dot{\bm{\Upsilon}}^*.
\end{equation} 
The reference accelerations $\dot{\bm{\Upsilon}}^*$ are computed using a simple PD control strategy for what concerns the linear terms. In parallel, the rotational terms are obtained by adopting a PD controller on the $SO(3)$ group, thus avoiding to use a parametrization like Euler angles. The implementation follows what is presented in \citep[Section 5.11.6, p.173]{olfati2001nonlinear}.

While the task defined in Eq. \eqref{eq:task_constraint} can be considered as \emph{high priority} tasks, we may conceive other possible objectives to shape the robot motion. One example regards the orientation of a particular frame. For instance, we would like to keep the torso in an upright posture. Similarly to what have been done for the other tasks, we exploit the frame Jacobian $\bm{J}_\text{frame}$. Since this task is considered as \emph{low priority}, we minimize the squared Euclidean distance of the frame acceleration from a desired value $\dot{\bm{\mathcal{T}}}^* \in \mathbb{R}^3$, i.e
\begin{equation}
\label{eq:torso_task}
	\minimize_{\bm{\tau}_s, \textbf{f}} \frac{1}{2}\norm{\dot{\bm{J}}_{\text{frame}} \bm{\nu} + \bm{J}_{\text{frame}} \dot{\bm{\nu}} - \dot{\bm{\mathcal{T}}}^*}^2.
\end{equation}
The reference acceleration $\dot{\bm{\mathcal{T}}}^*$ is obtained through a PD controller on $SO(3)$.

Finally, we also added another lower priority \emph{postural} task to track joint configurations. Similarly as before, the joint accelerations $\ddot{\bm{s}}(\tau,f_k)$, which depend upon the control inputs through the dynamics equations detailed in Eq. \eqref{eq:system}, are kept close to a reference $\ddot{s}^*$:
\begin{equation}
\label{eq:postural}
\minimize_{\bm{\tau}_s, \textbf{f}} \frac{1}{2}\norm{\ddot{\bm{s}}(\bm{\tau}_s, \textbf{f}) - \ddot{\bm{s}}^*}^2.
\end{equation}
The postural desired accelerations are defined through a \textit{PD+gravity compensation} control law, as in \citep{nava2016stability}.

Considering the contact wrenches $\textbf{f}$ as control inputs, we need to ensure their actual feasibility. To this end we apply the same constraints as in Sec. \ref{c6sec:constraints}. We can group Eq. \eqref{eq:task_constraint} and \eqref{c6eq:wrench_constr} in the following QP formulation:
\begin{IEEEeqnarray}{RRCL}
	\IEEEyesnumber \phantomsection	\label{eq:qp_balancing}
\minimize_{\bm{\tau}_s, \textbf{f}} &\IEEEeqnarraymulticol{3}{C}{\Gamma}\IEEEyessubnumber\\
	\text{subject to }: & \dot{\bm{\Upsilon}}(\bm{\tau}_s, \textbf{f}) &=& \dot{\bm{\Upsilon}}^* \IEEEyessubnumber \\[2pt]
		&\bm{A}_{st}\,\textbf{f} &\leq& {}_{st}\bm{b}, \IEEEyessubnumber \label{eq:contact_inequalities}
\end{IEEEeqnarray} 
where the cost function $\mathcal{J}$ is defined as:
\begin{equation*}
	\Gamma = \frac{1}{2}\left\|{\ddot{\bm{s}}(\bm{\tau}_s, \textbf{f}) - \ddot{\bm{s}}^*}\right\|^2 +  \frac{w_\mathcal{T}}{2}\left\|{\dot{\bm{J}}_{\text{frame}} \bm{\nu} + \bm{J}_{\text{frame}} \dot{\bm{\nu}} - \dot{\bm{\mathcal{T}}}^*}\right\|^2 +  \frac{w_\tau}{2}\|\bm{\tau}_s\|^2,
\end{equation*}
with $w_\mathcal{T}, w_\tau \in \mathbb{R}$ defining the relative priority of each task.
With respect to Eq.s \eqref{eq:torso_task} and \eqref{eq:postural}, an additional term is inserted, namely the 2-norm of the joint torques, which serves as a regularization. Among several feasible solutions, we are mostly interested in the one which requires the least effort. 

Let us focus on the foot contact wrenches ${}_l\textbf{f}$ and ${}_r\textbf{f}$. During the \textit{switch} phases we expect one of the two wrenches to vanish in order to smoothly deactivate the corresponding contact. In order to ease this process we added a cost term equal to
$
	\frac{w_f}{2}\left(\mathcal{F}_r\|{}_l\textbf{f}\| + \mathcal{F}_l\|{}_r\textbf{f}\| \right)
$
where $\mathcal{F}_r$ (respectively $\mathcal{F}_l$) is the normalized load that the right (respectively left) foot is supposed to carry. It is $1$ when the corresponding foot should hold the full weight of the robot, $0$ when unloaded. This information is provided by the planner described in Sec \ref{sec:footstepPlanner}. Hence, a high $\mathcal{F}_r$ would penalize the use of ${}_l\textbf{f}$, and vice-versa.

The QP problem of Eq. \eqref{eq:qp_balancing} is solved at a rate of 100Hz through qpOASES~\citep{Ferreau2014}. Even if this controller could run at a higher frequency, at the current stage 100Hz is the maximum rate at which we can set a torque reference to the robot.

To summarize, this QP controller adopts a mix of tasks with either weighted or strict priorities. Indeed, it is very similar to those presented in Sec. \ref{sec:whole-body-controllers}. During experiments, we tested different solutions, but this proved to adapt well with the preceding control layers and the low-level controller of the iCub humanoid robot.
\section{Validation and experimental results}\label{sec:iroswalking_results}
The presented architecture is composed by three different parts. In this section, we show three experiments whose aim is to test each part in an incremental way. All the experiments are performed on the iCub robot, controlling 23 Dofs either in position or in torque control. 

The snapshots presented in Fig.s \ref{fig:exp_straight} and \ref{fig:exp_unicycle} are obtained from the video included in the playlist: \url{https://www.youtube.com/playlist?list=PLBOchT-u69w9hJ6BmqPf06r0zWmungOrc}.
\subsection{Test of the predictive controller in position control}

\begin{figure}[tpb]
	\centering
	\def\svgwidth{\columnwidth}
	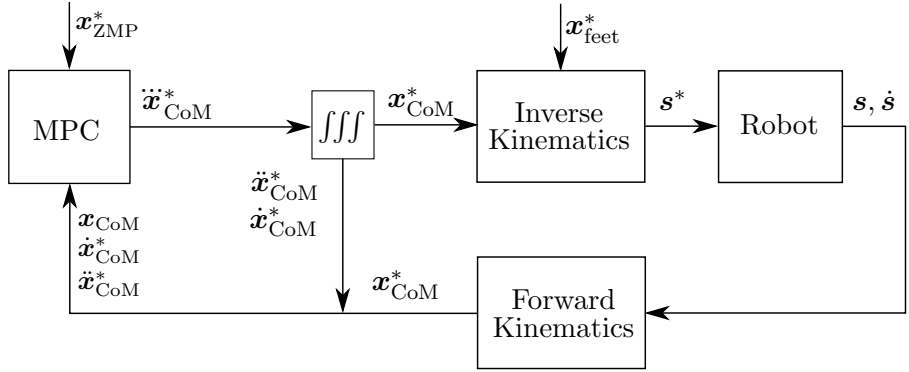
	\caption{A first skeleton of the architecture, composed only by the MPC (RH) controller connected to the inverse kinematics (IK). Their output is directly sent to the robot.} 
	\label{fig:mpc}
\end{figure}

\begin{figure}[tpb]
	\centering
	\subfloat[Tracking of the desired ZMP on the $xy$ plane.]{	\includegraphics[width=0.9\columnwidth]{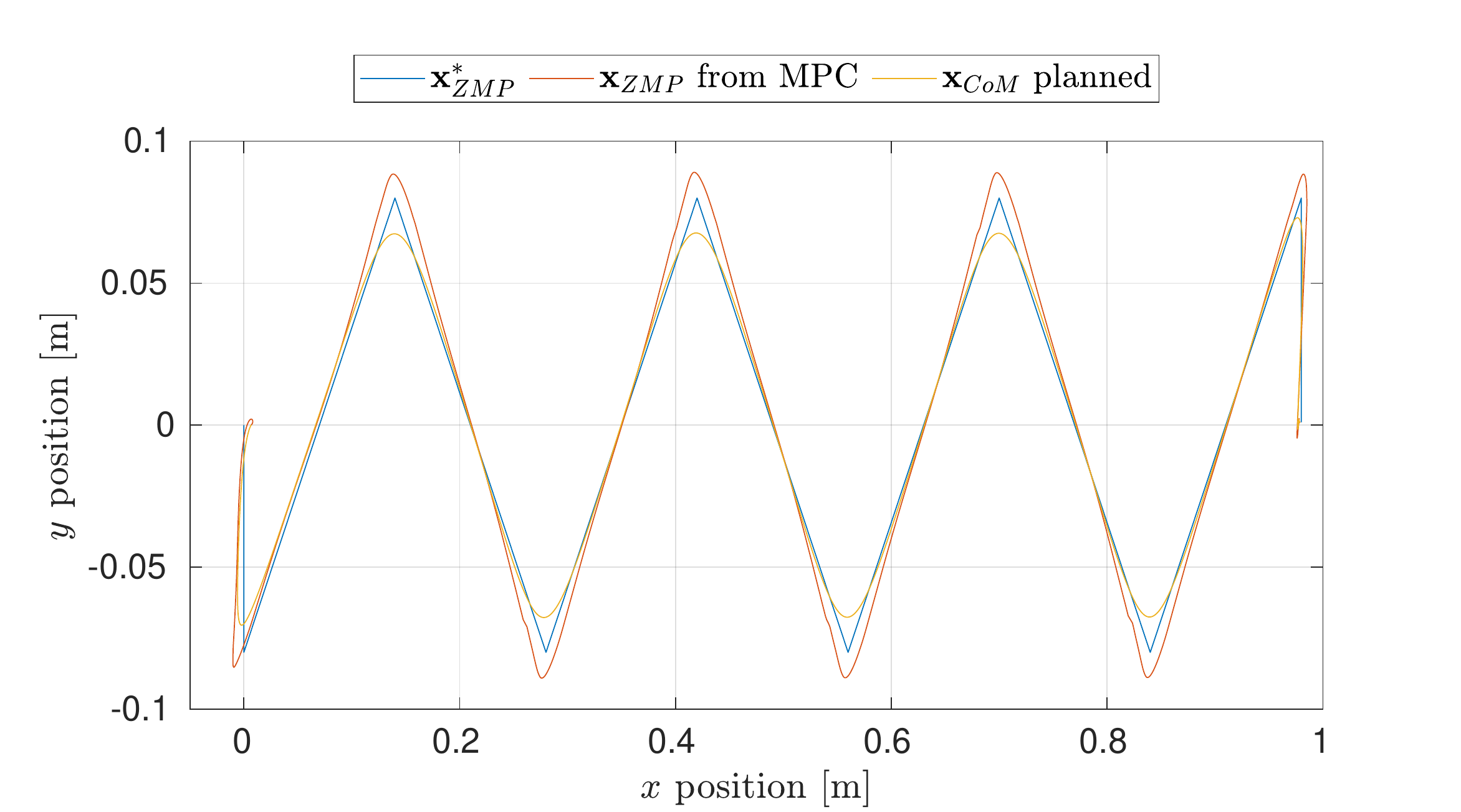}}
	
	\subfloat[Tracking of the desired ZMP along the $y$ direction in time.]{
		\includegraphics[width=0.9\columnwidth]{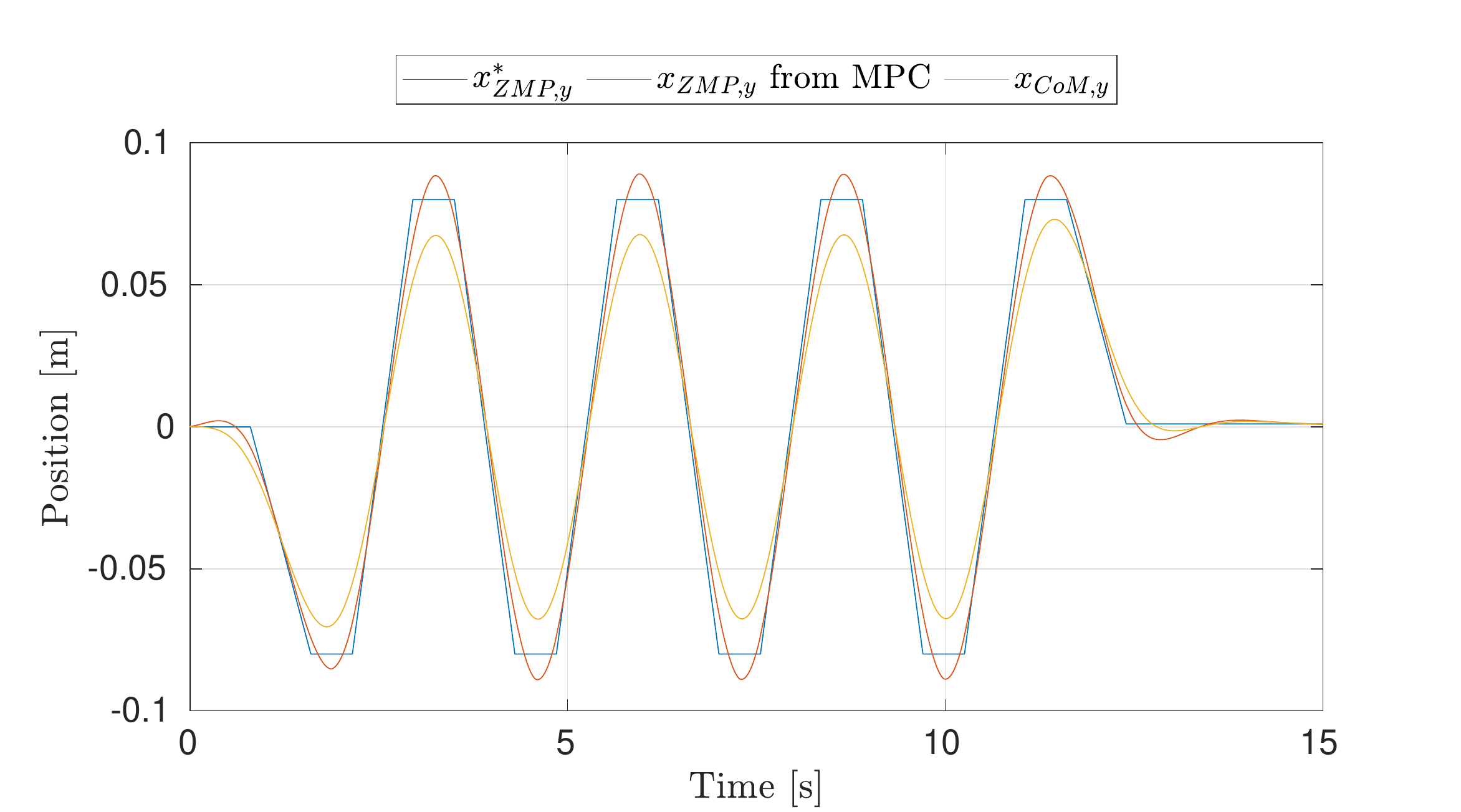}}
	\caption{Tracking of the desired ZMP trajectory by the MPC controller. It is possible to notice the overshoots in the $y$ direction. Constraints in ZMP avoid this overshoot to make the ZMP exiting the support polygon. }
	\label{fig:zmp_tracking}
\end{figure}

\begin{figure}[tpb]
	\centering
	\subfloat[$t=t_0$] {\includegraphics[width=.22\columnwidth]{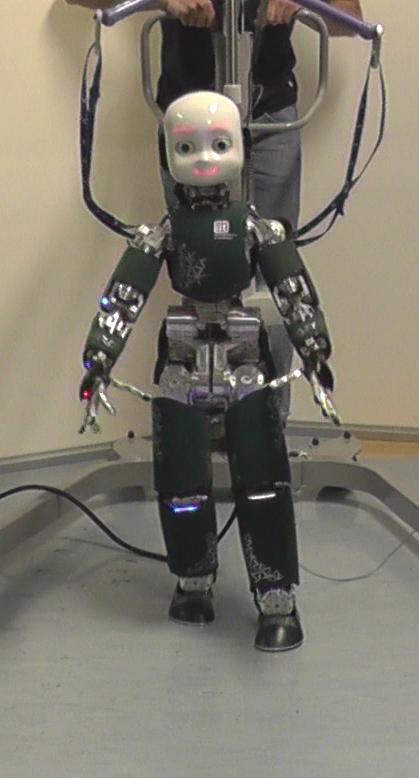}}
	\subfloat[$t=t_0+1s$] {\includegraphics[width=.22\columnwidth]{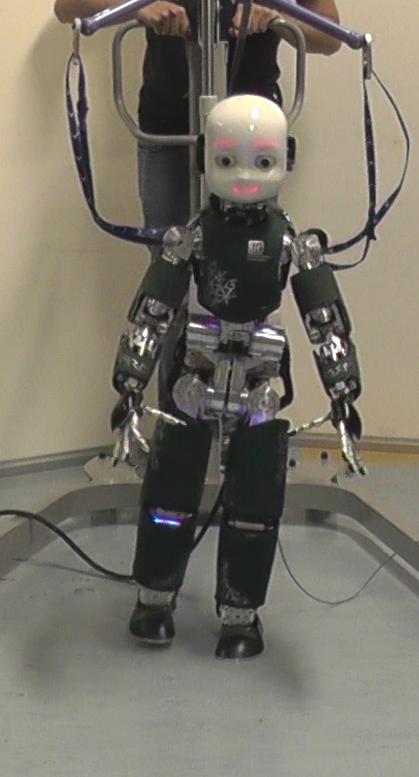}}
	\subfloat[$t=t_0+2s$] {\includegraphics[width=.22\columnwidth]{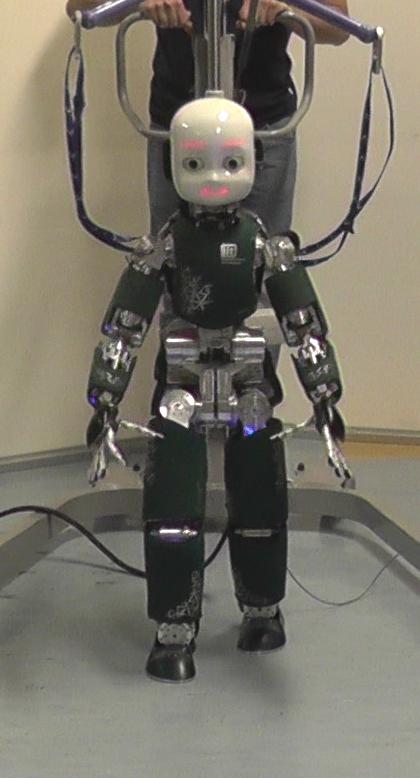}}
	\subfloat[$t=t_0+3s$] {\includegraphics[width=.22\columnwidth]{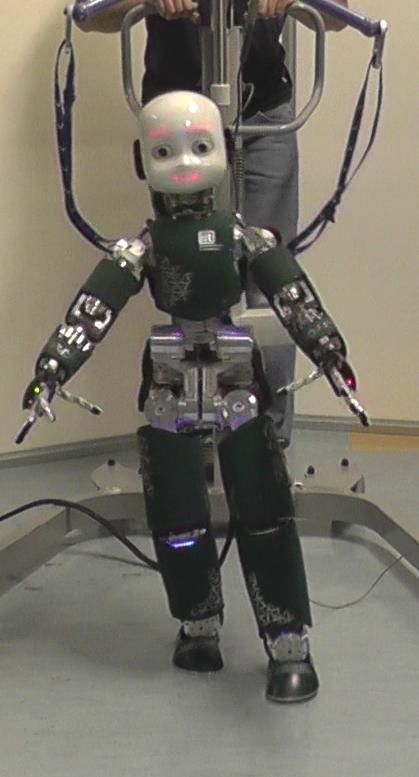}}
	\caption{Snapshots of the robot while walking straight in position control.}
	\label{fig:exp_straight}
\end{figure}

In the first experiment, we use the MPC module to control the robot joints in \emph{position control}. As depicted in Figure \ref{fig:mpc}, the control loop is closed on the CoM position only. In fact, we noticed that the noise introduced by joint velocity measurements can strongly affect the performance of the overall architecture.  The reference foot poses $\bm{x}_\text{feet}^*$ and the desired ZMP profile $\bm{x}_{\text{ZMP}}^*$ are obtained through an off-line planner \citep[Chapter 9]{hu2016walking}.

Both the MPC and the IK are running on the iCub head, which is equipped with a 4$^{th}$ generation Intel\textsuperscript{\textregistered} Core i7@1.7GHz and 8GB of RAM. The whole architecture takes in average less than 8$\mathrm{ms}$ for a control loop.

The robot is taking steps of $14\mathrm{cm}$ long in $1.35s$ (of which $0.8s$ in double support). Fig. \ref{fig:exp_straight} presents some snapshots of the robot while walking.

Fig. \ref{fig:zmp_tracking} presents the ZMP tracking performances of the MPC controller. The desired ZMP position is obtained as a set of straight lines connecting the feet's centers. It is possible to notice that the ZMP obtained from the MPC is smoother than the desired one. In addition, it presents some overshoots. The constraints applied on the ZMP position make sure that these overshoots are within the support polygon.

\subsection{Adding the unicycle planner}
\begin{figure}[tpb]
	\centering
	\def\svgwidth{\columnwidth}
	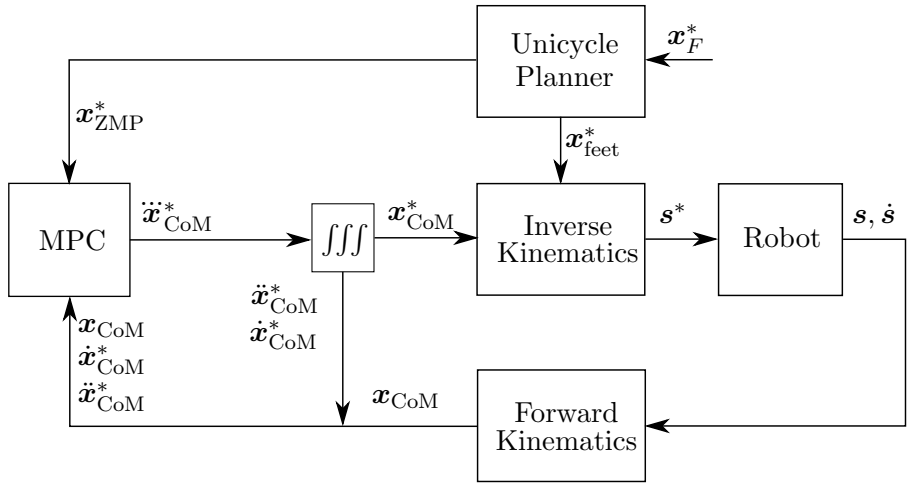
	\caption{The scheme of Figure \ref{fig:mpc} has been upgraded with the unicycle planner which is responsible of providing online references.}
	\label{fig:mpc+unicycle}
\end{figure}
\begin{figure}[tpb]
	\centering
	\subfloat[$t=t_0$] {\includegraphics[width=.22\columnwidth]{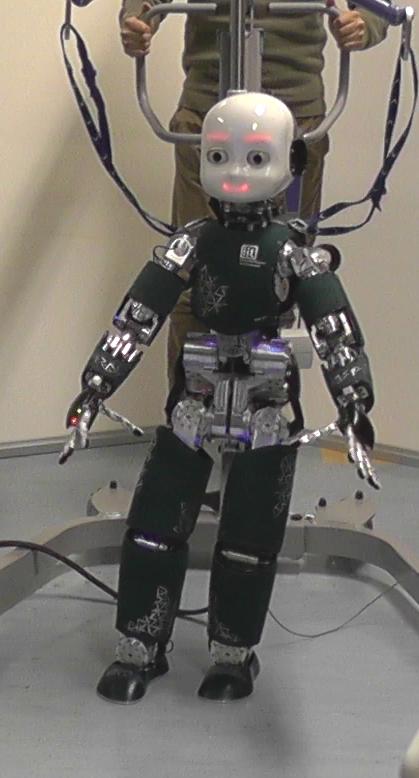}}
	\subfloat[$t=t_0+1s$] {\includegraphics[width=.22\columnwidth]{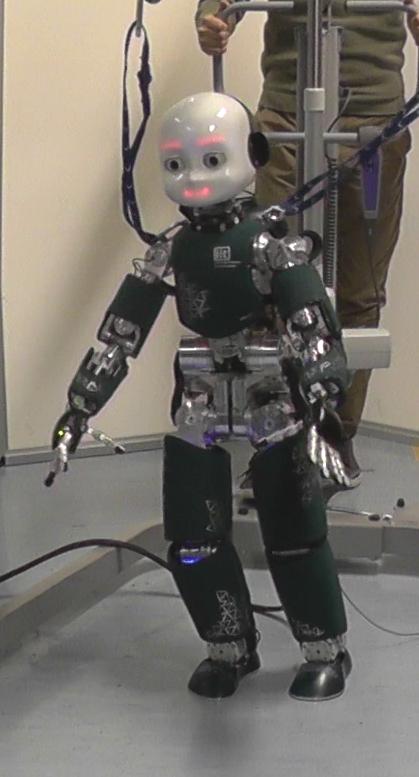}}
	\subfloat[$t=t_0+2s$] {\includegraphics[width=.22\columnwidth]{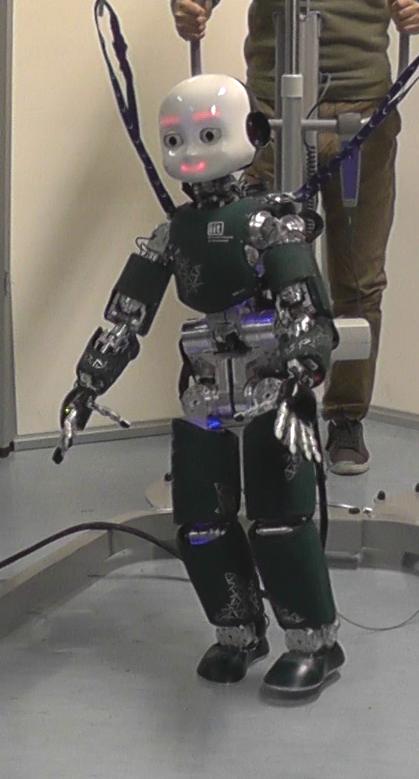}}
	\subfloat[$t=t_0+3s$] {\includegraphics[width=.22\columnwidth]{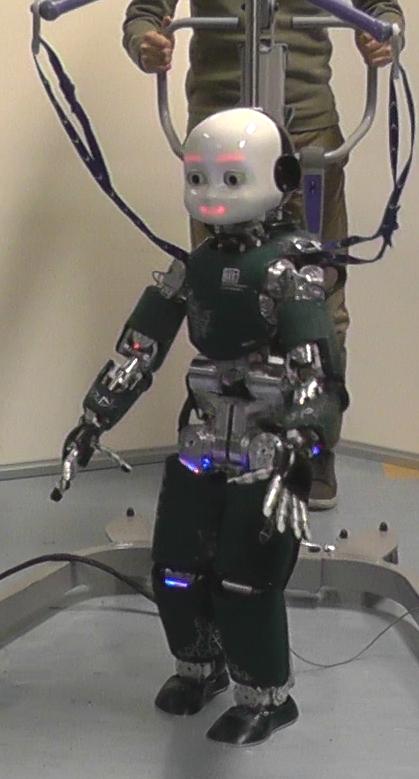}}
	\caption{With the unicycle planner, the robot advances while turning right.}
	\label{fig:exp_unicycle}
\end{figure}

Fig. \ref{fig:mpc+unicycle} shows that the controller has been improved by connecting the unicycle planner described in Sec. \ref{sec:footstepPlanner}. This allows us to adapt the robot walking direction in an \emph{online} fashion. As an example, by using a joystick we can drive a reference position $\bm{x}^*_F$ for the point $F$ attached to the unicycle. Depending on how much we press the joypad, this point moves further away from the robot, generating a faster walking motion.

Using the planner described in Sec. \ref{sec:footstepPlanner}, the step timings and dimensions are not fixed a priori. In this particular experiment, a single step could last between $1.3s$ and $5.0s$, while the swing foot can land in a radius of $0.175$m from the stance foot. Fig. \ref{fig:exp_unicycle} presents some snapshots of the robot while taking a right turn.

\subsection{Complete architecture}

\begin{figure}[tpb]
	\centering
	\def\svgwidth{\columnwidth}
	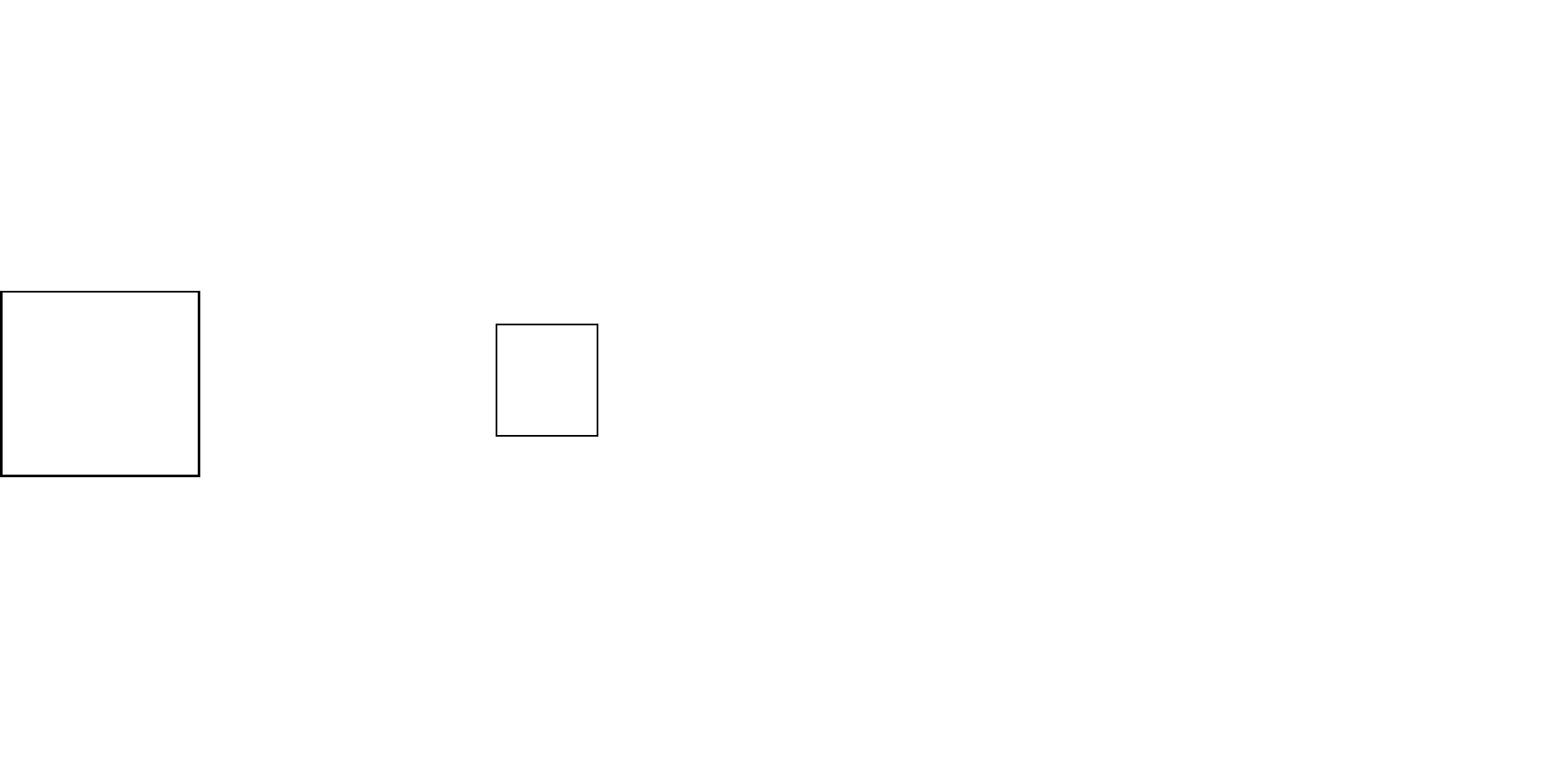
	\caption{The complete architecture. The output of the IK is no longer sent to the robot, but to the stack-of-task balancing controller, together with the desired CoM and foot positions. Joint positions and velocities are taken as feedback from the robot.}
	\label{fig:complete}
\end{figure}

\begin{figure}[tpb]
	\centering
	\includegraphics[width=0.9\columnwidth]{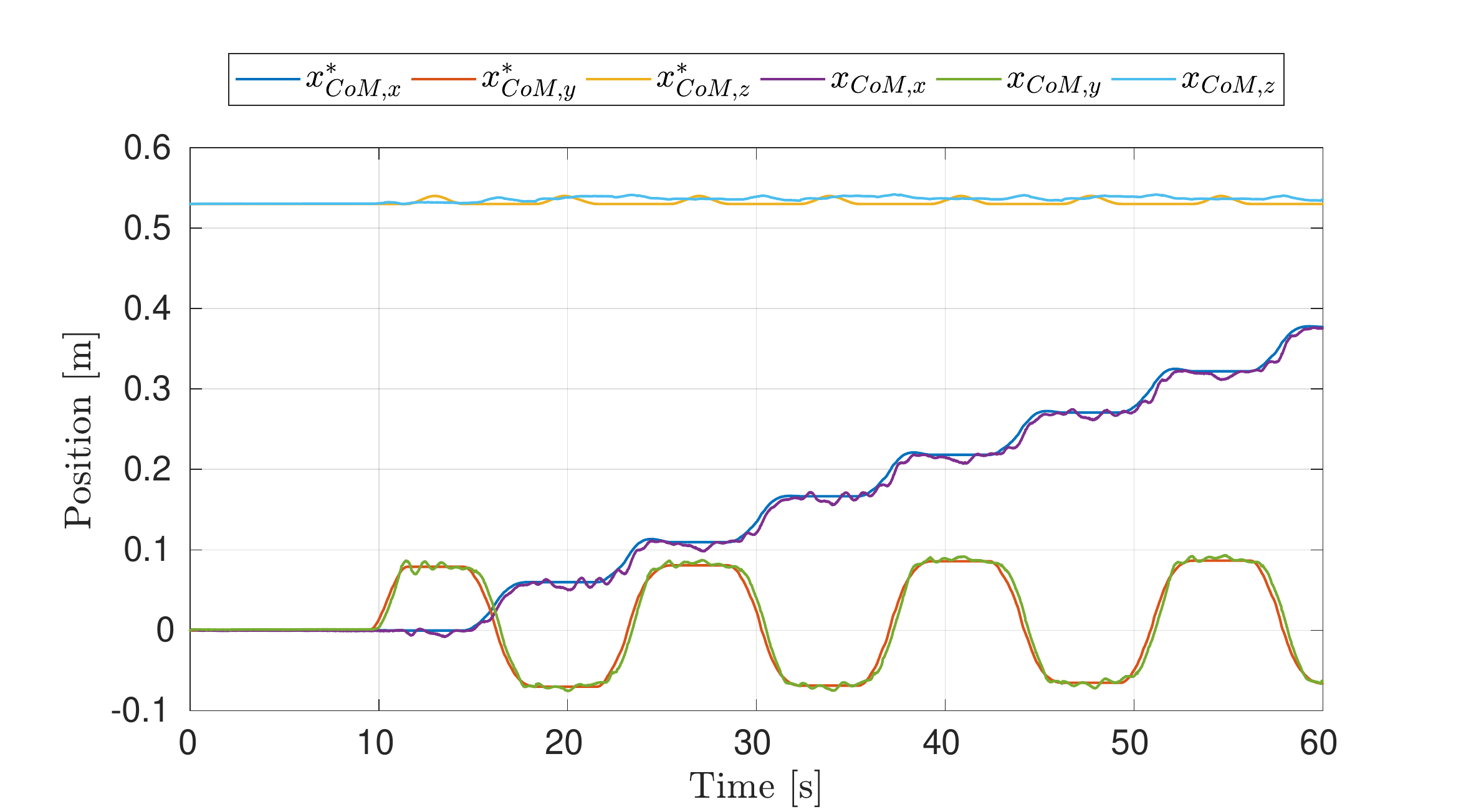}
	\caption{Center of Mass (CoM) tracking when taking some steps in torque control. The quantities are expressed in an inertial frame $\mathcal{I}$.}
	\label{fig:com}
\end{figure}

\begin{figure}[tpb]
	\centering
	\subfloat[Left foot]
	{\includegraphics[width=0.9\columnwidth]{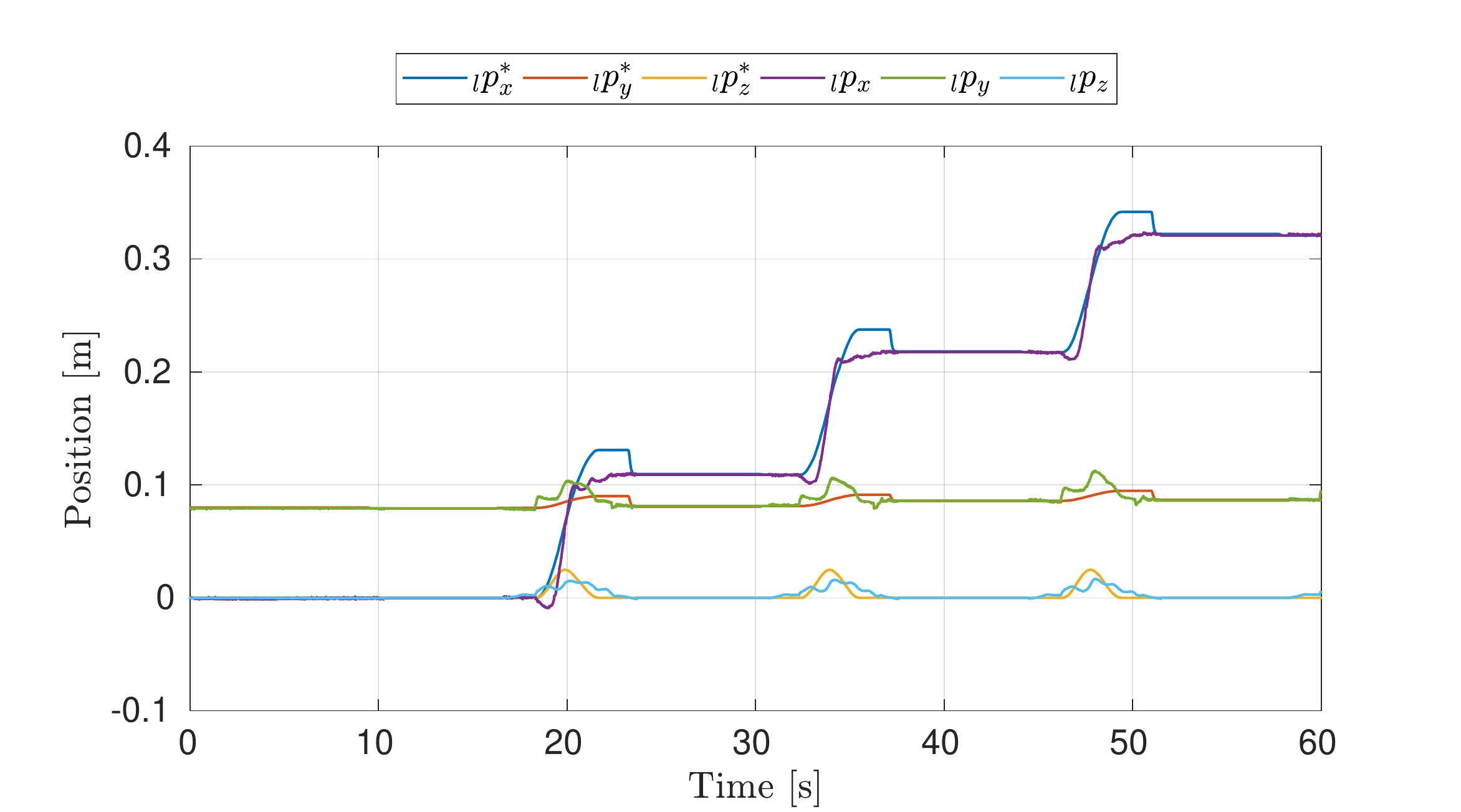}}
	
	\subfloat[Right foot]
	{\includegraphics[width=0.9\columnwidth]{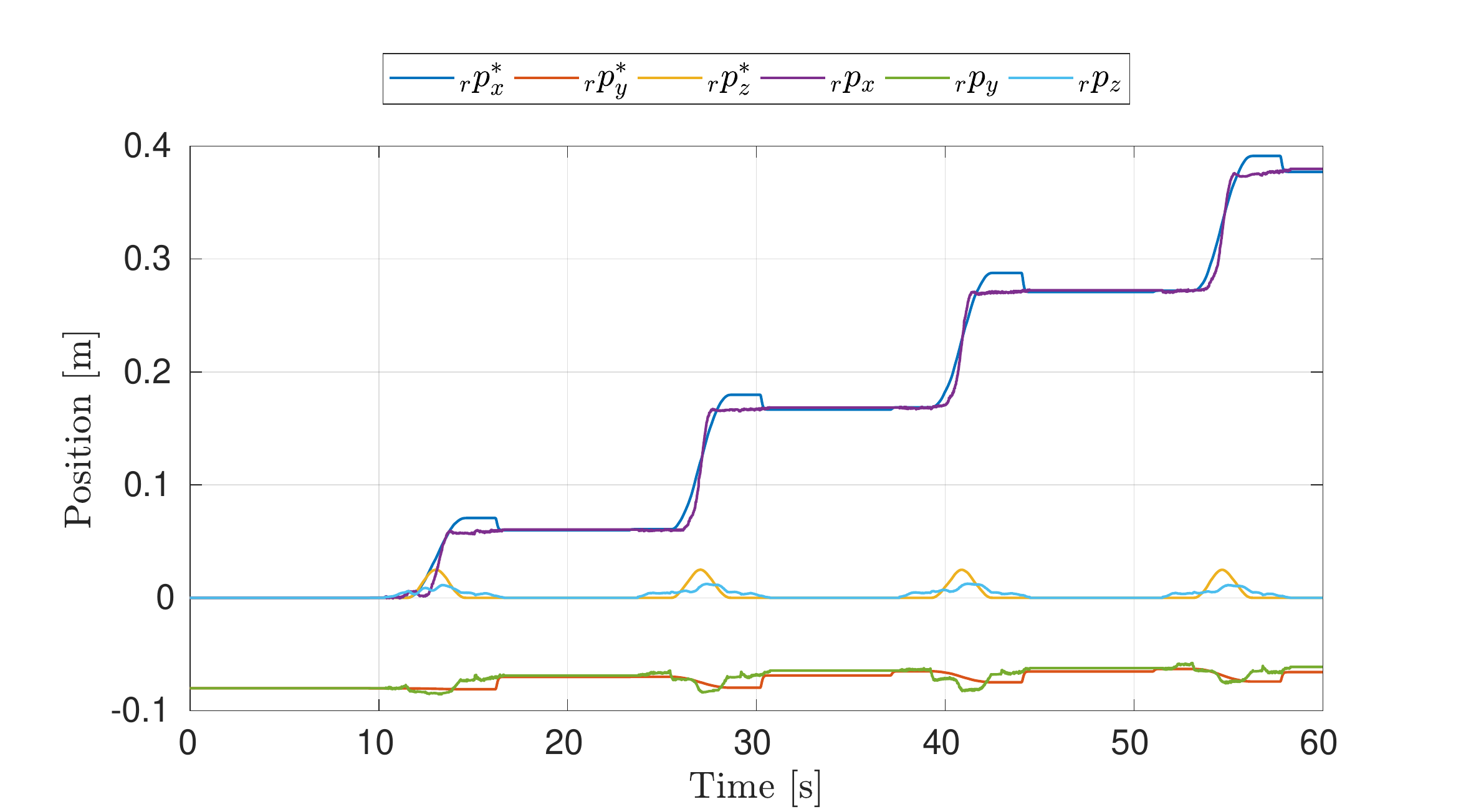}}
	\caption{Tracking of the foot positions. After a step, the \emph{desired} values are updated according to the measured position. This allows to cope with imprecise foot placements. This mechanism is the cause of the jumps on the desired values visible after the end of the step.}
	\label{fig:lFoot}
\end{figure}

Finally, we also plugged the task based balancing controller presented in Sec. \ref{sec:sot}. The overall architecture is depicted in Figure \ref{fig:complete}, highlighting that the stack of task balancing controller is now in charge of sending commands the robot. In order to draw comparisons with the first experiments, we followed again a straight trajectory. Even if the task is similar, it is necessary to use the unicycle planner now in order to cope with foot positioning errors (see Fig. \ref{fig:lFoot}). In fact, trajectories can be updated in order to take into account possible foot misplacements. Fig. \ref{fig:com} shows the CoM tracking capabilities of the presented balancing controller. During this experiment, the minimum stepping time needed to be increased to $3s$ (the maximum is still $5s$), while maintaining the same maximum step length of the previous experiment. 

\subsection{Comparison and discussion}

Torque control experiments have shown some bottlenecks that prevented achieving higher performances. These issues are mainly  related to the estimation of joints torques (limiting the performances of the joint torque controller) and to the floating base estimation. 
Regarding the first problem, the iCub humanoid robot does not have joint torque sensors, but it exploits 6 F/T sensors to estimate these values \citep{Fumagalli2012}. The estimations is provided at 100Hz, limiting the performances of the low-level torque controller. In addition, the F/T sensors are subject to noise and biases. This leads to a peak error in the torque tracking as high as $15\mathrm{Nm}$ (on the hip roll joint) during the experiment showed in this chapter. To give a comparison, the maximum desired torque required on the same joint is $25\mathrm{Nm}$.

Regarding the floating base estimation, we use a simple legged odometry algorithm where we keep track of the base pose only trough joint measurements. This algorithm causes unwanted discontinuities when the standing foot slightly rotates. This limitation affected the maximum velocity achievable by the robot.

In view of these limitations, walking in position control strongly outperforms the torque mode. The precise joint position tracking allows a more repeatable and faster walking. On the other hand, the robot is not able to adapt to any unevenness of the terrain, nor pushes.

\section{Conclusions}\label{iroswalking_conclusions}
This architecture is able to cope with variable walking speed, generating trajectories which don't require the robot to step in place while coping with \emph{online} changes of desired reference trajectories. The presented architecture is also flexible enough to allow controlling the robot by defining either joints positions or torques. 

Compared to Chapter \ref{chap:steprecovery}, this architecture uses more simplified models at the planning level. Nevertheless, in both cases the inner control loop uses the full robot model to compute the desired joint torques. The main difference between the controller presented in Sec. \ref{sec:sot} and the one of Sec. \ref{sec:momentum} is that the latter tracks the foot reference poses indirectly, through the definition of desired joints position in a low-priority task. Instead, the one presented in Sec. \ref{sec:sot} tracks a Cartesian reference for the feet as a high-priority task. Hence, the controller presented in this chapter is more precise and suitable to track a walking trajectory. 

The architecture presented in this chapter does not allow the robot to execute motions as wide as those shown in Sec. \ref{sec:simulation}. In addition, it does not allow the adaptation of desired footsteps in case of external perturbation.

In the next chapter, we still use a simplified model for the robot momentum, but more detailed. We remove the assumption of constant height, planning dynamic step-ups for the IHMC Atlas.

\chapter{Optimal Control of Large Step-Ups for the Atlas Robot}
\label{chap:stepups}
In this chapter, we present a non-linear trajectory optimization method for generating step-up motions. We adopt a simplified model of the centroidal dynamics to generate feasible Center of Mass trajectories. We describe this model in Sec. \ref{sec_ls: background}. The activation and deactivation of contacts at both feet are considered explicitly, assuming a fixed sequence.
A particular choice of cost function allows reducing the torques required for the step-up motion.
Sec. \ref{sec_ls:sup} presents the trajectory generation problem and its  transcription into an optimization problem.
The output of the planner is a Center of Mass trajectory plus an optimal duration for each walking phase. These desired values are stabilized by a QP-based whole-body controller, introduced in Sec. \ref{sec_ls: background}, which determines a set of desired joint torques.
Experiments are carried on the Atlas humanoid robot, showing a reduction of the maximum leading knee torque of about 20\%. They are presented in Sec. \ref{sec_ls:results}, while Sec. \ref{sec_ls:conclusions} contains concluding remarks.

\begin{figure}[tpb]
	\centering
	\includegraphics[width=0.8\columnwidth]{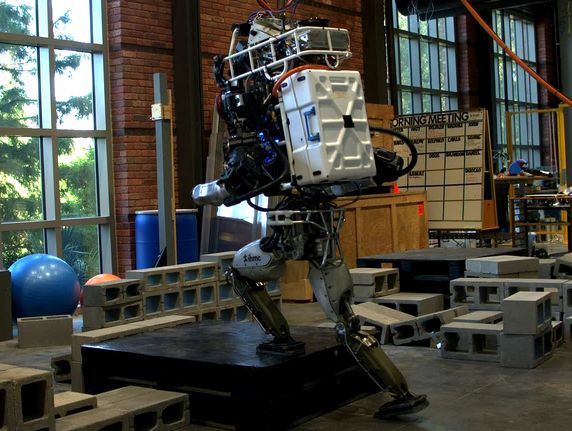}
	\caption{Atlas performing a 31$\mathrm{cm}$ tall step-up.}
	\label{fig:atlasShowingOff}
\end{figure}

\section{Background} \label{sec_ls: background}

\subsection{The Atlas QP-based whole-body controller} \label{sec_ls:qp_controller}
The IHMC Atlas humanoid robot is controlled by means of a QP-based whole-body controller \citep{koolen2016design}, which is able to maintain the robot's balance also in case the available contact patch reduces drastically \citep{wiedebach2016walking}. This controller is first introduced in Sec. \ref{sec:whole-body-controllers} and here presented with a much higher level of detail. This is to describe how the presented planner interfaces with the robot.

The Atlas QP controller determines a set of desired joint accelerations, including the spatial acceleration of the floating-base joint with respect to $\mathcal{I}$. We denote this quantity with $\dot{\bm{\nu}}_c \in \mathbb{R}^{n+6}$ where $n$ is the number of joints under control. In addition, the whole-body controller generates desired contact wrenches. 
The controller handles different motion tasks, which include Cartesian and center of mass (CoM) tasks.
The former are achieved by minimizing the following quantity:
\begin{equation}
    \left\|\bm{J}_t \dot{\bm{\nu}}_c - \left(\bm{\alpha}^* - \dot{\bm{J}}_t \bm{\nu}\right) \right\|^2,
\end{equation}
where $\bm{J}_t \in \mathbb{R}^{n_t \times (n+6)}$, is the task Jacobian, with $n_t$ being the number of degrees of freedom of task $t$. $\bm{\alpha}^* \in \mathbb{R}^{n_t}$ is the task desired acceleration, while $\bm{\nu}$ are the \emph{measured} base and joint velocities. The quantity $\left(\bm{\alpha}^* - \dot{\bm{J}}_t \bm{\nu}\right)$ is a vector which does not depend on the decision variables $\dot{\bm{\nu}}_c$, and it is referred with the symbol $\bm{\beta}_t$.

The CoM task is achieved by controlling the robot momentum ${}_{\bar{G}}\bm{h} \in \mathbb{R}^6$. In particular, we exploit Eq.s \eqref{eq:centroidal_momentum_dynamics}-\eqref{eq:momentum_derivative_cmm}:
\begin{equation}
    {}_{\bar{G}}\dot{\bm{h}} = \bm{J}_\text{CMM} \dot{\bm{\nu}} + \dot{\bm{J}}_\text{CMM} \bm{\nu} = m \bm{\bar{g}} + \sum_{k = 1}^{n_c} {}_{\bar{G}}\bm{X}^k {}_k\textbf{f},
\end{equation}
where each contact wrench is supposed to be applied on a 2D patch. A different point force is considered applied at each vertex $v$ of a contact patch $k$. In particular, these forces are defined through a coordinate vector $\bm{\varrho}_i \in \mathbb{R}^m$, whose basis are $m$ extreme rays of the friction cone. By constraining each component of $\bm{\varrho}_i$ in the range $\left[\varrho^\text{min}, \varrho^\text{max} \right]$, we avoid excessively high wrenches. More importantly, friction and Center of Pressure \citep{sardain2004forces} (CoP) constraints are implicitly satisfied \citep{pollard2001animation}. Trivially, also the resulting normal force is bounded to be positive. The corresponding wrench in the frame $\bar{G}$, is expressed as a function of ${}_{k_v}\bm{\varrho}$:
\begin{equation}
    {}_{\bar{G}}\bm{X}^k {}_k\textbf{f} = \sum_v \bm{Q}_{k_v}\, {}_{k_v}\bm{\varrho},
\end{equation}
where $\bm{Q}_{k_v} \in \mathbb{R}^{6\times n_\varrho}$ depends on the position of the contact vertex $v$ and on the basis vectors. By concatenating all the $\bm{Q}_{k_v}$ and ${}_{k_v}\bm{\varrho}$, we obtain 
\begin{equation}
	\sum_{k = 1}^{n_c} {}_{\bar{G}}\bm{X}^k {}_k\textbf{f} = \bm{Q} \bm{\varrho}.
\end{equation}

The Cartesian and CoM task are grouped as follows:
\begin{equation*}
        \hat{\bm{J}} = \begin{bmatrix}
                            \bm{J} \\ 
                            \bm{J}_\text{CMM}
                        \end{bmatrix}, \quad
        \hat{\bm{\beta}} = \begin{bmatrix}
                            \bm{\beta} \\ 
                            \dot{\bm{J}}_\text{CMM} \bm{\nu} - {}_{\bar{G}}\dot{\bm{h}}^*
                   \end{bmatrix},
\end{equation*}
with $\bm{J}$ and $\bm{\beta}$ being the concatenation of all the tasks Jacobians $\bm{J}_t$ and $\bm{\beta}_t$ respectively. ${}_{\bar{G}}\dot{\bm{h}}^*$ is a desired momentum rate of change.

At every control cycle, the solved QP problem writes:
\begin{IEEEeqnarray*}{LL}
\minimize_{\bm{\nu}_c,~ \bm{\varrho}} & \Gamma = \|\hat{\bm{J}} \dot{\bm{\nu}}_c - \hat{\bm{\beta}}\|^2_{\bm{C}_t} + \|\bm{\varrho}\|^2_{\bm{C}_\varrho} + \|\dot{\bm{\nu}_c}\|^2_{\bm{C}_{\dot{\bm{\nu}}}} \\
\text{subject to: } &\bm{J}_\text{CMM} \dot{\bm{\nu}}_c + \dot{\bm{J}}_\text{CMM} \bm{\nu} = m \bm{\bar{g}} + \bm{Q} \bm{\varrho} \\
&\bm{\varrho}^\text{min} \leq \bm{\varrho} \leq \bm{\varrho}^\text{max}.
\end{IEEEeqnarray*}
The desired joint accelerations and contact wrenches are used to compute a set of desired joint torques to be commanded to the robot.

Compared to the whole-body controller presented in Sec. \ref{sec:sot}, the one described here considers all the tasks at the same priority, relying to the weights in $\Gamma$ to determine the relative importance of each task. This approach has the advantage of avoiding tasks infeasibility. In other words, the resulting optimization problem cannot result unfeasible because of not attainable references. On the other hand, tuning is more complicated. Another difference is represented by the use of the contact force parametrization through $\bm{\varrho}$, which avoids the insertion of inequalities like Eq. \eqref{eq:contact_inequalities}. Finally, this controller outputs desired acceleration, while the one presented in Sec. \ref{sec:sot} defines reference joint torques.

\subsection{The variable height double pendulum}
The linear momentum rate of change can be written according to Newton's second law, similarly to what is done in Sec. \ref{sec:model}:
\begin{equation}\label{eq:com_point_dynamics}
    m \ddot{\bm{x}}_{\text{CoM}} = \sum_k {{}_k \bm{f}} + m \bm{g},
\end{equation}
where $\bm{g} \in \mathbb{R}^3$ corresponds to the first three rows of $\bar{\bm{g}}$. ${}_k\bm{f} \in \mathbb{R}^3$ is an external force expressed in $\mathcal{I}$.
We can express the wrenches applied at the feet in a frame parallel to $\mathcal{I}$, having the origin coincident with the corresponding CoP. We assume that no moment is exerted along the axis perpendicular to the foot, passing through the CoP. Thanks to these two choices, the moments are null. It is possible to enforce the angular momentum to be constant by constraining these forces along a line passing through the CoM, i.e.:
\begin{equation}\label{eq:simplified_force}
    {}_\circ \bm{f} = m \lambda_\circ \left(\bm{x}_\text{CoM} - {}_\circ^\mathcal{I}\bm{x}_\text{CoP} \right).
\end{equation}
The symbol $\circ$ is used as a placeholder for either the left $l$ or the right $r$ foot. ${}_\circ \bm{f}\in \mathbb{R}^3$ is the force applied on the foot, measured in $\mathcal{I}$ coordinates. $\lambda_\circ \in \mathbb{R}$ is a multiplier. Since the contacts are considered unilateral, $\lambda_\circ$ is constrained to be greater than zero for guaranteeing the positivity of normal forces. ${}_\circ^\mathcal{I}\bm{x}_\text{CoP} \in \mathbb{R}^3$ is the position of the foot CoP expressed in $\mathcal{I}$. It is a function of the foot position ${}_\circ\bm{p}$, its orientation, and of the CoP expressed in foot coordinates, ${}_\circ\bm{x}_\text{CoP}$:
\begin{equation}\label{eq:cop_in_foot}
    {}_\circ^\mathcal{I}\bm{x}_\text{CoP} = {}_\circ\bm{p} + ^\mathcal{I}\bm{R}_\circ {}_\circ\bm{x}_\text{CoP}.
\end{equation}
Notice that the $z$ coordinate of ${}_\circ\bm{x}_\text{CoP}$ is zero by construction. 
For the sake of simplicity, from now on we will drop the $\mathcal{I}$ superscript in front of the rotation matrix $^\mathcal{I}\bm{R}_\circ$. For the same reason, we drop the subscript CoM from the quantity expressing the CoM position. This results in the following variation of notation:
\begin{IEEEeqnarray}{RCL}
	\IEEEyesnumber
	\bm{R}_\circ & \coloneqq &  {}^\mathcal{I}\bm{R}_\circ \IEEEyessubnumber \\
	\bm{x} & \coloneqq & \bm{x}_\text{CoM}. \IEEEyessubnumber
\end{IEEEeqnarray}

Assuming the forces exerted on the feet to be the only external forces, we can rewrite Eq.\eqref{eq:com_point_dynamics} as:
\begin{equation}
    \begin{split}
        m \ddot{\bm{x}} = m  \bm{g} &+ m \lambda_l \left(\bm{x} - {}_l\bm{p} - \bm{R}_l ~ {}_l\bm{x}_\text{CoP} \right) \\
        &+ m \lambda_r \left(\bm{x} - {}_r\bm{p} - \bm{R}_r ~ {}_r\bm{x}_\text{CoP} \right).
    \end{split}
\end{equation}
Dividing by the total mass of the robot, we can finally obtain the equation of motion for the variable-height double pendulum:
\begin{equation}\label{eq:double_pendulum}
    \begin{split}
        \ddot{\bm{x}} = \bm{g} &+ \lambda_l \left(\bm{x} - {}_l\bm{p} - \bm{R}_l ~ {}_l\bm{x}_\text{CoP} \right) \\
        &+ \lambda_r \left(\bm{x} - {}_r\bm{p} - \bm{R}_r ~ {}_r\bm{x}_\text{CoP} \right).
    \end{split}
\end{equation}
This model have been studied and adopted both in the single contact \citep{koolen2016balance, posa2017balancing, caron2017dynamic,caron2018balance} and multi-contact scenarios \citep{perrin2018effective}, as in this chapter.
\section{Dynamic planning for large step-ups} \label{sec_ls:sup}
The step-up planner presented in this chapter leverages the idea of \emph{phase-based trajectory optimization} \citep{winkler18,carpentier2016versatile,caron2017make}. We assume to know the contact sequence beforehand, simplifying the handling of their activation and deactivation.
In addition, we also assume the foot positions to be known in advance. Given the application, this represents a safety measure too.

This section presents constraints, tasks and methodologies used to solve the corresponding non-linear trajectory optimization problem. 

We start by simplifying Eq. \eqref{eq:double_pendulum} considering a simple forced double integrator dynamics, as follows:
\begin{equation}\label{eq:dynamics_simple}
    \ddot{\bm{x}}  = \bm{a}_i.
\end{equation}
$\bm{a} \in \mathbb{R}^3$ assumes different values depending on the contact state of the corresponding phase $i$:
\begin{itemize}
    \item no contacts \begin{equation*}
        \bm{a}_i = \bar{\bm{g}};
    \end{equation*}
    \item one foot in contact ($\circ$ is either $l$ or $r$) \begin{equation*}
        \bm{a}_i = \bm{g} + \lambda_\circ \left(\bm{x} - {}_\circ\bm{p} - \bm{R}_\circ ~ {}_\circ\bm{x}_\text{CoP} \right);
    \end{equation*}
    \item both feet in contact \begin{equation*}
        \bm{a}_i = \bm{g} + \lambda_l \left(\bm{x} - {}_l\bm{p} - \bm{R}_l ~ {}_l\bm{x}_\text{CoP} \right) + \lambda_r \left(\bm{x} - {}_r\bm{p} - \bm{R}_r ~ {}_r\bm{x}_\text{CoP} \right).
    \end{equation*}
\end{itemize}
We can obtain an approximated discrete form of Eq. \eqref{eq:dynamics_simple} through a second order Taylor expansion: 
\begin{IEEEeqnarray}{RCL}
    \IEEEyesnumber \phantomsection \label{eq:discrete_double_pendulum}
    \bm{x}(k+1) &=& \bm{x}(k) + \mathrm{dt}_i ~ \bm{v}(k) + \frac{1}{2}\mathrm{dt}_i^2~\bm{a}_i(k), \IEEEyessubnumber\\
    \bm{v}(k+1) &=& \bm{v}(k) + \mathrm{dt}_i~\bm{a}_i(k), \IEEEyessubnumber
\end{IEEEeqnarray}
where $k \in \mathbb{R}$ is used to indicate a generic discrete instant, while $\bm{v}(k) \in \mathbb{R}^3$ is the CoM velocity at instant $k$. The duration $T_i \in \mathbb{R}$ of each phase $i$ is not defined a-priori, but it is considered as an optimization variable. Nevertheless, each phase consists of a fixed number of instants. Hence, the integration step can be easily computed, i.e. $\mathrm{dt}_i = T_i/N$, where $N \in \mathbb{N}^+$ is the number of instants per phase. In addition, each $T_i$ is bounded in $\left[T_{i}^\text{ min},~ T_{i}^\text{ max}\right]$.

\subsection{Force and leg constraints}\label{sec:force_constraints}
The forces applied at the feet need to satisfy some conditions in order to be attainable by the robot. These are embedded as constraints in the optimization problem.

First of all, the CoP has to lie inside the foot polygon. This can be obtained by specifying a set of linear inequalities:
\begin{equation}
    \bm{A}_\circ \, {}_\circ\bm{x}_\text{CoP} \leq \bm{b}_\circ,
\end{equation}
where the matrix $\bm{A}_\circ \in \mathbb{R}^{v \times 3}$ and the vector $\bm{b}_\circ \in \mathbb{R}^{v}$ can be computed according to the foot location of and shape. In particular, we assume it has a polygonal shape with $v$ vertices. Since ${}_\circ\bm{x}_\text{CoP}$ is defined in foot coordinates, $\bm{A}_\circ$ and $\bm{b}_\circ$ do not depend on the foot pose.

The contact forces are supposed to lie within the friction cone in order to avoid slipping. This can be achieved by imposing
\begin{equation}\label{eq:friction_in_foot}
    \sqrt{{}^\circ_\circ f_{x}^2 + {}^\circ_\circ f_{y}^2} \leq \mu_s ~ {}^\circ_\circ f_{z}.
\end{equation}
 $\mu_s \in \mathbb{R}$ is the static friction coefficient and ${}^\circ_\circ\bm{f} \in \mathbb{R}^3$ is the force exerted on a foot, expressed in foot coordinates, namely ${}^\circ_\circ\bm{f} = \bm{R}_\circ^\top {{}_\circ \bm{f}}$. We assume the foot frame to be parallel to a frame attached to the ground, with the $z-$ axis perpendicular to the walking surface. Using Eq. \eqref{eq:simplified_force} and \eqref{eq:cop_in_foot}, we can rewrite Eq. \eqref{eq:friction_in_foot} as follows:
\begin{equation} \label{eq:friction_full}
    \begin{bmatrix}
        1 & 1 & -\mu_s^2
    \end{bmatrix} \left(\bm{R}_\circ^\top m \lambda_\circ \left(\bm{x} - {}_\circ\bm{p} - \bm{R}_\circ ~ {}_\circ\bm{x}_\text{CoP} \right) \right)^2 \leq 0.
\end{equation}
Notice that when $\lambda_\circ = 0$, friction is automatically satisfied as every component goes to zero. Thus, we can simplify Eq. \eqref{eq:friction_full}:
\begin{equation}
    \begin{bmatrix}
        1 & 1 & -\mu_s^2
    \end{bmatrix}  \left(\bm{R}_\circ^\top \left(\bm{x} - {}_\circ\bm{p} - \bm{R}_\circ ~ {}_\circ\bm{x}_\text{CoP}\right)\right)^2 \leq 0.
\end{equation}

In order to consider the torsional friction, we impose the equivalent normal torque ${}_\circ\tau_z$, applied at the origin of the foot frame, to be bounded, i.e. 
\begin{equation}
	-\mu_t \, {}^\circ_\circ f_{z} \leq {}_\circ \tau_{z}\leq \mu_t \, {}^\circ_\circ f_{z},
\end{equation}
with $\mu_t \in \mathbb{R}$ the torsional friction coefficient. Given that the external force ${}^\circ_\circ\bm{f}$ is applied in ${}_\circ \bm{x}_\text{CoP}$, we can rewrite this constraint as follows:
\begin{equation}
    - \begin{bmatrix}
    0 \\ 0 \\ \mu_t
    \end{bmatrix}^\top {}^\circ_\circ\bm{f} \leq \begin{bmatrix}
                                     -{}_\circ {x}_{\text{CoP},y} \\ {}_\circ {x}_{\text{CoP},x} \\ 0
                                  \end{bmatrix}^\top {}^\circ_\circ\bm{f} \leq \begin{bmatrix}
                                                                        0 \\ 0 \\ \mu_t
                                                                    \end{bmatrix}^\top {}^\circ_\circ\bm{f}
\end{equation}
or equivalently
\begin{equation}
    -\bm{d}_\circ^\top {}^\circ_\circ\bm{f} \leq \bm{c}_\circ^\top {}^\circ_\circ\bm{f} \leq \bm{d}_\circ^\top {}^\circ_\circ\bm{f}.
\end{equation}
Notice that $\bm{c}_\circ^\top {}^\circ_\circ\bm{f}$ corresponds to the $z$ component of the cross product ${}_\circ \bm{p} \times {}^\circ_\circ\bm{f} $.
By substituting ${}^\circ_\circ\bm{f}$ as for the static friction constraint, we can finally write the following set of constraints:
\begin{equation}
     \bm{F}_\circ \bm{R}_\circ^\top \left(\bm{x} - {}_\circ\bm{p} - \bm{R}_\circ ~ {}_\circ\bm{x}_\text{CoP} \right) \leq 0,
\end{equation}
where
\begin{equation}
    \bm{F}_\circ = \begin{bmatrix}
                \left( \bm{c}_\circ - \bm{d}_\circ\right)^\top \\
                \left( -\bm{c}_\circ - \bm{d}_\circ\right)^\top
           \end{bmatrix}.
\end{equation}

When planning, we have to take into consideration the robot kinematic limits. We can approximate the robot leg length with the distance between the CoM and the corresponding foot position, thus limiting excessively wide motions or movements which could cause a leg to collapse. This can be achieved with the following constraint:
\begin{equation}
    -l^\text{min} \leq \|\bm{x} - {}_\circ\bm{p}\| \leq l^\text{max},
\end{equation}
with $l^\text{min}, l^\text{max} \in \mathbb{R}$ the lower and upper bound, respectively. These constraints can be applied on each leg separately.

\subsection{Tasks} \label{sec:double_pendulum_tasks}
The cost function is designed to generate a CoM trajectory for the robot to perform large step-ups, exploiting its momentum. It is composed of several tasks. The full cost function is shown at the top of Optimization Problem \ref{alg:step-up-planner}.

The first task weights the distance of the terminal states from the desired position $\bm{x}^*$ and velocity $\bm{v}^*$:
\begin{equation}
    \Gamma_{x^*} = w_{x^*} \sum_{k = \mathbb{K}_f} \left(\|\bm{x}(k) - \bm{x}^*\|^2 + \|\bm{v}(k) - \bm{v}^*\|^2\right)
\end{equation}
where $w_{x^*} \in \mathbb{R}^+$ is a tunable gain. $\mathbb{K}_f$ contains a certain portion of the last phase, including the terminal state. Thus, it is possible to generate trajectories which reach the end point in advance while defining control inputs able to maintain such position. 

The model defined by Eq. \eqref{eq:double_pendulum} does not carry any information about the joint torques necessary to achieve the planned motion. Nevertheless, with the aim of reducing the torques required at the leading knee to step-up, we introduce the following task:
\begin{equation*}
    \Gamma_\tau = w_\tau \left({}_l\uptau(k)^2 + {}_r\uptau(k)^2\right) + w_{\uptau_\text{max}}\left( \max_k {}_l\uptau(k)^2 + \max_k {}_r\uptau(k)^2 \right),
\end{equation*}
where
\begin{equation}
    {}_\circ\uptau(k) = \begin{cases}
                        \left(x_z(k) - {}_\circ p_{z} - {}_\circ\delta^* \right) \lambda_\circ(k), & \text{if $\circ$ in contact},\\
                        0, & \text{otherwise},
                  \end{cases}
\end{equation}
is used as a heuristic to reduce joint torques. ${}_\circ\delta^*$ is a reference height difference. Intuitively, the robot knees undergo the highest stress when they are highly bent. In this configuration, the height difference between the CoM and the foot gets small. This heuristic prevents the solver from requiring high forces (characterized by a high multiplier $\lambda_\circ$) on a collapsed leg. Thus $\Gamma_\tau$ weights the summation over each time instant of the this quantity squared. The second term is employed to prevent this quantity to have an impulsive behavior. $w_\tau \in \mathbb{R}^+$ and $w_{\uptau_\text{max}} \in \mathbb{R}^+$ are weights allowing to specify the relative priority for each task.

Considering the control inputs $\lambda_\circ(k)$ and ${}_\circ\bm{x}_\text{CoP}(k)$, it is preferable to generate trajectories which require small control variations. This reduces the need for large torque variations when tracking the trajectories on the real robot. To this end, we consider the following task:
\begin{equation}
    \Gamma_{\Delta_u} = w_{\Delta_u} \sum_{\circ, k=2\dots N \cdot P} \|{}_\circ\bm{u}(k) - {}_\circ\bm{u}(k-1)\|^2,
\end{equation}
where $w_{\Delta_u} \in \mathbb{R}^+$ is the task weight and ${}_\circ\bm{u} = \left[\lambda_\circ~{}_\circ\bm{x}_\text{CoP}^\top\right]^\top$.

Finally, we consider a series of regularization terms: 
\begin{equation}
    \Gamma_\text{reg} = w_t \sum_i \|T_i - T_{i, d}\|^2 + w_u \sum_{k, \circ} \|{}_\circ\bm{u}(k)\|^2.
\end{equation}
This task allows selecting solutions which minimize the use of control inputs and the duration of each phase is close to a desired one $T_{i, d}$. $w_t, w_u \in \mathbb{R}$ are the respective tasks weights. 

\subsection{The optimization problem}
The trajectory optimization problem constituted by the dynamic constraint defined in Eq.\eqref{eq:discrete_double_pendulum}, the constraints listed in Sec. \ref{sec:force_constraints} and the tasks of Sec. \ref{sec:double_pendulum_tasks}, is casted into an optimization problem via the Direct Multiple Shooting method, as described in Sec. \ref{sec:shooting}.
The optimization variables $\bm{\chi}$ correspond to the following set:
\begin{equation*}
    \bm{\chi} = \begin{cases}
                \bm{x}(k), &k = 0 \dots N \cdot P, \\
                \bm{v}(k), &k = 0 \dots N \cdot P, \\
                \bm{a}(k), &k=1 \dots N \cdot P, \\
                \lambda_\circ(k), &k=1 \dots N \cdot P,~ *=l,~ r, \\
                {}_\circ\bm{x}_\text{CoP}(k), &k=1 \dots N \cdot P,~ *=l,~ r,  \\
                T_i, &i=1\dots P,
           \end{cases}
\end{equation*}
with $P \in \mathbb{N}^+$ equal to the number of phases. $\bm{x}(k)$ and $\bm{v}(k)$ are state variables and need to be initialized with the measurements from the robot $\bm{x}_0$, $\bm{v}_0$:
\begin{align}
    \bm{x}(0) &= \bm{x}_0, \\
    \bm{v}(0) &= \bm{v}_0.
\end{align} 

Its complete formulation is shown in Optimization Problem \ref{alg:step-up-planner}. Note that index $i$ refers to a specific phase. Hence, quantities that depend on it are fixed for its entire duration and considered as datum. It is implemented using the \texttt{CasADi} \citep{Andersson2018} toolkit, where \texttt{Ipopt} \citep{IPOpt2006} is selected as solver.
\begin{algorithm}
\SetKwBlock{SubjectTo}{subject to:}{}
 \begin{flalign*}
     \minimizebf_{\bm{\chi}} & ~~\Gamma_{x^*} + \Gamma_\tau + \Gamma_{\Delta_u} + \Gamma_\text{reg}&&
 \end{flalign*}
 \SetAlgoLined
 \SubjectTo{
    $\bm{x}(0) = \bm{x}_0$ \\
    $\bm{v}(0) = \bm{v}_0$ \\
  \For{$i = 1 \dots P$} {
    $T_{i}^\text{ min} \leq T_i \leq T_{i}^\text{ max}$ \\
    $\mathrm{dt}_i := T_i/N$\\
    \For{$k = \left((i-1)P\right)$ $\dots$ $\left(i \cdot P - 1 \right) $ } 
		{
        $\bm{x}(k+1) = \bm{x}(k) + \mathrm{dt}_i ~ \bm{v}(k) + \frac{1}{2}\mathrm{dt}_i^2~\bm{a}(k)$\\
        $\bm{v}(k+1) = \bm{v}(k) + \mathrm{dt}_i~\bm{a}(k)$\\
        $\bm{\mathrm{a}} := \bm{g}$\\
        \If{left in contact} {
            $\bm{\mathrm{a}} := \bm{\mathrm{a}} + \lambda_l(k) \left(\bm{x}(k) - {}_l\bm{p}(i) - \bm{R}_{l}(i)\, {}_l\bm{x}_\text{CoP}(k) \right)$\\
            $\bm{A}_l(i) \, {}_l\bm{x}_\text{CoP}(k) \leq \bm{b}_l(i)$\\[2pt]
            $\left[1 ~ 1 ~ -\mu^2\right] \left(\bm{R}_{l}^\top(i) \left(\bm{x}(k) {-} {}_l\bm{p}(i) {-} \bm{R}_{l}(i) {}_l\bm{x}_\text{CoP}(k) \right)\right)^2 \leq 0$\\[2pt]
            $\bm{F}_{l}(i) \bm{R}_{l}^\top(i) \left(\bm{x}(k) - {}_l\bm{p}(i) - \bm{R}_{l}(i)\, {}_l\bm{x}_\text{CoP}(k) \right) \leq 0$\\[2pt]
            $l^\text{ min} \leq \|\bm{x}(k) - {}_l\bm{p}(i)\| \leq l^\text{ max}$\\
        }
        \If{right in contact} {
            $\bm{\mathrm{a}} := \bm{\mathrm{a}} + \lambda_r(k) \left(\bm{x}(k)- {}_r\bm{p}(i) - \bm{R}_{r}(i)\, {}_r\bm{x}_\text{CoP}(k) \right)$\\
            $\bm{A}_r(i) \, {}_r\bm{x}_\text{CoP}(k) \leq \bm{b}_r(i)$\\[2pt]
            $\left[1 ~ 1 ~ -\mu^2\right] \left(\bm{R}_{r}^\top(i) \left(\bm{x}(k) {-} {}_r\bm{p}(i) {-} \bm{R}_{r}(i) {}_r\bm{x}_\text{CoP}(k) \right)\right)^2 \leq 0$\\[2pt]
            $\bm{F}_{r}(i) \bm{R}_{r}^\top(i) \left(\bm{x}(k) - {}_r\bm{p}(i) - \bm{R}_{r}(i)\, {}_r\bm{x}_\text{CoP}(k) \right) \leq 0$\\[2pt]
            $l^\text{ min} \leq \|\bm{x}(k) - {}_r\bm{p}(i)\| \leq l^\text{max}$\\
        }
        $\bm{a}(k) = \bm{\mathrm{a}}$
    }
  }
 }
 \SetAlgorithmName{Optimization Problem}{ }
 
 \caption{ } 
 \label{alg:step-up-planner}
\end{algorithm}

The optimization problem is non-convex due to the non-linearities of the model equation, of the friction constraints and the chosen tasks. In order to facilitate the finding of a solution, we initialize the phases duration $T_i$ with their desired value. The CoM position trajectory is initialized to a linear interpolation from $\bm{x}_0$ to $\bm{x}^*$. All other variables are initialized to zero. 

We noticed that it is particularly important to initialize at least the CoM height with a value different from zero. In fact, in case the CoM and the feet belongs to the same plane, the forces would vanish, see Eq. \eqref{eq:simplified_force}, and gravity would be impossible to compensate. Similarly, $T_i$ should not be initialized to zero to avoid numerical problems in Eq. \eqref{eq:discrete_double_pendulum}.

\section{Validation and experimental results} \label{sec_ls:results}
\begin{figure*}[tpb]
    \centering
    \subfloat[$t=3.3s$] {\includegraphics[width=.23\textwidth]{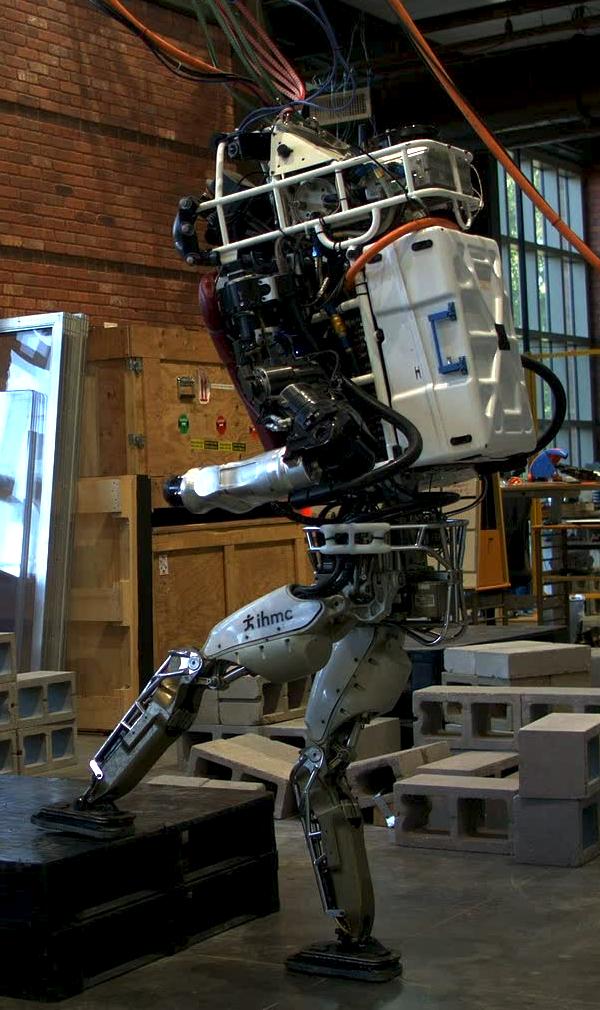}}
    \hspace{.01\textwidth}
    \subfloat[$t=4.2s$] {\includegraphics[width=.23\textwidth]{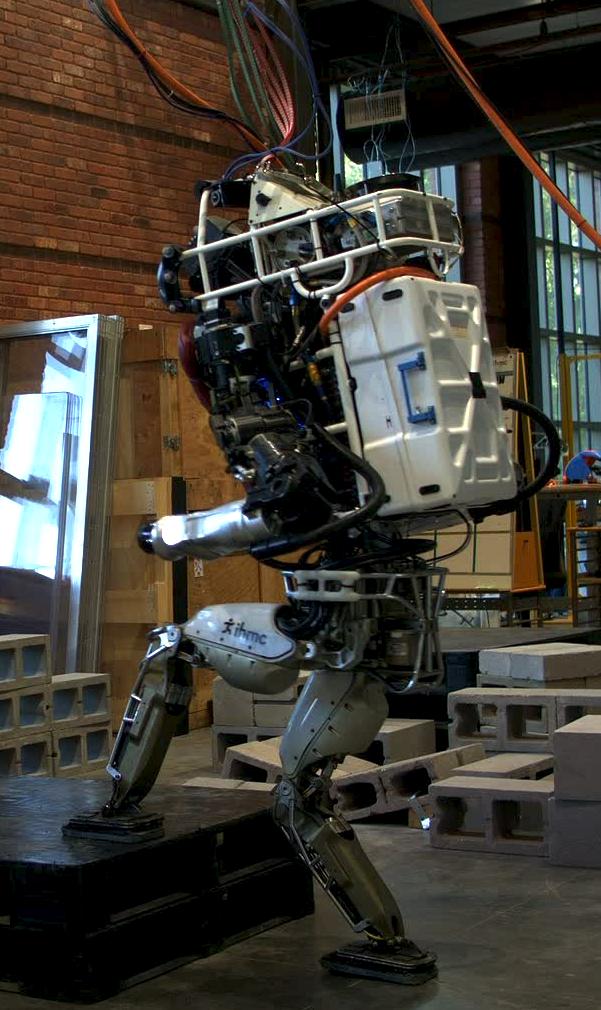}}
    \hspace{.01\textwidth}
    \subfloat[$t=5.6s$] {\includegraphics[width=.23\textwidth]{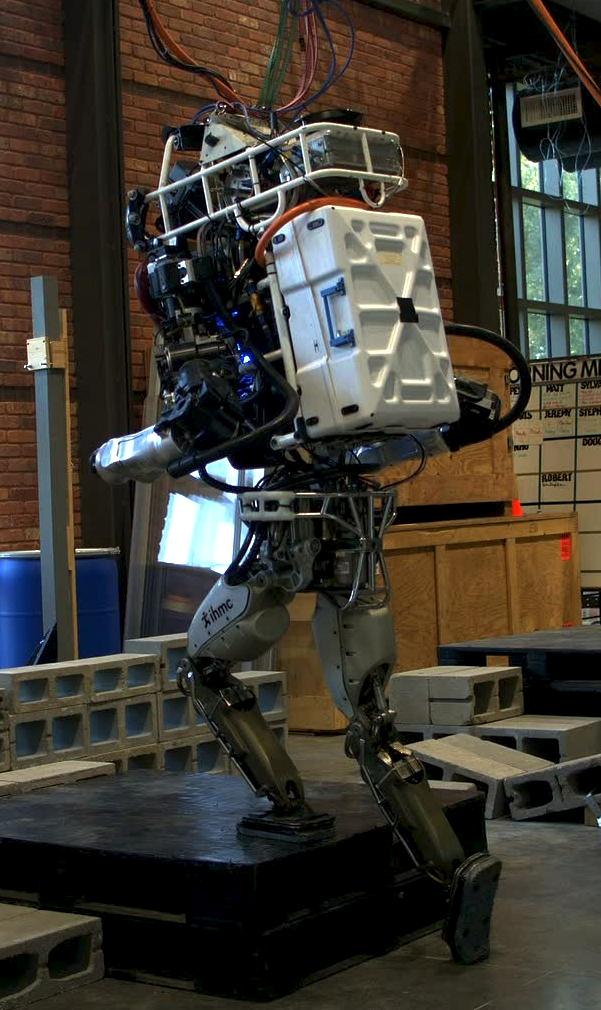}}
    \hspace{.01\textwidth}
    \subfloat[$t=8.8s$] {\includegraphics[width=.23\textwidth]{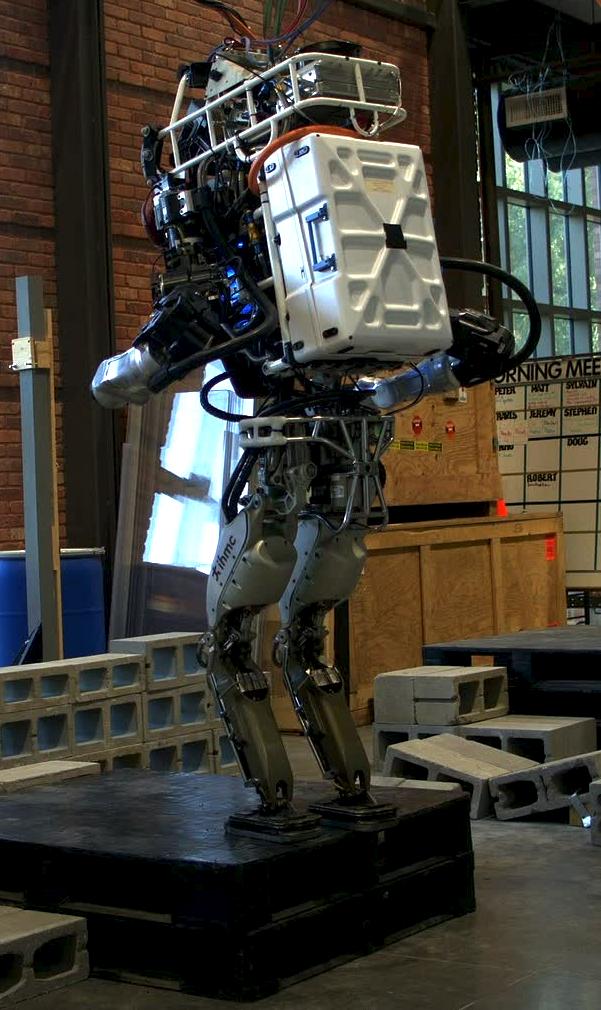}}
    \caption{Snapshots\protect\footnotemark ~of the step-up motion. These instances correspond to a switch of phases, except for the one at time $4.2s$ which corresponds to the lowest CoM height.}
    \label{fig_ls:snapshots}
\end{figure*}
\footnotetext{\url{https://www.youtube.com/playlist?list=PLBOchT-u69w9hJ6BmqPf06r0zWmungOrc}}
\begin{figure}[tpb]
    \centering
    \subfloat[$x$ direction]{\includegraphics[width=0.9\columnwidth]{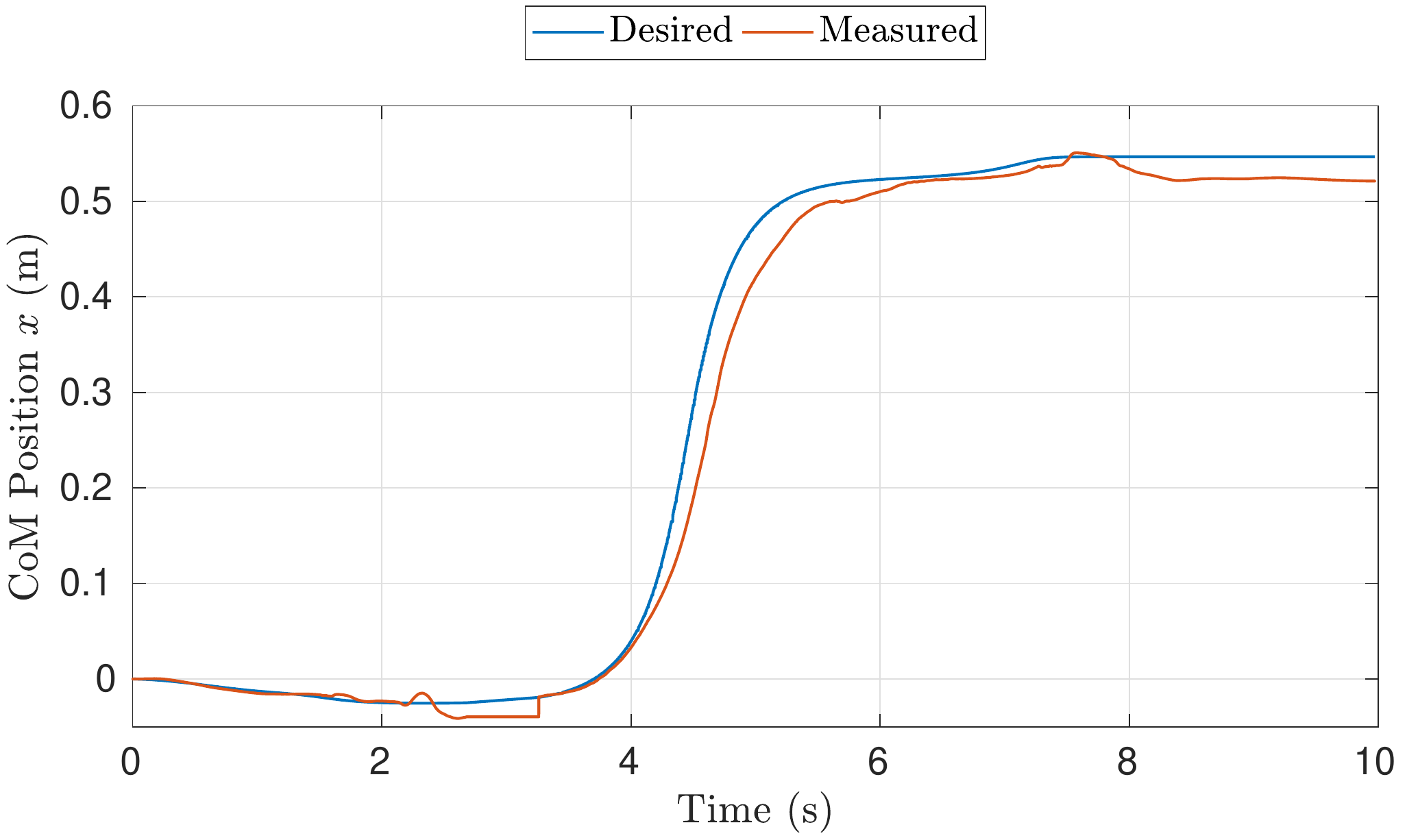} \label{fig_ls:jump_comx}}
    
    \subfloat[$y$ direction]{\includegraphics[width=0.9\columnwidth]{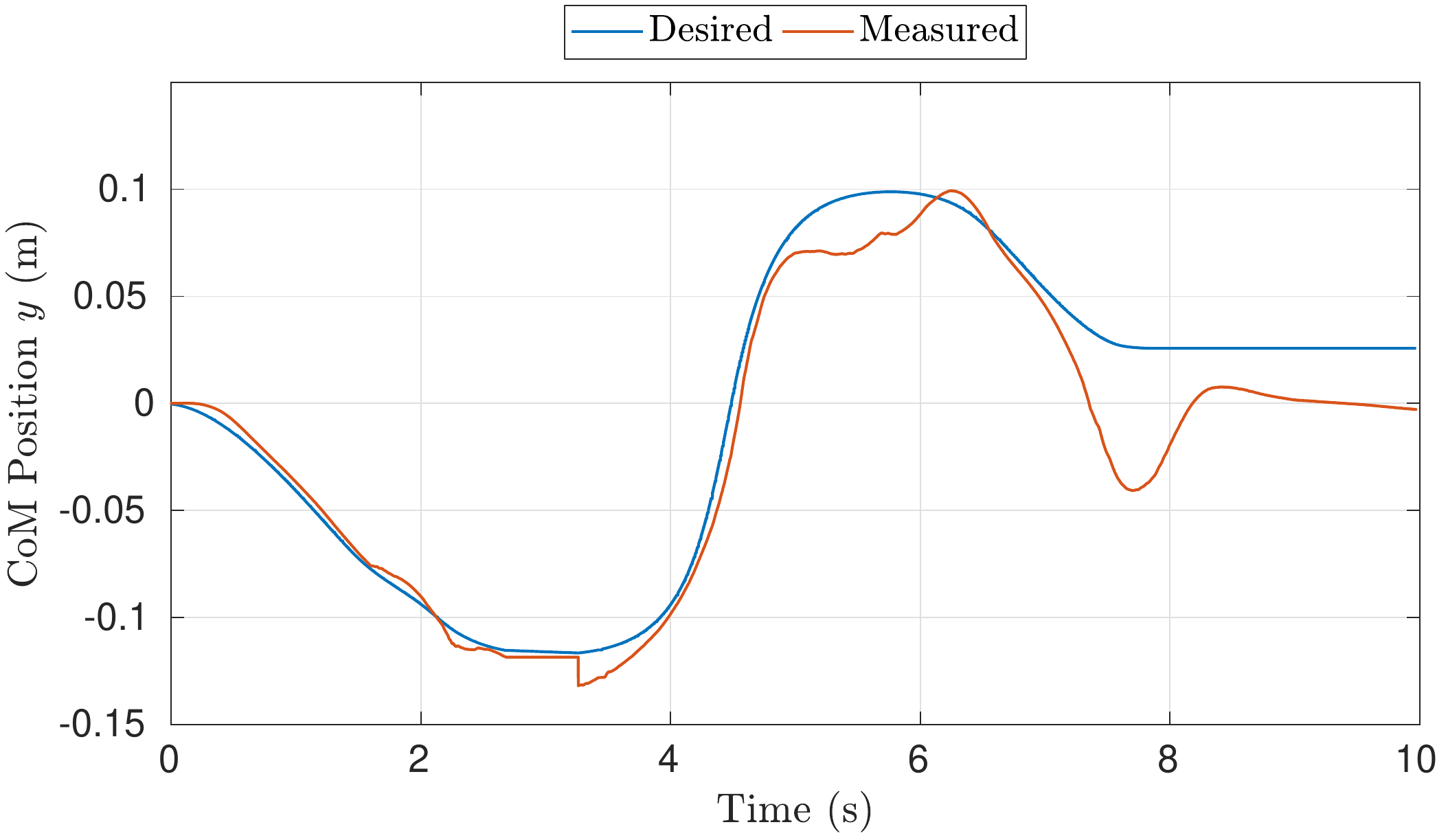} \label{fig_ls:jump_comy}}
    
    \caption{Tracking of the CoM position along the planar directions.}    
\end{figure}
\begin{figure}[tpb]
    \centering
    \includegraphics[width=0.9\columnwidth]{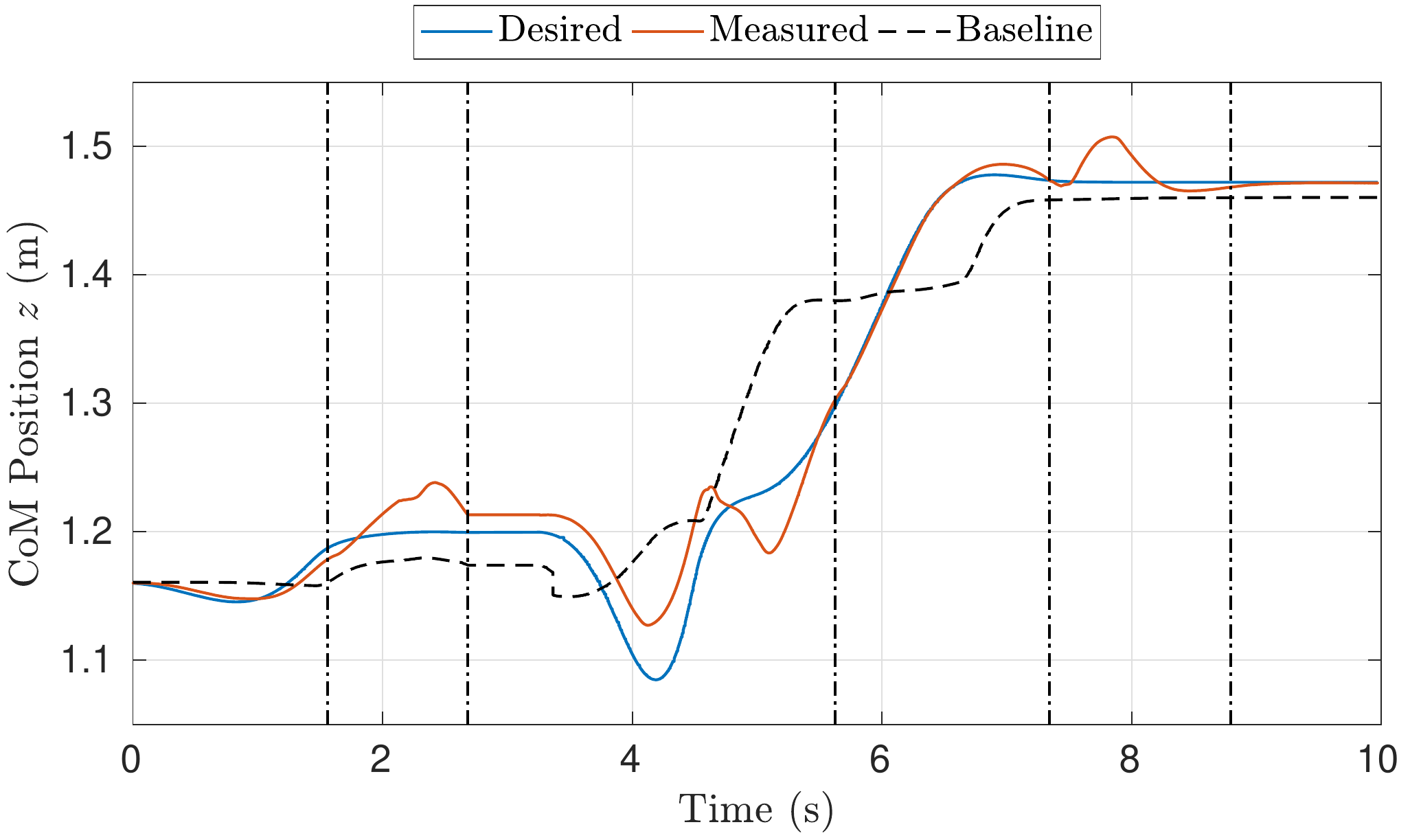}
    \caption{Tracking of the CoM height. The vertical dotted lines correspond to a change of phase.}
    \label{fig_ls:jump_comz}
\end{figure}
\begin{figure}[tpb]
    \centering
    \includegraphics[width=0.9\columnwidth]{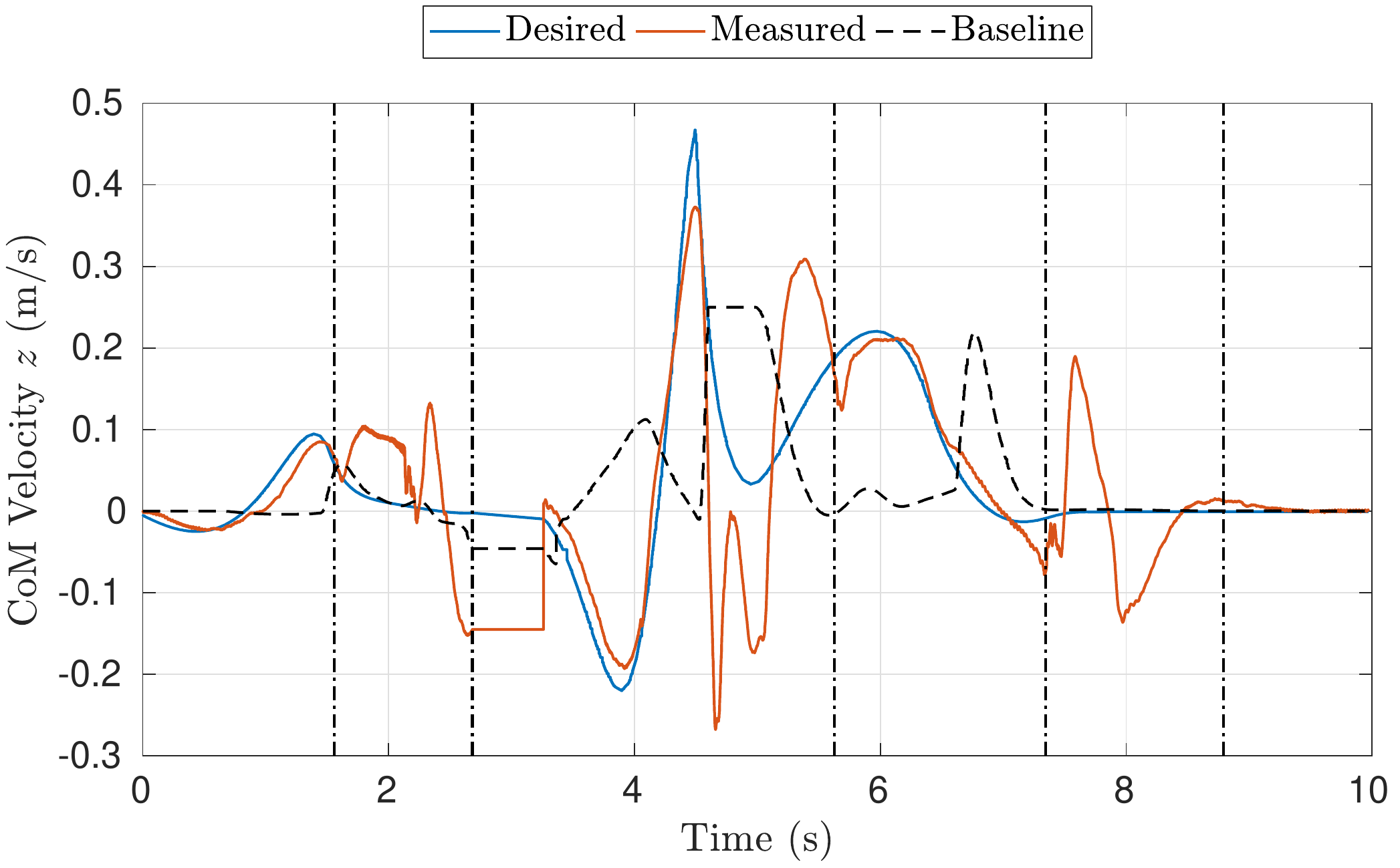}
    \caption{Tracking of the desired height velocity. The vertical dotted lines indicate the change of phases. It is possible to notice that the tracking worsens around $t=5s$, which is close to the end of the double support phase.}
    \label{fig_ls:jump_comVelocityz}
\end{figure}
\begin{figure}[tpb]
    \centering
    \includegraphics[width=0.9\columnwidth]{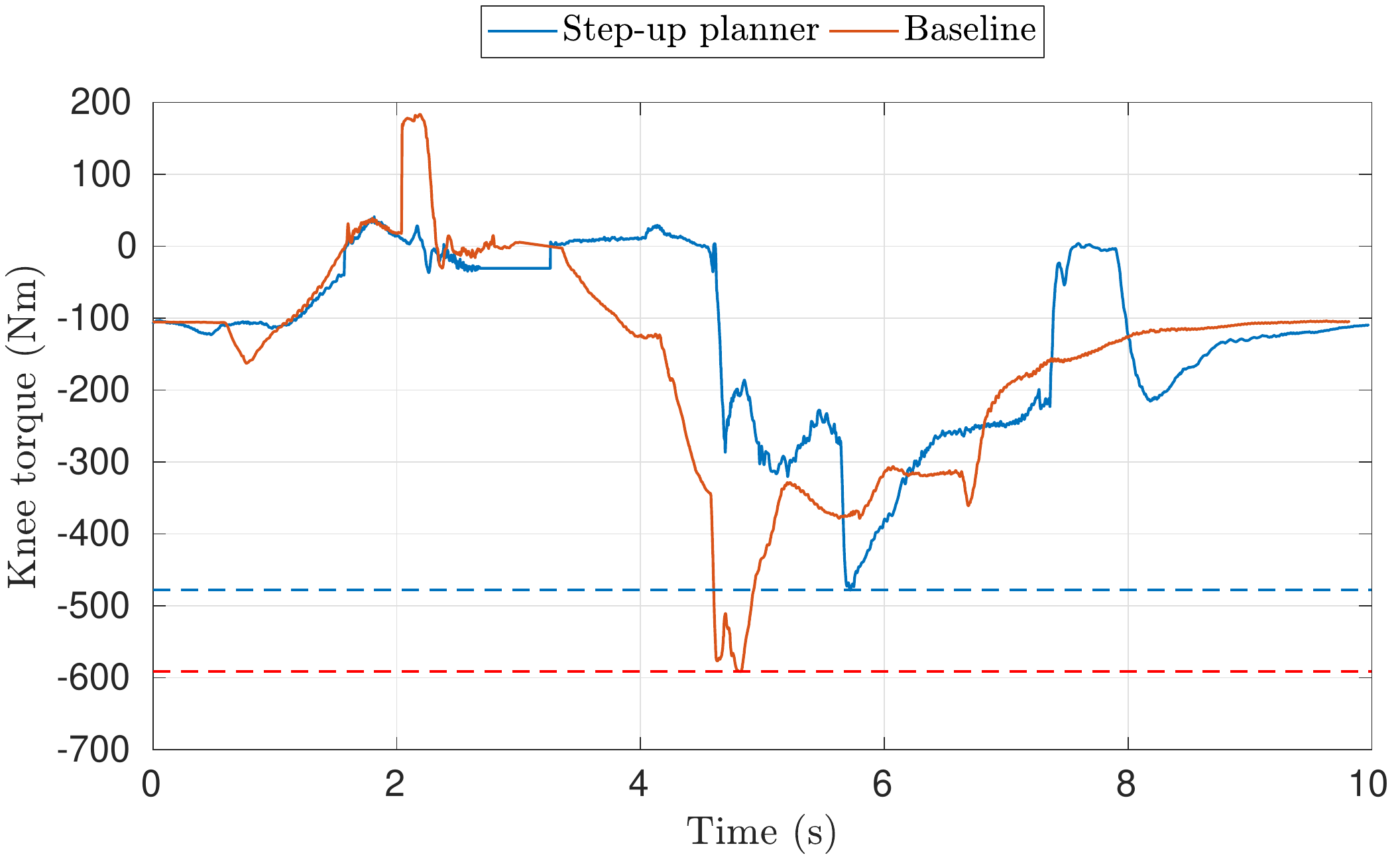}
    \caption{Comparison of measured knee torques. The baseline is obtained by letting the controller described in \citep{seyde2018inclusion} generating the motion automatically. 
    The duration of each phase is generated by planner in both cases for easing the comparison. 
    The dotted lines are the peak torques. The maximum torque required to the robot when using the step-up planner is $20\%$ lower than the baseline.}
    \label{fig_ls:jump_torques}
\end{figure}

The trajectories generated with the method described in Sec. \ref{sec_ls:sup} are stabilized by the controller presented in Sec. \ref{sec_ls:qp_controller}. The desired accelerations and contact wrenches are turned into a set of desired joint torques commanded to the Atlas humanoid robot.

The task consists in stepping onto a platform whose height is about 31$\mathrm{cm}$, as showed in Figure \ref{fig:atlasShowingOff}. We noticed that this height is already enough to have the robot pushing the leg joint limits during the swing phase. The robot starts at a distance of about 30$\mathrm{cm}$ from the platform and performs a 55$\mathrm{cm}$ step forward on the platform. Once the robot is close to the platform, the optimization is started using the current robot state as initialization. The optimization problem is solved using a desktop computer equipped with a 5$^{th}$ generation Intel\textsuperscript{\textregistered} Core i7@3.3GHz processor and 64GB RAM memory, running Ubuntu 16.04. \texttt{Ipopt} is set up to use the \texttt{mumps} \citep{amestoy2000mumps} solver. A solution is generated after about 1.5$\mathrm{s}$. This time can be reduced up to a factor 3 by using the linear solver \texttt{ma27} \citep{hsl2007collection} instead of \texttt{mumps}. Nevertheless, considering that the trajectories are generated only once before the beginning of the motion, this improvement is not necessary at this stage.

The length of each phase is set to $N=30$ instants. We noticed that this value is a good trade-off between the computational time and the smoothness of the generated trajectories. The maximum duration of a phase is equal to $2.3\mathrm{s}$, thus corresponding to a maximum $\mathrm{dt} = 77\mathrm{ms}$.

Amongst the set of variables $\bm{\chi}$ constituting the solution provided by the solver, we pack the CoM state into a desired trajectory. In addition, the timings $T_i$ are used to determine the single and double support phase durations. Fig. \ref{fig_ls:jump_comx} and \ref{fig_ls:jump_comy} show the CoM position tracking along the $x$ and $y$ direction. Figure \ref{fig_ls:jump_comz} presents the tracking of the CoM height. Note that the CoM is lowered right before performing the step-up. This instant is shown in the snapshots of Fig. \ref{fig_ls:snapshots}. We believe such desired motion arises to achieve sufficiently high vertical velocity, see Fig. \ref{fig_ls:jump_comVelocityz}, while considering the limits on the leg length. By performing this motion, the robot gains momentum, requiring a lower torque on the leading knee.

In order to provide a comparison, we performed the same task without prescribing any CoM trajectory, thus adopting the motion generation techniques presented in \citep{seyde2018inclusion}. To facilitate the comparison, we impose the same phase durations. Fig. \ref{fig_ls:jump_torques} shows the torque profile measured on the left knee when performing the step-up. We focus on this joint since is the one witnessing the highest torque expenditure during the step-up motion. In particular, using the controller presented in \citep{koolen2016balance}, the maximum value of this torque reaches $591\mathrm{Nm}$, while it is reduced to $478\mathrm{Nm}$ when using the method presented in this chapter. This corresponds to a nearly $20\%$ maximal torque reduction. Such result comes at the cost of a higher torque on the trailing knee. In order to facilitate the step-up motion, the right leg is much more exploited compared to the baseline, with a knee torque of nearly $430\mathrm{Nm}$.

Since the trajectories are computed offline, small disturbances can induce the robot to a fall. Robustness is achieved by limiting the CoP to a smaller region compared to the real robot foot in the planning phase. In this way, there is margin in the QP controller for counteracting disturbances while executing those trajectories. During the experiment described above, the foot dimension is set to $30\%$ of the original size. 

\begin{figure}[tpb]
 	\centering
 	\includegraphics[width=0.9\columnwidth]{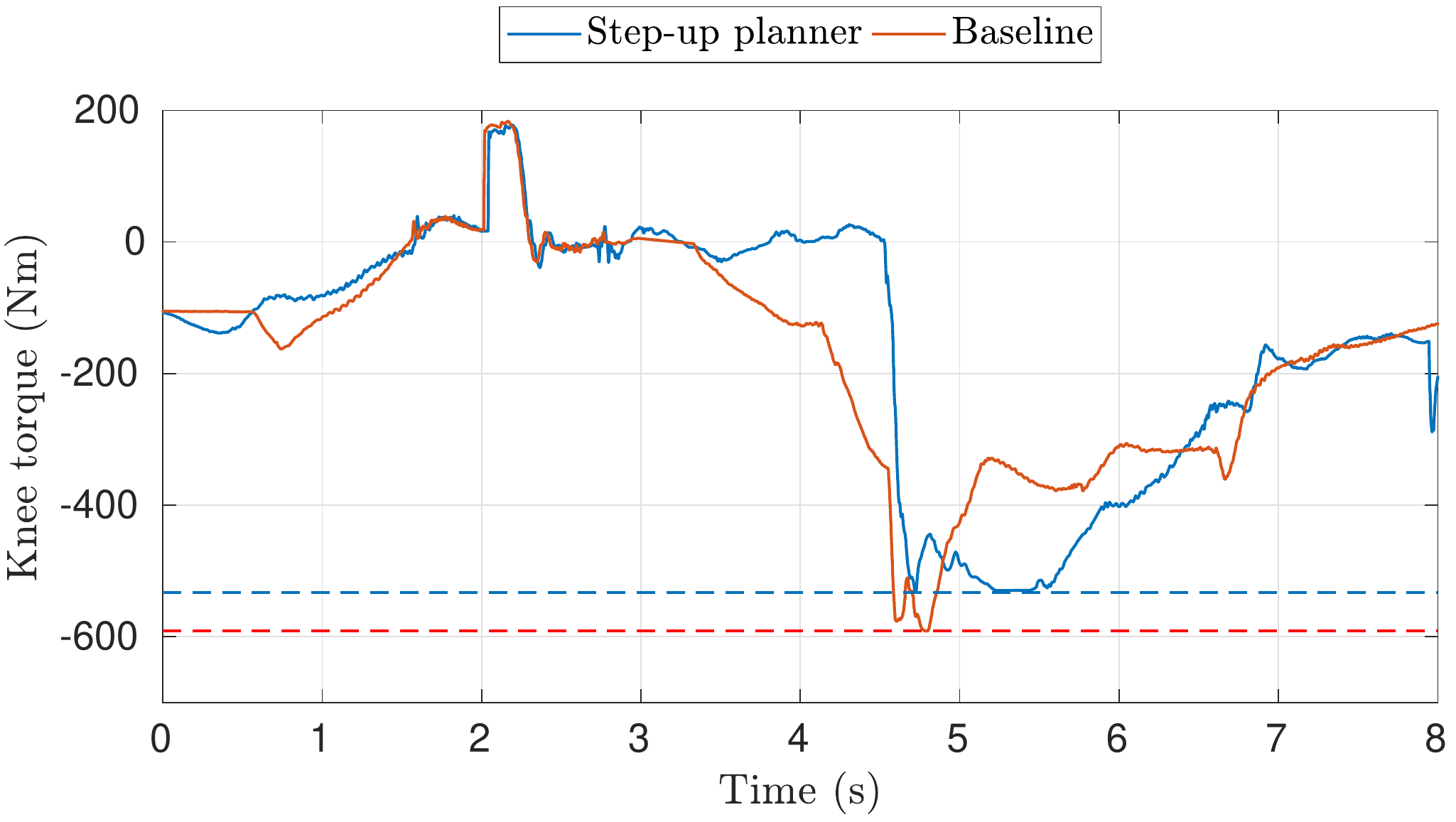}
 	\caption{Leading knee torques when the static friction is reduced to 0.5. Here the reduction of the maximum value is about 10\%.}
 	\label{fig_ls:reliable_torques}
 \end{figure}

The planned step-up is a dynamic motion which requires a high vertical velocity of the CoM. We noticed that the static friction coefficient $\mu_s$ has a particular effect in determining how ``dynamic'' is the planned motion. Intuitively, it limits the angle between the normal to the ground and the pendulum. A small angle corresponds to a conservative motion where the projection of the CoM on the ground lies well inside the support polygon. In the experiment shown above, the static friction coefficient is set to 0.7. By reducing it to $0.5$ the motion is more conservative and robust, even though the effectiveness of the method is reduced to a maximum torque reduction of 10\%. As shown in Fig. \ref{fig_ls:reliable_torques}, the knee torque reaches a peak of $532\mathrm{Nm}$.

 \begin{figure*}[tpb]
 	\centering
 	\subfloat[$x$ direction.]{ 	\includegraphics[width=0.9\columnwidth]{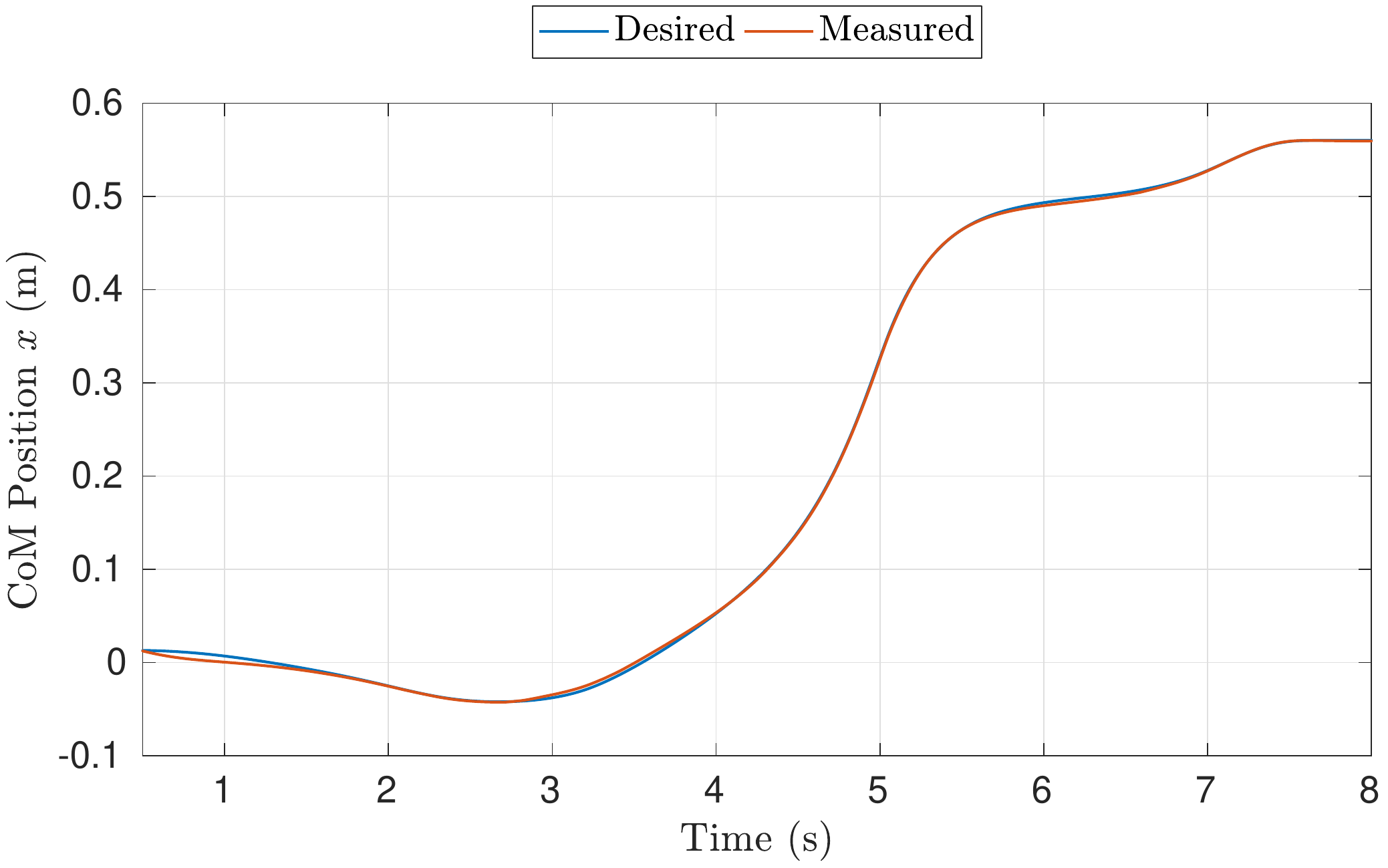} \label{fig_ls:sim_comx}}
 	
 	\subfloat[$y$ direction.]{ 	\includegraphics[width=0.9\columnwidth]{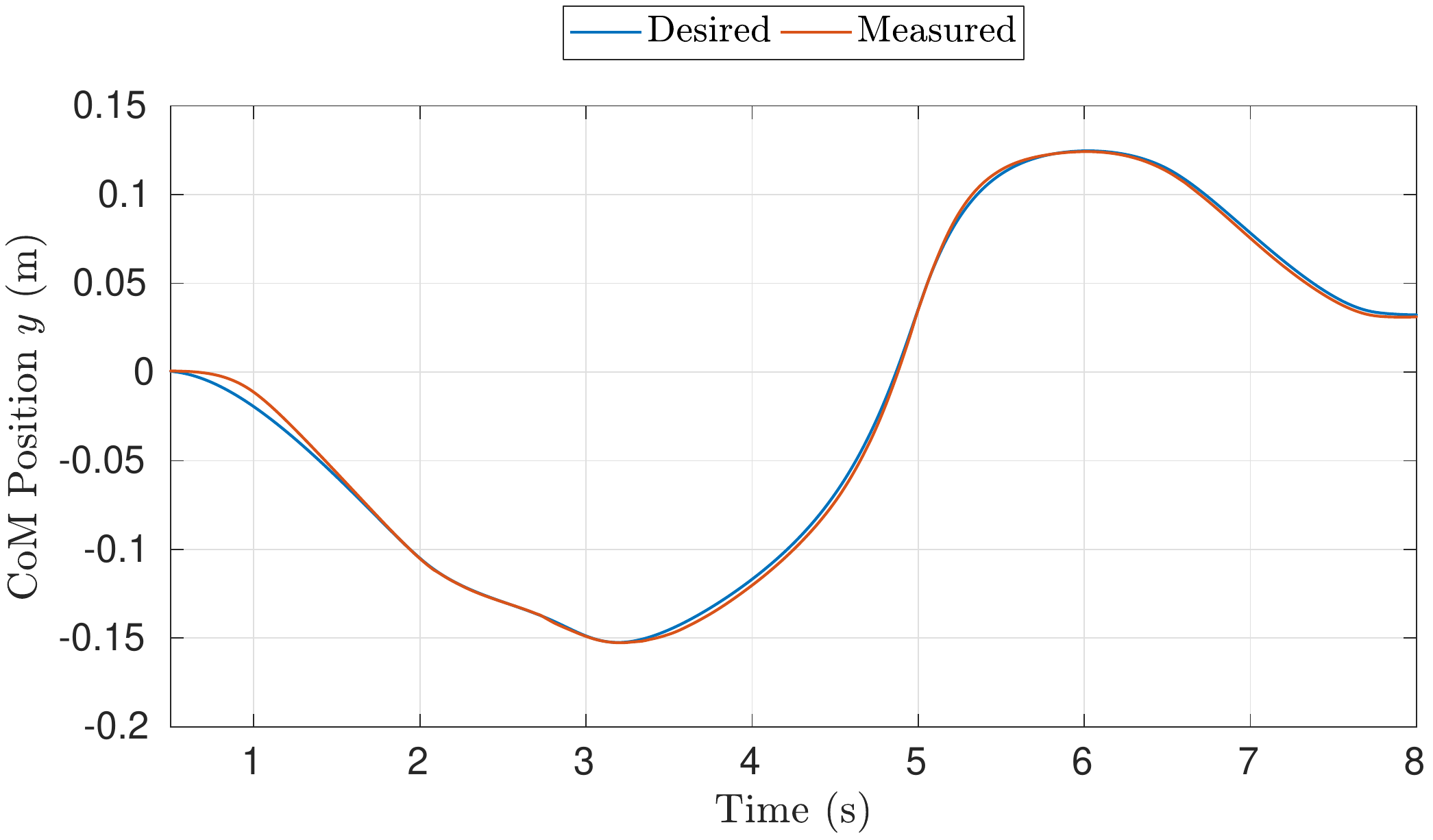}  \label{fig_ls:sim_comy}}
\end{figure*}
\begin{figure*}[tpb]
	\ContinuedFloat
 	\subfloat[$z$ direction.]{ 	\includegraphics[width=0.9\columnwidth]{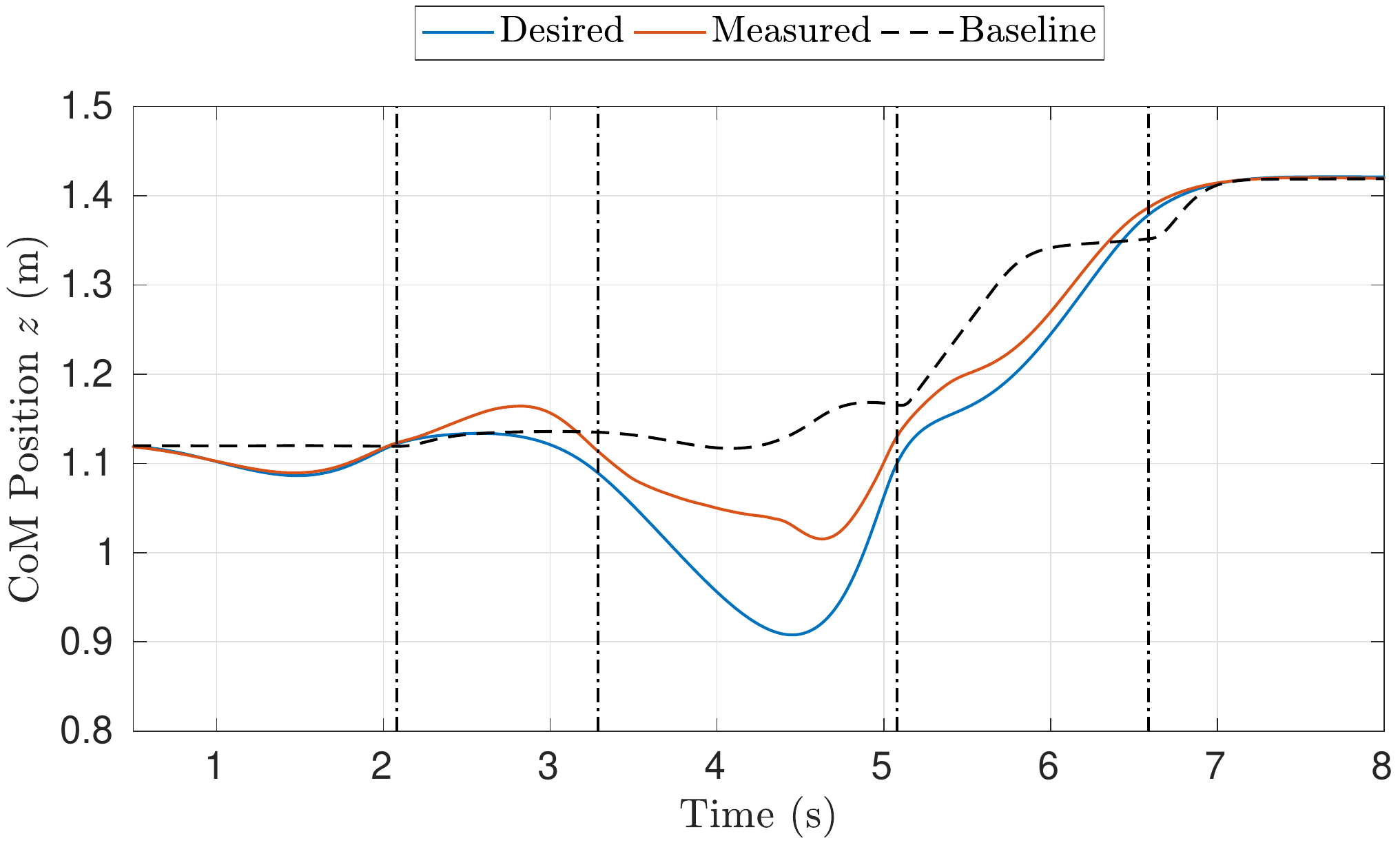}  	\label{fig_ls:sim_comz}}

 	\caption{Tracking of the CoM position during the simulation experiment.}
 	
 \end{figure*}
 \begin{figure}[tpb]
 	\centering
 	\includegraphics[width=0.9\columnwidth]{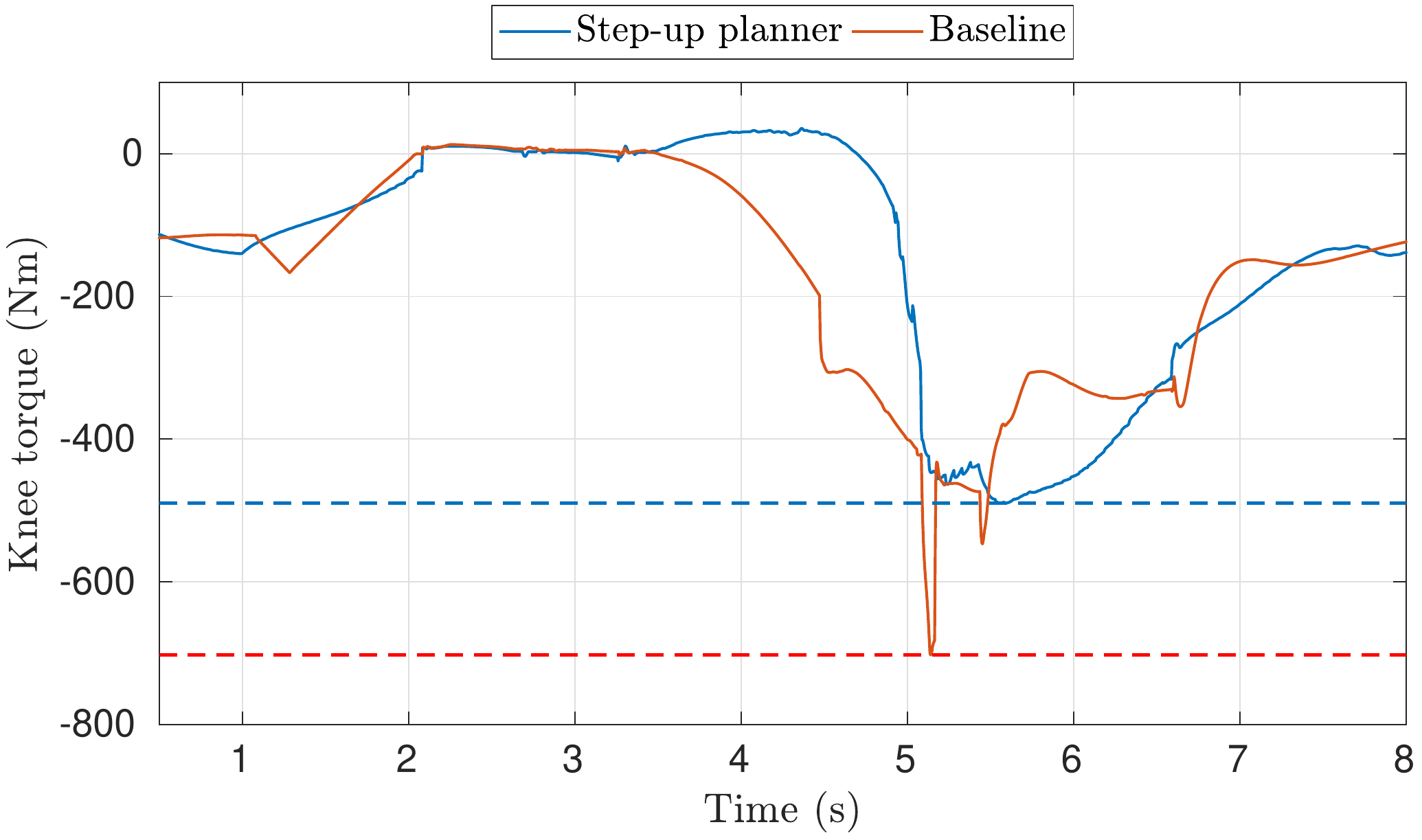}
 	\caption{Comparison of measured knee torques during the simulation experiment.}
 	\label{fig_ls:sim_torques}
 \end{figure}

When performing experiments on the real robot, the poor tracking of the desired trajectories limits the robustness and reproducibility of the results. Since the trajectories are computed offline, a poor tracking limits their efficacy. In the majority of the experiments, the reduction of the maximum torque consisted of about $10\%$. Anyhow, we believe this still represents an interesting achievement. It shows that trajectory optimization techniques have an effect in reducing the maximum effort required to a $150\mathrm{kg}$ robot when performing a motion at the limits of its reachable workspace.

When performing the same test in simulation, the tracking on the planar directions is much better, as shown in Fig. \ref{fig_ls:sim_comx} and \ref{fig_ls:sim_comy}, while along the $z$ direction, it is somehow comparable, Fig. \ref{fig_ls:sim_comz}. The improved tracking helps in maintaining balance, allowing more dynamic motions. The figures shown are generated with a static friction coefficient equal to 1.0, while the timings are kept equal to the real robot experiments. The baseline is obtained in the same way, leaving to the controller the generation of desired trajectories given a defined sequence of footsteps. As it can be noticed from Fig. \ref{fig_ls:sim_torques}, when the robot switches from double to single support the knee torque has a spike of $702\mathrm{Nm}$. Given the coincidence with the change of support configuration, the source of such spike is most probably the controller itself. Since, in simulation, the torque control is almost perfect, this spike is reflected also in the measured data. Conversely, on the real robot, any spike on the desired torques is smoothed due to the actuator dynamics. Nevertheless, when tracking the trajectories provided by the planner, there is no spike and the maximum torque is at $489\mathrm{Nm}$.

The weights adopted in the cost function are the same in simulation and on the real robot experiments. The tuning process consists in adding one task at a time starting from the most important, i.e. $\Gamma_{x_d}$, which is assigned an arbitrary weight. 
\section{Conclusions} \label{sec_ls:conclusions}
This chapter presents a method to generate trajectories for large step-up motions with humanoid robots. Its effectiveness has been tested on the IHMC Atlas humanoid robot. Experiments carried on the real platform showed that such trajectories can reduce the maximum torque required to the knee up to $20\%$. 
The method has been studied for large step-ups, but its applicability is not limited to this domain. Indeed, its formulation enables the planning of motions involving flight phases, hence requiring dynamic movements. This represents a fascinating future work.
The desired trajectories are computed offline right before starting the step-up motion. This reduces the robustness of the planned trajectories, strongly relying on the performances of the low-level controller. Poor tracking of the planned trajectories limits also their efficacy in reducing the maximum knee torque.

Both Chapter \ref{chap:steprecovery} and \ref{chap:iros_walking} use predictive techniques which are successful only if the output is applied on the robot at every control iteration. Instead, in this chapter we use Multiple Shooting in a trajectory optimization framework. Since the trajectories are computed off-line, before performing the motion, the computational time is not a major obstacle. Indeed, this planner is the slowest compared to those presented in the previous two chapters. Nevertheless, the time horizon is also consistently larger. In addition, the optimization problem is \emph{non-linear}, both in the costs and in the constraints. Thus, it cannot be reformulated as a QP problem, which is usually easier to solve. 

It is worth comparing the approach to angular momentum. The linear approximation adopted in Sec. \ref{sec:ang_mom} here is not applicable because of the much larger control horizon. As a consequence, we constrain its derivative to be null, similarly to the LIP model presented in Sec. \ref{sec:zmppreview}. 

In the next part, we perform a complete modeling of the humanoid robot in contact with the environment. In particular, we consider the full robot kinematics and its centroidal dynamics, without any assumption neither on the CoM height, nor on the angular momentum. The interaction between the robot and the walking surface is modeled explicitly through different contact parametrizations. The approach does not need a predefined contact sequence. Nonetheless, walking patterns emerge automatically by specifying a minimal set of references, such as a constant desired Center of Mass velocity and a reference point on the ground. This new planner is conceived to be used at low frequency, allowing the insertion of another control layer. 
At the same time, it can be iterated to produce off-line trajectories in joint space, which are tested on the iCub humanoid robot.

\epigraphhead[500]{\begin{quotation}
		{\footnotesize
			\noindent\emph{Your assumptions are your windows on the world. Scrub them off every once in a while, or the light won't come in.}
			\begin{flushright}
				Isaac Asimov
			\end{flushright}
		}
	\end{quotation}}
\part{Predictive Control Based on Complete Models} \label{part:dynamic_planner}

\chapter{Modeling of a Humanoid with Complementarity Conditions} \label{chap:modeling_dp}
In Part \ref{part:applications} the focus is on the application of predictive controllers to locomotion problems. Approximations are introduced at various stages. In this part, we aim at adopting a model as detailed as possible.
We start describing in this chapter the dynamical equations and the algebraic constraints which model the system under control, namely a legged robot instantiating contacts with the ground. It can be noticed that several inequality conditions are also employed. As a consequence, the resulting system is an inequality-constrained DAE, as introduced in Sec. \ref{sec:oc_basics}. 

Sec. \ref{sec:contacts_interface} introduce some preliminary concepts we adopt to describe a humanoid robot in contact with the environment. The definition of complementarity constraints, as introduced in Sec. \ref{sec:soa_trajectory}, is fundamental and tackled in Sec. \ref{sec:contact_parametrization}. The following sections are devoted to other aspects critical to define a meaningful model for whole-body trajectory generation. Finally, Sec. \ref{sec:complete_dae} presents the complete model.

\section{Preliminaries} \label{sec:contacts_interface}

Humanoid robots are generally equipped with feet of finite size. The availability of a flat surface in contact with the ground increases the capabilities of walking controllers to cope with external disturbances. On the other hand, when performing a step, the foot can impact the ground in a not flat configuration, reducing the amount of contact wrenches obtainable from the ground. Conversely, toe-off motions can be used to increase the work-space available during double support phases \citep{griffin2018straight}. Given these reasons, all the various contact configurations should be taken into account when planning step motions. By considering the foot as a flat surface, it would be required to model how it can land on the walking surface, also limiting the set of possible wrenches depending on the situation. As an example, in case of line contact, no torque can be exerted along that line, as in a classical hinge mechanism. 

In order to reduce the complexity, it is possible to consider the foot as composed by a set of points, for example (but not limited to) four points located at the corners of the foot. As an example, this approach is adopted in \citep{wensing2013generation,dai2014whole, caron2017make} and in Chapter \ref{chap:stepups}. The advantage is that the several contact configurations can be modeled independently, depending on the number of points in contact.

A pure force is supposed to be applied on each of the contact points. In case of four points, twelve variables are used to define a six dimensional quantity, i.e. the resulting contact wrench acting on the foot. This is a drawback that will be addressed later in Sec. \ref{sec:forceRegularization}. Appendix \ref{ap:four_forces} is devoted to the computation of these forces starting from a generic wrench.

Define ${}_i\bm{p} \in \mathbb{R}^3$ as the $i-$th contact point location in an inertial frame $\mathcal{I}$, and ${}_i\bm{f} \in \mathbb{R}^3$ as the force exerted on that point. Such force is expressed on a frame located in ${}_i\bm{p}$ and with orientation parallel to $\mathcal{I}$. We assume to have full control over the derivative of both these quantities:
\begin{IEEEeqnarray}{RCL}
	\phantomsection \IEEEyesnumber \label{eq:force_velocity_control}
	{}_i\dot{\bm{p}} &=& \bm{u}_{{}_ip}, \IEEEyessubnumber \label{eq:velocity_control}\\
	{}_i\dot{\bm{f}} &=& \bm{u}_{{}_if}, \IEEEyessubnumber
\end{IEEEeqnarray}
where $\bm{u}_{{}_ip}, ~ \bm{u}_{{}_if} \in \mathbb{R}^3$ are the control inputs for the $i$-th contact point. Since contact points are not supposed to penetrate the walking ground, we can impose $h({}_i \bm{p}) \geq 0$.

The force ${}_i\bm{f}$ results from the interaction of the contact point with the ground, hence it is limited. Being a reaction force, its normal component with respect to the walking ground is supposed to be non-negative. In particular,
\begin{equation}
	\mathbf{n}({}_i\bm{p})^\top{}_i\bm{f} \geq 0.
\end{equation}
Additionally, we focus on the case where the contact force is not enough to overcome the static friction:
\begin{equation} \label{eq:friction}
	\|\mathbf{t}({}_i\bm{p})^\top {}_i\bm{f} \| \leq \mu_s ~\mathbf{n}({}_i\bm{p})^\top{}_i\bm{f},
\end{equation}
where $\mu_s$ is the static friction coefficient. The meaning of functions $h(\cdot)$, $\mathbf{n}(\cdot)$ and $\mathbf{t}(\cdot)$ is available in Sec. \ref{sec:notation}.

For what concerns the robot control, we consider having full joint controllability. In particular, we assume the joint velocities as a control input:
\begin{equation}
\dot{\bm{s}} = \bm{u}_s.
\end{equation}
This assumption may seem unrealistic, but it allows for the insertion of an additional control loop. Joint velocities, together with contact vertex positions and forces, can be considered as a reference to one of the whole-body controllers presented in Sections \ref{sec:sot} or \ref{sec_ls:qp_controller}.

Finally, the base rotation included in $\bm{q}$ (see Sec. \ref{sec:modelling}) is vectorized using the quaternion parametrization. The corresponding unitary quaternion is called ${}^\mathcal{I}\bm{\rho}_B \in \mathbb{H}$. The base position is indicated with the symbol ${}^\mathcal{I}\bm{p}_B \in \mathbb{R}^3$. The equations governing the dynamical evolution of the base are as follows:
\begin{IEEEeqnarray}{RCL}
	\IEEEyesnumber \phantomsection
	{}^\mathcal{I}\dot{\bm{p}}_B &=& {}^\mathcal{I}\bm{R}_B {}^B\bm{v}_{\mathcal{I},B} ,\IEEEyessubnumber \label{eq:quaternionLeftDer}\\
	{}^\mathcal{I}\dot{\bm{\rho}}_B &=& \bm{u}_\rho. \IEEEyessubnumber \label{eq:quaternionRotationDerivative}
\end{IEEEeqnarray}
${}^B\bm{v}_{\mathcal{I},B} \in \mathbb{R}^3$, $\bm{u}_\rho \in \mathbb{R}^4$ and $\bm{u}_s \in \mathbb{R}^n$ are control inputs, defining the base linear velocity, the quaternion derivative and the joints velocity, respectively. More specifically, ${}^B\bm{v}_{\mathcal{I},B}$ is the linear part of ${}^B\bm{V}_{\mathcal{I},B} \in \mathbb{R}^6$ the \emph{left-trivialized} (i.e. measured in body coordinates, see Sec. \ref{sec:trivializatons} for details) base velocity.

\section{Contact parametrization} \label{sec:contact_parametrization}

Given its reactive nature, the contact force ${}_i\bm{f}$ applied to the $i-$th contact point is supposed to be not-null only if the point is in contact with the walking surface. This condition could be represented by the following equality:
\begin{equation}\label{eq:complementarity}
	h({}_i \bm{p}) ~\mathbf{n}({}_i\bm{p})^\top{}_i\bm{f} = 0.
\end{equation}
Such a constraint can be difficult to tackle in an optimization framework. This is due to the fact that the feasible set is only constituted by two lines, namely $h({}_i \bm{p}) = 0$ and $\mathbf{n}({}_i\bm{p})^\top{}_i\bm{f} = 0$, which are intersecting in the origin. In particular, at this point, the constraint Jacobian is singular, thus violating the \emph{linear independence constraint qualification} (LICQ), on which most off-the-shelf solvers rely upon, Sec. \ref{sec:nlp}.

We present three approaches aimed at mitigating this problem: 
\begin{itemize}
	\item Relaxed complementarity;
	\item Dynamically enforced complementarity;
	\item Hyperbolic secant in control bounds.
\end{itemize}

\subsection{Relaxed complementarity}\label{sec:relaxed_complementarity}
The condition defined in Eq. \eqref{eq:complementarity} can be \emph{relaxed} by noticing that both $h({}_i \bm{p})$ and $\mathbf{n}({}_i\bm{p})^\top{}_i\bm{f}$ are positive quantities. Hence, instead of using an equality condition, we can upper-bound their product with a small positive constant $\epsilon \in \mathbb{R}^+$:
\begin{equation}\label{eq:classical_complementarity}
	h({}_i \bm{p}) ~\mathbf{n}({}_i\bm{p})^\top{}_i\bm{f} \leq \epsilon.
\end{equation}
Through this simple modification, the feasibility region increases in dimension. In addition, if the inequality constraint is active at the optimum, the Jacobian is no longer singular. This approach is pretty common in literature, corresponding to the use of ``bounded'' slack variables \citep{betts2010practical}.

\subsection{Dynamically enforced complementarity}\label{sec:dynamical_complementarity}
The complementarity constraints are algebraic conditions applied to the dynamical system described in Eq. \eqref{eq:force_velocity_control}, thus they can be enforced using a Baumgarte stabilization method \citep{baumgarte1972stabilization}. In particular, we impose the complementarity constraints to converge dynamically to zero. This result is achieved by setting the constraint \emph{dynamics} as follows:
\begin{equation}
	\frac{\dif}{\dif t} \left(h({}_i \bm{p}) ~\mathbf{n}({}_i\bm{p})^\top{}_i\bm{f}\right) = -K_\text{bs}\left(h({}_i \bm{p}) ~\mathbf{n}({}_i\bm{p})^\top{}_i\bm{f}\right)
\end{equation}
with $K_\text{bs} \in \mathbb{R}^+$ a positive gain. Hence, the product $h({}_i \bm{p}) ~\mathbf{n}({}_i\bm{p})^\top{}_i\bm{f}$ would exponentially decrease to zero at a rate dependent on $K_\text{bs}$. We can expand the time derivative as follows:
\begin{equation}
	\frac{\dif}{\dif t}\Big(\cdot\Big) =  \frac{\dif}{\dif t}\left(h({}_i \bm{p})\right)\mathbf{n}({}_i\bm{p})^\top{}_i\bm{f} + h({}_i \bm{p}){}_i\bm{f}^\top \frac{\dif}{\dif t}\left(\mathbf{n}({}_i\bm{p})\right) + h({}_i \bm{p}) ~\mathbf{n}({}_i\bm{p})^\top{}_i\dot{\bm{f}},
\end{equation}
where we exploited the fact that Eq. \eqref{eq:classical_complementarity} is scalar to ease the computation of each term.
We can substitute the time derivative of the $h(\cdot)$ and $\mathbf{n}(\cdot)$ functions with the following relations:
\begin{IEEEeqnarray}{RCL}
	\phantomsection \IEEEyesnumber
	\frac{\dif}{\dif t}\left(h({}_i \bm{p})\right) &=& \frac{\partial}{\partial {}_i \bm{p}}\left(h({}_i \bm{p})\right){}_i\bm{\dot{p}}, \IEEEyessubnumber \\
	\frac{\dif}{\dif t}\left(\mathbf{n}({}_i\bm{p})\right) &=& \frac{\partial}{\partial {}_i \bm{p}}\left(\mathbf{n}({}_i\bm{p})\right){}_i\bm{\dot{p}}. \IEEEyessubnumber
\end{IEEEeqnarray}

For simplicity, let us define $\zeta$ as follows:
\begin{equation*}
	\zeta := \frac{\partial}{\partial {}_i \bm{p}}\left(h({}_i \bm{p})\right){}_i\bm{\dot{p}}~\mathbf{n}({}_i\bm{p})^\top{}_i\bm{f} + h({}_i \bm{p}){}_i\bm{f}^\top\frac{\partial}{\partial {}_i \bm{p}}\left(\mathbf{n}({}_i\bm{p})\right){}_i\bm{\dot{p}} + h({}_i \bm{p}) ~\mathbf{n}({}_i\bm{p})^\top{}_i\dot{\bm{f}}.
\end{equation*}
Finally, we obtain the condition
\begin{equation}\label{eq:dynamic_complementarity_equality}
	\zeta = -K_\text{bs}\left(h({}_i \bm{p}) ~\mathbf{n}({}_i\bm{p})^\top{}_i\bm{f}\right).
\end{equation}
In case of planar ground, we have the following relations:
\begin{IEEEeqnarray}{RCL}
	\phantomsection \IEEEyesnumber
	h({}_i \bm{p}) &=& \bm{e}_3^\top {}_i \bm{p}, \IEEEyessubnumber\\
	\frac{\partial}{\partial {}_i \bm{p}}\left(h({}_i \bm{p})\right) &=& \bm{e}_3^\top, \IEEEyessubnumber\\
	\mathbf{n}({}_i\bm{p}) &=& \bm{e}_3, \IEEEyessubnumber \\
	\frac{\partial}{\partial {}_i \bm{p}}\left(\mathbf{n}({}_i\bm{p})\right) &=& \bm{0}_{3\times 3}. \IEEEyessubnumber	
\end{IEEEeqnarray}
In fact, in this specific case, the normal direction and the height of the point coincide with the $z$-direction and the corresponding coordinate, respectively.
Hence, in the planar case, $\zeta$ would reduce to 
\begin{equation}
	\zeta_\text{planar} = {}_i\dot{\bm{p}}_z \cdot{}_i\bm{f}_z + {}_i\bm{p}_z \cdot {}_i\dot{\bm{f}}_z.
\end{equation}
We can relax Eq. \eqref{eq:dynamic_complementarity_equality} by exploiting again the fact that the product $h({}_i \bm{p}) ~\mathbf{n}({}_i\bm{p})^\top{}_i\bm{f}$ is positive by construction. Henceforth, if we impose
\begin{equation}
	\zeta \leq -K_\text{bs}\left(h({}_i \bm{p}) ~\mathbf{n}({}_i\bm{p})^\top{}_i\bm{f}\right),
\end{equation}
we still have exponential convergence, at a rate which is higher or equal to the one specified by $K_\text{bs}$. Finally, similarly to Eq. \eqref{eq:classical_complementarity}, we add a further relaxation through a positive number $\varepsilon \in \mathbb{R}$ to increase the feasibility region,
\begin{equation}
	\zeta \leq -K_\text{bs}\left(h({}_i \bm{p}) ~\mathbf{n}({}_i\bm{p})^\top{}_i\bm{f}\right) + \varepsilon,
\end{equation}
obtaining the final version of the complementarity condition.

\subsection{Hyperbolic secant in control bounds}\label{sec:hyperbolic_secant}
We can impose Eq. \eqref{eq:complementarity} dynamically by enforcing the following set of constraints on the control input:
\begin{IEEEeqnarray}{RCLR}
	\phantomsection \IEEEyesnumber \label{eq:force_control_cases}
	-\bm{M}_f \leq &\bm{u}_{{}_if} & \leq \bm{M}_f     & \quad \text{if } h({}_i \bm{p}) = 0 \IEEEyessubnumber \label{eq:force_control_bounds}\\
			  &\bm{u}_{{}_if} & = -\bm{K}_f {}_i\bm{f} & \quad \text{if } h({}_i \bm{p}) \neq 0 \IEEEyessubnumber \label{eq:force_control_dissipation}
\end{IEEEeqnarray}
meaning that when the point is in contact, $\bm{u}_{{}_if}$ is free to take any value in $\left[-\bm{M}_f, \bm{M}_f\right]$ with $\bm{M}_f \in \mathbb{R}^3$ a (non-negative) control bound. On the other hand, if the contact point is not on the walking surface, the control input makes the contact force decreasing exponentially (Eq. \eqref{eq:force_control_dissipation}) at a rate depending on the positive definite control gain $\bm{K}_f \in \mathbb{R}^{3\times 3}$. Defining $\delta^*({}_i\bm{p})$ as a binary function such that
\begin{equation}
	\delta^*({}_i\bm{p}) = 
	\begin{cases}
	1 & \quad \text{if } h({}_i \bm{p}) = 0, \\
	0 & \quad h({}_i \bm{p}) \neq 0,
	\end{cases} 
\end{equation}
it is possible to write Eq. \eqref{eq:force_control_cases} as a set of two inequalities:
\begin{IEEEeqnarray}{RCL}
	\IEEEyesnumber \phantomsection \label{eq:force_control_full}
- \bm{K}_f \left(1 - \delta^*({}_i\bm{p})\right) {}_i\bm{f} - \delta^*({}_i\bm{p})\bm{M}_f &\leq& \bm{u}_{{}_if}, \IEEEyessubnumber \label{eq:force_control_lb}\\
- \bm{K}_f \left(1 - \delta^*({}_i\bm{p})\right) {}_i\bm{f} + \delta^*({}_i\bm{p})\bm{M}_f &\geq& \bm{u}_{{}_if}. \IEEEyessubnumber \label{eq:hyperbolic_complementarity_lb}
\end{IEEEeqnarray}
\begin{figure}[tpb]
	\centering
	\includegraphics[width=0.85\textwidth]{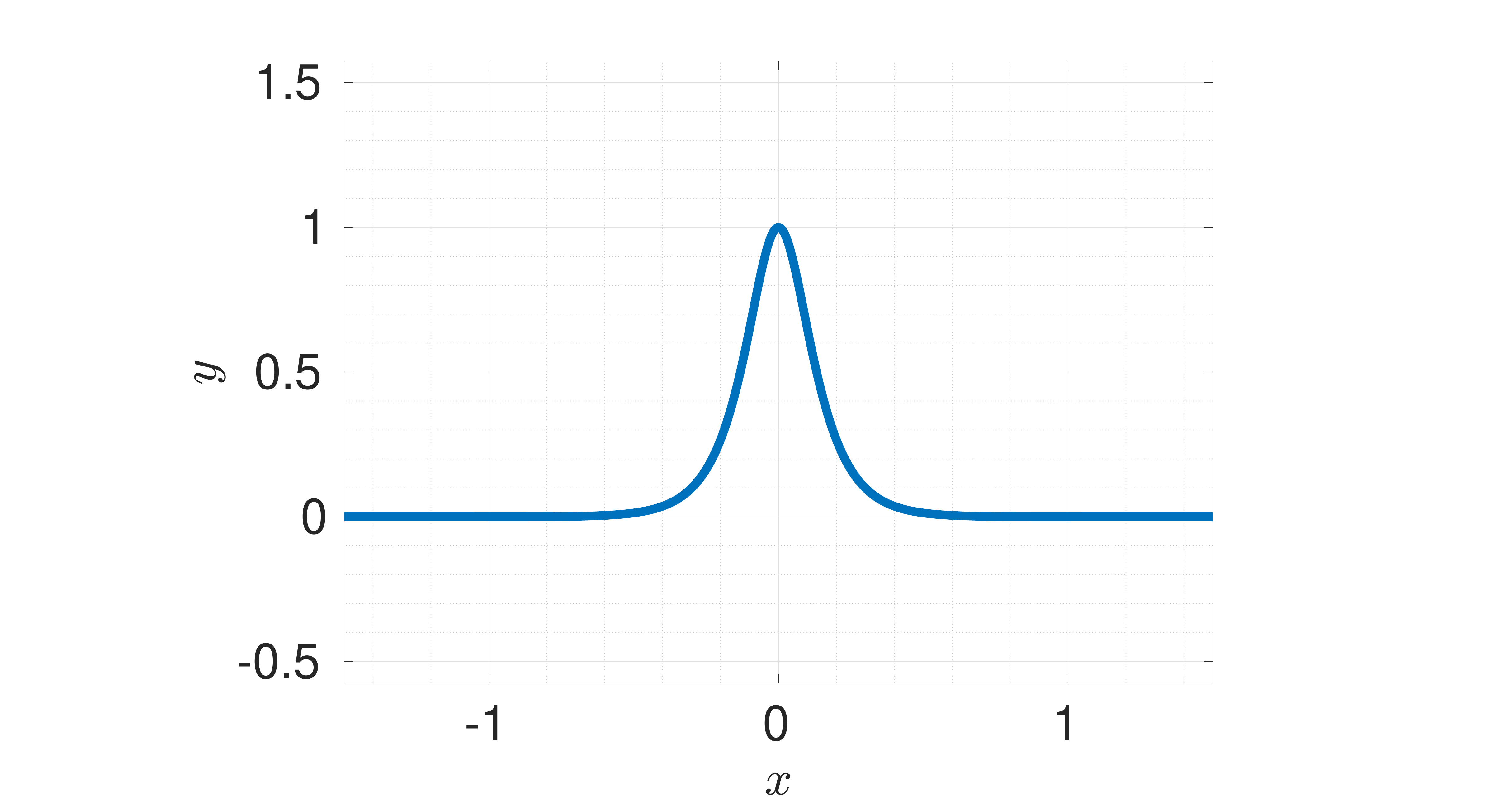} 
	\caption{The plot of $\text{sech}(10x)$.}
	\label{fig:sech}
\end{figure}
Even if $\delta^*({}_i\bm{p})$ would require the adoption of integer variables, it is possible to use a continuous approximation, $\delta({}_i\bm{p})$, namely the hyperbolic secant, an example of which is shown in Fig. \ref{fig:sech}:
\begin{equation}\label{eq:sech}
    \delta({}_i\bm{p}) = \text{sech}\left(k_h~h({}_i \bm{p})\right),
\end{equation}
where $k_h$ is a user-defined scaling factor. Notice that, when $\delta^*({}_i\bm{p}) = 0$, the bounds coincide and are equal to $-\bm{K}_f {}_i\bm{f}$.

As discussed in Sec. \ref{sec:dynamical_complementarity}, we can simplify the lower bound defined in Eq. \eqref{eq:force_control_lb}, allowing the force to decrease at a higher rate than the one given by Eq. \eqref{eq:hyperbolic_complementarity_lb}. Hence, we can rewrite Eq. \eqref{eq:force_control_full} as:
\begin{equation}\label{eq:force_control_final}
-\bm{M}_f \leq \bm{u}_{{}_if} \leq - \bm{K}_f \left(1 - \delta({}_i\bm{p})\right) {}_i\bm{f} + \delta({}_i\bm{p})\bm{M}_f.
\end{equation}
Given Eq. \eqref{eq:friction}, it is enough to apply any of these equations only to the force component normal to the ground: if it decreases to zero, also planar components have to vanish to satisfy friction constraints. Hence, we can simplify Eq. \eqref{eq:force_control_final} as follows:
\begin{equation}
	-\bm{e}_3^\top\bm{M}_f \leq \bm{e}_3^\top\bm{u}_{{}_if} \leq - K_{f,z} \left(1 - \delta({}_i\bm{p})\right) \mathbf{n}({}_i\bm{p})^\top {}_i\bm{f} + \delta({}_i\bm{p})\bm{e}_3^\top\bm{M}_f,
\end{equation}
with $K_{f,z}$ the corresponding element of $\bm{K}_f$.

\subsection{Summing up} \label{sec:complementarity_list}
We analyze different methods for expressing the complementarity constraints defined in Eq. \eqref{eq:complementarity}. In particular, we have:
\begin{itemize}
	\item $h({}_i \bm{p}) ~\mathbf{n}({}_i\bm{p})^\top{}_i\bm{f} \leq \epsilon$;
	\item $\zeta \leq -K_\text{bs}\left(h({}_i \bm{p}) ~\mathbf{n}({}_i\bm{p})^\top{}_i\bm{f}\right) + \varepsilon$;
	\item $-\bm{e}_3^\top\bm{M}_f \leq \bm{e}_3^\top\bm{u}_{{}_if} \leq - K_{f,z} \left(1 - \delta({}_i\bm{p})\right) \mathbf{n}({}_i\bm{p})^\top {}_i\bm{f} + \delta({}_i\bm{p})\bm{e}_3^\top\bm{M}_f$.
\end{itemize}

It is also realistic to assume the control inputs to be bounded, i.e. $-\bm{M}_f \leq \bm{u}_{{}_if} \leq \bm{M}_f$, while this is not necessary when using Eq. \eqref{eq:hyperbolic_complementarity_lb} since bounds are already included in the constraint.

Note that these conditions do not depend on the type of ground, which is considered rigid. The parameters involved do not have a direct physical meaning (like in compliant contact models), but rather determine the ``accuracy'' of the simulated behavior.

\section{Prevention of unrealizable motions for the contact points}
The complementarity conditions presented above cannot prevent the contact points to move on the walking plane when in contact. In fact, even if friction constraints defined in Eq. \eqref{eq:friction} are satisfied, the contact points are still free to move on the walking surface. Force and position variables are until now almost independent. It is possible to prevent planar motions when in contact by limiting the effect of the control input $\bm{u}_{{}_ip}$ along the planar components:
\begin{equation}\label{eq:planarControl}
\mathbf{t}({}_i\bm{p})^\top {}_i \dot{\bm{p}} = \tanh\left(k_t~h({}_i \bm{p})\right) \left[\bm{e}_1 ~ \bm{e}_2\right]^\top \bm{u}_{{}_ip}
\end{equation}
where $k_t \in \mathbb{R}$ is a user-defined scaling factor. Eq. \eqref{eq:planarControl} multiplies the control input along the planar direction to zero when $h({}_i\bm{p})$ is null and, at the same time, it reduces the velocity when the contact point is approaching the ground. It is possible to rewrite Eq. \eqref{eq:planarControl} as ${}_i\dot{\bm{p}} =  \bm{\tau}({}_i\bm{p}) \bm{u}_{{}_ip}$, where
\begin{equation}
\bm{\tau}({}_i\bm{p}) ={}^\mathcal{I}\bm{R}_{plane} ~ diag\left(\begin{bmatrix}
\tanh\left(k_t ~ h({}_i \bm{p})\right) \\
\tanh\left(k_t ~ h({}_i \bm{p})\right) \\
1
\end{bmatrix}\right).
\end{equation}
Note that, from now on, $\bm{u}_{{}_ip}$ is assumed to be defined in $plane$ coordinates. Thus, the normal component of the velocity is directly affected by $\bm{e}_3^\top \bm{u}_{{}_ip}$. Also, it is necessary to bound this control input, $\bm{u}_{{}_ip} \in \left[-\bm{M}_V, \bm{M}_V\right], \bm{M}_V \in \mathbb{R}^3$, to properly exploit the effect of the hyperbolic tangent. 

While each contact point is supposed to be independent from the control point of view, they all need to remain on the same surface and maintain a constant relative distance, since they belong to the same rigid body. At the same time, we want them to be within the workspace reachable by the robot legs. We can achieve both objectives with the following algebraic condition acting on each of the contact points:
\begin{equation}\label{eq:dp_point_consistency}
	{}_i\bm{p} = {}^\mathcal{I} \bm{H}_\text{foot} {}^\text{foot}{}_i \bm{p},
\end{equation} 
where ${}^\text{foot}{}_i \bm{p}$ is the (fixed) position of the contact point within the foot surface, expressed in foot coordinates. 
Here, the transformation matrix ${}^\mathcal{I} \bm{H}_\text{foot}$ would depend on the base position ${}^\mathcal{I}\bm{p}_B$, the base quaternion ${}^\mathcal{I}\bm{\rho}_B$ and the joint configuration $\bm{s}$. As a consequence, the full kinematics of the robot is taken into consideration.

\section{Momentum dynamics}
In Sec. \ref{sec:contacts_interface}, we consider the contact points as if they have the possibility of exerting a force with the environment. Here, we describe the effect of this forces on the robot through the centroidal dynamics introduced in Sec. \ref{sec:intro_momentum}. This choice is supported by the fact that the momentum dynamics depend only on the contact forces, their location and on the CoM position:
\begin{equation}
	{}_{\bar{G}} \dot{\bm{h}} = m\bar{\bm{g}} + \sum_i
	{}_{\bar{G}}\bm{X}^i \begin{bmatrix}
	{}_i\bm{f}\\
	\bm{0}_3
	\end{bmatrix} = m\bar{\bm{g}} + \sum_i
	\begin{bmatrix}
	\bm{\mathds{1}}_3 \\
	({}_i\bm{p} - \bm{x}_{\text{CoM}})^\wedge
	\end{bmatrix} {}_i\bm{f}.
\end{equation}

Compared to Eq. \eqref{eq:centroidal_momentum_dynamics}, we simplify the matrix ${}_{\bar{G}}\bm{X}^i$ since no torque is applied at the contact points.
We also need to make sure that the CoM position obtained by integrating Eq. \eqref{eq:com_from_momentum} corresponds to the one obtained via the joints variables. This is done through the following algebraic equation:
\begin{equation}\label{eq:comConsistency}
\bm{x}_\text{CoM} = \text{CoM}({}^\mathcal{I}\bm{p}_B, {}\mathcal{I}\bm{\rho}_B, s)
\end{equation} 
where $\text{CoM}({}^\mathcal{I}\bm{p}_B, {}\mathcal{I}\bm{\rho}_B, s)$ is the function mapping base pose and joints position to the CoM position, i.e. the right hand side of Eq. \eqref{eq:com_definition}. While this equation defines a link between the linear momentum and the joint variables, the same would not hold for the angular part. In other words, we need to model how the angular momentum affects the evolution of joints. To this end, we can exploit the Centroidal Momentum Matrix $\bm{J}_\text{CMM}$. In fact, we can extract the angular part of Eq. \eqref{eq:cmm_intro}, obtaining the following condition:
\begin{equation}\label{constr:CMM}
{}_{\bar{G}} \bm{h}^\omega = \left[\bm{0}_{3 \times 3} ~ \bm{\mathds{1}}_3\right]\bm{J}_\text{CMM} \bm{\nu}.
\end{equation}

Here, the system velocity $\bm{\nu}$ contains the base angular velocity ${}^B \bm{\omega}_{\mathcal{I},B}$. It can be substituted with the quaternion derivative through the map $\bm{\mathcal{G}}$. It is defined in \citep[Section 1.5.4]{graf2008quaternions} as 
\begin{equation} \label{eq:g_map_quaternion} 
\bm{\mathcal{G}}({}\mathcal{I}\bm{\rho}_B) = \begin{bmatrix} {-{}^\mathcal{I}\rho_{B,1}} & {{}^\mathcal{I}\rho_{B,0}} & {{}^\mathcal{I}\rho_{B,3}} & {-{}^\mathcal{I}\rho_{B,2}} \\ {-{}^\mathcal{I}\rho_{B,2}} & {-{}^\mathcal{I}\rho_{B,3}} & {{}^\mathcal{I}\rho_{B,0}} & {{}^\mathcal{I}\rho_{B,1}} \\ {-{}^\mathcal{I}\rho_{B,3}} & {{}^\mathcal{I}\rho_{B,2}} & {-{}^\mathcal{I}\rho_{B,1}} & {{}^\mathcal{I}\rho_{B, 0}} \end{bmatrix}
\end{equation}
such that 
\begin{equation} \label{eq:base_angular_velocity}
	{}^B \bm{\omega}_{\mathcal{I},B} = 2 \bm{\mathcal{G}}({}^\mathcal{I}\rho_B)\bm{u}_\rho.
\end{equation}
Hence, it depends on the same control input introduced in Eq. \eqref{eq:quaternionRotationDerivative}.

\section{The complete differential-algebraic system of equations} \label{sec:complete_dae}
By summarizing all the ODEs and algebraic conditions introduced in this chapter, we obtain the following inequality constrained DAE. \\
$\bullet$ Dynamical Constraints 
\begin{IEEEeqnarray}{RCLL}
	\IEEEyesnumber \phantomsection \label{constr:system_dynamics}
	{}_i\dot{\bm{f}} &=& \bm{u}_{{}_if}, & \forall \text{ contact point}, \IEEEyessubnumber \label{constr:force_derivative}\\
	{}_i\dot{\bm{p}} &=& \bm{\tau}({}_i\bm{p}) \bm{u}_{{}_ip}, & \forall \text{ contact point},  \IEEEyessubnumber \label{constr:position_derivative}\\
	{}_{\bar{G}} \dot{\bm{h}} &=& \IEEEeqnarraymulticol{2}{L}{m\bar{\bm{g}} + \sum_i
		\begin{bmatrix}
			1_3 \\
			({}_i\bm{p} - \bm{x}_{\text{CoM}})^\wedge
		\end{bmatrix} {{}_i\bm{f}}}, \IEEEyessubnumber\label{constr:momentum_derivative}\\	
	\dot{\bm{x}}_\text{CoM} &=& \IEEEeqnarraymulticol{2}{L}{\frac{1}{m}  \left({}_{\bar{G}} \bm{h}^p\right),}  \IEEEyessubnumber \label{constr:com_derivative}\\
	{}^\mathcal{I}\dot{\bm{p}}_B &=& {}^\mathcal{I}\bm{R}_B {}^B\bm{v}_{\mathcal{I},B}, \IEEEyessubnumber \label{constr:basePosDerivative}\\
	{}^\mathcal{I}\dot{\bm{\rho}}_B &=& \bm{u}_\rho, \IEEEyessubnumber \label{constr:quatDerivative}\\
	\dot{\bm{s}} &=& \bm{u}_s. \IEEEyessubnumber
\end{IEEEeqnarray}
$\bullet$ Equality Constraints 
\begin{IEEEeqnarray}{RCL}
	\IEEEyesnumber \phantomsection \label{constr:equalities}
	{}_i\bm{p} &=& {}^A \bm{H}_\text{foot} {}^\text{foot}{}_i \bm{p}, \quad \forall \text{ contact point}, \IEEEyessubnumber \label{constr:pointPositions}\\
	\bm{x}_\text{CoM} &=& \text{CoM}({}^\mathcal{I}\bm{p}_B, {}^\mathcal{I}\bm{\rho}_B, \bm{s}), \IEEEyessubnumber \\
	{}_{\bar{G}} \bm{h}^\omega &=& \left[0_{3 \times 3} ~ \mathds{1}_3\right]\bm{J}_\text{CMM} \begin{bmatrix}
		{}^B\bm{v}_{\mathcal{I},B} \\
		2 \mathcal{\bm{G}}({}^\mathcal{I}\bm{\rho}_B)\bm{u}_\rho\\
		\bm{u}_s
	\end{bmatrix}, \IEEEyessubnumber  \\
	\|{}^\mathcal{I}\bm{\rho}_B\|^2 &=& 1. \IEEEyessubnumber 
\end{IEEEeqnarray}
$\bullet$ Inequality Constraints, applied for each contact point
\begin{IEEEeqnarray}{RCL}
	\IEEEyesnumber \phantomsection \label{constr:inequalities}
	\mathbf{n}({}_i\bm{p})^\top{}_i\bm{f} &\geq& 0, \IEEEyessubnumber\\
	\|\mathbf{t}({}_i\bm{p})^\top {}_i\bm{f} \| &\leq& \mu_s ~\mathbf{n}({}_i\bm{p})^\top{}_i\bm{f}, \IEEEyessubnumber\\
	-\bm{M}_V &\leq& \bm{u}_{{}_ip} \leq \bm{M}_V, \IEEEyessubnumber\\
	-\bm{M}_f &\leq& \bm{u}_{{}_if} \leq \bm{M}_f ,\IEEEyessubnumber\\
	h({}_i \bm{p}) &\geq& 0, \IEEEyessubnumber\\
	\IEEEeqnarraymulticol{3}{C}{\text{Complementarity, see Sec. \ref{sec:complementarity_list}}}.\IEEEyessubnumber
\end{IEEEeqnarray}

\chapter{The Whole-Body Non-Linear Predictive Controller} \label{chap:tasks}
Given the model presented in Chapter \ref{chap:modeling_dp}, we defin[] in Sec. \ref{sec:additional_constraints} a set of additional constraints, while, in the following sections, we describe a set of tasks aimed at the generation of walking trajectories. They compose the full optimal control problem presented in Sec. \ref{sec:oc}. More specifically, the additional constraints presented in this chapter are inequalities which prevent the generation of undesirable or unfeasible motions for the robot. On the other hand, the tasks are the mathematical description of the goals we want to achieve when planning for the walking motions. Finally, Sec. \ref{sec:dp_software_infrastructure} presents the software infrastructure used to solve the optimal control problem.

\section{Walking specific constraints} \label{sec:additional_constraints}
While taking steps, we need to make sure that the robot legs do not collide with each other. Self collisions constraints are usually hard to consider and may slow down consistently the determination of a solution. A simpler solution to avoid self-collisions between legs consists of avoiding the left one to be on the right of the other leg, even if cross steps are then forbidden. We assume the frame attached to the right foot to have the positive $y-$direction pointing toward left. In this case, it would be sufficient to impose the $y-$component of the ${}^{r}\bm{x}_{l}$ (i.e. the relative position of the left foot expressed in the right foot frame) to be greater than a given quantity, i.e.:
\begin{equation}
\bm{e}_2^\top {}^{r}\bm{x}_{l} \geq d_\text{min}.
\end{equation}

In other cases, we want to prevent the solver to find a solution requiring the robot to raise the swing leg too much. Once the leg is raised, there are infinite motions which generate the same centroidal momentum. Too wide motions of the swing leg may cause other self-collision, especially between the arms and the thigh. Hence, we set an upper-bound on the difference between the height of the two feet. In particular, we can consider the mean position of every contact point to simplify the definition of the constraint:
\begin{equation}
-M_{hf} \leq e_3^\top\left({}_{\#}\bm{p}_l - {}_{\#}\bm{p}_r\right) \leq M_{hf}
\end{equation}
where ${}_{\#}\bm{p}_l$ and ${}_{\#}\bm{p}_r$ are the mean positions of all the contact points of the left and right foot respectively, i.e. ${}_\# \bm{p}_\circ =\frac{1}{n_p} \sum^{n_p}_i {}_i\bm{p}_\circ$. $n_p$ is the number of contact points in a single foot and $M_{hf} \in \mathbb{R}^+$ is the constraint upper-bound. 

Some additional constraints can be considered:
\begin{IEEEeqnarray}{RCL}
	\IEEEyesnumber
	x_{\text{CoM},z \text{ min}} &\leq& \bm{e}_3^\top{\bm{x}_{\text{CoM}}},  \IEEEyessubnumber \label{constr:com_height_limit}\\
	-\bm{M}_{h_\omega} &\leq& {}_{\bar{G}} \bm{h}^\omega \leq \bm{M}_{h_\omega}. \IEEEyessubnumber \label{constr:angular_momentum_bounds}
\end{IEEEeqnarray}
Eq \eqref{constr:com_height_limit} avoids picking solutions which would bring the CoM position too close or even below the ground. Eq. \eqref{constr:angular_momentum_bounds} set a bound $\bm{M}_{h_\omega} \in \mathbb{R}^3$ to the angular momentum. These constraints avoid considering trajectories that would cause excessive motions or let the robot falling.

\section{Tasks in Cartesian space} \label{sec:cartesian_tasks}
In order to make the robot move toward a desired position, it is necessary to specify tasks in Cartesian space. 
\subsection{Contact point centroid position task}\label{sec:centroid_task}
We define as a task the L2 norm of the error between a point attached to the robot and its desired position in an absolute frame. Suppose we choose the CoM position as a target point. By moving its desired value forward in space, the robot could simply lean toward the goal without moving the feet. This undesired behavior may lead the robot to fall. It is possible to avoid the robot leaning forward by locating the target point on the feet instead of the CoM. In particular, we select the centroid of the contact points as target, thus avoiding specifying a desired placement for each foot:
\begin{equation}
\Gamma_{{}_\# p} = \frac{1}{2} \|{}_\# \bm{p} - {}_\# \bm{p}^*\|^2_{\bm{W}_\#},
\end{equation}
where ${}_\# \bm{p} =\frac{1}{2} ({}_{\#}\bm{p}_l + {}_{\#}\bm{p}_r)$ and ${}_\# \bm{p}^* \in \mathbb{R}^2$ is its desired value. 

\subsection{CoM linear velocity task}\label{sec:com_velocity_cost}
While walking, we want the robot to keep a constant forward motion. In fact, since foot positions are not scripted, it may be possible to plan two consecutive steps with the same foot. Requiring a constant forward velocity can help avoiding such phenomena. This task can be defined as:
\begin{equation}\label{eq:momentum_cost}
\Gamma_{{}_{\bar{G}} h^p} = \frac{1}{2} \|{}_{\bar{G}} \bm{h}^p - m  \dot{\bm{x}}_{\text{CoM}}^*\|^2_{\bm{W}_v}
\end{equation}
with $\dot{\bm{x}}_{\text{CoM}}^*$ a desired CoM velocity. It is possible to select and weigh the different directions separately through the weights $\bm{W}_v$.

\subsection{Foot yaw task}
The task on the centroid of the contact points, defined in Sec. \ref{sec:centroid_task} allows defining the direction the robot has to step. With this task, we specify at which angle the foot should be oriented with respect to the $z$-axis, in brief the foot yaw angle. Define $\gamma^*_\circ$ as the desired yaw angle for either the left or the right foot ($\circ$ is a placeholder). We construct a unitary vector $\bm{\ell}^*_{\circ} \in \mathbb{R}^2$ belonging to the $xy$-plane (of $\mathcal{I}$), oriented such that the angle with the $x$-axis of $\mathcal{I}$ corresponds to  $\gamma^*_\circ$. Its components are: 
\begin{equation}
\bm{\ell}^*_\circ = \begin{bmatrix}
\cos(\gamma^*_\circ) \\
\sin(\gamma^*_\circ)
\end{bmatrix}.
\end{equation}
Similarly, the vector $\bm{\ell}_{\circ} \in \mathbb{R}^2$ is fixed to the foot and parallel to the foot $x$-axis. This vector can be easily obtained as a linear combination of the contact points position. The goal of this task is to align $\bm{\ell}_{\circ}$ to $\bm{\ell}^*_{\circ}$. This can be achieved by minimizing their cross-product, which corresponds to the following task:
\begin{equation}\label{eq:yaw_task_partial}
\Gamma^{\prime}_\text{yaw} = \sum_{l, r} \frac{1}{2} \left\| \begin{bmatrix}
-\sin(\gamma^*_\circ) & \cos(\gamma^*_\circ)
\end{bmatrix}\bm{\ell}_{\circ} \right\|^2.
\end{equation}
Notice that Eq. \eqref{eq:yaw_task_partial} has a minimum also when $\bm{\ell}_{\circ}$ is null. In other words, $\Gamma^{\prime}_\text{yaw}$ can be minimized by shrinking the projection on the $xy$-plane of the vector attached to the robot foot. This is undesired because it would set the foot to be perpendicular to the ground. Hence, we consider also a second vector attached to the foot and perpendicular to $\bm{\ell}_{\circ}$, called $\bm{\ell}^\bot_{\circ}$. This vector is parallel, and have the same direction of the foot $y$-axis. Hence, the final task has the following form:
\begin{equation} \label{eq:yaw_task}
\begin{split}
\Gamma_\text{yaw} = &\sum_{l, r} \frac{1}{2} \left\| \begin{bmatrix}
-\sin(\gamma^*_\circ) & \cos(\gamma^*_\circ)
\end{bmatrix}\bm{\ell}_{\circ} \right\|^2 \\
+ &\sum_{l, r} \frac{1}{2} \left\| \begin{bmatrix}
\cos(\gamma^*_\circ) & \sin(\gamma^*_\circ)
\end{bmatrix}\bm{\ell}^\bot_{\circ} \right\|^2.
\end{split}
\end{equation}
Eq. \eqref{eq:yaw_task} does not impede the foot roll and pitch motions during swing.

\section{Regularization tasks}
The dynamical system of Eq. \eqref{sec:complete_dae} depends on a high number of variables. Despite the additional constraints of Sec. \ref{sec:additional_constraints} and the Cartesian tasks of Sec. \ref{sec:cartesian_tasks}, a consistent part of the dynamics is not taken into consideration nor constrained. For this reason, it is necessary to introduce regularization tasks which contribute in generating walking trajectories.

\subsection{Frame orientation task}\label{sec:orientation_task}
While moving, we want a robot frame to be oriented in a specific orientation ${}^\mathcal{I}\bm{R}^*_\text{frame}$. We thus need to define an orientation error measure. In particular, we weight the distance of the rotation matrix ${}^\mathcal{I}\tilde{\bm{R}}_\text{frame} = {}^\mathcal{I}\bm{R}^{*\top}_\text{frame}{}^\mathcal{I}\bm{R}_\text{frame}$ from the identity. Having to express this task in vector form, we can convert ${}^\mathcal{I}\tilde{\bm{R}}_\text{frame}$ into a quaternion (through a function \texttt{quat}, which implements the Rodrigues formula, see Eq. \eqref{eq:rodriguesFormula}) and weight its difference from the identity quaternion $\bm{I}_q$. Namely:
\begin{equation}\label{cost:frameOrientation}
\Gamma_\text{frame} = \frac{1}{2}\left\|\texttt{quat}\left({}^A\tilde{\bm{R}}_\text{frame}\right) - \bm{I}_q\right\|^2.
\end{equation}
This task can be applied on multiple bodies, like the robot torso and waist.
\subsection{Force regularization task} \label{sec:forceRegularization}
While considering each single contact force in a foot as independent, they still belong to a single body part. Thus, we prescribe the contact forces in a foot to be as similar as possible, refraining from using partial contacts if not strictly necessary. This can be obtained through the following:
\begin{equation}\label{cost:forceRegularization}
\Gamma_{\text{reg} f} = \sum_{l, r}\sum^{n_p}_i \frac{1}{2} \left\|{}_i\bm{f} - diag({}_i\bm{\alpha}^*)\sum^{n_p}_j {}_j\bm{f}\right\|^2_{\bm{W}_f}.
\end{equation}
Here ${}_i\bm{\alpha}^* \in \mathbb{R}^3$ determines the desired ratio for force $i$ with respect to the total force. For example, if we want all the forces in a foot to be equal, it is sufficient to select all the components of ${}_i\bm{\alpha}^* \in \mathbb{R}^3$ equal to $\frac{1}{n_p}$. In this case, the corresponding CoP is the centroid ${}_\# \bm{p}$. In other cases, it may be helpful to move the CoP somewhere else in the foot. In this case is sufficient to compute the corresponding ${}_i\bm{\alpha}^*$, for example by using the technique presented in Appendix \ref{ap:four_forces}.

\subsection{Joint regularization task} \label{sec:regularization}
In order to avoid the planner to provide solutions which require huge joint variations, we can introduce a regularization task for the joint variables:
\begin{equation}\label{cost:jointsRegularization}
\Gamma_{\text{reg} s} = \frac{1}{2}\left\|\dot{\bm{s}} + \bm{K}_s(\bm{s} - \bm{s}^*)\right\|^2_{\bm{W}_j}
\end{equation}
with $\bm{s}^*$ a desired joints configuration and $\bm{W}_j$ a weight matrix. The minimum for this cost is achieved when $\dot{\bm{s}}= -\bm{K}_s(\bm{s} - \bm{s}^*)$, with $\bm{K}_s \in \mathbb{R}^{n\times n}, \bm{K}_s \succcurlyeq 0$. When this equality holds, joint values converge exponentially to their desired values $\bm{s}^*$. In this way, both joint velocities and joint positions are regularized.

\subsection{Swing height task}
When performing a step, the swing foot clearance usually ensures some level of robustness with respect to ground asperity. Nevertheless, since the soil profile is supposed to be known in advance, a solution satisfying all the equations described in Chap. \ref{chap:modeling_dp} may require the swing foot to be raised just few millimeters from the ground. In order to specify a desired swing height, we can adopt the following task:
\begin{equation}
	\Gamma_{\text{swing}} = \sum_{l, r}\sum^{n_p}_i\frac{1}{2}\left(\bm{e}_3^\top {}_i\bm{p} - {}_sh^*\right)^2\left\|\left[\bm{e}_1 ~ \bm{e}_2\right]^\top \bm{u}_{{}_ip}\right\|^2.
\end{equation}
It penalizes the distance between the $z$-component of each contact point position from a desired height ${}_sh^* \in \mathbb{R}$ when the corresponding planar velocity is not null. Trivially, this cost has two minima: when the planar velocity is zero (thus the point is not moving) or when the height of the points is equal to the desired one.

\section{The complete optimal control problem} \label{sec:oc}
Given the set of equations listed in Chapter \ref{chap:modeling_dp} and the tasks described in Chapter \ref{chap:tasks} it is possible to fill an optimal control problem, whose complete formulation is presented below. Here the vector $\mathbf{w}$ contains the set of weights defining the relative ``importance'' of each task.

\begin{IEEEeqnarray}{CRCLL}
	\IEEEyesnumber \phantomsection
	\minimize_{\bm{\chi},\, \bm{\mathcal{U}}} & \IEEEeqnarraymulticol{4}{C}{\mathbf{w}^\top 
	\begin{bmatrix*}[l]
		\Gamma_{{}_\# p} \\
		\Gamma_{{}_{\bar{G}} h^p}\\
		\Gamma_\text{frame} \\
		\Gamma_{\text{reg} f} \\
		\Gamma_{\text{reg} s} \\
		\Gamma_{\text{swing}} \\
		\Gamma_\text{yaw}
	\end{bmatrix*}, }\label{costFunction} \IEEEyessubnumber\\
	\text{subject to:}&  \nonumber\\
	& \dot{\bm{\chi}} &=& \bm{f}(\bm{\chi}, \bm{\mathcal{U}}), & \text{ see Eq. \eqref{constr:system_dynamics}}, \IEEEyessubnumber\\
	& \mathbf{l} &\leq& \bm{g}\left(\bm{\chi}, \bm{\mathcal{U}}\right) \leq \mathbf{u},\,\, &\text{ see Eq.s \eqref{constr:equalities}$-$\eqref{constr:inequalities}}, \IEEEyessubnumber\\
	& \bm{e}_2^\top{{}^{r}\bm{x}_{l}} &\geq& d_\text{min}, \IEEEyessubnumber \label{constr:minDistance}\\
	& x_{\text{CoM},z \text{ min}} &\leq& \bm{e}_3^\top{\bm{x}_{\text{CoM}}},  \IEEEyessubnumber\\
	& -\bm{M}_{h_\omega} &\leq& {}_{\bar{G}} \bm{h}^\omega \leq \bm{M}_{h_\omega}, \IEEEyessubnumber\\
	& -M_{hf} &\leq& \IEEEeqnarraymulticol{2}{L}{e_3^\top\left({\#}\bm{p}_l - {\#}\bm{p}_r\right) \leq M_{hf}.} \IEEEyessubnumber
\end{IEEEeqnarray}
Here, the state variables $\bm{\mathcal{X}}$ are those derived in time, while $\bm{\mathcal{U}}$ contains all the control inputs. Thus:
\begin{equation} \label{eq:dp_state_control}
\bm{\chi} = 
\begin{bmatrix}
{}_i\bm{f} \\
{}_i\bm{p} \\
\vdots	\\
{}_{\bar{G}} \bm{h} \\
\bm{x}_\text{CoM} \\
{}^\mathcal{I} \bm{p}_B \\
{}^\mathcal{I}\bm{\rho}_B \\
\bm{s}
\end{bmatrix}, \quad
\bm{\mathcal{U}} = 
\begin{bmatrix}
\bm{u}_{{}_if} \\
\bm{u}_{{}_ip} \\
\vdots \\
{}^B\bm{v}_{\mathcal{I},B} \\
\bm{u}_\rho \\
\bm{u}_s
\end{bmatrix},
\end{equation}
where the symbol $\vdots$ represents the repetition of the corresponding variables for each contact point. 

\section{Software infrastructure} \label{sec:dp_software_infrastructure}
\begin{figure}[tpb]
	\centering
	\includegraphics[width=\columnwidth]{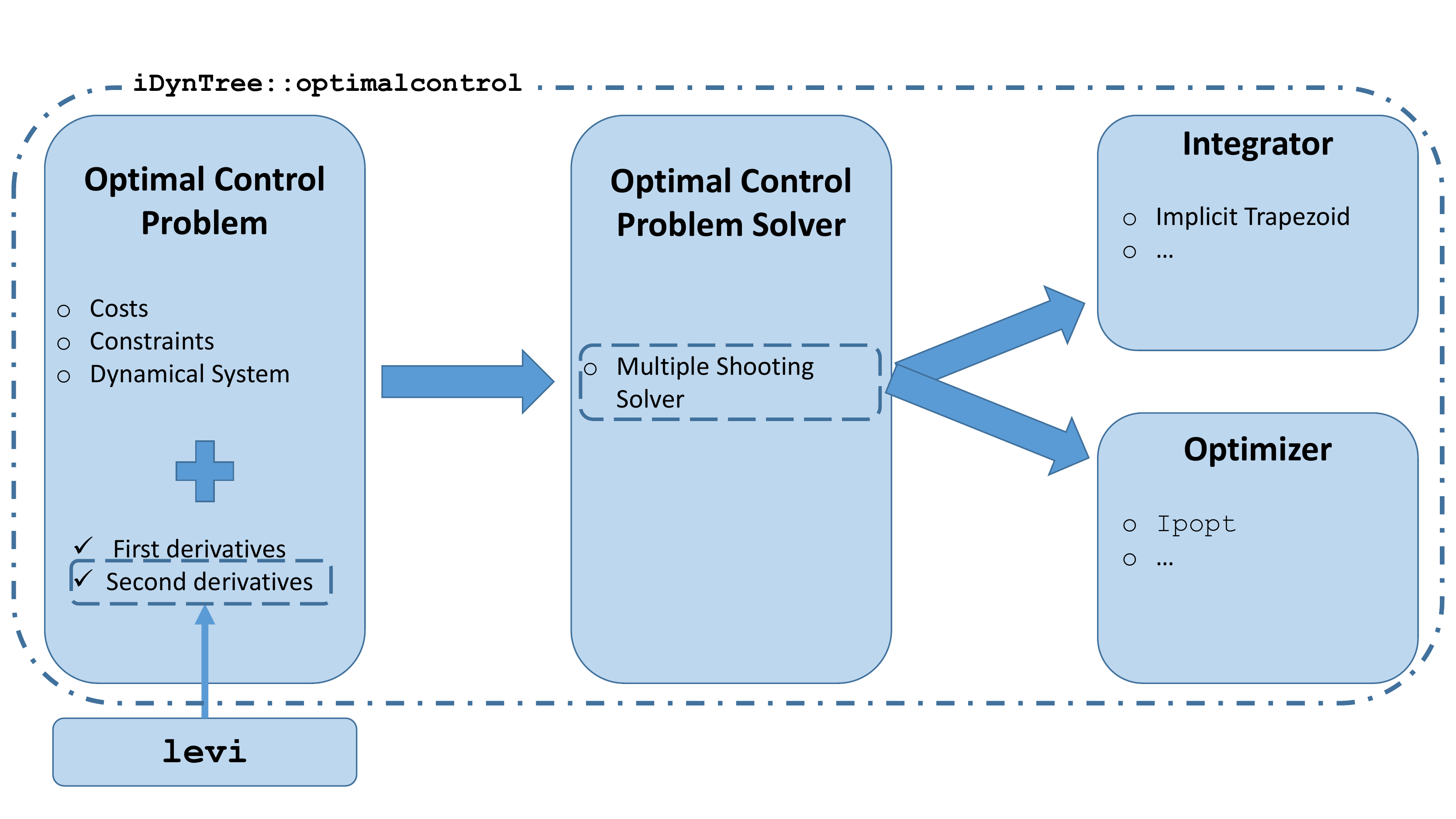} 
	\caption{A graphic description of the software architecture.}
	\label{fig:architecture}
\end{figure}
The optimal control problem described in Sec. \ref{sec:oc} is solved using a Direct Multiple Shooting method, Sec. \ref{sec:shooting}. The system dynamics, defined in Eq. \eqref{constr:system_dynamics}, is discretized adopting an implicit trapezoidal method with a fixed integration step, as described in Sec. \ref{sec:integration_methods}. The corresponding optimization problem is solved thanks to \texttt{Ipopt} \citep{IPOpt2006}. 

The pipeline from the problem definition to its solution is implemented using the \texttt{iDynTree::optimalcontrol}\footnote{\url{https://github.com/robotology/idyntree/tree/devel/src/optimalcontrol}} library, allowing for easy testing of other integrators or solvers. The computation of the Hessian of the Lagrangian is done by adopting \texttt{levi}\footnote{\url{https://github.com/S-Dafarra/levi}}, a recently developed library. More details can be found in Appendices \ref{chap:jacobians} and \ref{chap:hessian}.  Fig. \ref{fig:architecture} presents a graphic description of the software architecture.

The walking trajectories are generated according to the Receding Horizon Principle presented in Sec. \ref{sec:receding_horizon}, adopting a fixed prediction window of 2 seconds. The horizon is large enough to predict at least one full step. This planner is conceived for being adopted in conjunction with a whole-body controller or to be used in an off-line fashion.

\chapter{Validation and Experimental Results} \label{chap:experiments}
In this chapter, we present the results obtained when solving the optimal control problem described in Chapter \ref{chap:tasks}. In particular, we test its capabilities to generate whole-body walking trajectories for a flat ground using the model of the iCub humanoid robot introduced in Sec. \ref{sec:icub}. The chapter is introduced by Sec. \ref{sec:considerations} with some considerations about the planner. It is validated in Sec. \ref{sec:validation} by visualizing the generated trajectories. Sec. \ref{sec:complementarity_comparison} compares the different complementarity constraints in terms of accuracy and computational overhead. Sec. \ref{sec:dp_robot_experiments} presents the results of applying the generated trajectories on the robot. Finally, Sec. \ref{sec:conclusions} concludes the chapter.

\section{Considerations} \label{sec:considerations}
The optimal control problem described in Sec. \ref{sec:oc} is built such that (almost) no constraint is task specific. As a consequence, it is particularly important to define the cost function carefully since the solution will be a trade-off between all the various tasks. On the other hand, the detailed model of the system allows achieving walking motions without specifying a desired CoM trajectory or by fixing the angular momentum to zero.
Nevertheless, due to the limited time horizon, it is better to prevent the solver from finding solutions which would bring to unfeasible states in future planner iterations. For this reason, Eq. \eqref{constr:com_height_limit} and Eq. \eqref{constr:angular_momentum_bounds} are added, even if the bounds are relatively large. 

Another possible effect resulting from the application of the Receding Horizon principle is the emergence of ``procrastination'' phenomena. Due to the moving horizon, the solver may continuously delay in actuating motions, since the task keeps being shifted in time. A simple fix to this phenomena is to increase the weights $\mathbf{w}$ with time, such that it is more convenient to reach a goal position earlier.

Finally, given that the problem under consideration is non-convex, the optimizer will find a local minimum. This may result in a sub-optimal solution for the given tasks, but this fact does not limit the applicability of the results to the robot. During the first iteration, the solver is initialized by simply translating the whole robot in the desired position. In successive iterations, the solver is warm-started with the solution previously computed.

All the tests presented in the next sections have been carried on a 7$^{th}$ generation Intel\textsuperscript{\textregistered} Core i7@2.8GHz laptop.

\section{Validation}\label{sec:validation}

\begin{figure}[tpb]
    \centering
    \subfloat[$t=0.5s$] {\includegraphics[width=.22\columnwidth]{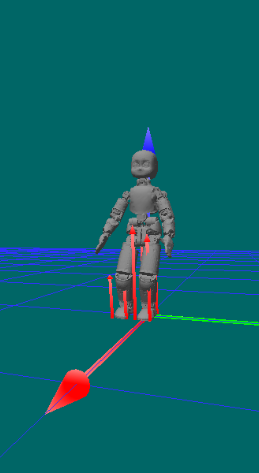}}
    \hspace{.001\columnwidth}
    \subfloat[$t=1.5s$] {\includegraphics[width=.22\columnwidth]{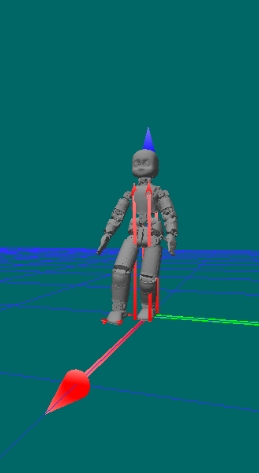}}
    \hspace{.001\columnwidth}
    \subfloat[$t=2.5s$] {\includegraphics[width=.22\columnwidth]{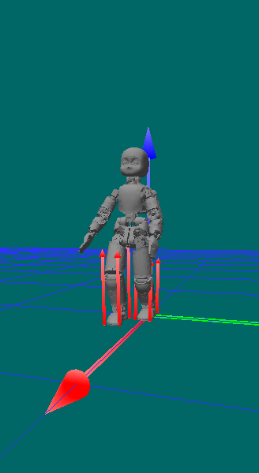}}
    \hspace{.001\columnwidth}
    \subfloat[$t=3.5s$] {\includegraphics[width=.22\columnwidth]{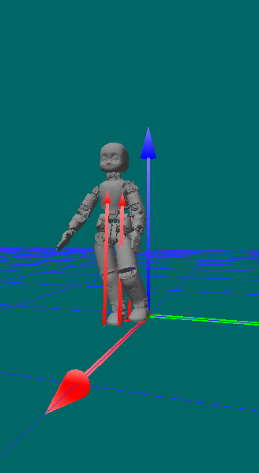}}
    \caption{Snapshots\protect\footnotemark\,of the generated walking motion. The red arrows indicate the force required at each contact point scaled by a factor of 0.01. These images have been obtained using the complementarity constraints of Sec. \ref{sec:dynamical_complementarity}, but using the other methods, the result is visually identical.}
    \label{fig:slow_straight}
\end{figure}
\footnotetext{\url{https://www.youtube.com/playlist?list=PLBOchT-u69w9hJ6BmqPf06r0zWmungOrc}}
\begin{figure}[tpb]
	\centering
	\subfloat[Relaxed complementarity] {\includegraphics[height=.25\textheight]{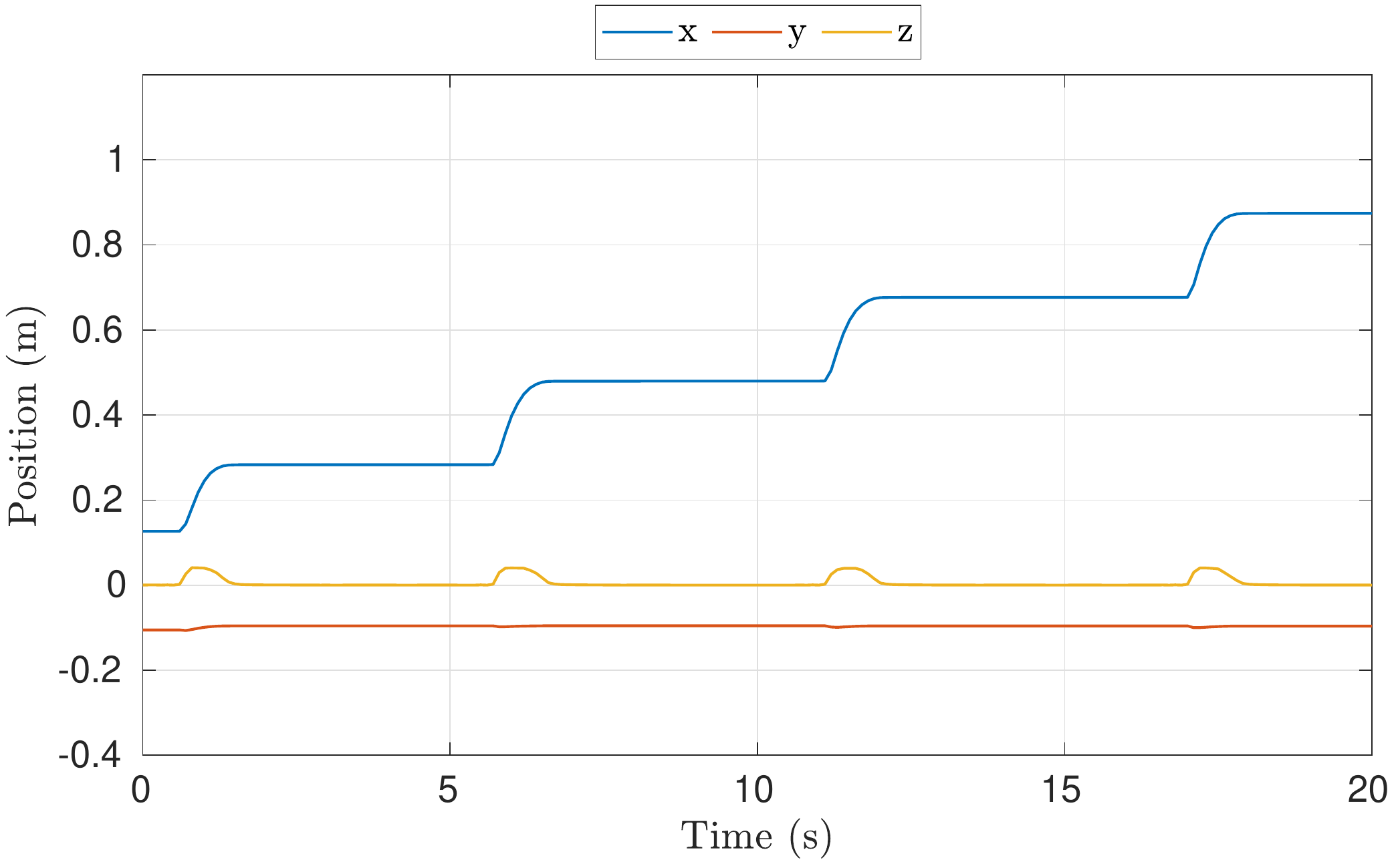}}

	\subfloat[Dynamically enforced complementarity] {\includegraphics[height=.25\textheight]{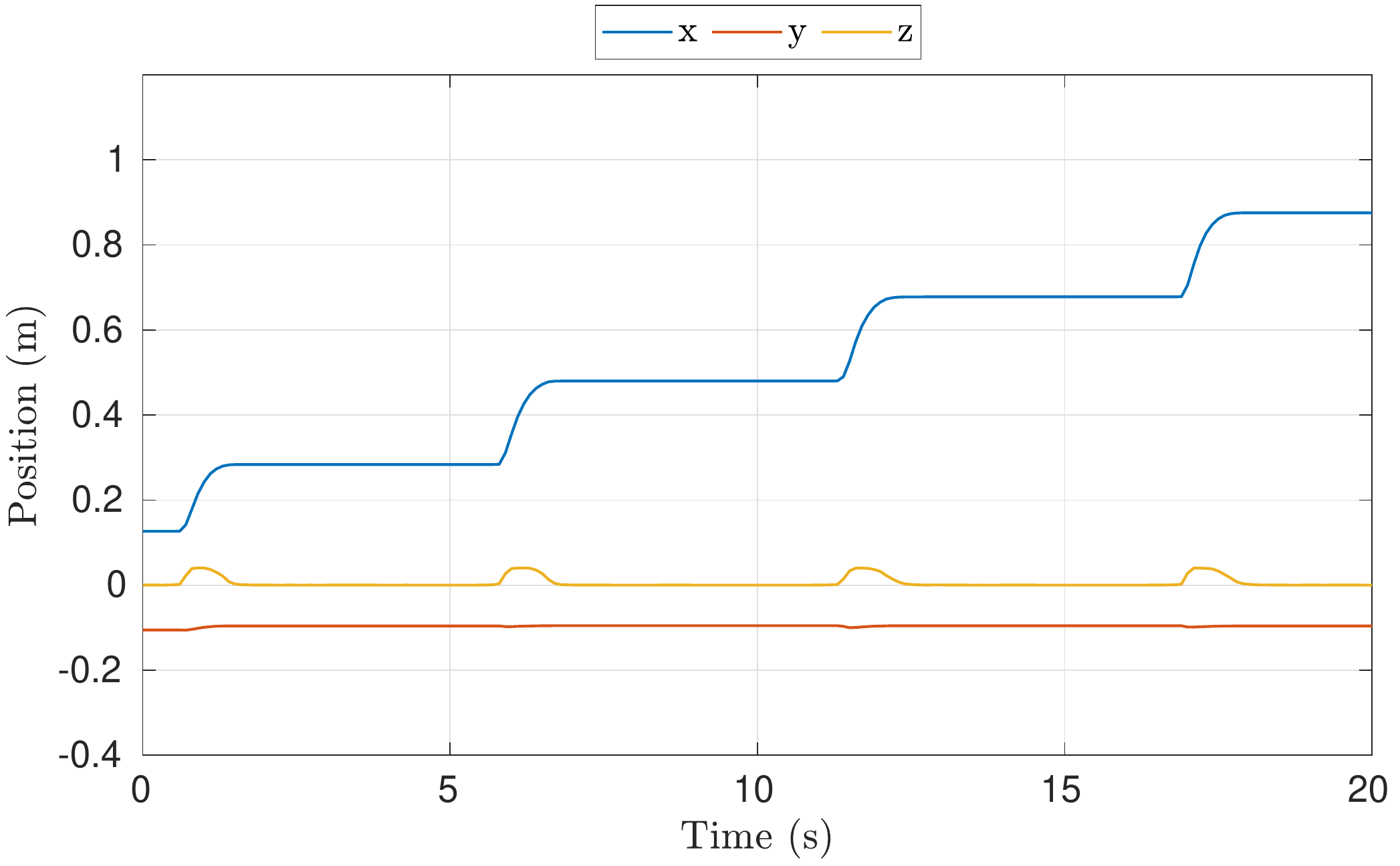}}
	
	\subfloat[Hyperbolic secant in control bounds] {\includegraphics[height=.25\textheight]{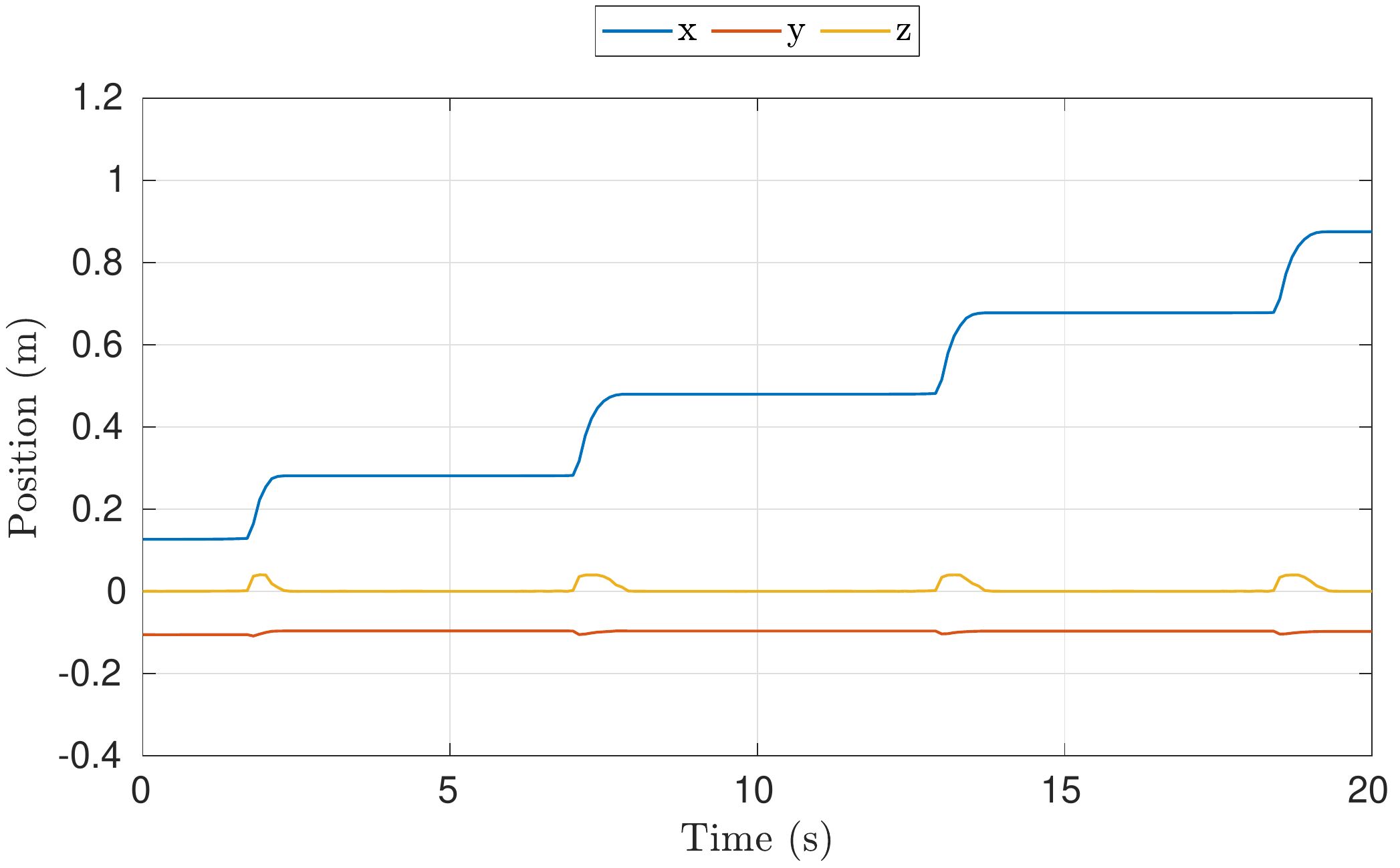}}
	\caption{Planned position of one of the right foot contact points using different complementarity constraints. The walking phases are recognizable, but they are not defined beforehand. The controller does not specify directly when a phase begins and ends.}
	\label{fig:point_position}
\end{figure}

\begin{figure}[tpb]
	\centering
	\subfloat[Relaxed complementarity] {\includegraphics[height=.25\textheight]{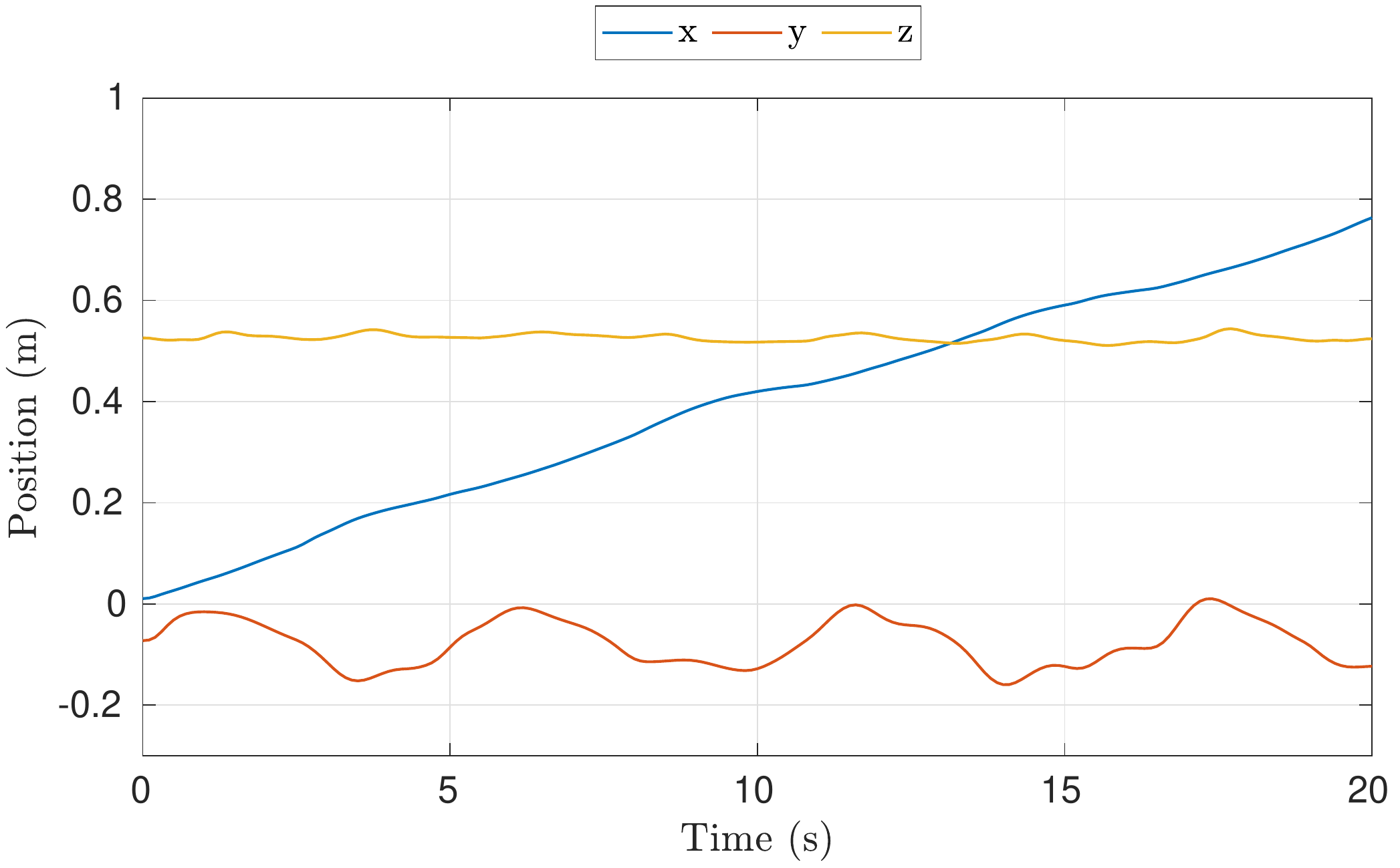}}

	\subfloat[Dynamically enforced complementarity] {\includegraphics[height=.25\textheight]{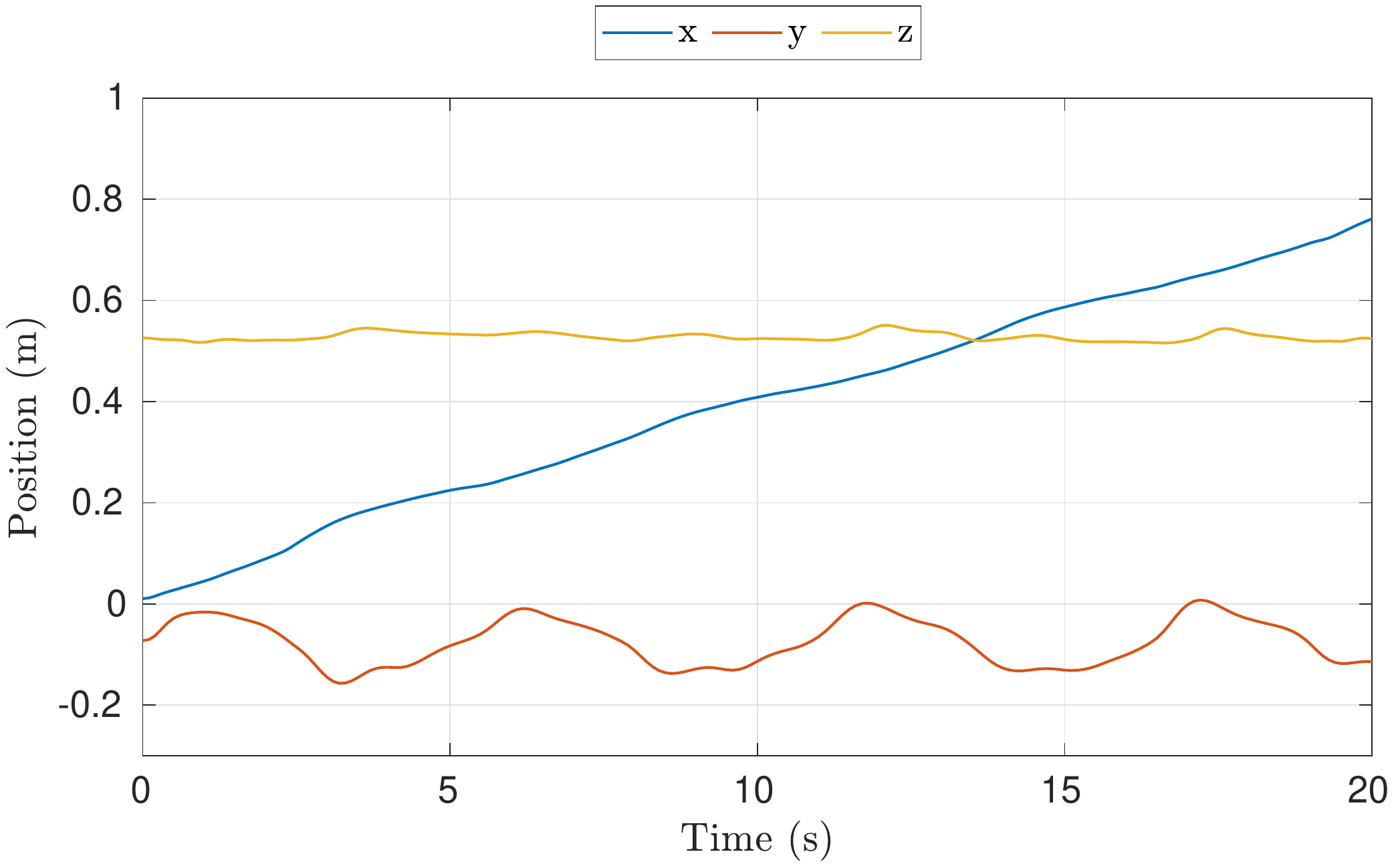}}

	\subfloat[Hyperbolic secant in control bounds] {\includegraphics[height=.25\textheight]{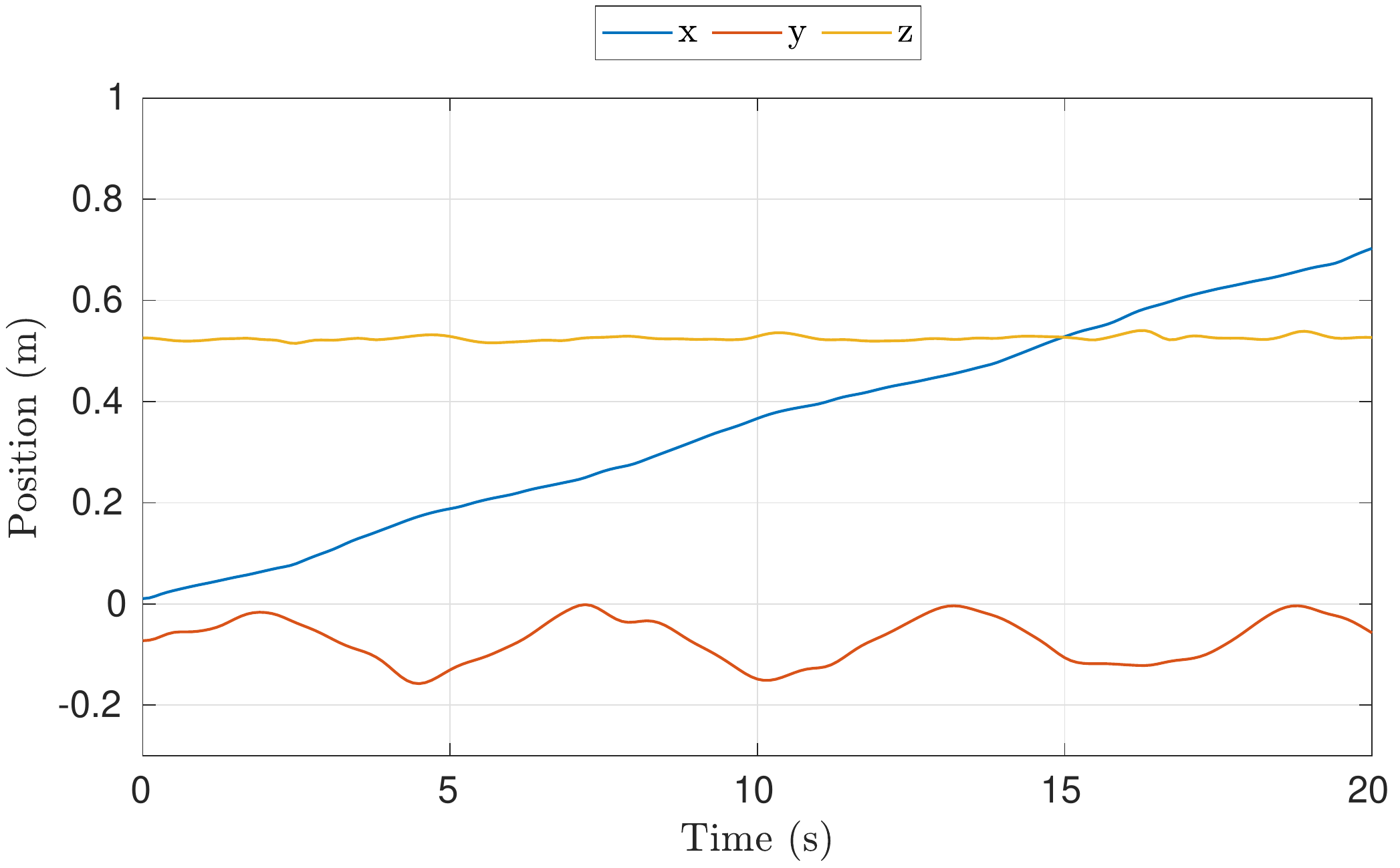}}
	\caption{Planned CoM position using different complementarity constraints. It is possible to notice a continuous velocity on the $x$ direction. The plots appear a little irregular. This may be a consequence of the chosen time step.}
	\label{fig:dp_com_position}
\end{figure}

\begin{figure}[tpb]
	\centering
	\subfloat[Relaxed complementarity] {\includegraphics[height=.25\textheight]{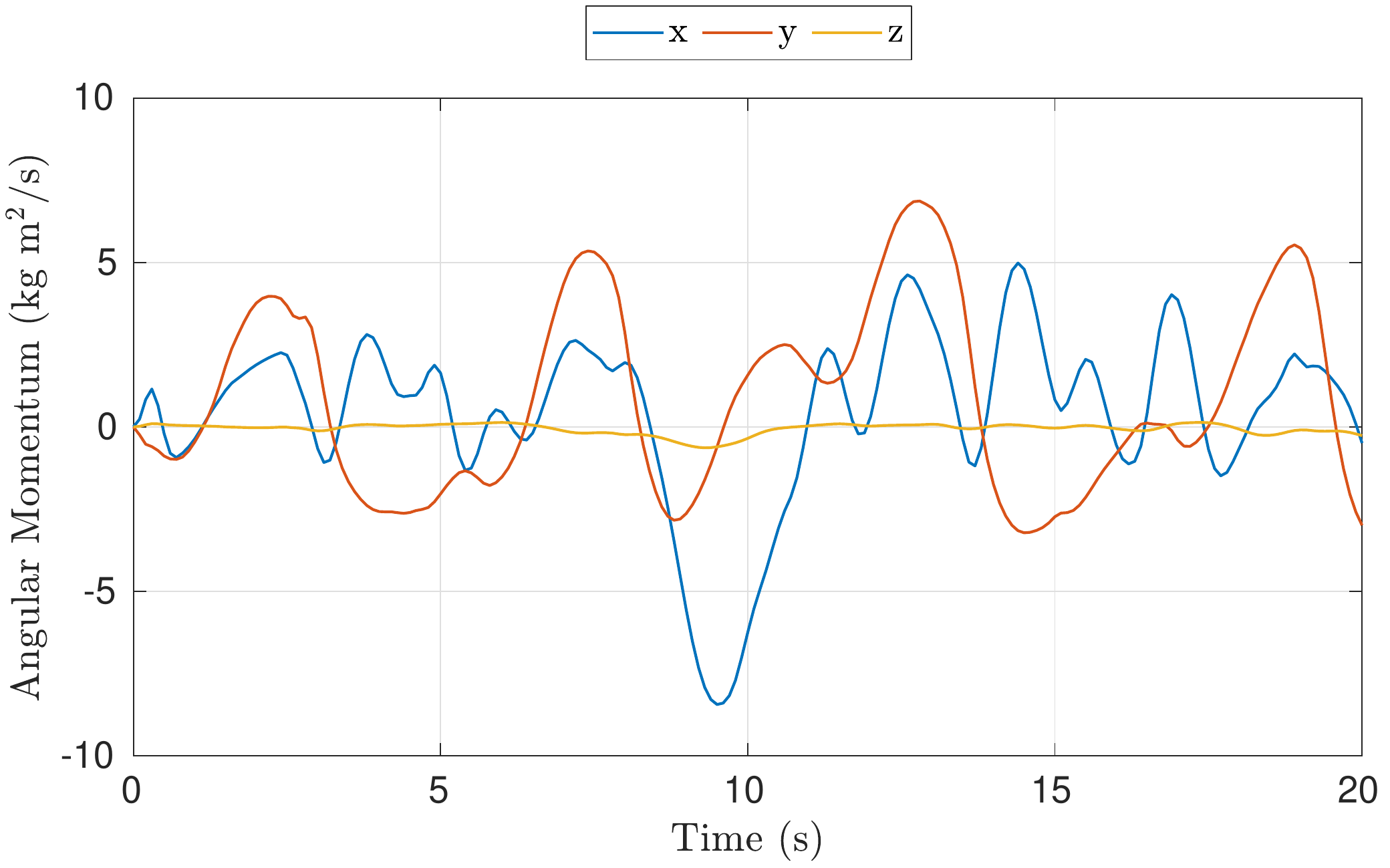}}

	\subfloat[Dynamically enforced complementarity] {\includegraphics[height=.25\textheight]{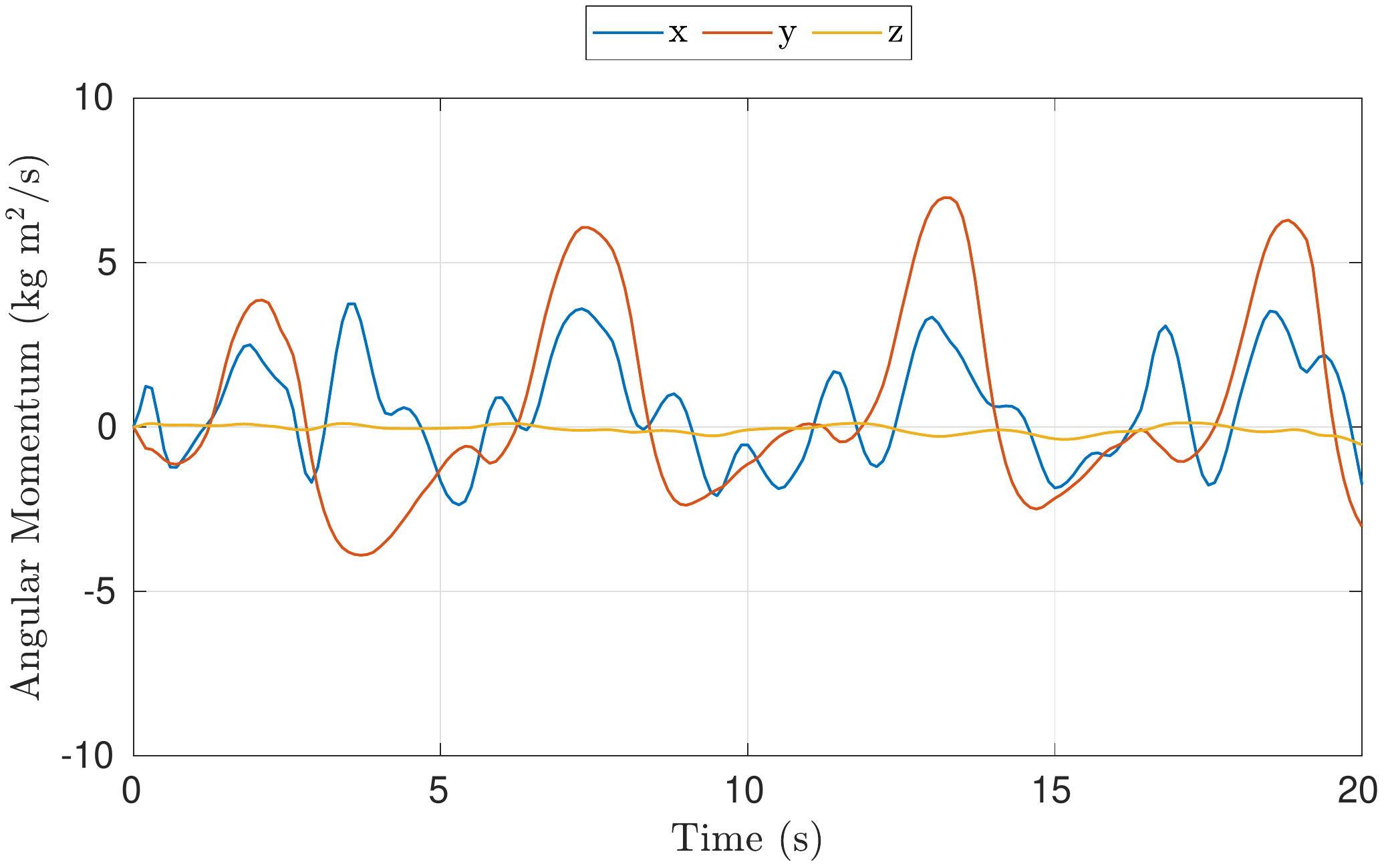}}
	
	\subfloat[Hyperbolic secant in control bounds] {\includegraphics[height=.25\textheight]{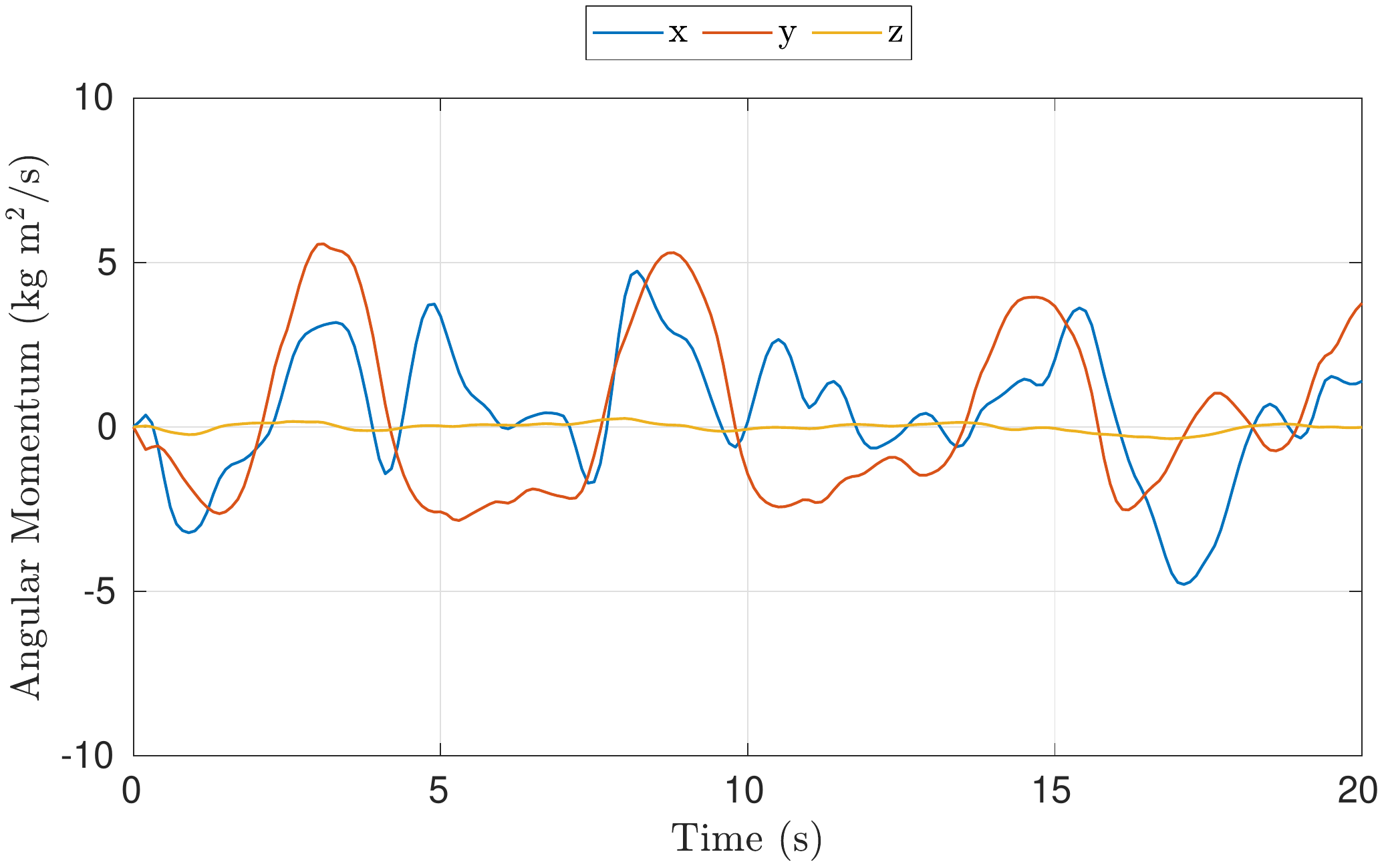}}
	\caption{Planned angular momentum obtained adopting different complementarity constraints. Interestingly, in (a) there is an overshoot of the angular momentum on the $x$ axis which is not present in the other two methods.}
	\label{fig:dp_angular_momentum}
\end{figure}

The planner is set up using an integration step of $100 \mathrm{ms}$ , while the time horizon is $2\mathrm{s}$. The choice of a large integration step serves for two reasons. First, it reduces the number of variables used by the optimization problem (fixing the time horizon). Secondly, it allows inserting another control loop at higher frequency. After each iteration of the planner, the first state is used as a feedback to a new planner iteration in a \emph{receding horizon} fashion. 

When planning, we control 23 of the robot joints. We consider four contact points for each foot, located at the vertices of the rectangle enclosing the robot foot. Concerning the references, the desired position for the centroid of the contact points is moved $10 \texttt{cm}$ along the walking direction every time the robot performs a step. A simple state machine, where the reference is moved as soon as a step is completed, is enough to generate a continuous walking pattern. The speed is modulated by prescribing a desired CoM forward velocity equal to $5 \texttt{cm/s}$. As discussed in the following, we are able to obtain very similar results when using any of the complementarity methods described in Sec. \ref{sec:complementarity_list}.

Fig. \ref{fig:slow_straight} shows some snapshots of the first generated step, while Fig. \ref{fig:point_position} shows the position of one of the right contact points. It is possible to recognize the different walking phases, though they are not planned a priori. Nevertheless, the controller does not specify explicitly when a phase begins and ends. Notice that the stance phases obtained using the hyperbolic secant method, Sec. \ref{sec:hyperbolic_secant}, are slightly longer than those produced with the other two methods, which are instead very similar.

Figure \ref{fig:dp_com_position} presents the planned CoM position. Here, it is possible to notice that the position along the $x$ direction grows at a constant rate. This is a direct consequence of the task on the CoM velocity presented in Sec. \ref{sec:com_velocity_cost}. 
Notice that the bound on the CoM height, $x_{\text{CoM},z \text{ min}}$, is set to half of the initial robot height, but such constraint is never activated. The plots appear a little irregular. This is probably due to the choice of time step. In addition, these are the results of consecutive runs of the optimal control problem described in Sec. \ref{sec:oc}. From one iteration to another, the solver may find slightly different solutions because of the shift in the prediction horizon, causing the irregularities showed in the figures.

Figure \ref{fig:dp_angular_momentum} shows the planned angular momentum, which is not fixed to zero. Although it is limited to $10~\texttt{kg}~{\texttt{m}^2}/{\texttt{s}}$, such limit is never reached. Interestingly, the trajectories computed with the relaxed complementarity constraints showed in Sec. \ref{sec:relaxed_complementarity}, produce a moderate overshoot in the component along $x$ that is not present with the other two methods. This is probably due to a little wider motion while swinging a leg. Amongst the three, the dynamically enforced complementarity method is the one producing the smoothest result. 

It is worth stressing that none of the tasks described above define how and when to raise the foot. By prescribing a reference for the centroid of the contact points and by preventing the motion on the contact surface, swing motions are planned automatically. Nevertheless, this advantage comes with a cost. It is difficult to define a desired swing time and, more importantly, the relative importance of each task, i.e. the values of $\mathbf{w}$, must be chosen carefully. During experiments, we adopted an incremental approach. We added the tasks one by one, starting from $\Gamma_{{}_\# p}$ and then we gradually refine the walking motion by tuning a cost at a time. 

\section{Complementarity constraints comparison}\label{sec:complementarity_comparison}

\begin{figure}[tbp]
	\centering
	\subfloat[Relaxed complementarity]{\includegraphics[width=\columnwidth]{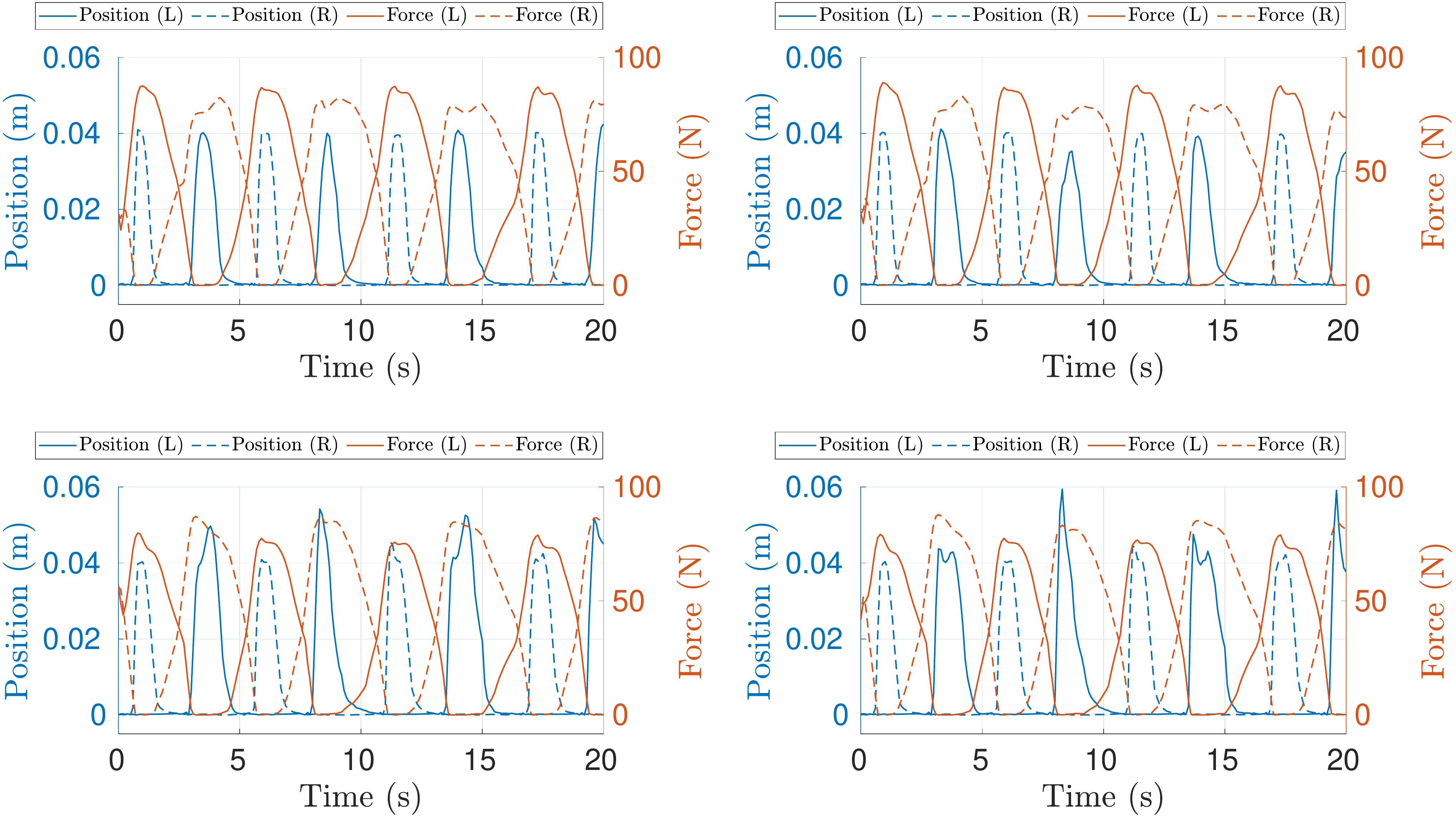}}
		
	\subfloat[Dynamically enforced complementarity]{\includegraphics[width=\columnwidth]{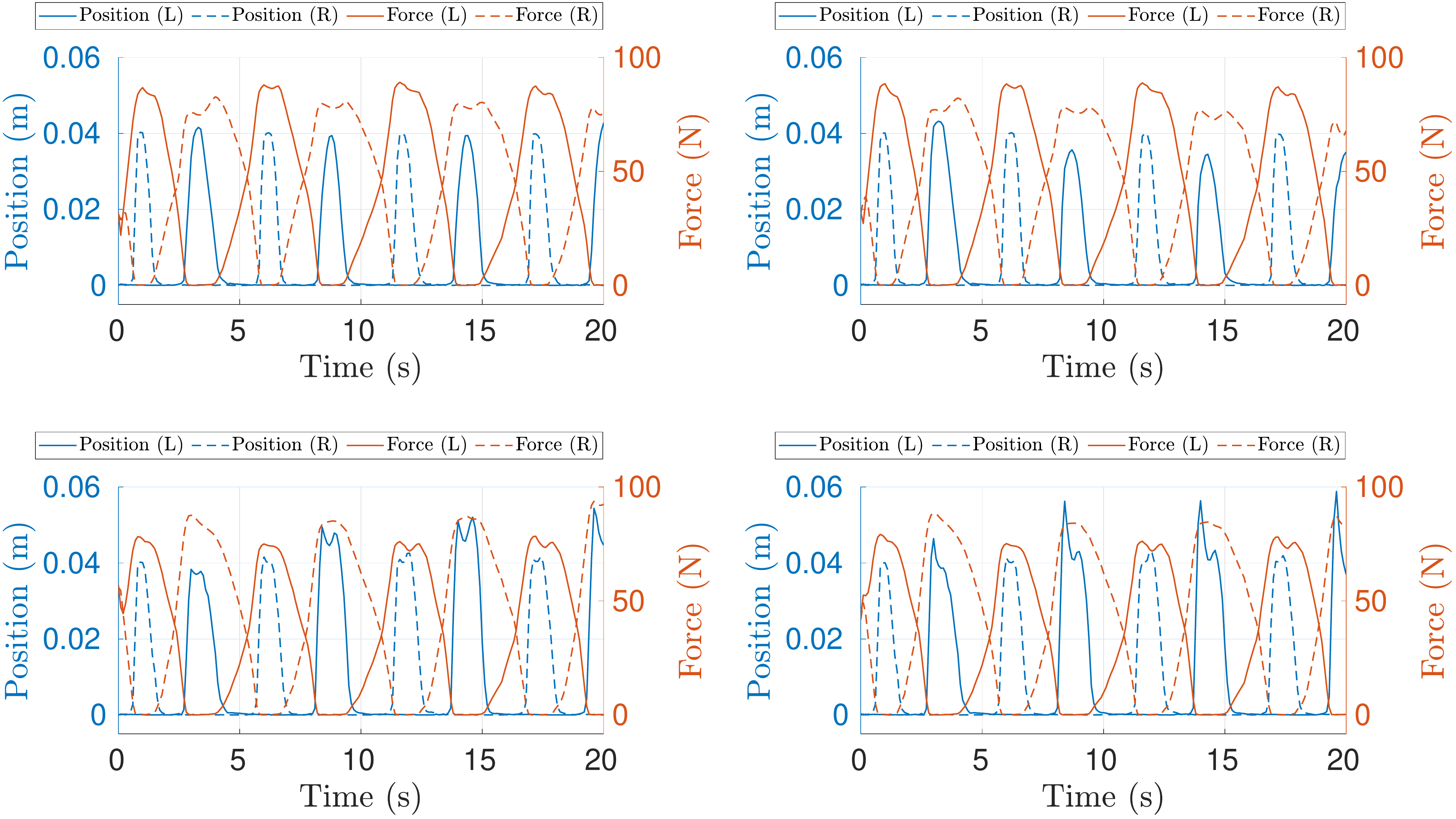}}
	\caption{Vertical position and normal force of each contact point. The quantities relative to a point are plotted together with those of the corresponding point in the other foot. The dashed lines are relative to the right foot. The main differences are visible around the zero axis, where it is possible to notice that, depending on the method, some normal force can be requested even if the height is not zero. Plots from all the points are shown for completeness.}
\end{figure}

\begin{figure}[tbp]\ContinuedFloat
	\centering
	\subfloat[Hyperbolic secant in control bounds]{\includegraphics[width=\columnwidth]{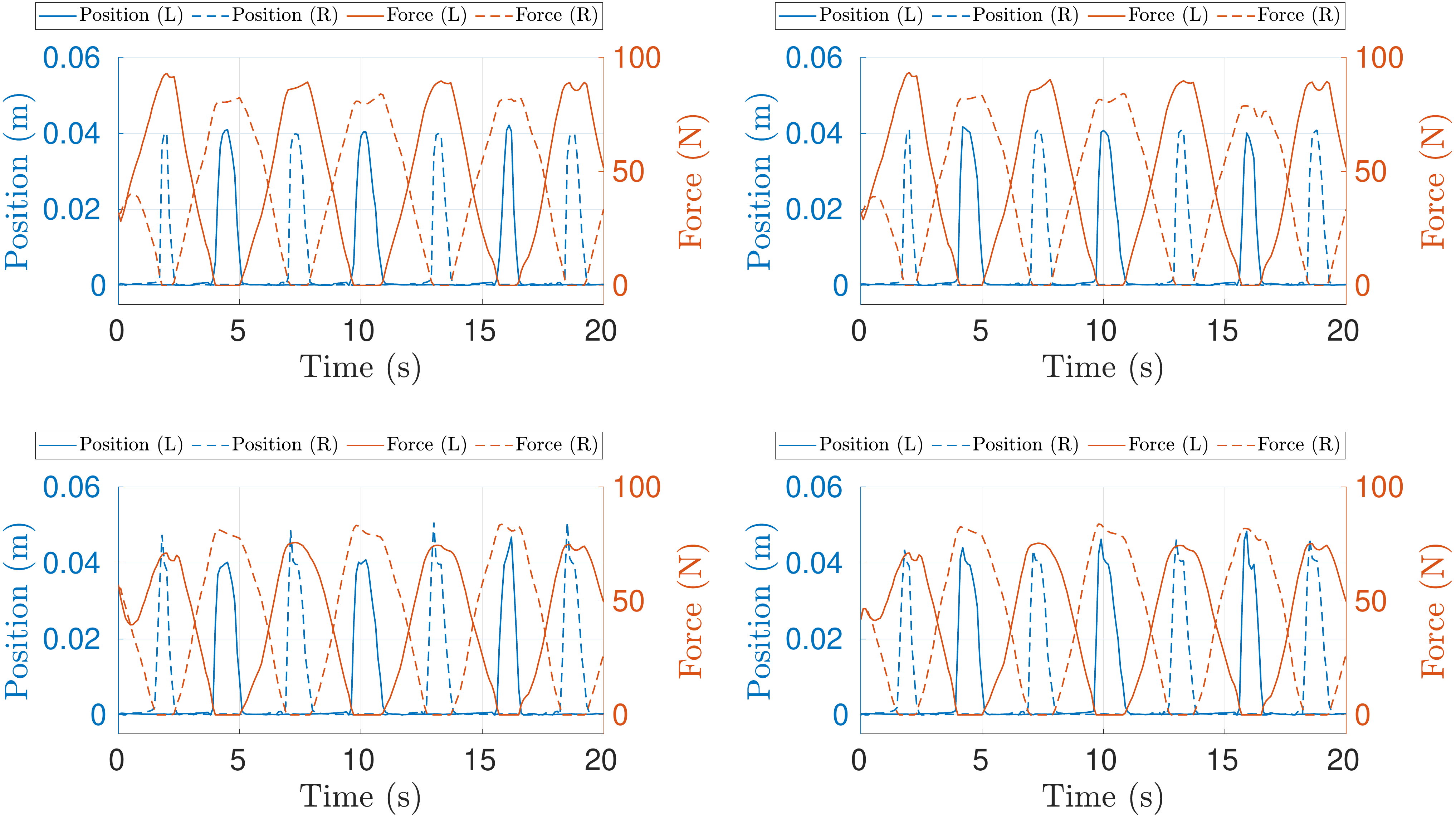}}
	\caption{(cont.) Vertical position and normal force of each contact point. The quantities relative to a  point are plotted together with those of the corresponding point in the other foot. The dashed lines are relative to the right foot. The main differences are visible around the zero axis, where it is possible to notice that, depending on the method, some normal force can be requested even if the height is not zero. Plots from all the points are shown for completeness.}
	\label{fig:pointVsForce}
\end{figure}

\begin{figure}[tpb]
	\centering
	\subfloat[Relaxed complementarity] {\includegraphics[width=\columnwidth]{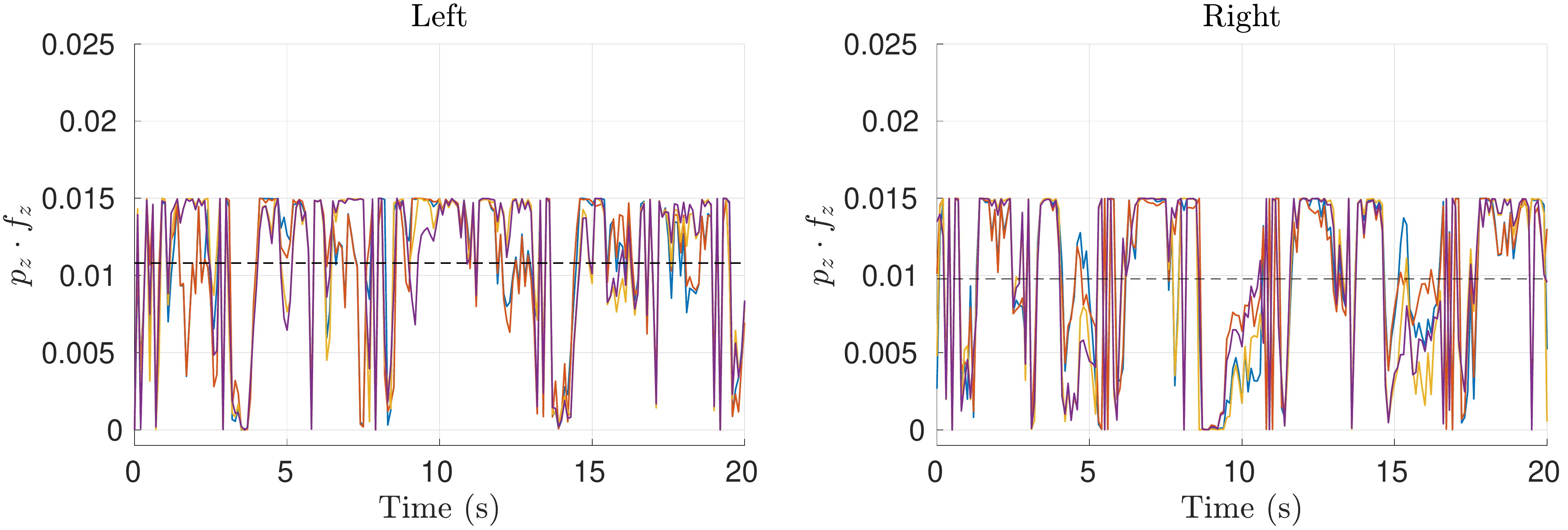}}

	\subfloat[Dynamically enforced complementarity] {\includegraphics[width=\columnwidth]{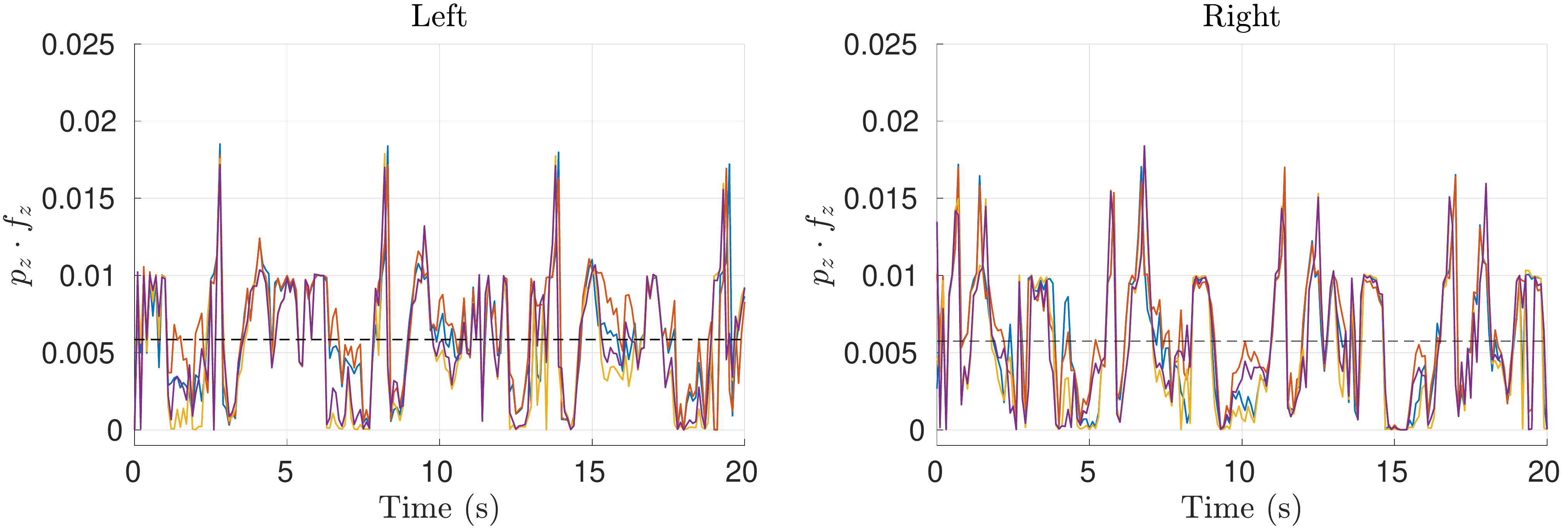}}
	
	\subfloat[Hyperbolic secant in control bounds] {\includegraphics[width=\columnwidth]{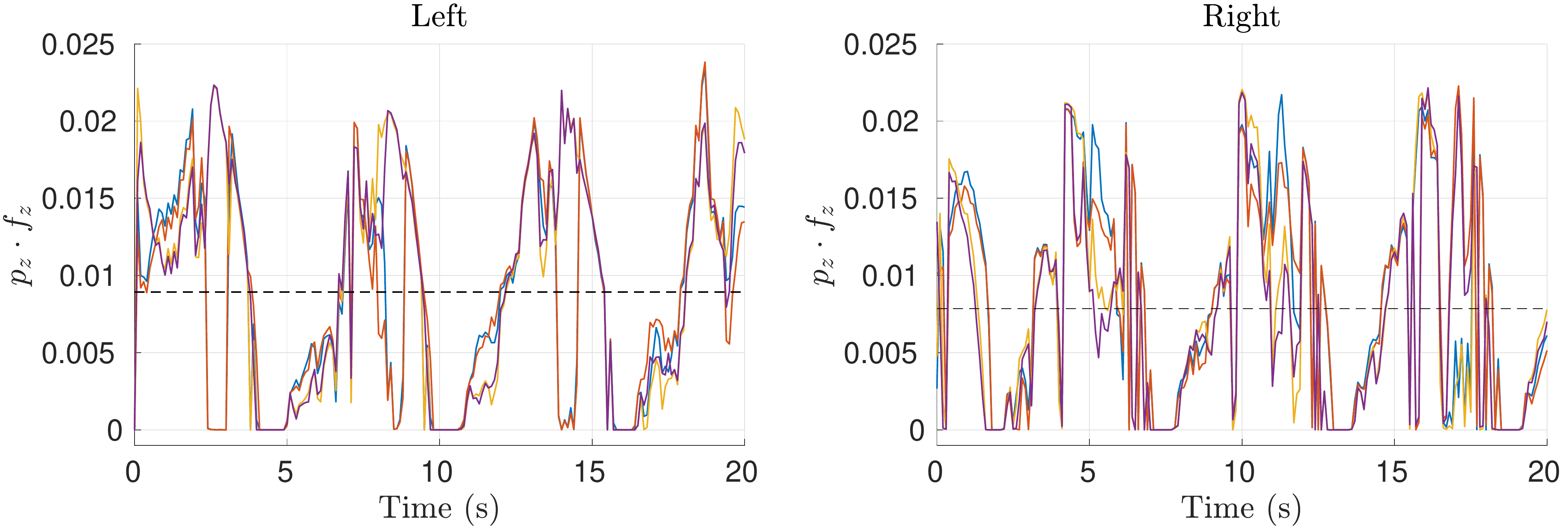}}
	\caption{Product between the vertical position and the normal force of each contact point, separated by foot, when using the different complementarity methods summarized in Sec. \ref{sec:complementarity_list}. The black dashed lines indicate the mean values. By plotting the result of $p_z\cdot f_z$ for each point, we show the accuracy of each method in simulating a rigid contact.}
	\label{fig:complementarity_level}
\end{figure}

In this section, we analyze the differences amongst the complementarity methods described in Sec. \ref{sec:contact_parametrization}. As a measure of performance, we adopt the product between the normal force and the height of the contact point from the ground, i.e. ${}_ip_z\cdot{}_if_z$. In other words, we test the accuracy with which Eq. \eqref{eq:complementarity} is satisfied, simplifying the formulation thanks to the planar ground assumption. Results are shown in Fig. \ref{fig:complementarity_level}. We tune the various methods in order to maximize such accuracy. Indeed, parameters that are too ``restrictive'' (e.g. an $\epsilon$ too small) may prevent the solver from finding a walking strategy because the points are not able to move. On the other hand, low accuracy may mean the solver requires a force of considerable magnitude when the point is still raised from the ground. Given that robot weighs only 30$\mathrm{kg}$, even a small force affects the robot's dynamics. Expectedly, higher accuracy involves more computational time.

The parameters are chosen as follows:
\begin{itemize}
	\item Relaxed complementarity: $\epsilon = 0.015[\mathrm{N\cdot m}]$.
	\item Dynamically enforced complementarity: $K_\text{bs} = 20\left[\frac{1}{\mathrm{s}}\right]$, $\varepsilon = 0.2\left[\frac{\mathrm{N\cdot m}}{\mathrm{s}}\right]$.
	\item Hyperbolic secant in control bounds: $K_{f,z} = 300\left[\frac{1}{\mathrm{s}}\right]$, $k_h = 350\left[\frac{1}{\mathrm{m}}\right]$.
\end{itemize}

$\bm{M}_f$ is set to 100$\mathrm{N}/\mathrm{s}$ for all the components and it is common for all the methods. Fig. \ref{fig:pointVsForce} shows the contact points' height compared to the normal force applied to them, for both the two feet and for all the three methods. By superimposing quantities from the left and the right foot, it is possible to notice the different contact phases. The normal force applied to a point drops to zero when this moves, and vice-versa. Nonetheless, it is possible to spot millimetric motions even when a force is applied. This measure of accuracy is depicted in Fig. \ref{fig:complementarity_level}. 

The accuracy is the major difference between the three methods. The relaxed complementarity method is the one providing the smallest maximum value of ${}_ip_z\cdot{}_if_z$. This is a direct consequence of the choice of $\epsilon$. Nevertheless, it is the one with the highest mean value. In this case, the dynamically enforced complementarity method is the one providing the best results. Amongst the three, the hyperbolic secant method is the only one where the applied force consistently drops to zero in all the contact points when the foot is raised. In fact, with this method we force the normal force derivative to be strongly negative when the point is not on the ground. At the same time, this method does not prevent the point to move when a force is applied. For this reason, this method provides the highest peaks.

Table \ref{tab:dp_timings} compares the three methods in terms of time required to find a solution. The values are obtained by measuring the time required to obtain the 200 points composing the trajectories shown in the figures of this chapter. The dynamically enforced complementarity method is the fastest one, while the hyperbolic secant one is the best in the other metrics. 

The variance is extremely high for all the methods. We noticed that the iterations with the highest duration were occurring when moving the reference for the centroid of the contact points. In fact, this is the moment in which the planner has to predict a full new step. Since we initialize the planner with the previously computed solution, in this instant the optimal point is far from the initialization. Hence, more time is required to find a solution. This issue can be addressed by providing the planner with a more effective initialization, for example by exploiting some of the methods presented in Part \ref{part:applications}.

Overall, the dynamically enforced complementarity method is the one providing the best performances.

\begin{table}[tbp]
	\centering
	\begin{tabular}{ c || c | c | c }
		[\texttt{s}]& Relaxed & Dynamically  enforced & Hyperbolic secant \\
		\hline
		Average & 11.91 & 10.53 & 13.19 \\
		Variance & 15.62 & 13.05 & 12.39 \\
		Minimum & \hphantom{0}1.00 & \hphantom{0}0.98 & \hphantom{0}0.77 \\
		Maximum & 83.50 & 68.90 & 55.08 \\
		\hline
	\end{tabular}
	\caption{Time performances using different complementarity methods. All times are showed in seconds and obtained after 200 runs of the solver.}
	\label{tab:dp_timings}
\end{table}

\section{Experiments on the robot} \label{sec:dp_robot_experiments}

\begin{figure*}[tpb]
	\centering
	\subfloat[$t=t_0s$] {\includegraphics[width=.23\textwidth]{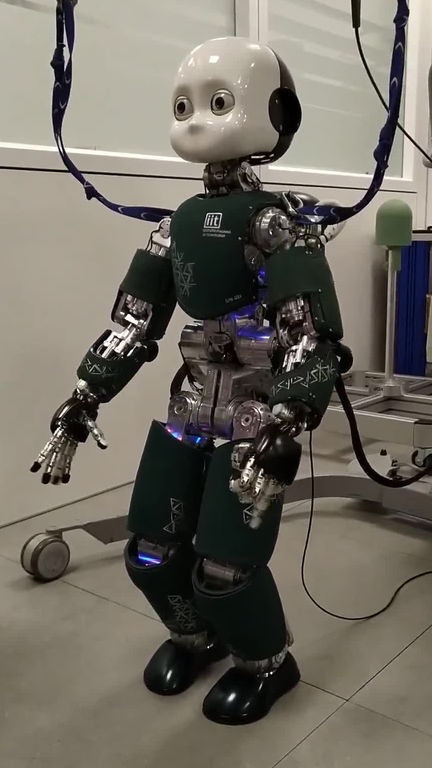}}
	\hspace{.01\textwidth}
	\subfloat[$t=t_0 + 1s$] {\includegraphics[width=.23\textwidth]{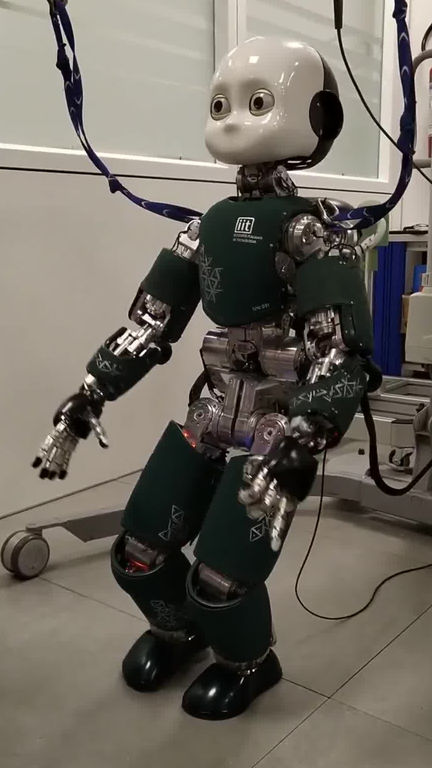}}
	\hspace{.01\textwidth}
	\subfloat[$t=t_0 + 2s$] {\includegraphics[width=.23\textwidth]{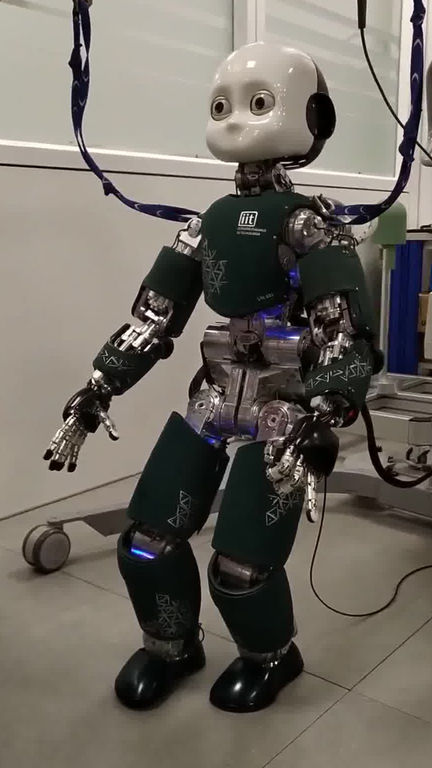}}
	\hspace{.01\textwidth}
	\subfloat[$t=t_0 + 3s$] {\includegraphics[width=.23\textwidth]{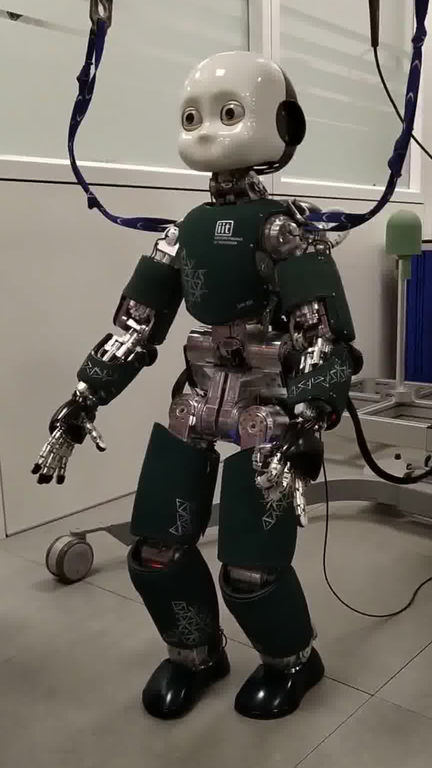}}
	
	\subfloat[$t=t_0 + 4s$] {\includegraphics[width=.23\textwidth]{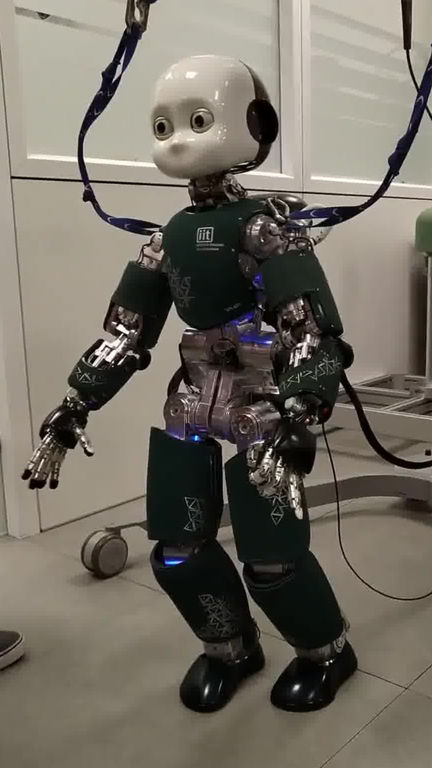}}
	\hspace{.01\textwidth}
	\subfloat[$t=t_0 + 5s$] {\includegraphics[width=.23\textwidth]{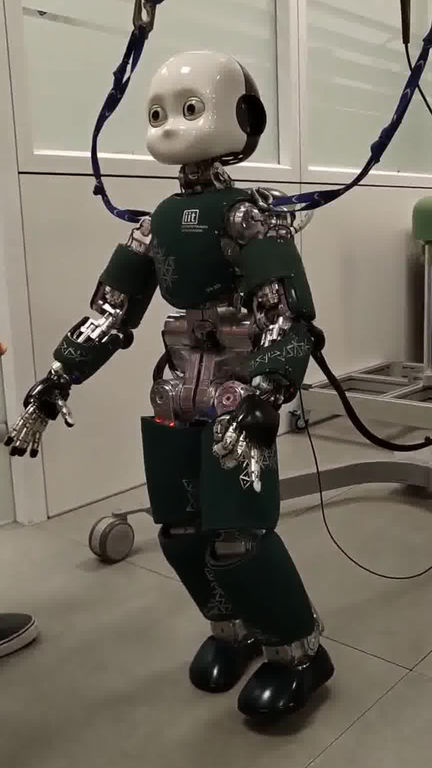}}
	\hspace{.01\textwidth}
	\subfloat[$t=t_0 + 6s$] {\includegraphics[width=.23\textwidth]{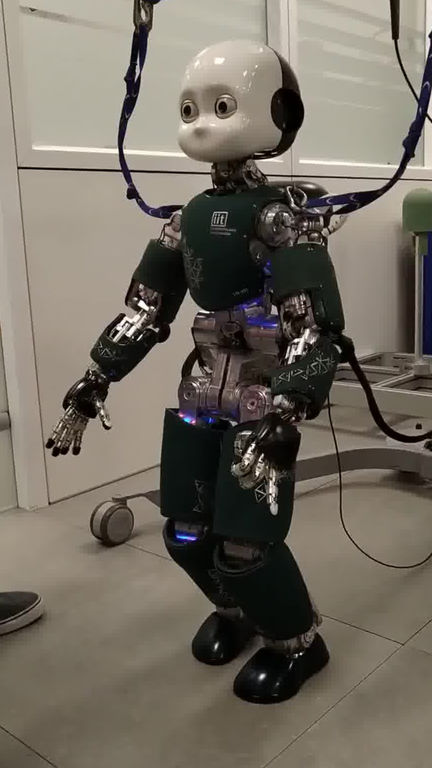}}
	\hspace{.01\textwidth}
	\subfloat[$t=t_0 + 7s$] {\includegraphics[width=.23\textwidth]{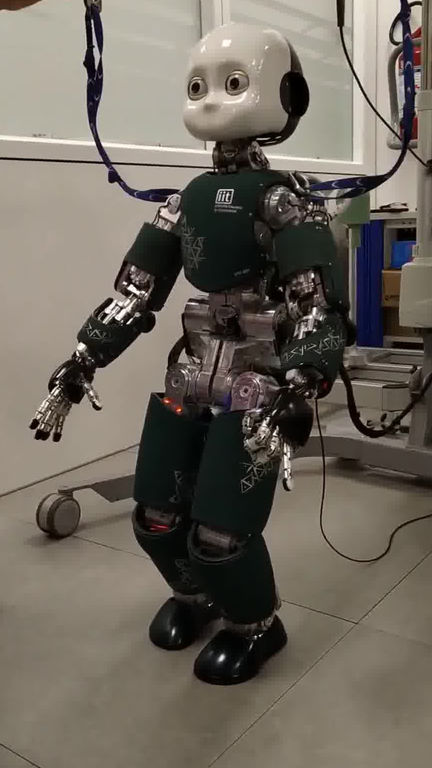}}
	\caption{Snapshots\protect\footnotemark\,of the robot walking using the planned trajectories. The planner generates joint references which are interpolated and stabilized through a joint position controller.}
	\label{fig_dp:snapshots}
\end{figure*}
\footnotetext{\url{https://www.youtube.com/playlist?list=PLBOchT-u69w9hJ6BmqPf06r0zWmungOrc}}
	
To further validate the output of the planner presented in this part, we tested the generated trajectories on the iCub humanoid robot. In particular, they are used as a reference for a joint position controller. Since their frequency is at 10$\mathrm{Hz}$, we interpolate them to have a reference point every 10$\mathrm{ms}$. Hence, the trajectories are replayed on the robot closing the loop only at joint level. The robot was able to perform several steps, as shown in Fig. \ref{fig_dp:snapshots}, demonstrating the feasibility of the generated trajectories.

As a result of Sec. \ref{sec:complementarity_comparison}, we decided to test only the best performing method, i.e. the dynamically-enforced complementarity method. In addition, compared to results shown in Sec. \ref{sec:validation}, we reduce the forward speed to $3\mathrm{cm}/\mathrm{s}$, advancing the mean point reference of $6\mathrm{cm}$ at every step. In this case, it has been also useful to move the desired CoP position, as anticipated in Sec. \ref{sec:forceRegularization}. In particular, by moving it toward the inner foot edge, the robot walks more robustly. 

The trajectories are generated off-line for a span of twenty seconds, after 200 runs of the solver. They are tested in open-loop, thus limiting the maximum velocity achievable by the robot. 
\section{Conclusions} \label{sec:conclusions}

This chapter validates what presented in Chapters \ref{chap:modeling_dp} and \ref{chap:tasks}. It shows that walking trajectories can emerge by specifying a moving reference for the contact points' centroid and the desired CoM velocity only.

The planner considers relatively large time-steps. This enables the insertion of another control loop at higher frequency, whose goal is to stabilize the planned trajectories. 
The code is implemented entirely in C++.
The main bottleneck is represented by the computational time. A single planner iteration may take from slightly less than a second to more than a minute. This prevents an on-line implementation on the real robot. Nevertheless, the continuous formulation of the problem allows the application of techniques, like those presented in \citep{neunert2018whole, farshidian2017real}, which solve the problem through the iterative application of fast LQR solvers.

Finally, given the non-convex nature of the problem defined in Sec. \ref{sec:oc}, it is fundamental to provide a meaningful initial guess. Indeed, local minima may bring the planner to ``procrastinate'', as anticipated in Sec. \ref{sec:considerations}. 
An interesting future work consists in adopting Reinforcement Learning techniques, like \citep{peng2018deepmimic} to warm start the optimization problem.
In addition, the definition of references can affect the time necessary to find a solution. In future works, we will investigate both the adoption of faster solvers and the definition of a more sophisticated and efficient way of providing references.

\bookmarksetup{startatroot}
\chapter*{Epilogue}
\addcontentsline{toc}{chapter}{Epilogue}
The thesis presented the application of predictive techniques to tackle motion generation problems. In particular, Part \ref{part:applications} is devoted to specific applications of model predictive controllers and trajectory optimization. Chapter \ref{chap:steprecovery} uses an MPC controller as an emergency measure, to perform a step in case of strong external perturbation. Results are shown only in simulation. In fact, this controller is effective only if it can determine a reference for the contact wrenches at every control loop. The computational time required is an order of magnitude too high to be applied real-time. The MPC controller exploits the momentum dynamics to generate a control action, approximating the angular momentum to maintain linearity. 

In Chapter \ref{chap:iros_walking}, we further simplify the centroidal momentum dynamics, enabling the adoption of on-line MPC for stabilizing the robot walking when controlled both in position and torque control. Desired footsteps are planned by approximating the robot as a unicycle. The problem of deciding where to step and at what time is solved by framing it as an optimization problem.

Chapter \ref{chap:stepups} enriches the model used to plan walking trajectories, planning dynamic motions exploiting the robot momentum. We consider explicitly the forces applied at the two feet, constrained to produce zero angular momentum. The different walking phases are defined beforehand, but their duration is a decision variable. Noticeably, by using the generated trajectories we are able to reduce up to 20\% the maximum torque required to the leading knee when performing a large step-up.

Motivated by the performance gain obtained with a more detailed model, in Part \ref{part:dynamic_planner} we explore whether it is possible to generate walking motions by specifying a minimal set of references, like a desired center of mass velocity and a reference position for the centroid of the contact points. We also compare how different ways of specifying the complementarity conditions affect the overall planner performances. The generated trajectories are played on the robot to validate their efficacy.

In this last application, computational time represents again a major obstacle limiting this application to offline trajectory generation. Summarizing, experiments on the real robot could be achieved in two cases: the considered model is simple enough to swiftly find a solution to the optimal control problem, or the model is sufficiently detailed to make the generated trajectories meaningful enough to be played on the robot off-line.

The definition of contacts and their planning plays a majors role when generating locomotion trajectories. In all the applications presented in Part \ref{part:applications}, contacts are planned without taking into consideration the state of the robot. In Part \ref{part:dynamic_planner}, we try to address this issue letting the optimizer determining where and when to instantiate a contact. Nevertheless, this was possible through a particular definition of cost function, tuning, and at the price of a higher computational complexity. The problem of deciding a convenient placement of contacts simultaneously to the generation of trajectories is far from being solved, especially in this thesis. Things get even more complicated and less tractable when several parts of the robot can produce a contact. In literature, authors are starting to use new techniques, like Reinforcement Learning, to exploit contacts on the whole robot body for generating complex motions \citep{hwangbo2019learning}, representing an interesting future direction.

The use of multiple limbs affects the definition and detection of ``fall state''. For example, the \emph{Capture Point} concept introduced in Sec. \ref{sec:capture_point}, determines a simple and powerful condition to detect when a robot is about to fall. It applies when the robot can be approximated as a pendulum, but such approximation would be too coarse when multiple contacts are instantiated in non-coplanar surfaces. Hence, in this case, the usual fall detection techniques would fail. In addition, if we consider the robot pelvis or the torso as viable bodies to instantiate a contact, the definition of fall state would complicate too. The robot laying on the floor would correspond to a situation where it is using multiple contacts, back included, to sustain its weight. It is an extreme, but we can argue that such configuration represents a fall state only if the robot was not supposed to reach it. As a consequence, we can propose the undesired loss of gravitational potential energy as a possible definition of ``fall state''. At the same time, a fall can be detected when, giving the current contacts configuration, we are not able to control the robot in the neighborhood of a desired trajectory. To this end, we can exploit a prediction horizon, similarly to Sec. \ref{trigger}, to determine when this situation occurs. A future work consists in extending this concept, thus drawing capturability conditions \citep{koolen2012capturability} also in the multi-contact scenario.

In the Prologue we mentioned that many humanoids fell during the DRC. Most probably, none of the methods presented here, if applied during the Challenge, would have changed that figure. The tuning and testing process, together with the abilities of the team of scientists controlling the robot, play the most important role for what concerns robustness. Nonetheless, we hope that the development of the techniques adopted in this thesis may help in designing controllers with higher degree of robustness requiring a smaller developer effort.

\vspace*{\fill} 
\begin{quotation}
	{\footnotesize
		\noindent\emph{Confusion is better than stupid conclusions. In confusion, there is still a possibility. In stupid conclusions, there is no possibility.}
		\begin{flushright}
			Jaggi Vasudev
		\end{flushright}
	}
\end{quotation}

\vspace*{\fill}

\bookmarksetupnext{level=-1}
\begin{appendices}
	\appendixtitleon
	\appendixtitletocon
	\chapter{Equivalence Between a Contact Wrench and Four Forces at the Vertices of a Rectangular Foot}\label{ap:four_forces}
Let us consider a foot of rectangular shape, having length $l$ and width $d$. We define the foot frame as located at the center of the foot, oriented such that the $x$-axis is parallel with the side edge (of length $l$). The $z$-axis is perpendicular to the foot surface. With this choice, the four corners have the following coordinates:
\begin{equation}
	\begin{array}{ l l }
	{}_1\bm{p} = \left(\hphantom{-}\frac{l}{2}, \hphantom{-}\frac{d}{2}, \hphantom{-}0\right), & {}_2\bm{p} = \left(\hphantom{-}\frac{l}{2}, -\frac{d}{2}, \hphantom{-}0\right), \\
	{}_3\bm{p} = \left(-\frac{l}{2}, \hphantom{-}\frac{d}{2}, \hphantom{-}0\right), & {}_4\bm{p} = \left(-\frac{l}{2}, -\frac{d}{2}, \hphantom{-}0\right).
	\end{array}
\end{equation}
We suppose a 3D force ${}_i\bm{f}$ is applied at each corner $i$.
Assuming we have a 6D wrench ${}_\text{ext}\textbf{f} = \left[{}_\text{ext}\bm{f}^\top \, {}_\text{ext}\bm{\tau}^\top\right]^\top$ applied in the foot frame, we want to determine the corresponding value for the four corner forces. In particular, the following two conditions hold:
\begin{IEEEeqnarray}{RCL}
	\IEEEyesnumber
{}_1\bm{f} + {}_2\bm{f}+ {}_3\bm{f}+ {}_4\bm{f} &=& {}_\text{ext}\bm{f} \IEEEyessubnumber\\
\sum {}_i\bm{p} \times {}_i\bm{f} &=& {}_\text{ext}\bm{\tau}. \IEEEyessubnumber
\end{IEEEeqnarray}

Let us start by considering only the forces normal component. These have to match the total normal force and the torque applied along $x$ and $y$. Hence, we can extract the following three equations: 
\begin{IEEEeqnarray}{RCL}
\IEEEyesnumber \phantomsection \label{eq:corner_constraints}
{}_1f_z + {}_2f_z + {}_3f_z + {}_4f_z &=& {}_\text{ext}f_z, \IEEEyessubnumber \label{eq:summation_contraint}\\
\frac{d}{2}({}_1f_z + {}_3f_z) - \frac{d}{2}({}_2f_z + {}_4f_z) &=& {}_\text{ext}{\tau}_x, \IEEEyessubnumber \\ 
\frac{l}{2}({}_3f_z + {}_4f_z) - \frac{l}{2}({}_1f_z + {}_2f_z) &=& {}_\text{ext}{\tau}_y. \IEEEyessubnumber 
\end{IEEEeqnarray}
From the last two equalities, we can obtain:
\begin{IEEEeqnarray}{RCL}
	\IEEEyesnumber
{}_3f_z = {}_2f_z + \frac{{}_\text{ext}{\tau}_x}{d} + \frac{{}_\text{ext}{\tau}_y}{l}, \IEEEyessubnumber \label{eq:f_3 partial}\\
{}_1f_z = {}_4f_z + \frac{{}_\text{ext}{\tau}_x}{d} - \frac{{}_\text{ext}{\tau}_y}{l}.\IEEEyessubnumber \label{eq:f_1 partial}
\end{IEEEeqnarray}
We can now use Eq. \eqref{eq:summation_contraint} to substitute ${}_2f_z$, i.e.
\begin{equation*}
{}_2f_z = {}_\text{ext}f_z - {}_1f_z - {}_3f_z - {}_4f_z
\end{equation*}
into Eq. \eqref{eq:f_3 partial}, which leads to
\begin{IEEEeqnarray}{RCL}
	\IEEEyesnumber
{}_3f_z &=& {}_\text{ext}f_z - {}_1f_z - {}_3f_z - {}_4f_z + \frac{{}_\text{ext}\tau_x}{d} + \frac{{}_\text{ext}\tau_y}{l}, \nonumber\\
&=& \frac{{}_\text{ext}f_z}{2} - \frac{{}_1f_z}{2} - \frac{{}_4f_z}{2} + \frac{{}_\text{ext}\tau_x}{2d} + \frac{{}_\text{ext}\tau_y}{2l}.
\end{IEEEeqnarray}
Now, we can also substitute Eq. \eqref{eq:f_1 partial} to obtain
\begin{IEEEeqnarray}{RCL}
	\IEEEyesnumber
{}_3f_z &=& \frac{{}_\text{ext}f_z}{2} - \frac{{}_4f_z + \frac{{}_\text{ext}\tau_x}{d} - \frac{{}_\text{ext}\tau_y}{l}}{2} - \frac{{}_4f_z}{2} + \frac{{}_\text{ext}\tau_x}{2d} + \frac{{}_\text{ext}\tau_y}{2l}, \nonumber\\
&=& \frac{{}_\text{ext}f_z}{2} -{}_4f_z + \frac{{}_\text{ext}\tau_y}{l}.
\end{IEEEeqnarray}
Thus, we can finally write ${}_1f_z$, ${}_2f_z$ and ${}_3f_z$ as a function of ${}_4f_z$:
\begin{IEEEeqnarray}{RCL}
	\IEEEyesnumber
{}_1f_z &=& {}_4f_z + \frac{{}_\text{ext}\tau_x}{d} - \frac{{}_\text{ext}\tau_y}{l}, \IEEEyessubnumber\\
{}_2f_z &=& \frac{{}_\text{ext}f_z}{2} -{}_4f_z - \frac{{}_\text{ext}\tau_x}{d}, \IEEEyessubnumber\\
{}_3f_z &=&  \frac{{}_\text{ext}f_z}{2} -{}_4f_z + \frac{{}_\text{ext}\tau_y}{l}.\IEEEyessubnumber
\end{IEEEeqnarray}
There may be infinite solutions according to the choice of ${}_4f_z$. Nevertheless, if we impose these normal forces to be positive, the set of feasible solutions may considerably shrink. 

In fact, by considering ${}_if_z \geq 0$, we obtain the following set of inequalities:
\begin{IEEEeqnarray}{RCL}
	\IEEEyesnumber \phantomsection \label{eq:4_forces_inequality}
{}_4f_z & \geq & 0, \IEEEyessubnumber\\
{}_4f_z & \geq & -\frac{{}_\text{ext}\tau_x}{d} + \frac{{}_\text{ext}\tau_y}{l}, \IEEEyessubnumber\\
{}_4f_z & \leq & \frac{{}_\text{ext}f_z}{2} - \frac{{}_\text{ext}\tau_x}{d}, \IEEEyessubnumber\\
{}_4f_z & \leq & \frac{{}_\text{ext}f_z}{2} + \frac{{}_\text{ext}\tau_y}{l}\IEEEyessubnumber.
\end{IEEEeqnarray}

Let us divide all the inequalities above by ${}_\text{ext}f_z$ and define ${}_i\alpha$ as the ratio of the total normal force each corner is carrying, i.e., ${}_i\alpha = ({}_if_z)/({}_\text{ext}f_z)$. In addition, the CoP measured in the foot frame has the following coordinates:
\begin{equation*}
\text{CoP}_x = -\frac{{}_\text{ext}\tau_y}{{}_\text{ext}f_z}, \quad \text{CoP}_y = \frac{{}_\text{ext}\tau_x}{{}_\text{ext}f_z}.
\end{equation*}
We can finally rewrite Eq. \eqref{eq:4_forces_inequality} accordingly:
\begin{IEEEeqnarray}{RCL}
	\IEEEyesnumber
{}_4\alpha & \geq & 0, \IEEEyessubnumber\\[1pt]
{}_4\alpha & \geq & -\frac{\text{CoP}_y}{d} - \frac{\text{CoP}_x}{l}, \IEEEyessubnumber\\[1pt]
{}_4\alpha & \leq & \frac{1}{2} -\frac{\text{CoP}_y}{d}, \IEEEyessubnumber\\[1pt]
{}_4\alpha & \leq & \frac{1}{2}  - \frac{\text{CoP}_x}{l}. \IEEEyessubnumber 
\end{IEEEeqnarray}
Since the CoP is constrained inside the foot polygon, i.e. $\text{CoP}_x \in [-l/2, l/2]$ and $\text{CoP}_y \in [-d/2, d/2]$, we also obtain that ${}_4\alpha \in [0, 1]$. In fact, a negative $\alpha$ would correspond to a negative normal force. On the contrary, any $\alpha$ greater than 1 would impose a normal force in another corner to be negative to satisfy Eq. \eqref{eq:summation_contraint}. We can condensate the bounds as follows:
\begin{equation}
\max \left(0, -\frac{\text{CoP}_y}{d} - \frac{\text{CoP}_x}{l}\right) \leq {}_4\alpha \leq \min\left(\frac{1}{2} -\frac{\text{CoP}_y}{d}, \frac{1}{2}  - \frac{\text{CoP}_x}{l}\right).
\end{equation}
Hence, when the CoP is in position ${}_4\bm{p}$, the two bounds will coincide and equal to 1. If the CoP is in the opposite corner, ${}_1\bm{p}$, then ${}_4\alpha$ needs to be 0. In all the other cases, the bounds do not coincide, obtaining infinite possible values for ${}_4\alpha$. A simple solution is to choose ${}_4\alpha$ as the mean point of the bounds. This leads to the following:
\begin{IEEEeqnarray}{RCL}
	\IEEEyesnumber \phantomsection
{}_4\alpha &=& \frac{\left(\max \left(0, -\frac{\text{CoP}_y}{d} - \frac{\text{CoP}_x}{l}\right) + \min\left(\frac{1}{2} -\frac{\text{CoP}_y}{d}, \frac{1}{2}  - \frac{\text{CoP}_x}{l}\right)\right)}{2}, \IEEEeqnarraynumspace\IEEEyessubnumber\\
{}_1\alpha &=& \hphantom{ - }{}_4\alpha + \frac{\text{CoP}_x}{l} + \frac{\text{CoP}_y}{d}, \IEEEyessubnumber\\
{}_2\alpha &=& -{}_4\alpha + \frac{1}{2} -\frac{\text{CoP}_y}{d}, \IEEEyessubnumber\\
{}_3\alpha &=& -{}_4\alpha +  \frac{1}{2}  - \frac{\text{CoP}_x}{l}.  \IEEEyessubnumber
\end{IEEEeqnarray}
Interestingly, with this choice, when the CoP is in the center of the foot, all ${}_i\alpha$ are equal to $1/4$.

Once all the normal forces have been identified, the other components, i.e. ${}_if_x$ and ${}_if_y$, can be obtained via pseudo-inverse starting from Eq. \eqref{eq:corner_constraints}. In particular, we have to satisfy the following three conditions
\begin{IEEEeqnarray}{RCL}
	\IEEEyesnumber
	{}_1f_x + {}_2f_x + {}_3f_x + {}_4f_x &=& {}_\text{ext}f_x, \IEEEyessubnumber \\
	{}_1f_y + {}_2f_y + {}_3f_y + {}_4f_y &=& {}_\text{ext}f_y, \IEEEyessubnumber \\
	\sum \begin{bmatrix}
		-{}_ip_y & {}_ip_x
	\end{bmatrix} 
	\begin{bmatrix}
		{}_if_x \\ {}_if_y
	\end{bmatrix}&=& {}_\text{ext}{\tau}_z, \IEEEyessubnumber 
\end{IEEEeqnarray}
using eight unknowns. Hence there are infinite solutions. Notice that friction constraints may not be satisfied by a corner force, even if the original contact wrench does. 
\chapter{Jacobians of Kinematic and Dynamic Quantities}\label{chap:jacobians}

In this appendix, we present the partial derivatives of several quantities exploited in Sec. \ref{sec:oc}. As discussed in Sec. \ref{sec:dp_software_infrastructure}, the optimal control problem is turned into an optimization problem and solved through off-the-shelf solvers. Such tools need to evaluate the partial derivatives of cost and constraint functions. To this end, we define an expression for those derivatives which are not trivially computable, usually involving non-linear functions of the variables involved in Eq. \eqref{eq:dp_state_control}, namely in $\bm{\mathcal{X}}$ and $\bm{\mathcal{U}}$. In this appendix, we use extensively the velocity trivializations presented in Sec. \ref{sec:trivializatons}. The dissertation shown here complements the results of \citep{carpentier2018analytical}.

\section{Partial derivative of a rotated vector} \label{sec:rotationPartialDer}
In this section, we face the partial derivatives computation of expressions involving a rotation matrix. A possible example is given by Eq. \eqref{eq:quaternionLeftDer}, where we seek to find the partial derivative of ${}^\mathcal{I}\bm{R}_B {}^B\bm{v}_{\mathcal{I},B}$ with respect to the base quaternion $^{\mathcal{I}} \bm{\rho}_B$. The computation of  $\frac{\partial}{\partial ^{\mathcal{I}} \bm{\rho}_B} {}^\mathcal{I}\bm{R}_B$ would result in a tensor, difficult to be handled. Indeed, it is easier to consider the case where the rotation matrix is multiplied by a generic vector $\bm{\chi} \in \mathbb{R}^3$.

Define $\bm{R} \in SO(3)$ as a generic rotation matrix and $\bm{\rho} \in \mathbb{H}$ the corresponding quaternion. We can rewrite the product $\bm{R} \bm{\chi}$ by using the Rodrigues formula \citep{murray2017mathematical}:
\begin{equation} \label{eq:rodriguesFormula}
\bm{R} \bm{\chi} = \bm{\chi} + 2\rho_w \bm{r}^\wedge \bm{\chi} + 2\bm{r}^\wedge \bm{r}^\wedge\bm{\chi},
\end{equation}
where
\begin{equation}
\bm{r} = \begin{bmatrix}
\rho_i \\
\rho_j \\
\rho_k
\end{bmatrix}
\end{equation}
contains the imaginary part of the quaternion and $\rho_w$ is its real part.
Thus:
\begin{IEEEeqnarray}{RCL}
	\IEEEyesnumber \phantomsection
\frac{\partial}{\partial \rho_w} \bm{R} \bm{\chi} &=& 2\bm{r}^\wedge\bm{\chi}, \IEEEyessubnumber\\
\frac{\partial}{\partial \bm{r}} \bm{R} \bm{\chi} &=& -2\rho_w \bm{\chi}^\wedge + 2 \frac{\partial}{\partial \bm{r}}\left(\bm{r}^\wedge\right) \bm{r}^\wedge \bm{\chi} + 2 \bm{r}^\wedge \frac{\partial}{\partial \bm{r}}\left(\bm{r}^\wedge\right)\bm{\chi}, \IEEEyessubnumber\\
&=& -2\rho_w\bm{\chi}^\wedge - 2 \left(\bm{r}^\wedge \bm{\chi }\right)^\wedge - 2 \bm{r}^\wedge
\bm{\chi}^\wedge, \IEEEyessubnumber \\
\frac{\partial}{\partial \bm{\rho}} \bm{R} \bm{\chi} &=& \begin{bmatrix}
 2\bm{r}^\wedge\bm{\chi} & -2\rho_w\bm{\chi}^\wedge - 2 \left(\bm{r}^\wedge \bm{\chi }\right)^\wedge - 2 \bm{r}^\wedge
 \bm{\chi}^\wedge
\end{bmatrix}. \IEEEyesnumber \label{eq:rotationDerivative}
\end{IEEEeqnarray}
Here we extensively used the property $(\bm{x})^\wedge \bm{y} = -(\bm{y})^\wedge \bm{x}$.

\section{Partial derivatives of a non unitary quaternion}\label{sec:normalizedQuaternion}
The quaternion modulus constraint can be broken across solver iterations. In fact, a generic solver can search over an unfeasible region while reaching the optimal solution. When computing the partial derivatives involving the base quaternion, we are implicitly assuming that it expresses a rotation (see for example Eq. \eqref{eq:rodriguesFormula}). This is true only if it has unit modulus. In order to satisfy this assumption, we have to normalize the quaternion before using it:
\begin{equation}
\bm{\rho}^N = \frac{\bm{\rho}}{\|\bm{\rho}\|} = \frac{\bm{\rho}}{\sqrt{\bm{\rho}^\top\bm{\rho}}}.
\end{equation}
This means that in order to compute the partial derivatives with respect to the not unitary quaternion, we have to perform the following operations:
\begin{equation}
\frac{\partial (\cdot)}{\partial \bm{\rho}} = \frac{\partial (\cdot)}{\partial \bm{\rho}^N} \frac{\partial \bm{\rho}^N}{\partial \bm{\rho}}.
\end{equation}
The first component, $\frac{\partial (\cdot)}{\partial \bm{\rho}^N}$, is the partial derivative of a generic function considering a normalized quaternion. For what concerns the second part instead:
\begin{IEEEeqnarray}{RCL}
	\phantomsection \IEEEyesnumber \label{eq:normalizedQuaternonDerivative}
	\frac{\partial \bm{\rho}^N}{\partial \bm{\rho}} &=& \frac{1}{\|\bm{\rho}\|}\bm{I}_4 - \left(\bm{\rho}^\top\bm{\rho}\right)^{-\frac{3}{2}}\bm{\rho} \bm{\rho}^\top, \IEEEyessubnumber \\
	&=& \frac{1}{\left(\bm{\rho}^\top\bm{\rho}\right)\|\bm{\rho}\|}\left(\left(\bm{\rho}^\top\bm{\rho}\right)\bm{I}_4 - \bm{\rho}\bm{\rho}^\top\right). \IEEEyessubnumber \label{eq:quaternionCorrection}
\end{IEEEeqnarray}
Thus, Eq. \eqref{eq:quaternionCorrection} must be multiplied on the right of all derivatives which consider an unit quaternion.

\section{Partial derivatives of adjoint matrices}\label{sec:adjoint_derivatives}
In this section, we present the partial derivative of an adjoint transform with respect to joint values. Let us consider two frames $P$ and $C$ connected through a single joint $j$. The time derivative of the adjoint matrix ${}^P\bm{X}_C$ is as follows
\citep[Eq. (2.36)]{traversaro2017thesis}:
\begin{equation}
{}^P\dot{\bm{X}}_C = {}^P\bm{X}_C \left({}^C\bm{V}_{P,C}\times \right).
\end{equation}
The operator $\times$ is defined such that, given ${}^C\textbf{s}_{P,C} = \left[{}^C \bm{v}_{P,C}^\top {}^C \bm{\omega}_{P,C}^\top\right]^\top$ we obtain:
\begin{equation}
{}^C\textbf{s}_{P,C}\times := 
\begin{bmatrix}
{}^C \bm{\omega}^\wedge_{P,C} & {}^C \bm{v}^\wedge_{P,C} \\
0_{3\times3} & {}^C \bm{\omega}^\wedge_{P,C}
\end{bmatrix}.
\end{equation}
From Eq. \eqref{eq:velocity_expanded} we have
\begin{equation}
{}^C\textbf{V}_{P,C} = {}^C\textbf{s}_{P,C} ~\dot{s}_j,
\end{equation}
thus we finally obtain
\begin{equation}\label{eq:adjoint_partial_derivative}
\frac{\partial}{\partial s_j} {}^P\bm{X}_C = {}^P\bm{X}_C \left({}^C\textbf{s}_{P,C}\times \right).
\end{equation}

Let us consider now a more generic case where ${}^D\bm{X}_E$ defines a transformation between two frames, $D$ and $E$, rigidly attached to the robot. The partial derivative with respect to $s_j$ depends on whether joint $j$ belongs to $\pi_D(E)$. If this is the case, we have
\begin{equation}
	{}^D\bm{X}_E = {}^D\bm{X}_P {}^P\bm{X}_C {}^C\bm{X}_E,
\end{equation}	
where only the transformation ${}^P\bm{X}_C$ depends on $s_j$. As a consequence
\begin{equation}\label{eq:multiple_adjoints_derivative}
	\begin{split}
		\frac{\partial}{\partial s_j} {}^D\bm{X}_E &=  {}^D\bm{X}_P {}^P\bm{X}_C \left({}^C\textbf{s}_{P,C}\times \right){}^C\bm{X}_E, \\
		&= {}^D\bm{X}_C \left({}^C\textbf{s}_{P,C}\times \right){}^C\bm{X}_E.
	\end{split}
\end{equation}
If instead $j \notin \pi_D(E)$ the partial derivative is null. To summarize
\begin{equation}\label{eq:adjoint_derivative}
	\frac{\partial}{\partial s_j} {}^P\bm{X}_C = \begin{cases}
	{}^D\bm{X}_C \left({}^C\textbf{s}_{P,C}\times \right){}^C\bm{X}_E, &\text{if } j \in \pi_D(E),\\
	0 &\text{otherwise}.
	\end{cases} 
\end{equation}

Analogously, it is possible to define the partial derivative of the 6D force coordinate transformation ${}_P\bm{X}^C$.
In particular,
\begin{equation}\label{eq:adjoint_wrench_derivative}
\frac{\partial}{\partial s_j} \left({}_P \bm{X}^C\right) = {}_{P} \bm{X}^C {}^C\textbf{s}_{P,C}\bar{\times}^*.
\end{equation}
The operator $\bar{\times}^*$ is defined such that, given ${}^C\textbf{s}_{P,C} = \left[{}^C \bm{v}_{P,C}^\top {}^C \bm{\omega}_{P,C}^\top\right]^\top$ we obtain \citep[Eq. (2.50)]{traversaro2017thesis}:
\begin{equation}
{}^C\textbf{s}_{P,C}\bar{\times}^* := 
\begin{bmatrix}
{}^C \bm{\omega}^\wedge_{P,C} & \bm{0}_{3\times3}  \\
{}^C \bm{v}^\wedge_{P,C} & {}^C \bm{\omega}^\wedge_{P,C}
\end{bmatrix}.
\end{equation}

\section{Partial derivatives of transformation matrices}\label{sec:rel_transf_derivative}

The relative transform ${}^D\bm{H}_E$ determines a change of coordinate for a generic vector ${}^E\bm{\chi}$, such that:
\begin{equation}\label{eq:generic_transform}
	{}^D\bm{\chi} = {}^D\bm{H}_E {}^E\bm{\chi}.
\end{equation}
Assuming ${}^E\bm{\chi}$ to be constant, this equation would depend only on the joint values connecting the two frames. In order to compute the partial derivative with respect to a generic joint value $s_j$, first we can rewrite Eq. \eqref{eq:generic_transform} as 
\begin{equation}
	{}^D\bm{\chi} = {}^D\bm{R}_E {}^E\bm{\chi} + {}^D\bm{o}_E.
\end{equation}

\subsection{Relative position}
The partial derivative of a relative position with respect to joint values can be obtained through the relative Jacobian. Denoting ${}^E\bm{J}_{D,E}$ as the \emph{left-trivialized} Jacobian, the relative \emph{left-trivialized} velocity is equal to
\begin{equation}
	{}^E\bm{V}_{D,E} = \begin{bmatrix}
	{}^D\bm{R}_E^\top ~ {}^D\dot{\bm{o}}_E \\
	{}^E\bm{\omega}_{D,E}
	\end{bmatrix} = {}^E\bm{J}_{D,E} \dot{\bm{s}},
\end{equation}
from which we obtain 
\begin{equation}
	{}^D\dot{\bm{o}}_E = {}^D\bm{R}_E\begin{bmatrix}
	\mathds{1}_3 & \bm{0}_{3\times3}
	\end{bmatrix} {}^D\bm{J}_{D,E}~\dot{\bm{s}} = \frac{\partial {}^D\bm{o}_E}{\partial s} \frac{\partial s}{\partial t},
\end{equation}
hence
\begin{equation}
	\frac{\partial}{\partial \bm{s}} {}^D\bm{o}_E = {}^D\bm{R}_E\begin{bmatrix}
	\mathds{1}_3 & \bm{0}_{3\times3}
	\end{bmatrix} {}^D\bm{J}_{D,E}.
\end{equation}
Assuming $j \in \pi_D(E)$, the partial derivative w.r.t joint $j$ consists in
\begin{equation}
	\frac{\partial}{\partial s_j} {}^D\bm{o}_E = {}^D\bm{R}_E\begin{bmatrix}
	\mathds{1}_3 & \bm{0}_{3\times3}
	\end{bmatrix}{}^E \bm{X}_C {}^C\textbf{s}_{P,C}.
\end{equation}

\subsection{Relative rotation}
The partial derivative of ${}^D\bm{R}_E {}^E\bm{\chi}$ instead requires some manipulation. In particular it is possible to exploit the result of Eq. \eqref{eq:adjoint_derivative} by noting in Eq. \eqref{eq:from_left_to_right_trivialization} that the upper left block of the adjoint matrix contains the rotation between the two frames. Thus, we have
\begin{equation}
	{}^D\bm{R}_E {}^E\bm{\chi} = \begin{bmatrix}
	\mathds{1}_3 & \bm{0}_{3\times3}
	\end{bmatrix} {}^D\bm{X}_E \begin{bmatrix}
	{}^E\bm{\chi} \\
	\bm{0}_{3\times 1}
	\end{bmatrix},
\end{equation}
or alternatively
\begin{equation}
{}^D\bm{R}_E {}^E\bm{\chi} = \begin{bmatrix}
\mathds{1}_3 & \bm{0}_{3\times3}
\end{bmatrix} {}^D\bm{X}_E \begin{bmatrix}
\mathds{1}_3 \\ \bm{0}_{3\times3}
\end{bmatrix} {}^E\bm{\chi}.
\end{equation}
Assuming $j \in \pi_D(E)$, we obtain
\begin{equation}
	\frac{\partial}{\partial s_j} {}^D\bm{R}_E{}^E\bm{\chi} = \begin{bmatrix}
	\mathds{1}_3 & \bm{0}_{3\times3}
	\end{bmatrix} {}^D\bm{X}_C \left({}^C\textbf{s}_{P,C}\times \right){}^C\bm{X}_E \begin{bmatrix}
	{}^E\bm{\chi} \\
	\bm{0}_{3\times 1}
	\end{bmatrix},
\end{equation}
allowing us to build $\frac{\partial}{\partial \bm{s}} {}^D\bm{R}_E{}^E\bm{\chi}$ by column.

\subsection{Transformations relative to the inertial frame} \label{sec:base_transforms_derivative}
Eq. \eqref{eq:dp_point_consistency} presents a coordinate transformation of a (fixed) vector ${}^\text{foot}{}_i \bm{p}$. The transformation matrix ${}^\mathcal{I} \bm{H}_\text{foot}$ depends on the base pose and on all the joints connecting the foot frame to the base. It is possible to split this relation in the following two equations:
\begin{IEEEeqnarray}{RCL}
	\phantomsection \IEEEyesnumber
	{}_i \bm{p} &=& {}^\mathcal{I} \bm{H}_B {}^B{}_i \bm{p}, \IEEEyessubnumber \label{eq:transform_base_to_I}\\
	{}^B{}_i \bm{p} &=& {}^B \bm{H}_\text{foot} {}^\text{foot}{}_i \bm{p}. \IEEEyessubnumber \label{eq:transform_point_to_base}
\end{IEEEeqnarray}
Eq. \eqref{eq:transform_point_to_base} depends only on joint positions and its partial derivative can be computed as showed in Sec. \ref{sec:rel_transf_derivative}. On the other hand, Eq. \eqref{eq:transform_base_to_I} depends only on the base position ${}^\mathcal{I}\bm{p}_B$ and its orientation, expressed through the quaternion ${}^\mathcal{I}\bm{\rho}_B$. By rewriting it as
\begin{equation}
	{}_i \bm{p} = {}^\mathcal{I} \bm{R}_B {}^B{}_i \bm{p} + {}^\mathcal{I}\bm{p}_B,
\end{equation}
the derivative with respect to the base position can be computed trivially, while $\frac{\partial}{\partial {}^\mathcal{I}\bm{\rho}_B} \left({}^\mathcal{I} \bm{R}_B {}^B{}_i \bm{p}\right)$ can be computed as in Sec. \ref{sec:rotationPartialDer}.

\section{Partial derivatives of the base velocity}
The \emph{left-trivialized} base velocity is defined as 
\begin{equation}
	{}^B\bm{V}_{\mathcal{I},B} = \begin{bmatrix}
	{}^B\bm{v}_{\mathcal{I},B} \\
	{}^B\bm{\omega}_{\mathcal{I},B}
	\end{bmatrix}.
\end{equation}
While the linear part is contained in the optimization variables, the angular part is not. In fact, the quaternion time derivative is used instead inside $\bm{\mathcal{U}}$. As shown in Eq. \eqref{eq:base_angular_velocity}, the base angular velocity depends linearly on the quaternion derivative, while the map $\bm{\mathcal{G}}$ depends on the base quaternion ${}^\mathcal{I}\bm{\rho}_B$. For simplicity, let us drop the superscripts and define $\bm{\omega}$ as the \emph{left-trivialized} angular velocity. $r$ is the imaginary part of the base quaternion and $\rho_w$ is its real part while $\dot{\bm{r}}$ and $\dot{\rho}_w$ compose the quaternion derivative. Then, from \citep[Sec. 1.5.4]{graf2008quaternions}, we have
\begin{equation}
	\bm{\omega} = 2\left( \rho_w\dot{\bm{r}} - \dot{\rho}_w\bm{r} - \bm{r}^\wedge\dot{\bm{r}} \right).
\end{equation}
We finally obtain
\begin{equation}
	\frac{\partial}{\partial \bm{\rho}}\bm{\omega} = \begin{bmatrix}
	\dot{\bm{r}} & \dot{\bm{r}}^\wedge - \rho_w \mathds{1}_3
	\end{bmatrix}.
\end{equation}

\section{Partial derivatives of a link velocity}\label{sec:velocity_derivative}
Let us consider the \emph{left-trivialized} velocity of a generic link $L$.
This element depends on a series of quantities, namely:
\begin{itemize}
	\item joint positions $\bm{s}$,
	\item joint velocities $\dot{\bm{s}}$,
	\item base velocity ${}^B\bm{V}_{\mathcal{I},B}$.
\end{itemize}
The partial derivative with respect to $\dot{\bm{s}}$ and ${}^B\bm{V}_{\mathcal{I},B}$ are trivially obtained from the corresponding \emph{left-trivialized} Jacobian. Contrarily, the partial derivative with respect to joint values requires some manipulation.

At first, we compute the partial derivative of a velocity relative to the robot base (hence not depending on base quantities). Subsequently, the base velocity is added. 
In Eq. \eqref{eq:velocity_expanded}, the matrix ${}^L\bm{X}_{m_B(k)}$ depends on (a subset of) the joint variables.
By considering one single joint at a time, the partial derivative of ${}^L\bm{V}_{B,L}$ is:
\begin{equation}
\frac{\partial}{\partial s_j} {}^L\bm{V}_{B,L} = \sum_{k \in \pi_B(L)} \left({}^L\bm{X}_C \left(\frac{\partial}{\partial s_j} {}^C\bm{X}_P \right)  {}^P\bm{X}_{m_B(k)}  {}^{m_B(k)}\textbf{s} \right) \dot{s}_k,
\end{equation}
where $C = m_B(j)$, i.e. the child of joint $j$ and $P = \lambda_B\left(j\right)$ its parent.
By applying Eq. \eqref{eq:adjoint_partial_derivative} we obtain:
\begin{equation}
\frac{\partial}{\partial s_j} {}^L\bm{V}_{B,L} = \sum_{k \in \pi_B(L)} \left( {}^L\bm{X}_C \left({}^C\bm{X}_P {}^P\textbf{s}_{C,P} \times \right)  {}^P\bm{X}_{m_B(k)}  {}^{m_B(k)}\textbf{s} \right) \dot{s}_k.
\end{equation}

Note that, compared to Eq. \eqref{eq:adjoint_partial_derivative}, the order of $P$ and $C$ is inverted. We can restore the previous order by noting that 
\begin{equation}
\begin{split}
	{}^P\textbf{s}_{P,C} &= -{}^P\textbf{s}_{C,P}, \\
	{}^C\textbf{s}_{P,C} &= {}^C \bm{X}_P {}^P\textbf{s}_{P,C}.
\end{split}
\end{equation}
In addition, it is possible to exploit the following property:
\begin{equation}
	\left( \left( {}^P\bm{X}_C{}^C\textbf{s}_{P,C} \right) \times\right) =  {}^P\bm{X}_C\left({}^C\textbf{s}_{P, C} \times\right) {}^C\bm{X}_P.
\end{equation}
so that
\begin{equation}
	{}^C\bm{X}_P {}^P\textbf{s}_{C,P} \times = -{}^C\textbf{s}_{P,C} \times {}^C\bm{X}_P.
\end{equation}
As a consequence, we obtain
\begin{IEEEeqnarray}{RCL}
	\IEEEyesnumber \phantomsection
	\frac{\partial}{\partial s_j} {}^L\bm{V}_{B,L} &=& -\sum_{k \in \pi_B(L)} \left( {}^L\bm{X}_C \left({}^C\textbf{s}_{P,C} \times {}^C\bm{X}_P\right)  {}^P\bm{X}_{m_B(k)}  {}^{m_B(k)}\textbf{s} \right) \dot{s}_k, \nonumber \\
	&=& -\sum_{k \in \pi_B(L)} \left( {}^L\bm{X}_C {}^C\textbf{s}_{P,C} \times {}^C\bm{X}_{m_B(k)}  {}^{m_B(k)}\textbf{s} \right) \dot{s}_k. 
\end{IEEEeqnarray}
Joint $j$ must also belong to $\pi_B(L)$ and $k \leq j$, i.e. joint $j$ must be in the chain from $k$ to $L$ (using $B$ as base), otherwise the matrix ${}^L\bm{X}_{m_B(k)}$ does not depend on $s_j$. Henceforth, the summation is on the part of the kinematic tree which starts from $B$ and ends in $C$. This set corresponds to $\pi_B(C)$. Thus, we have:
\begin{equation}
\frac{\partial}{\partial s_j} {}^L\bm{V}_{B,L} =  -{}^L\bm{X}_C {}^C\textbf{s}_{P,C} \times  \sum_{k \in \pi_B(C)} \left({}^C\bm{X}_{m_B(k)}  {}^{m_B(k)}\textbf{s} \right) \dot{s}_k.
\end{equation}
Applying Eq. \eqref{eq:velocity_expanded} to the summation on the right hand side, we can write  
\begin{equation}\label{eq:velocityPartialDerivativeCompact}
\frac{\partial}{\partial s_j} {}^L\bm{V}_{B,L} = -{}^L\bm{X}_C {}^C\textbf{s}_{P,C} \times {}^C \bm{V}_{B,C}.
\end{equation}
	
Let's consider now the velocity of link $L$ with respect to an inertial frame $\mathcal{I}$, obtained through Eq. \eqref{eq:velocity_inertial}:
\begin{equation}\label{eq:velocityExpansion}
\frac{\partial}{\partial s_j} {}^L\bm{V}_{\mathcal{I},L} = \frac{\partial}{\partial s_j}  {}^L\bm{X}_B {}^B\bm{V}_{\mathcal{I},B} + \frac{\partial}{\partial s_j} {}^L\bm{V}_{B,L}.
\end{equation}
The second part is obtained through Eq. \eqref{eq:velocityPartialDerivativeCompact}, while the first is computed as in Eq. \eqref{eq:adjoint_derivative} obtaining
\begin{equation}\label{eq:base_part_Velocity_derivative}
\frac{\partial}{\partial s_j}  {}^L\bm{X}_B {}^B\bm{V}_{\mathcal{I},B} =
\begin{cases}
-{}^L\bm{X}_C {}^C\textbf{s}_{P,C} \times {}^C\bm{X}_B {}^B\bm{V}_{\mathcal{I},B}, & j \in \pi_B(L), \\
0 & \text{otherwise}.
\end{cases}
\end{equation}
Finally, substituting Eq. \eqref{eq:velocityPartialDerivativeCompact} and Eq. \eqref{eq:base_part_Velocity_derivative} into Eq. \eqref{eq:velocityExpansion}, we obtain
\begin{equation}\label{eq:velocity_inertial_derivative}
\frac{\partial}{\partial s_j} {}^L\bm{V}_{\mathcal{I},L} = \begin{cases}
{}^L\bm{X}_C {}^C\textbf{s}_{P,C} \times \left(-{}^C\bm{V}_{A,C}\right), & j \in \pi_B(L) \\
0 & \text{otherwise}.
\end{cases}
\end{equation}

\section{Partial derivatives of the CoM position}\label{sec:com_jacobian}
Eq. \eqref{eq:comConsistency} imposes that $\bm{x}_\text{CoM}$ should match the CoM position computed from forward kinematics, $\text{CoM}({}^\mathcal{I}\bm{p}_B, {}\mathcal{I}\bm{\rho}_B, s)$.
In order to compute the partial derivative of such function w.r.t base and joint variables, we can resort to the CoM jacobian $\bm{J}_\text{CoM}$. We can split $\bm{J}_\text{CoM}$ as follows:
\begin{equation}
\bm{J}_\text{CoM} = \begin{bmatrix}
\bm{J}_\text{CoM}^{v_b} & \bm{J}_\text{CoM}^{\omega_b} & \bm{J}_\text{CoM}^{\dot{s}}
\end{bmatrix},
\end{equation}
such that
\begin{equation}
	\bm{v}_\text{CoM} = \begin{bmatrix}
	\bm{J}_\text{CoM}^{v_b} & \bm{J}_\text{CoM}^{\omega_b} & \bm{J}_\text{CoM}^{\dot{s}}
	\end{bmatrix} \begin{bmatrix}
		{}^\mathcal{I}\dot{\bm{p}}_{B} \\
		{}^\mathcal{I}\bm{\omega}_{\mathcal{I},B} \\
		\dot{\bm{s}}
	\end{bmatrix}.
\end{equation}
where $\bm{v}_\text{CoM}$ is the CoM velocity relative to the inertial frame $\mathcal{I}$, expressed in a frame located on the CoM having orientation parallel to $\mathcal{I}$ (\emph{mixed} representation). Thanks to this choice of frame, the CoM jacobian $\bm{J}_\text{CoM}$ maps the base position time derivative ${}^\mathcal{I}\dot{\bm{p}}_{B}$, the \emph{right-trivialized} base angular velocity ${}^\mathcal{I}\bm{\omega}_{\mathcal{I},B}$ and the joint time derivative into the CoM velocity. Thus, we have:
\begin{IEEEeqnarray}{RCL}
	\frac{\partial}{\partial s} \text{CoM}({}^\mathcal{I}\bm{p}_B, {}\mathcal{I}\bm{\rho}_B, s) &=& \bm{J}_\text{CoM}^{\dot{s}}, \\
	\frac{\partial}{\partial {}^\mathcal{I}\bm{p}_B} \text{CoM}({}^\mathcal{I}\bm{p}_B, {}\mathcal{I}\bm{\rho}_B, s) &=& \bm{J}_\text{CoM}^{v_b}.
\end{IEEEeqnarray}
For what concerns the rotation part, we need the partial derivative with respect to the quaternion. Given that we are considering a Jacobian in mixed representation, we need to map a quaternion derivative into an angular velocity expressed in an inertial frame (\emph{right-trivialized}). This relation is defined in \citep[Sec. 1.5.3]{graf2008quaternions}:
\begin{equation}\label{eq:quaternionMap}
{}^\mathcal{I}\bm{\omega}_{\mathcal{I},B} = 2\bm{E}~{}^\mathcal{I}\dot{\bm{\rho}}_B.
\end{equation}
Finally we have
\begin{equation} \label{eq:comJacQuaternion}
\frac{\partial}{\partial {}^\mathcal{I}\bm{\rho}_B} \text{CoM}({}^\mathcal{I}\bm{p}_B, {}\mathcal{I}\bm{\rho}_B, s) = 2\bm{J}_\text{CoM}^{\omega_b}\bm{E}.
\end{equation}

Alternatively, we can obtain these partial derivative by starting from the following relation:
\begin{equation}
	\bm{x}_\text{CoM} = {}^\mathcal{I} \bm{H}_B {}^B\bm{p}_\text{CoM},
\end{equation}
where ${}^B\bm{p}_\text{CoM} \in \mathbb{R}^3$ is the CoM position expressed in base coordinates. In this way, the partial derivatives w.r.t base position and quaternion can be computed as in Sec. \ref{sec:base_transforms_derivative} and we are left with the computation of $\frac{\partial}{\partial \bm{s}} {}^B \bm{p}_\text{CoM}$. Exploiting the results of Sec. \ref{sec:rel_transf_derivative} and the CoM definition of Eq. \eqref{eq:com_definition}, the partial derivative w.r.t. joint $j$ is the following
\begin{equation}\label{eq:base_com_derivative}
	\begin{split}
			\frac{\partial}{\partial s_j} {}^B \bm{p}_\text{CoM} =& \sum_{i : j \in \pi_B(i)} \frac{m_i}{m}\left({}^B\bm{R}_i\begin{bmatrix}
				\mathds{1}_3 & \bm{0}_{3\times3}
			\end{bmatrix}{}^i \bm{X}_C {}^C\textbf{s}_{P,C} + \right. \\
			&+\left. \begin{bmatrix}
				\mathds{1}_3 & \bm{0}_{3\times3}
			\end{bmatrix} {}^B\bm{X}_C \left({}^C\textbf{s}_{P,C}\times \right){}^C\bm{X}_i \begin{bmatrix}
			{}^i\bm{\rho}_{\text{CoM}} \\
			\bm{0}_{3\times 1}
		\end{bmatrix}\right),
	\end{split}
\end{equation}
where the summation is over those joints having $j$ in their path to the base.

\section{Partial derivatives of centroidal momentum}

In this section, we analyze Eq. \eqref{constr:CMM}. The partial derivative w.r.t the (\emph{left-trivialized}) base velocity and joint velocities consist in selecting the corresponding columns of the Centroidal Momentum Matrix $\bm{J}_\text{CMM}$, first presented in Sec. \ref{sec:intro_momentum}. In the following, we examine the dependency from the joint positions and base pose.

\subsection{Partial derivative with respect to CoM position and base pose}\label{sec:momentum_derivative_com}
Equation \eqref{eq:momentumExpanded} depends upon the base pose and CoM position only through matrix ${}_{\bar{G}} \bm{X}^B$.
We can rewrite Eq. \eqref{eq:momentumExpanded} as:
\begin{equation} \label{eq:momentumInBase}
{}_{\bar{G}} \bm{h} = {}_{\bar{G}} \bm{X}^\mathcal{I} {}_{\mathcal{I}} \bm{X}^B {}_{B} \bm{h}
\end{equation}
where ${}_{B} \bm{h}$ is the centroidal momentum expressed in base coordinates. We can expand Eq. \eqref{eq:momentumInBase}:
\begin{equation}
{}_{\bar{G}} \bm{h} = 
\begin{bmatrix}
\mathds{1}_3 & 0 \\
-\bm{x}_\text{CoM}^\wedge & \mathds{1}_3
\end{bmatrix}
\begin{bmatrix}
{}^{\mathcal{I}} \bm{R}_B & 0 \\
{}^{\mathcal{I}} \bm{p}^\wedge_B {}^{\mathcal{I}} \bm{R}_B  & {}^{\mathcal{I}} \bm{R}_B
\end{bmatrix}
\begin{bmatrix}
{}_{B} \bm{h}^{v} \\
{}_{B} \bm{h}^{\omega}
\end{bmatrix},
\end{equation}
obtaining
\begin{equation}
{}_{\bar{G}} \bm{h} = 
\begin{bmatrix}
{}^{\mathcal{I}} \bm{R}_B & 0 \\
{}^{\mathcal{I}} \bm{p}^\wedge_B {}^{\mathcal{I}} \bm{R}_B -\bm{x}_\text{CoM}^\wedge{}^{\mathcal{I}} \bm{R}_B  & {}^{\mathcal{I}} \bm{R}_B
\end{bmatrix}
\begin{bmatrix}
{}_{B} \bm{h}^{v} \\
{}_{B} \bm{h}^{\omega}
\end{bmatrix}.
\end{equation}
By substituting $\bm{x}_\text{CoM} = {}^{\mathcal{I}} \bm{R}_B {}^B\bm{p}_{\text{CoM}} + {}^\mathcal{I}\bm{p}_B$, exploiting the additivity property of the $\wedge$ operator and the rotational invariance of cross products, we obtain:
\begin{equation}
	{}_{\bar{G}} \bm{h} = 
	\begin{bmatrix}
	{}^{\mathcal{I}} \bm{R}_B & 0 \\
	-{}^B\bm{p}_{\text{CoM}}^\wedge  & {}^{\mathcal{I}} \bm{R}_B
	\end{bmatrix}
	\begin{bmatrix}
	{}_{B} \bm{h}^{v} \\
	{}_{B} \bm{h}^{\omega}
	\end{bmatrix}.
\end{equation}
Hence the centroidal momentum ${}_{\bar{G}} \bm{h}$ does not depend on the base position:
\begin{equation}
	\frac{\partial}{\partial {}^\mathcal{I}\bm{p}_B} {}_{\bar{G}} \bm{h} = 0.
\end{equation}

In order to compute the derivative w.r.t. the CoM position and the base quaternion, we can express the linear and angular momentum separately:
\begin{IEEEeqnarray}{RCL}
	\IEEEyesnumber \phantomsection\label{eq:AXBDerivative}
	{}_{\bar{G}} \bm{h}^v &=& {}^{\mathcal{I}} \bm{R}_B {}_{B} \bm{h}^{v}, \IEEEyessubnumber \\
	{}_{\bar{G}} \bm{h}^\omega &=& -{}^B\bm{p}_{\text{CoM}}^\wedge {}_B\bm{h}^{v} + {}^{\mathcal{I}} \bm{R}_B {}_{B} \bm{h}^{\omega} \IEEEyessubnumber 
\end{IEEEeqnarray}
For what concerns the partial derivatives of the rotation matrix with respect to the base quaternion, we can use the results of Section \ref{sec:rotationPartialDer}. Regarding the CoM position, we have: 
\begin{equation}
	\frac{\partial}{\partial {}^B\bm{p}_{\text{CoM}}}{}_{\bar{G}} \bm{h} = \begin{bmatrix}0_{3\times 3} \\ {}_{B} \bm{h}^{v\wedge}  \end{bmatrix}.
\end{equation}
The partial derivative w.r.t $\bm{x}_\text{CoM}$, i.e. $\frac{\partial {}^B\bm{p}_{\text{CoM}}}{\partial \bm{x}_\text{CoM}}$ can be easily computed by composition. Otherwise, it is possible to exploit Eq.\eqref{eq:base_com_derivative} to compute directly the partial derivative w.r.t joint variables for this component.

\subsection{Partial derivative with respect to joint values} \label{sec:CMMjoints}
Let us consider now the partial derivative w.r.t joint values:
\begin{equation} \label{eq:momentumJointDerivative}
\frac{\partial}{\partial s_j} {}_{\bar{G}} \bm{h} =\frac{\partial}{\partial s_j}\left({}_{\bar{G}} \bm{X}^B\right){}_{B} \bm{h} + {}_{\bar{G}} \bm{X}^B \frac{\partial}{\partial s_j} \left({}_{B} \bm{h} \right).
\end{equation}
The first derivative can be computed from the results of Sec. \ref{sec:momentum_derivative_com}. In this section we focus on $\frac{\partial}{\partial s_j} \left({}_{B} \bm{h} \right)$. 

The centroidal momentum expressed in base coordinates is the sum of all the link momenta expressed in the same frame. Hence:
\begin{equation}\label{eq:momentum_initial_derivative}
\frac{\partial}{\partial s_j} {}_{B} \bm{h} = \sum_i \frac{\partial}{\partial s_j} \left({}_{B} \bm{X}^i \right)  \bm{I}_i {}^i \bm{V}_{\mathcal{I},i} + \sum_i {}_{B} \bm{X}^i \bm{I}_i \frac{\partial}{\partial s_j}\left( {}^i \bm{V}_{\mathcal{I},i} \right).
\end{equation}
The derivative $\frac{\partial}{\partial s_j} \left({}_{B} \bm{X}^i \right)$ is obtained as in Eq. \eqref{eq:adjoint_wrench_derivative}, while Eq. \eqref{eq:velocity_inertial_derivative} presents the partial derivative of the link velocity. Thus, we can rewrite Eq. \eqref{eq:momentum_initial_derivative} as follows:
\begin{IEEEeqnarray}{RCL}
	\IEEEyesnumber \phantomsection
	\frac{\partial}{\partial s_j} {}_{B} \bm{h} &=& \sum_{i : j \in \pi_B(i)}{}_{B} \bm{X}^C {}^C\textbf{s}_{P,C}\bar{\times}^* {}_C \bm{X}^i  \bm{I}_i {}^i \bm{V}_{\mathcal{I},i} - {}_{B} \bm{X}^i \bm{I}_i {}^i\bm{X}_C {}^C\textbf{s}_{P,C} \times \left({}^C\bm{V}_{\mathcal{I},C}\right) \nonumber \\
	&=& {}_{B} \bm{X}^C {}^C\textbf{s}_{P,C}\bar{\times}^*  \left(\sum_{i : j \in \pi_B(i)} {}_C \bm{X}^i  \bm{I}_i {}^i \bm{V}_{\mathcal{I},i}\right) + \nonumber \\
	&&- \left(\sum_{i : j \in \pi_B(i)} {}_{B} \bm{X}^i \bm{I}_i {}^i\bm{X}_C\right) {}^C\textbf{s}_{P,C} \times \left({}^C\bm{V}_{\mathcal{I},C}\right),
\end{IEEEeqnarray}
with $j$ connecting the child link $C$ to its parent $P$ and the sums being on the links which have $j$ in their path to the base.

\subsection{Partial derivative with respect to base and joint velocities}
Eq. \eqref{eq:momentumExpanded} depends upon the base velocity ${}^B\bm{V}_{\mathcal{I},B}$ through ${}^i\bm{V}_{\mathcal{I},i}$. By using Eq. \eqref{eq:velocity_inertial}, we can write:
\begin{equation}
	\frac{\partial}{\partial {}^B\bm{V}_{\mathcal{I},B}} {}^i\bm{V}_{\mathcal{I},i} = {}^i\bm{X}_B,
\end{equation}
hence
\begin{equation}
	\frac{\partial}{\partial {}^B\bm{V}_{\mathcal{I},B}} {}_{\bar{G}} \bm{h} = {}_{\bar{G}} \bm{X}^B \sum_i {}_{B} \bm{X}^i \bm{I}_i {}^i\bm{X}_B.
\end{equation}
This corresponds to the first $6$ columns of the CMM matrix. The other rows corresponds to the partial derivative with respect to joint velocities. Given a generic joint $j$, the partial derivative w.r.t. $\dot{s}_j$ is:
\begin{equation}
	\frac{\partial}{\partial \dot{s}_j} {}_{\bar{G}} \bm{h} = {}_{\bar{G}} \bm{X}^B \sum_{i : j \in \pi_B(i)} {}_{B} \bm{X}^i \bm{I}_i {}^i\bm{X}_{C}  {}^{C}\textbf{s}_{P,C},
\end{equation}
where we used the relation presented in Eq. \eqref{eq:velocity_expanded}.

\section{Partial derivatives of a quaternion as rotation error}

Sec. \ref{sec:orientation_task} adopts a task which minimizes the rotation error of a generic frame attached to the robot. Since the reference orientation is expressed in the inertial frame $\mathcal{I}$, this error depends on the kinematic status of the robot, viz. joint positions and base pose.

To compute the partial derivatives of the function $\texttt{quat}({}^A\tilde{\bm{R}}_\text{frame})$, let us start by considering the time derivative of the quaternion error $\bm{\rho}_{\text{err}}$ as in \citep[Section 1.5.4]{graf2008quaternions}:
\begin{equation}
\dot{\bm{\rho}}_\text{err} = \frac{1}{2} \bm{\mathcal{G}}(\bm{\rho}_{\text{err}})^\top \bm{\omega}_\text{err},
\end{equation}
where $\bm{\omega}_\text{err}$ is the \emph{left-trivialized} angular velocity. $\bm{\rho}_\text{err}$ is the quaternion associated to the rotation error matrix ${}^\mathcal{I}\tilde{\bm{R}}_\text{frame}={}^\mathcal{I}\bm{R}^{*\top}_\text{frame}{}^\mathcal{I}\bm{R}_\text{frame}$, while the \emph{left-trivialized} velocity $\bm{\omega}_\text{err}$ is obtained as:
\begin{equation}
\bm{\omega}_\text{err} = \left({}^\mathcal{I}\tilde{\bm{R}}_\text{frame}^\top{}^\mathcal{I}\dot{\tilde{\bm{R}}}_\text{frame}\right)^\vee.
\end{equation}
By substituting we have
\begin{IEEEeqnarray}{RCL}
	\IEEEyesnumber \phantomsection
	\dot{\bm{\rho}}_\text{err} &=& \frac{1}{2} \bm{\mathcal{G}}(\bm{\rho}_{\text{err}})^\top\left({}^\mathcal{I}\tilde{\bm{R}}_\text{frame}^\top{}^\mathcal{I}\dot{\tilde{\bm{R}}}_\text{frame}\right)^\vee, \IEEEyessubnumber \\
	&=& \frac{1}{2} \bm{\mathcal{G}}(\bm{\rho}_{\text{err}})^\top\left({}^\mathcal{I}\bm{R}^\top_\text{frame} {}^\mathcal{I}\bm{R}^{*}_\text{frame} \frac{\dif}{\dif t}\left({}^\mathcal{I}\bm{R}^{*\top}_\text{frame}{}^\mathcal{I}\bm{R}_\text{frame}\right)\right)^\vee, \IEEEyessubnumber \\
	&=& \frac{1}{2} \bm{\mathcal{G}}(\bm{\rho}_{\text{err}})^\top\left({}^\mathcal{I}\bm{R}^\top_\text{frame} {}^\mathcal{I}\bm{R}^{*}_\text{frame} {}^\mathcal{I}\bm{R}^{*\top}_\text{frame}\frac{\dif}{\dif t}\left({}^\mathcal{I}\bm{R}_\text{frame}\right)\right)^\vee, \IEEEyessubnumber \\
	&=& \frac{1}{2} \bm{\mathcal{G}}(\bm{\rho}_{\text{err}})^\top\left({}^\mathcal{I}\bm{R}^\top_\text{frame}{}^\mathcal{I}\dot{\bm{R}}_\text{frame}\right)^\vee, \IEEEyessubnumber \\
	&=& \frac{1}{2} \bm{\mathcal{G}}(\bm{\rho}_{\text{err}})^\top {}^\text{frame}\bm{\omega}_{\mathcal{I}, \text{frame}},
\end{IEEEeqnarray}
with ${}^\text{frame}\bm{\omega}_{\mathcal{I}, \text{frame}}$ the \emph{left-trivialized} frame angular velocity. This velocity can be computed using the bottom three rows of the corresponding \emph{left-trivialized} Jacobian ${}^\omega \bm{J}$, where we drop the subscripts for the sake of simplicity:
\begin{equation}
\begin{split}
	{}^\text{frame}\bm{\omega}_{\mathcal{I}, \text{frame}} &= {}^\omega \bm{J} \begin{bmatrix}
	{}^\mathcal{I}\dot{\bm{p}}_{B} \\
	{}^B\bm{\omega}_{\mathcal{I},B} \\
	\dot{\bm{s}}
	\end{bmatrix},\\
	&= \begin{bmatrix}
	{}^\omega \bm{J}^{v_b} & {}^\omega \bm{J}^{\omega_b} & {}^\omega \bm{J}^{\dot{s}}
	\end{bmatrix} \begin{bmatrix}
	{}^\mathcal{I}\dot{\bm{p}}_{B} \\
	2\bm{\mathcal{G}}({}^\mathcal{I}\bm{\rho}_{B}){}^\mathcal{I}\dot{\bm{\rho}}_B \\
	\dot{\bm{s}}
	\end{bmatrix}.
\end{split}
\end{equation}
Henceforth, the following derivatives are easily obtained:
\begin{IEEEeqnarray}{RCL}
	\frac{\partial}{\partial {}^\mathcal{I}\bm{p}_{B}} \bm{\rho}_\text{err} &=& \frac{1}{2} \bm{\mathcal{G}}(\bm{\rho}_{\text{err}})^\top {}^\omega \bm{J}^{v_b}, \\
	\frac{\partial}{\partial {}^\mathcal{I}\bm{\rho}_{B}} \bm{\rho}_\text{err} &=& \bm{\mathcal{G}}(\bm{\rho}_{\text{err}})^\top {}^\omega \bm{J}^{\omega_b}\bm{\mathcal{G}}({}^\mathcal{I}\bm{\rho}_{B}), \\
	\frac{\partial}{\partial \bm{s}} \bm{\rho}_\text{err} &=& \frac{1}{2} \bm{\mathcal{G}}(\bm{\rho}_{\text{err}})^\top {}^\omega \bm{J}^{\dot{s}}.
\end{IEEEeqnarray}

\chapter{Methods for Computing the Hessian of the Lagrangian Involving Kinematic and Dynamic Quantities}\label{chap:hessian}
In this appendix, we compute the Hessian of the Lagrangian related to the optimal control problem presented Sec. \ref{sec:oc}. In particular, in Sec. \ref{sec:hessian_preliminaries} we introduce few mathematical tools useful to simplify the calculations. Most of the derivatives require the mechanical application of the results presented in Appendix \ref{chap:jacobians}, thus they are not presented. In this appendix, we show those derivatives which are not trivial.

\section{Preliminaries}\label{sec:hessian_preliminaries}
Let us introduce few concepts that are helpful in the computation of the Hessian of the Lagrangian.
\subsection{Lagrangian Hessian by columns}
Let us modify the Lagrangian presented in Eq. \eqref{eq:lagrangian} as follows: 
\begin{equation}
	\mathcal{L} = \Gamma(\bm{\chi}) + \bm{\lambda}^\top \bm{h}(\bm{\chi}),
\end{equation}
where we stacked together both equality and inequality constraints and their respective multipliers. Hence the Hessian, i.e. its second partial derivative w.r.t. $\bm{\chi}$, is obtained as
\begin{equation}\label{eq:hessian_lagrangian_first}
	\frac{\partial^2}{\partial \bm{\chi}^2} \mathcal{L} = \frac{\partial^2}{\partial \bm{\chi}^2} \Gamma(\bm{\chi}) + \sum_i \lambda_i \frac{\partial^2}{\partial \bm{\chi}^2} h_i(\bm{\chi}).
\end{equation}
The second part of Eq. \eqref{eq:hessian_lagrangian_first} involves a linear combination of the partial derivatives of each constraint row, using the Lagrangian multipliers as coefficients. Nevertheless, as shown in Appendix \ref{chap:jacobians}, we are able to define the partial derivative of kinematic and dynamic quantities w.r.t a single variable, in an easier way compared to the derivative of a specific row w.r.t all the variables. For example, we are able to define the partial derivative of the velocity of a link w.r.t a single joint, but not the partial derivative of the $x$-component of the velocity w.r.t all the joint variables. In other words, it is easier to define analytically
\begin{equation*}
	\frac{\partial}{\partial \chi_i}\left(\frac{\partial}{\partial \chi_j}\bm{h}(\bm{\chi})\right) 
\end{equation*}
 rather than $\frac{\partial^2}{\partial \bm{\chi}^2} h_i(\bm{\chi})$. With this rationale, we can notice the following 
\begin{IEEEeqnarray}{RCL}
	\IEEEyesnumber
	\frac{\partial}{\partial \bm{\chi}} \mathcal{L} &=& \frac{\partial}{\partial \bm{\chi}} \Gamma(\bm{\chi}) {+} \begin{bmatrix}
		\bm{\lambda}^\top \frac{\partial}{\partial \chi_1}\bm{h}(\bm{\chi}) & \bm{\lambda}^\top \frac{\partial}{\partial \chi_2}\bm{h}(\bm{\chi}) & \cdots & \bm{\lambda}^\top \frac{\partial}{\partial \chi_n}\bm{h}(\bm{\chi})
	\end{bmatrix}, \IEEEeqnarraynumspace \IEEEyessubnumber\\
	&=& \nabla_{\bm{\chi}} \Gamma(\bm{\chi}) + \bm{\lambda}^\top\begin{bmatrix}   \left(\nabla_{\bm{\chi}}\bm{h}(\bm{\chi})\right)^{(1)} &  \cdots & \left(\nabla_{\bm{\chi}}\bm{h}(\bm{\chi})\right)^{(n)}
	\end{bmatrix}, \IEEEyessubnumber
\end{IEEEeqnarray}
where the apex $(j)$ indicates the $j$th column. This highlights that $\frac{\partial}{\partial \chi_i}\bm{h}(\bm{\chi}) = \left(\nabla_{\bm{\chi}}\bm{h}(\bm{\chi})\right)^{(i)}$. Hence, we have
\begin{equation}
	\frac{\partial^2}{\partial \chi_i \partial \chi_j} \mathcal{L} = \frac{\partial^2}{\partial \chi_i \partial \chi_j} \Gamma(\bm{\chi}) + \bm{\lambda}^\top \frac{\partial}{\partial \chi_j} \left(\nabla_{\bm{\chi}}\bm{h}(\bm{\chi})\right)^{(i)},
\end{equation}
thus building the Hessian by elements.

This simple rearrangement allows us to reuse the results of Appendix \ref{chap:jacobians}, and focus only on the computation of a generic column of each constraint Jacobian, i.e. $\left(\nabla_{\bm{\chi}}\bm{h}(\bm{\chi})\right)^{(i)}$. In case of sum of two matrices, the derivative of a column is simply the sum of the derivatives. In the following, we analyze the derivative of a column resulting from the product of two matrices.

\subsection{Partial derivative of a matrix product's column}

Consider two generic matrices $\bm{A}$ and $\bm{B}$ of suitable dimensions, both depending on $\bm{\chi}$. Our goal is to define 
\begin{equation}
	\frac{\partial}{\partial \bm{\chi}}\left(\bm{A}\bm{B}\right)^{(j)}.
\end{equation}
The $j$-th column of the product is equal to
\begin{equation}
    \left(\bm{A}\bm{B}\right)^{(j)} = \bm{A}\left(\bm{B}\right)^{(j)},
\end{equation}
i.e. it is equal to $\bm{A}$ times the $j$-th column of $\bm{B}$. Hence
\begin{equation}\label{eq:product_derivative_initial}
\frac{\partial}{\partial \bm{\chi}}\left(\bm{A}\bm{B}\right)^{(j)} = \bm{A}\frac{\partial}{\partial \bm{\chi}}\left(\bm{B}^{(j)}\right) + \frac{\partial}{\partial \bm{\chi}}\left(\bm{A}\right)\bm{B}^{(j)}.
\end{equation}
In order to compute the second term, we can write the product between $\bm{A}$ times and the $j$-th column of $\bm{B}$ as
\begin{equation}\label{eq:product_derivative_sum}
	\bm{A}\left(\bm{B}\right)^{(j)} = \sum_i b_{i,j} A^{(i)},
\end{equation}
where $b_{i,j}$ is the element $(i, j)$ of $\bm{B}$. In other words, the product between matrix $\bm{A}$ and vector $\bm{B}^{(j)}$ can be seen as the linear combination of all the columns of $\bm{A}$, using the components of $\bm{B}^{(j)}$ as coefficients. We can exploit this property to finally write
\begin{equation}\label{eq:product_rule}
	\frac{\partial}{\partial \bm{\chi}}\left(\bm{A}\bm{B}\right)^{(j)} = \bm{A}\frac{\partial}{\partial \bm{\chi}}\left(\bm{B}^{(j)}\right) + \sum_i b_{i,j}\frac{\partial}{\partial \bm{\chi}}\left(\bm{A}^{(i)}\right).
\end{equation}

It is worth mentioning the following property:
\begin{equation}
\frac{\partial}{\partial \bm{\chi}}\left(\bm{A}^{(i)}\right) \bm{e}_j = \frac{\partial}{\partial \chi_j}\left(\bm{A}\right)\bm{e}_i.
\end{equation}
This is trivially obtained by noticing that the partial derivative of the $i$-th column of $\bm{A}$ w.r.t $\bm{\chi}$ corresponds to 
\begin{equation}
	\frac{\partial}{\partial \bm{\chi}}\left(\bm{A}^{(i)}\right) = \begin{bmatrix}
	\frac{\partial}{\partial \chi_1}\left(\bm{A}^{(i)}\right) & \frac{\partial}{\partial \chi_2}\left(\bm{A}^{(i)}\right) & \cdots & \frac{\partial}{\partial \chi_n}\left(\bm{A}^{(i)}\right)
	\end{bmatrix},
\end{equation}
while
\begin{equation}
	\frac{\partial}{\partial \chi_j} \bm{A} = \begin{bmatrix}
	\frac{\partial}{\partial \chi_j}\left(\bm{A}^{(1)}\right) & \frac{\partial}{\partial \chi_j}\left(\bm{A}^{(2)}\right) & \cdots & \frac{\partial}{\partial \chi_j}\left(\bm{A}^{(n_A)}\right)
	\end{bmatrix},
\end{equation}
with $n_A$ being the number of columns of $\bm{A}$. Hence, the $j$-th column of $\frac{\partial}{\partial \bm{\chi}}\left(\bm{A}^{(i)}\right)$ corresponds to the $i$-th column of the partial derivative of $\bm{A}$ with respect to $j$.
We can exploit this property when selecting a column of Eq. \eqref{eq:product_derivative_sum}'s derivative:
\begin{IEEEeqnarray}{RCL}
	\IEEEyesnumber \phantomsection
	\sum_i b_{i,j}\frac{\partial}{\partial \bm{\chi}}\left(\bm{A}^{(i)}\right) \bm{e}_k &=& \sum_i b_{i,j}\frac{\partial}{\partial \chi_k}\left(\bm{A}\right)\bm{e}_i, \nonumber \\
	&=& \frac{\partial}{\partial \chi_k}\left(\bm{A}\right) \bm{B}^{(j)}.
\end{IEEEeqnarray}
Hence, it is of no surprise that 
\begin{equation*}
	\sum_i b_{i,j}\frac{\partial}{\partial \bm{\chi}}\left(\bm{A}^{(i)}\right) = \begin{bmatrix}
	    \frac{\partial}{\partial \chi_1}\left(\bm{A}\right) \bm{B}^{(j)} & \frac{\partial}{\partial \chi_2}\left(\bm{A}\right) \bm{B}^{(j)} & \cdots & \frac{\partial}{\partial \chi_n}\left(\bm{A}\right) \bm{B}^{(j)}
	\end{bmatrix}.
\end{equation*}
This means that the second component of Eq. \eqref{eq:product_derivative_initial} also corresponds to the derivative of $\bm{A}$ by all the components of $\bm{\chi}$, each one multiplied by the $j-$th column of $\bm{B}$.
\subsection{Derivative of a skew symmetric matrix's column}
Given $\bm{x} \in \mathbb{R}^3$, the skew symmetric matrix associated to it is the following:
\begin{equation}
	\bm{x}^\wedge = \begin{bmatrix}
	0 & -x3 & x_2 \\ 
	x_3 & 0 & -x_1 \\
	-x_2 & x_1 & 0
	\end{bmatrix}.
\end{equation}
The partial derivative of each column are easily computed:
\begin{IEEEeqnarray}{RCL}
	\IEEEyesnumber \phantomsection \label{eq:skew_derivative}
	\frac{\partial}{\partial \bm{x}} \left(\bm{x}^{\wedge(1)}\right) &=& \begin{bmatrix}
			0 & 0 & 0 \\ 
			0 & 0 & 1 \\
			0 & -1 & 0
	\end{bmatrix}, \nonumber\\
	\frac{\partial}{\partial \bm{x}} \left(\bm{x}^{\wedge(2)}\right) &=& \begin{bmatrix}
		0 & 0 & -1 \\ 
		0 & 0 & 0 \\
		1 & 0 & 0
	\end{bmatrix}, \\
	\frac{\partial}{\partial \bm{x}} \left(\bm{x}^{\wedge(3)}\right) &=& \begin{bmatrix}
		0 & 1 & 0 \\ 
		-1 & 0 & 0 \\
		0 & 0 & 0
	\end{bmatrix}. \nonumber
\end{IEEEeqnarray}
Interestingly, these corresponds to the three-dimensional Levi-Civita symbol, or permutation symbol.

The results obtained in the previous sections represent building blocks for computing the Hessian of the full optimization problem. In particular, we can notice that most of the derivatives shown in Appendix \ref{chap:jacobians} are sum or products of matrices. 

\section{Partial derivatives with respect to joint variables}
In the following, we show the second order derivative of the quantities which may not be trivial or mechanically obtained using the properties showed above. As an example, the second derivative of a rotation matrix with respect to the associated quaternion can be computed applying the properties from the two subsections above, and it is not developed here. On the contrary, we show the second derivatives of an adjoint matrix and of centroidal quantites w.r.t joint variables.
\subsection{Second partial derivative of a adjoint matrix}\label{sec:adjoint_hessian}
The adjoint matrix is the most common object in the derivatives of Appendix \ref{chap:jacobians}.
The derivative of a generic adjoint matrix with respect to joint $j$ is defined in Eq. \eqref{eq:multiple_adjoints_derivative}. First of all, by noticing that 
\begin{equation}
	\frac{\partial^2}{\partial s_j \partial s_k}(\cdot) = \frac{\partial^2}{\partial s_k \partial s_j}(\cdot),
\end{equation}
we can assume, without loss of generality, that $k \in \pi_D(C)$. In other words, we assume joint $k$ to be either equal to $j$ or closer to the base link $D$. With this assumption, ${}^C\bm{X}_E$ does not depend on $s_k$. Define $\varphi$ and $\theta$ as the parent and child link of joint $k$, respectively. We finally obtain 
\begin{equation}
	\frac{\partial^2}{\partial s_k \partial s_j}{}^D\bm{X}_E = {}^D\bm{X}_\theta \left({}^\theta\textbf{s}_{\varphi,\theta}\times \right) {}^\theta\bm{X}_C\left({}^C\textbf{s}_{P,C}\times \right){}^C\bm{X}_E.
\end{equation}

\subsection{Second partial derivative of the center of mass position}\label{sec:com_double_derivative}
We focus on the center of mass position measured in the base frame, thus removing the dependency from the base pose. Starting from Eq.\eqref{eq:base_com_derivative}, we can rewrite it as
\begin{equation*}
	\begin{split}
		\frac{\partial}{\partial s_j} {}^B \bm{p}_\text{CoM} =\begin{bmatrix}
		\mathds{1}_3 & \bm{0}_{3\times3}
		\end{bmatrix}& \sum_{i : j \in \pi_B(i)} \frac{m_i}{m}\left({}^B\bm{X}_i\begin{bmatrix}
		\mathds{1}_3 & \bm{0}_{3\times3} \\
		\bm{0}_{3\times3} & \bm{0}_{3\times3}
		\end{bmatrix}{}^i \bm{X}_C {}^C\textbf{s}_{P,C} + \right. \\
		&+\left. {}^B\bm{X}_C \left({}^C\textbf{s}_{P,C}\times \right){}^C\bm{X}_i \begin{bmatrix}
		{}^i\bm{\rho}_{\text{CoM}} \\
		\bm{0}_{3\times1}
		\end{bmatrix}\right),
	\end{split}
\end{equation*}
where we substituted the rotation matrix ${}^B\bm{R}_i$ with the corresponding adjoint. We move the selector and the transform ${}^B\bm{X}_C$ outside of the summation:
\begin{equation*}
\begin{split}
\frac{\partial}{\partial s_j} {}^B \bm{p}_\text{CoM} =\begin{bmatrix}
\mathds{1}_3 & \bm{0}_{3\times3}
\end{bmatrix}& {}^B\bm{X}_C \sum_{i : j \in \pi_B(i)} \frac{m_i}{m}\left({}^C\bm{X}_i\begin{bmatrix}
\mathds{1}_3 & \bm{0}_{3\times3} \\
\bm{0}_{3\times3} & \bm{0}_{3\times3}
\end{bmatrix}{}^i \bm{X}_C {}^C\textbf{s}_{P,C} + \right. \\
&+\left.\left({}^C\textbf{s}_{P,C}\times \right){}^C\bm{X}_i \begin{bmatrix}
{}^i\bm{\rho}_{\text{CoM}} \\
\bm{0}_{3\times 1}
\end{bmatrix}\right).
\end{split}
\end{equation*}
As Sec. \ref{sec:adjoint_hessian}, we compute the partial derivative w.r.t another joint $k$, such that $k \in \pi_B(C)$. Also in this case, there is no loss of generality. Under this assumption, ${}^C\bm{X}_i$ does not depend on $s_k$. In fact, the summation is over all the links which have $j$ in their path to the base, and, because of the assumption, the subtree between $C$ (the child link of $j$) and $i$ cannot contain joint $k$. Henceforth, only matrix ${}^B\bm{X}_C$ depends on joint $k$ and the derivative corresponds to:
\begin{equation} \label{eq:com_in_base_double_derivative}
\begin{split}
\frac{\partial}{\partial s_j \partial s_k}& {}^B \bm{p}_\text{CoM} =\begin{bmatrix}
\mathds{1}_3 & \bm{0}_{3\times3}
\end{bmatrix} {}^B\bm{X}_\theta \left({}^\theta\textbf{s}_{\varphi,\theta}\times \right) {}^\theta\bm{X}_C \sum_{i : j \in \pi_B(i)} \frac{m_i}{m}\Bigg({}^C\bm{X}_i\cdot \\
&\cdot\left.\begin{bmatrix}
\mathds{1}_3 & \bm{0}_{3\times3} \\
\bm{0}_{3\times3} & \bm{0}_{3\times3}
\end{bmatrix}
{}^i \bm{X}_C {}^C\textbf{s}_{P,C} +\left({}^C\textbf{s}_{P,C}\times \right){}^C\bm{X}_i \begin{bmatrix}
{}^i\bm{\rho}_{\text{CoM}} \\
\bm{0}_{3\times 1}
\end{bmatrix}\right).
\end{split}
\end{equation}

If $k \notin \pi_B(C)$ and $j \notin \pi_B(\theta)$, i.e. joint $j$ and joint $k$ belong to two different subtrees (depending on the choice of base), then the derivative is null.

\subsection{Second partial derivative of the centroidal momentum}
Starting from Eq. \eqref{eq:momentumJointDerivative}, we write the second order partial derivative of the centroidal momentum as follows:
\begin{equation}
\begin{split}
	\frac{\partial^2}{\partial s_j \partial s_k} {}_{\bar{G}} \bm{h} =&\frac{\partial^2}{\partial s_j\partial s_k}\left({}_{\bar{G}} \bm{X}^B\right){}_{B} \bm{h} + \frac{\partial}{\partial s_j}\left({}_{\bar{G}} \bm{X}^B\right)\frac{\partial}{\partial s_k}\left({}_{B} \bm{h}\right) + \\
	&+ \frac{\partial}{\partial s_k}\left({}_{\bar{G}} \bm{X}^B\right)\frac{\partial}{\partial s_j}\left({}_{B} \bm{h}\right) + {}_{\bar{G}} \bm{X}^B \frac{\partial^2}{\partial s_j\partial s_k} \left({}_{B} \bm{h} \right).
\end{split}
\end{equation}
While $\frac{\partial}{\partial s_j}\left({}_{\bar{G}} \bm{X}^B\right)\frac{\partial}{\partial s_k}\left({}_{B} \bm{h}\right)$ and $\frac{\partial}{\partial s_k}\left({}_{\bar{G}} \bm{X}^B\right)\frac{\partial}{\partial s_j}\left({}_{B} \bm{h}\right)$ can be easily computed from the results of Sections \ref{sec:momentum_derivative_com} and \ref{sec:CMMjoints}, in the following we focus on the first and fourth terms.

We can rewrite ${}_{\bar{G}} \bm{X}^B$ as
\begin{equation}
	{}_{\bar{G}} \bm{X}^B = \begin{bmatrix}
	{}^{\mathcal{I}}\bm{R}_B & 0 \\
	0 & {}^{\mathcal{I}}\bm{R}_B
	\end{bmatrix} \begin{bmatrix}
	\mathds{1}_3 & \bm{0}_{3\times3} \\
	-{}^B\bm{p}_{\text{CoM}}^\wedge & \mathds{1}_3
	\end{bmatrix}.
\end{equation}
Hence, by exploiting Eq.s \eqref{eq:com_in_base_double_derivative} and \eqref{eq:skew_derivative}, we can compute its second order derivative w.r.t joint values.

For what concerns the second order derivative of the centroidal momentum expressed in the base frame, we can use the same assumption we do in Sec. \ref{sec:com_double_derivative} about the relative ordering between joint $j$ and joint $k$. As a consequence, the matrices ${}_C\bm{X}^i$ and ${}^i\bm{X}_C$ do not depend on $s_k$. Hence
\begin{IEEEeqnarray}{RCL}
	\IEEEyesnumber \phantomsection \label{eq:momentum_second_derivative_initial}
	\frac{\partial^2}{\partial s_j \partial s_k} {}_{B} \bm{h} &=&  \frac{\partial}{\partial s_k}\left({}_{B} \bm{X}^C\right) \left[{}^C\textbf{s}_{P,C}\bar{\times}^*  \left(\sum_{i : j \in \pi_B(i)} {}_C \bm{X}^i  \bm{I}_i  \frac{\partial}{\partial s_k}\left({}^i \bm{V}_{\mathcal{I},i}\right)\right) + \right. \nonumber \\
	&& \left. - \left(\sum_{i : j \in \pi_B(i)} {}_{C} \bm{X}^i \bm{I}_i {}^i\bm{X}_C\right) {}^C\textbf{s}_{P,C} \times  \frac{\partial}{\partial s_k}\left({}^C\bm{V}_{\mathcal{I},C}\right)\right].
\end{IEEEeqnarray}
The three derivatives included in Eq. \eqref{eq:momentum_second_derivative_initial} are readily computed as in Sec.s \ref{sec:adjoint_derivatives} and \ref{sec:velocity_derivative}, obtaining the following result:
\begin{IEEEeqnarray}{RCL}
	\IEEEyesnumber \phantomsection 
	\frac{\partial^2}{\partial s_j \partial s_k} {}_{B} \bm{h} &=&  {}_B\bm{X}^\theta \left({}^\theta\textbf{s}_{\varphi,\theta}\bar{\times}^* \right) {}_\theta\bm{X}^C \left[{}^C\textbf{s}_{P,C}\bar{\times}^*  \left(\sum_{i : j \in \pi_B(i)} {}_C \bm{X}^i  \bm{I}_i  {}^i \bm{X}_C\right) + \right. \nonumber \\
	&& \left. - \left(\sum_{i : j \in \pi_B(i)} {}_{C} \bm{X}^i \bm{I}_i {}^i\bm{X}_C\right) {}^C\textbf{s}_{P,C} \times  \right]{}^C \bm{X}_\theta{}^\theta\textbf{s}_{\varphi,\theta}\times{}^\theta \bm{V}_{\mathcal{I}, \theta}.
\end{IEEEeqnarray}

As for the center of mass position derivative, if $k$ and $j$ are on two different subtrees the derivative is null.

\section{The \texttt{levi} software library}

Appendices \ref{chap:jacobians} and \ref{chap:hessian} present derivatives of several quantities which can be used to generate complex expressions. Nevertheless, combining these equations ``by hand'' may be complex and error prone. For example, consider the case where several matrices are multiplied together. Their derivative would involve many combinations of the rule showed in Eq. \eqref{eq:product_rule}, complicating the code implementation. For this reason we implemented a library able to automatize this process. 

The \texttt{levi} software library is based on the concept of \texttt{evaluable}. It is an object comparable to a black-box. Nevertheless, we can retrieve its value and the derivatives. Common mathematical operators are \texttt{evaluable}s as well, holding the \texttt{expression}s involved in the operation. \texttt{expression}s are objects containing a \emph{shared pointer} to an \texttt{evaluable}. This allows us to easily handle the lifetime of an \texttt{evaluable}: if no \texttt{expression} points to it, the \texttt{evaluable} is deallocated because it means that this element is not involved into any mathematical expression.

For example, consider the following relation:
\begin{equation}
	X := A \times (B + C).
\end{equation}
$A$, $B$ and $C$ are \texttt{expression}s pointing to \texttt{evaluable}s defining their value. They can be constants or complicated custom functions. Then, $(B + C)$ is also an \texttt{expression} pointing to an \texttt{evaluable} which contain both $B$ and $C$. Finally, $X$ is again an \texttt{expression} pointing to the \texttt{evaluable} performing the product between the two \texttt{expression}s $A$ and $(B + C)$.

The derivative of an \texttt{evaluable} can be obtained by specifying the column to be derived and the variable with respect to which the derivative has to be retrieved. The output of this call is another \texttt{expression} containing a pointer to the \texttt{evaluable} corresponding to the derivative. This allows us to combine derivatives seamlessly with other \texttt{expression}s. For example, when requesting the derivative of the sum of two expressions, we obtain yet another sum between the respective column derivative of the two addends. Hence, we can easily obtain the symbolic expression starting from complicated equations. At the same time, given the black-box paradigm, entire formulas can be substituted with a single custom \texttt{evaluable} fastening the evaluation of the \texttt{expression}. This mechanism allows us to use the analytical formula of a derivative when available, or to compute it by symbolic manipulation.

In brief, \texttt{levi} allows evaluating the derivative of complex \texttt{expression}s by mixing symbolic computation with custom made \texttt{evaluable}s.

\end{appendices}
\bookmarksetup{startatroot}


\begin{spacing}{0.9}

\cleardoublepage
\phantomsection
\addcontentsline{toc}{chapter}{Bibliography}
\bibliography{References/PHDThesis} 
\bibliographystyle{plainnat}

\end{spacing}



\end{document}